\documentclass[conference]{IEEEtran}
\IEEEoverridecommandlockouts

\usepackage[square,sort&compress,comma,numbers]{natbib}

\usepackage[utf8]{inputenc} 
\usepackage[T1]{fontenc}    
\usepackage{url}            
\usepackage{booktabs}       
\usepackage{amsfonts}       
\usepackage{nicefrac}       
\usepackage{microtype}      
\usepackage{soul}
\usepackage[utf8]{inputenc}
\usepackage{graphicx}
\usepackage{amsmath}
\usepackage{booktabs}

\usepackage[colorlinks=true,
citecolor=blue,
filecolor=black,
linkcolor=blue,
urlcolor=blue]{hyperref}

\usepackage{microtype}
\usepackage{graphicx}
\usepackage{booktabs} 

\usepackage{amsmath,amsfonts,bm}

\usepackage{xcolor}
\usepackage{colortbl}

\def\eqref#1{equation~\ref{#1}}

\def\1{\bm{1}}

\DeclareMathAlphabet{\mathsfit}{\encodingdefault}{\sfdefault}{m}{sl}
\SetMathAlphabet{\mathsfit}{bold}{\encodingdefault}{\sfdefault}{bx}{n}

\newcommand{\E}{\mathbb{E}}

\newcommand{\R}{\mathbb{R}}

\usepackage{multirow}
\usepackage{paralist}

\usepackage{booktabs} 
\usepackage{multicol}
\usepackage{diagbox}

\usepackage[ruled,noend]{algorithm2e}

\SetCommentSty{mycommfont}

\usepackage{here}

\usepackage{amsmath,amssymb,amsfonts,amsbsy,amsfonts,latexsym}
\usepackage{multirow}
\usepackage{makecell}
\usepackage[labelfont=bf,textfont=it,belowskip=0pt,aboveskip=5pt,tableposition=top]{caption}
\usepackage{xcolor}
\usepackage{colortbl}

\definecolor{colorA}{RGB}{189,201,225}
\definecolor{colorB}{RGB}{103,169,207}
\definecolor{colorC}{RGB}{ 28,144,153}
\definecolor{colorD}{RGB}{  1,108, 89}

\newcolumntype{R}{>{\columncolor{gray!40}}r}
\newcolumntype{L}{>{\columncolor{gray!40}}l}
\newcolumntype{C}{>{\columncolor{gray!40}}c}

\usepackage{tabularx,colortbl,xcolor}
\usepackage{multirow}
\usepackage[normalem]{ulem}
\useunder{\uline}{\ul}{}

\usepackage{enumitem}

\usepackage{xparse}

\DeclareGraphicsExtensions{.pdf,.png}

\SetKwInput{KwInput}{Input}

\usepackage{longtable}
\usepackage{pgfplots}

\NewDocumentCommand{\var}{O{s} m O{}}{%
  \ensuremath{#1_{#2}^{#3}}
}
\usepackage{siunitx}

\newcommand{\commentout}[1]{}

\definecolor{light-gray}{gray}{0.80}

\newcommand\aref{Algorithm~\ref}

\newcommand\eref{Eq.~\ref}
\newcommand\fref{Figure~\ref}
\newcommand\tref{Table~\ref}
\newcommand\sref{Section~\ref}

\newcommand\ha{ \rowcolor{orange!0}}

\newcommand\hc{ \rowcolor{orange!40}}

\def\0{{\bf 0}}

\def\R{{\mathbb R}}

\usepackage{amsthm}

\usepackage{amsthm}

\newtheorem{mytheorem}{Theorem}
\newtheorem{assumption}[mytheorem]{Assumption}
\newtheorem{lemma}[mytheorem]{Lemma}
\newtheorem{remk}[mytheorem]{Remark}

\usepackage[font=scriptsize]{subfig}

\newcommand{\ResNetBN}{ResNet$_{-BN}$\xspace}
\newcommand{\ResNetRes}{ResNet$_{-Res}$\xspace}
\newcommand{\ResNet}{ResNet\xspace}
\newcommand{\OURS}{\textsc{PyHessian}\xspace}

\usepackage{chngcntr}

\begin{document}

\title{\OURS: Neural Networks Through the \\ Lens of the Hessian}
\author{
Zhewei Yao$^{*}$\thanks{$^{*}$Equal contribution.}, Amir Gholami$^{*}$, Kurt Keutzer, Michael W. Mahoney\\
University of California, Berkeley\\
{\tt\small \{zheweiy, amirgh, keutzer, mahoneymw\}@berkeley.edu}

}

\maketitle
\pagestyle{plain} 



\begin{abstract}
We present \OURS, a new scalable framework that enables fast computation of Hessian (i.e., second-order derivative) information for deep neural networks.
\OURS enables fast computations of the top Hessian eigenvalues, the Hessian trace, and the full Hessian eigenvalue/spectral density, and it supports distributed-memory execution on cloud/supercomputer systems
and is available as open source~\cite{pyhessian}.
This general framework can be used to analyze neural network models, including the topology of the loss landscape (i.e., curvature information) to gain insight into the behavior of different models/optimizers. 
To illustrate this, we analyze
the effect of residual connections and Batch Normalization layers on the trainability of neural networks.
One recent claim, based on simpler first-order analysis, is that residual connections and Batch Normalization make the loss landscape ``smoother'', thus making it easier for Stochastic Gradient Descent to converge to a good solution.
Our extensive analysis shows new finer-scale insights, demonstrating that, while conventional wisdom is sometimes validated, in other cases it is simply incorrect.
In particular, we find that Batch Normalization does not necessarily make the loss landscape smoother, especially for shallower networks. 
\end{abstract}

\section{Introduction}
\label{sec:intro}

Residual neural networks~\cite{he2016deep} (ResNets) are widely used Neural Networks (NNs) for various learning tasks.
The two main architectural components of ResNets are residual connections~\cite{he2016deep} and Batch Normalization (BN) layers~\cite{ioffe2015batch}. 
However, going beyond motivating stories to characterize precisely when and why these two popular architectural ingredients help or hurt training/generalization---especially in terms of \emph{measurable} properties of the model---is still largely unsolved.
Relatedly, characterizing whether other suggested architectural changes will help or hurt training/generalization is still done in a largely ad hoc manner.
For example, it is often motivated by plausible but untested intuitions, and it is not characterized in terms of measurable properties of the model.


In this work, we present and apply \OURS, an open source scalable framework with which one can directly analyze Hessian information, i.e., second-derivative information w.r.t. model parameters, in order to address these and related~questions.
\OURS computes Hessian information by applying known techniques from Numerical Linear Algebra (NLA)~\cite{bai1996some,
golub2009matrices,
lin2016approximating} 
and Randomized NLA (RandNLA)~\cite{Mah-mat-rev_BOOK,
RandNLA_PCMIchapter_chapter,
drineas2006fast,
yao2018hessian,
avron2011randomized,
ubaru2017fast} 
(that are approximate but come with rigorous theory).
\OURS enables computing Hessian information---including top Hessian eigenvalues, Hessian trace, and Hessian eigenvalue spectral density (ESD), and it supports  distributed implementation---allowing distributed-memory execution on both cloud (e.g., AWS, Google Cloud) and supercomputer systems, for fast and efficient Hessian computation. 

As an application of \OURS, we use it to analyze the impact of residual connections and BN on the trainability of NNs, leading
to new insights. In more detail, our main contributions are the following:


\begin{figure*}[!ht]
\centering
\includegraphics[width=0.32\textwidth]{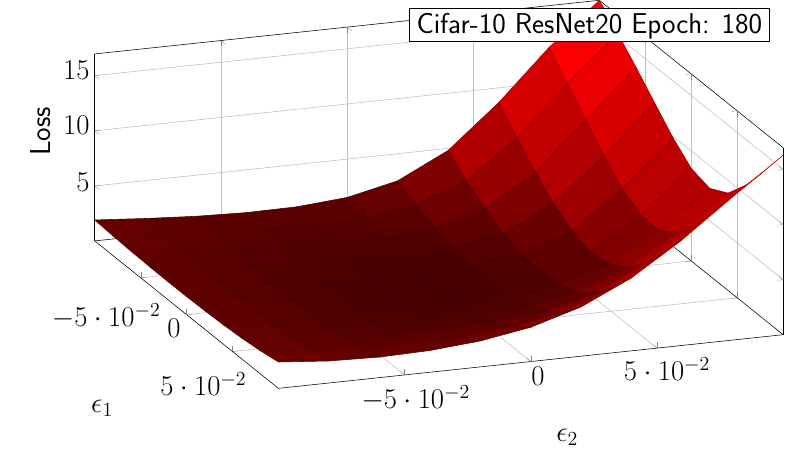}
\includegraphics[width=0.32\textwidth]{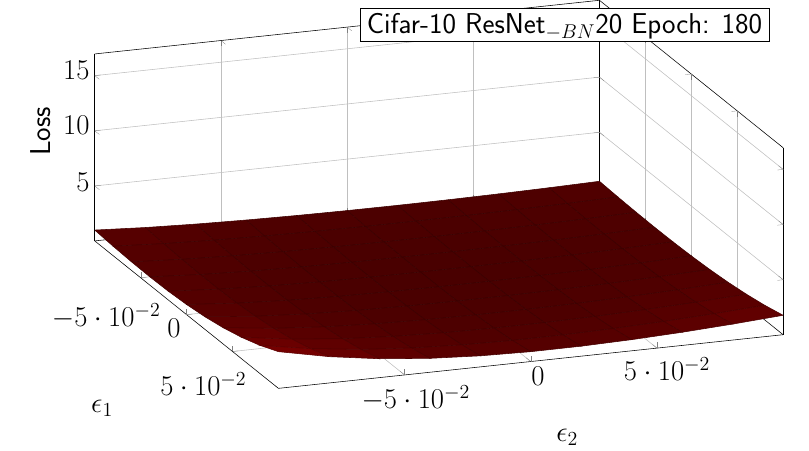}
\includegraphics[width=0.32\textwidth]{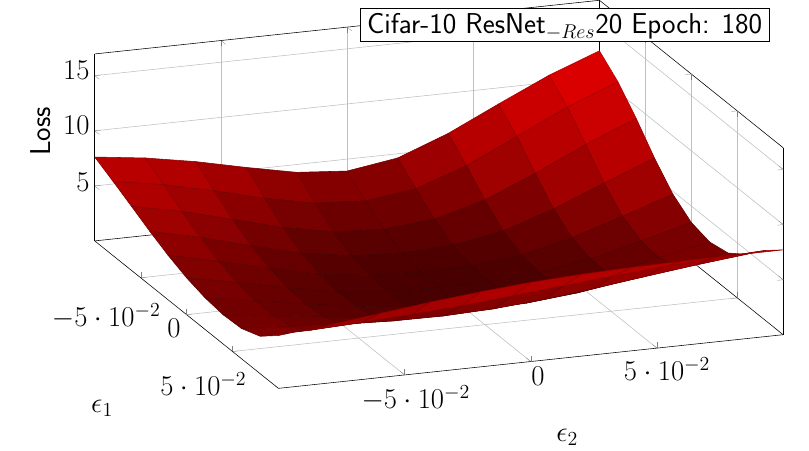}\\
\includegraphics[width=0.32\textwidth]{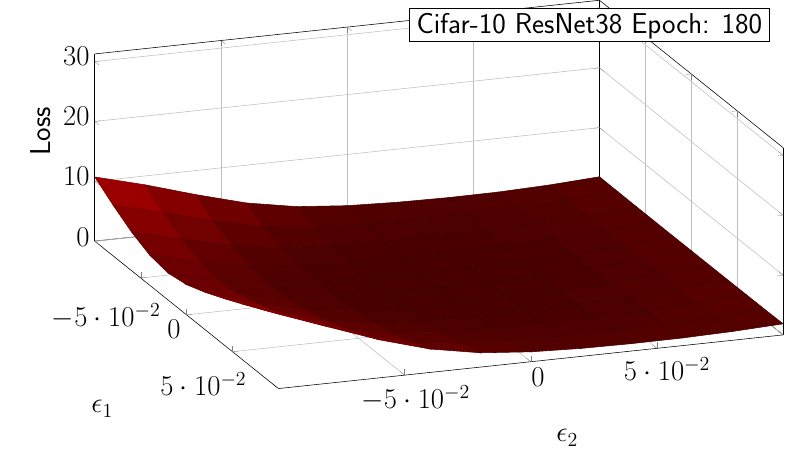}
\includegraphics[width=0.32\textwidth]{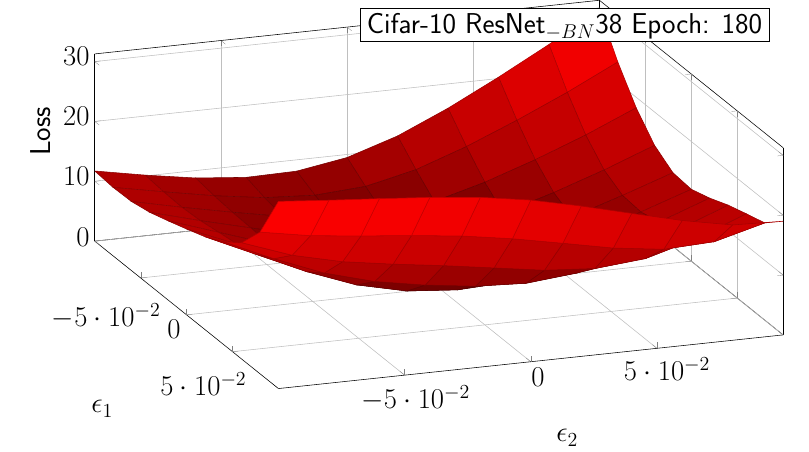}
\includegraphics[width=0.32\textwidth]{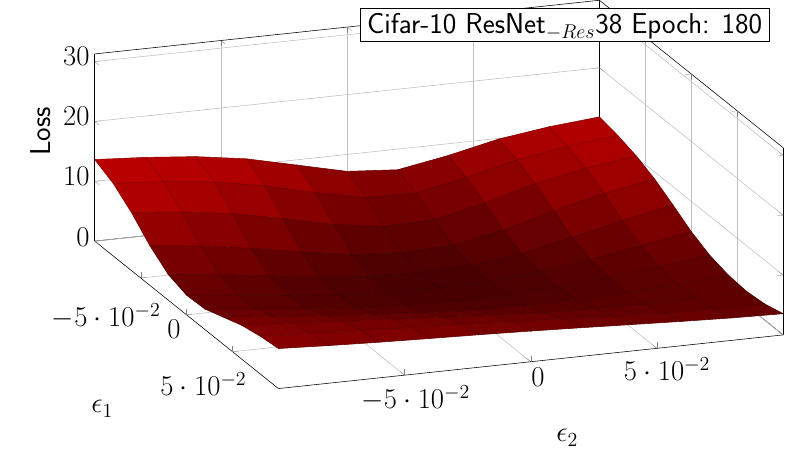}
\caption{
The parametric loss landscapes of ResNet20 (top) and ResNet38 (bottom) on Cifar-10 is plotted by perturbing the model parameters at the end of training across the first and second Hessian eigenvector.
Results for the original ResNet architecture (left), ResNet without BN (middle; denoted as \ResNetBN), and ResNet without residual connection (right; denoted as \ResNetRes).
It can be clearly seen that removing BN from ResNet20 actually leads to a smoother loss landscape, which is opposite to the common belief that adding BN leads to a smoother loss landscape~\cite{santurkar2018does}. 
We only observed the claimed smoothness property for the deeper ResNet38 model (second row).
This smoothness can be quantified by measuring the trace of the Hessian operator, reported in~\fref{fig:resnet20/32/38-hut-full-net}, as well as the full Hessian ESD, shown in~\fref{fig:resnet20-slq-full-net-part} and~\ref{fig:resnet38-slq-full-net-all}.
We also visualize the loss landscape throughout training for different epochs as shown in~\fref{fig:resnet20-loss-landscape-all} and~\ref{fig:resnet38-loss-landscape-all},
which provide further evidence.
Models trained on Cifar-100 also exhibit a similar behavior (as shown in~\fref{fig:resnet20-loss-landscape-all-cifar100}, \ref{fig:resnet32-loss-landscape-all-cifar100} and~\ref{fig:resnet38-loss-landscape-all-cifar100}). 
}
  \label{fig:resnet38-loss-landscape-part}
\end{figure*}

\begin{itemize}
    \item 
    We introduce \OURS, a new framework for direct and efficient computation of Hessian information, including the top eigenvalue, the trace, and the full ESD~\cite{pyhessian}. 
    We also apply \OURS to study how residual connections and BN affect training.
    \item
    We observe that removing BN layers from ResNet (denoted below as \ResNetBN) leads to \textbf{rapid increase
    of the Hessian spectrum} (the top eigenvalue, the trace, and the ESD support range).
    This increase is significantly more rapid for deeper models.
    See~\fref{fig:resnet20/32/38-hut-full-net}, \ref{fig:resnet20-slq-full-net-all},~\ref{fig:resnet32-slq-full-net-all}, and~\ref{fig:resnet38-slq-full-net-all} on Cifar-10
     as well as~\fref{fig:resnet20/32/38-hut-full-net-cifar100},~\ref{fig:resnet20-slq-full-net-all-cifar100},~\ref{fig:resnet32-slq-full-net-all-cifar100}, and~\ref{fig:resnet38-slq-full-net-all-cifar100} on Cifar-100.
    \item 
    We observe that, for shallower networks (ResNet20), removing the BN layer results in a flatter Hessian spectrum, as compared to standard ResNet20 with BN.  
    See~\fref{fig:resnet20/32/38-hut-full-net} and~\ref{fig:resnet20-slq-full-net-part} on Cifar-10 
    and~\fref{fig:resnet20/32/38-hut-full-net-cifar100} and~\ref{fig:resnet20-slq-full-net-all-cifar100} on Cifar-100. 
    This observation is the \textbf{opposite of the common belief} that the addition of BN layers make the loss landscape smoother (which we observe to hold only for deeper~networks).
    \item 
    We observe that, for deeper networks (in our case, ResNet32/38), removing BN results in converging to sharper local minima, as compared to ResNet with BN.
    See~\fref{fig:resnet20/32/38-hut-full-net}, \ref{fig:resnet20-slq-full-net-part}, \ref{fig:resnet32-loss-landscape-all} and~\ref{fig:resnet38-loss-landscape-all} on Cifar-10 as well as~\fref{fig:resnet20/32/38-hut-full-net-cifar100},~\ref{fig:resnet20-slq-full-net-all-cifar100},~\ref{fig:resnet32-loss-landscape-all-cifar100} and~\ref{fig:resnet38-loss-landscape-all-cifar100} on Cifar-100.
    \item 
    We show that removing residual connections from ResNet generally makes the top eigenvalue, the trace, and the Hessian ESD support range increase slightly. 
    This increase is consistent for both shallower and deeper models (ResNet20/32/38/56).
    See~\fref{fig:resnet20/32/38-hut-full-net}, \ref{fig:resnet20-slq-full-net-part}, ~\ref{fig:resnet56-slq-full-net-all},~\ref{fig:resnet20-loss-landscape-all},~\ref{fig:resnet32-loss-landscape-all},~\ref{fig:resnet38-loss-landscape-all}, and~\ref{fig:resnet56-loss-landscape-all} on Cifar-10  as well as~\fref{fig:resnet20/32/38-hut-full-net-cifar100},~\ref{fig:resnet20-slq-full-net-all-cifar100},~\ref{fig:resnet20-loss-landscape-all-cifar100},~\ref{fig:resnet32-loss-landscape-all-cifar100}, and~\ref{fig:resnet38-loss-landscape-all-cifar100} on Cifar-100.
    \item 
    We perform Hessian analysis for different stages of ResNet models (details in~\sref{sec:experiment_setting_and_convergence}), and we find that generally \textbf{BN is more important for the final stages} than for earlier stages.
    In particular, removing BN from the last stage significantly degrades testing performance, with a strong correlation with the Hessian trace.
    See the comparison between the orange curve and the blue curve in~\fref{fig:resnet32_stagewise_hut} and \ref{fig:resnet20/38_stagewise_hut}, the accuracy reported in~\tref{tab:model_acc_rm_bn} on Cifar-10 (\fref{fig:resnet20/38_stagewise_hut-cifar100}, and the accuracy reported  on Cifar-100 in~\tref{tab:model_acc_rm_bn-cifar100}).
\end{itemize}


\section{Related work}
\label{sec:related_work}

Here, we review work related to Hessian-based
analysis for  NN training and inference, as well as work that studies the impact of different architectural components on the topology of the NN loss landscape. 

\textbf{Hessian and Large-scale Hessian Computation:}
Hessian-based analysis/computation is widely used in scientific computing.
However, due to the (incorrect) belief that Hessian-based computations are infeasible for large NN problems, the majority of work in ML (except for quite small problems) performs only first-order analysis.%
\footnote{The na\"{\i}ve view arises since the Hessian matrix is of size (say) $m\times m$. Thus, like most linear algebra computations, exact full spectral computations (which are sufficient but never necessary) cost $\mathcal{O}(m^3)$ time.}  
However, using implicit or matrix-free methods, \emph{it is not even necessary to form the Hessian matrix explicitly in order to extract second-order information}.
Instead, it is possible to use stochastic methods from RandNLA to extract this information, without explicitly forming the Hessian matrix. 
For example, \cite{bai1996some,avron2011randomized} proposed fast algorithms for trace computation; and \cite{lin2016approximating,ubaru2017fast} provided efficient randomized algorithms to estimate the ESD of a positive semi-definite matrix. 
These algorithms only require an oracle for computing the product of the Hessian matrix with a given random vector. 
\emph{It is possible to compute this so-called ``matvec'' and extract Hessian information without explicitly forming the Hessian~\cite{becker1988improving,martens2010deep}.
In particular, using the so-called R-operator, the Hessian matvec can be computed with the same computational graph used for backpropagating the gradient~\cite{martens2010deep}.} 

Hessian eigenvalues of small NN models were analyzed~\cite{sagun2016eigenvalues,sagun2017empirical}; and the work of~\cite{pennington2017geometry} studied the geometry of NN loss landscapes by computing the distribution of Hessian eigenvalues at critical points.
More recently,~\cite{yao2018hessian} used a deflated power-iteration method to compute the top eigenvalues for deep NNs during training.
Moreover, the work of~\cite{ghorbani2019investigation} measured the Hessian ESD, based on the Stochastic Lanczos algorithm of~\cite{lin2016approximating,ubaru2017fast}.
Here, we extend the analysis of~\cite{ghorbani2019investigation,yao2018hessian} by studying how the depth of the NN model as well as its architecture affect the Hessian spectrum (in terms of top eigenvalue, trace, and full ESD).
Furthermore, we also perform block diagonal Hessian spectrum analysis, and we observe a fine-scale relationship between the Hessian spectrum and the impact of adding/removing residual connections and BN. 

Hessian-based analysis has also been used in
the context of NN training and inference.
For example, \cite{lecun1991second} analytically computes Hessian information for a single linear layer and uses the Hessian spectrum to determine the optimal learning rate to accelerate training. 
In~\cite{lecun1990optimal}, the authors approximated the Hessian as a diagonal operator and used the inverse of this diagonal matrix to prune NN parameters. 
Subsequently,
\cite{hassibi1993second} used the inverse of the full Hessian matrix to develop an ``Optimal Brain Surgeon'' method for pruning NN parameters. 
The authors argued that a diagonal approximation may not be very accurate, as off-diagonal elements of the Hessian are important; and they showed that capturing these off-diagonal elements does indeed lead to better performance, as compared to~\cite{lecun1990optimal}.
In the recent work of~\cite{dong2017learning}, a layer-wise pruning method was proposed.
This restricts the Hessian computations to each layer, and it provides bounds on the performance drop after pruning.
More recently,~\cite{dong2019hawq,shen2019q,dong2019hawqv2} proposed a Hessian-based method for quantizing%
\footnote{Quantization is a process in which the precision of the parameters is reduced from single precision (32-bits) to a lower precision (such as 8-bits).}
NN models, achieving significantly better performance, as compared to first-order based methods.

(Quasi-)Newton (second-order) methods~\cite{agarwal2016second,dembo1982inexact,pearlmutter1994fast,pilanci2017newton,pratt1998gauss,amari1998natural,bottou2018optimization} have been extensively explored for convex optimization problems~\cite{boyd2004convex}.
In particular, in the seminal work of~\cite{nocedal1980updating,liu1989limited}, a Quasi-Newton method was proposed to accelerate first-order based optimization methods.
The idea is to precondition the gradient vector with the inverse of the Hessian. 
However, instead of directly using the Hessian, a series of approximate rank-1 updates are used instead.
Follow up work of~\cite{schraudolph2007stochastic} extended this method and proposed a stochastic BFGS algorithm. 
More recently, the work of~\cite{bollapragada2018progressive} proposed an adaptive batch size Limited-memory BFGS method~\cite{liu1989limited} for large-scale machine learning problems; and an adaptive batch size method based on measuring directly the spectrum of the Hessian has been proposed~\cite{yao2018large} for large-scale NN training.

Hessian-based methods have also been explored for non-convex problems, including trust-region (TR)~\cite{conn2000trust}, cubic regularization (CR)~\cite{nesterov2006cubic}, and its adaptive variant (ARC) ~\cite{cartis2011adaptiveI,cartis2011adaptiveII}. 
For these problems, \cite{byrd2011use,erdogdu2015convergence,roosta2016sub,xu2016sub} provide sketching/sampling techniques for Newton methods, where guarantees are established for sampling size and convergence rates; and \cite{xu2017newton,xu2016sub,xu2017second,yao2018inexact} show that sketching/sampling methods can significantly reduce the need for data in approximate Hessian computation. 

One important concern for applying second-order methods to training is the cost of computing Hessian information at every iteration.
The work of \cite{martens2015optimizing} proposed the so-called Kronecker-Factored Approximations (K-FAC) method, which approximates the Fisher information matrix into a Kronecker product. 
However, the approach comes
 with several new hyperparameters, which can actually be more expensive to tune, compared to first-order methods~\cite{ma2019inefficiency}.

A major limitation in most of this prior work is that tests are typically restricted to small/simple NN models that may not be representative of NN workloads that are encountered in practice. 
This is in part due to the lack of a scalable and easily programmable framework that could be used to test second-order methods for a wide range of state-of-the-art models. 
Addressing this is the main motivation behind our development of \OURS, which is released as open-source software and is available to researchers~\cite{pyhessian}. 
In this paper, we illustrate how \OURS can be used for analyzing the NN behaviour during training, even for very deep state-of-the-art models. 
Future work includes using this framework for second-order based optimization, by testing it on modern NN models, as well as fairly gauging the benefit that may arise from such methods, in light of the cost for any extra hyperparameter tuning that may be needed~\cite{ma2019inefficiency}.

\textbf{Residual Connections and Batch Normalization:} 
Residual connections~\cite{he2016deep} and BN~\cite{ioffe2015batch} are two of the most important ingredients in modern convolutional NNs. 
There have been different hypothesis offered for why these two components help training/generalization.
First, the original motivation for residual connections was that
they allow gradient information to flow to earlier layers of the NN,
thereby reducing the vanishing gradient problem during training. 
The empirical study of~\cite{li2018visualizing} found that deep NNs with residual connections exhibit a 
significantly smoother loss landscape, as compared to models without residual connections. 
This was achieved by the so-called filter-normalized random direction method to plot 3D loss landscapes, i.e., not through direct analysis of the Hessian spectrum.
This result is interesting, but it is hard to draw conclusions with perturbations in two directions, for a model that has millions of parameters (and thus millions of possible perturbation directions).

Second, the original motivation for why BN helps training/generalization was originally attributed to reducing the so-called Internal Covariance Shift (ICS)~\cite{ioffe2015batch}.
However, this was disputed in the recent study of~\cite{santurkar2018does}.
In particular, the work of~\cite{santurkar2018does} used first-order analysis to analyze the loss landscape, and found that adding a BN layer results in a smoother loss landscape. 
Importantly, they found that adding BN does not reduce the so called ICS.
Again, while interesting, such first-order analysis may not fully capture the topology of the landscape; and, as we will show with our second-order analysis, \textbf{this smoothness claim is not correct in general}.

The work of~\cite{santurkar2018does} also performed an interesting theoretical analysis, showing a connection between adding the BN layer and the Lipschitz constant of the gradient (i.e., the top Hessian eigenvalue). 
It was argued that adding the BN layer leads to a smaller Lipschitz constant.
However, the theoretical analysis is only valid for per-layer Lipschitz constant, as it ignores the complex interaction between different layers.
It cannot be extended to the Lipschitz constant of the entire model (and, as we will show, this result does not hold for shallow networks).

\section{Methodology}
\label{sec:methodology}

For a supervised learning problem, we seek to minimize:
\begin{equation}
    \min_\theta L(\theta) = \frac1N \sum_{i=1}^N l(M(x_i), y_i, \theta),
    \label{e:loss}
\end{equation}
where $\theta\in \R^m$ is the learnable weight parameter, $l(M(x), y, \theta)$ is the loss function, $(x, y)$ is the input pair, $M$ is the NN architecture, and $N$ is the size of training data. 
Below we first discuss how \OURS computes the second-order statistics, and we then discuss
the impact of architectural components on the trainability of the model.

\subsection{Neural Network Hessian Matvec}

For a NN with $m$ parameters, the gradient of the loss w.r.t. model parameters is a vector 
$$
\frac{\partial L}{\partial\theta}=g_\theta\in\R^m,
$$ 
and the second derivative of the loss is a matrix, 
$$
H=\frac{\partial^2 L}{\partial\theta^2}=\frac{\partial g_\theta}{\partial\theta}\in\R^{m\times m},
$$
commonly called the Hessian.
A typical NN model involves millions of parameters, and thus even forming the Hessian is computationally infeasible.
However, it is possible to compute properties of the Hessian spectrum \emph{without} explicitly forming the Hessian matrix. 
Instead, all we need is an oracle to compute the application of the Hessian to a random vector $v$.
This can be achieved by observing the following:

\begin{equation}
    \frac{\partial g_\theta^Tv}{\partial \theta} = \frac{\partial g_\theta^T}{\partial \theta}v + g_\theta^T\frac{\partial v}{\partial \theta} = \frac{\partial g_\theta^T}{\partial \theta}v = H v.
    \label{e:hessian_matvec}
\end{equation}
Here, the first equality is the chain rule, the second is due to the independence of $v$ to $\theta$,
and the third equality is the definition of the Hessian.
Importantly, note that the cost of this Hessian matrix-vector multiply  (hereafter referred to as Hessian matvec) is the same as one gradient backpropagation.
Having this oracle, we can easily compute the top $k$ Hessian eigenvalues using power iteration~\cite{yao2018hessian}; see~\aref{alg:power_iteration}.
However, for a typical NN with millions of parameters, the top eigenvalues may not be representative of how the loss landscape behaves. Therefore, we also compute the trace and ESD of the Hessian, as described~below.


\subsection{Hutchinson Method for Hessian Trace Computation}
\label{sec:trace}

The trace of the Hessian can be computed using RandNLA, and in particular with Hutchinson's method~\cite{bai1996some,avron2011randomized} for the fast computation of the trace, using only Hessian matvec computations (as given in~\eref{e:hessian_matvec}).
In particular, since we are interested in the Hessian, i.e., a symmetric matrix, suppose we have a random vector $v$, whose components are i.i.d. sampled from a Rademacher distribution (or Gaussian distribution with mean 0 and variance 1).
Then, we have the identity 
\begin{equation}
\begin{aligned}
    Tr(H) &= Tr(HI) = Tr(H\E[vv^T]) = \E[Tr(Hvv^T)]  \\
             &= \E[v^THv],
\end{aligned}
\end{equation}
where $I$ is the identity matrix of appropriate size. 
That is, the trace of $H$ can be estimated by computing $\E[v^THv]$, where we compute the expectation by drawing multiple random samples.
Note that $Hv$ can be efficiently computed from~\eref{e:hessian_matvec}, and then $v^THv$ is simply a dot product between
the Hessian matvec and the original vector $v$.
See~\aref{alg:hutchinson} for a description.


\begin{figure*}[!ht]
\centering
\includegraphics[width=0.45\textwidth]{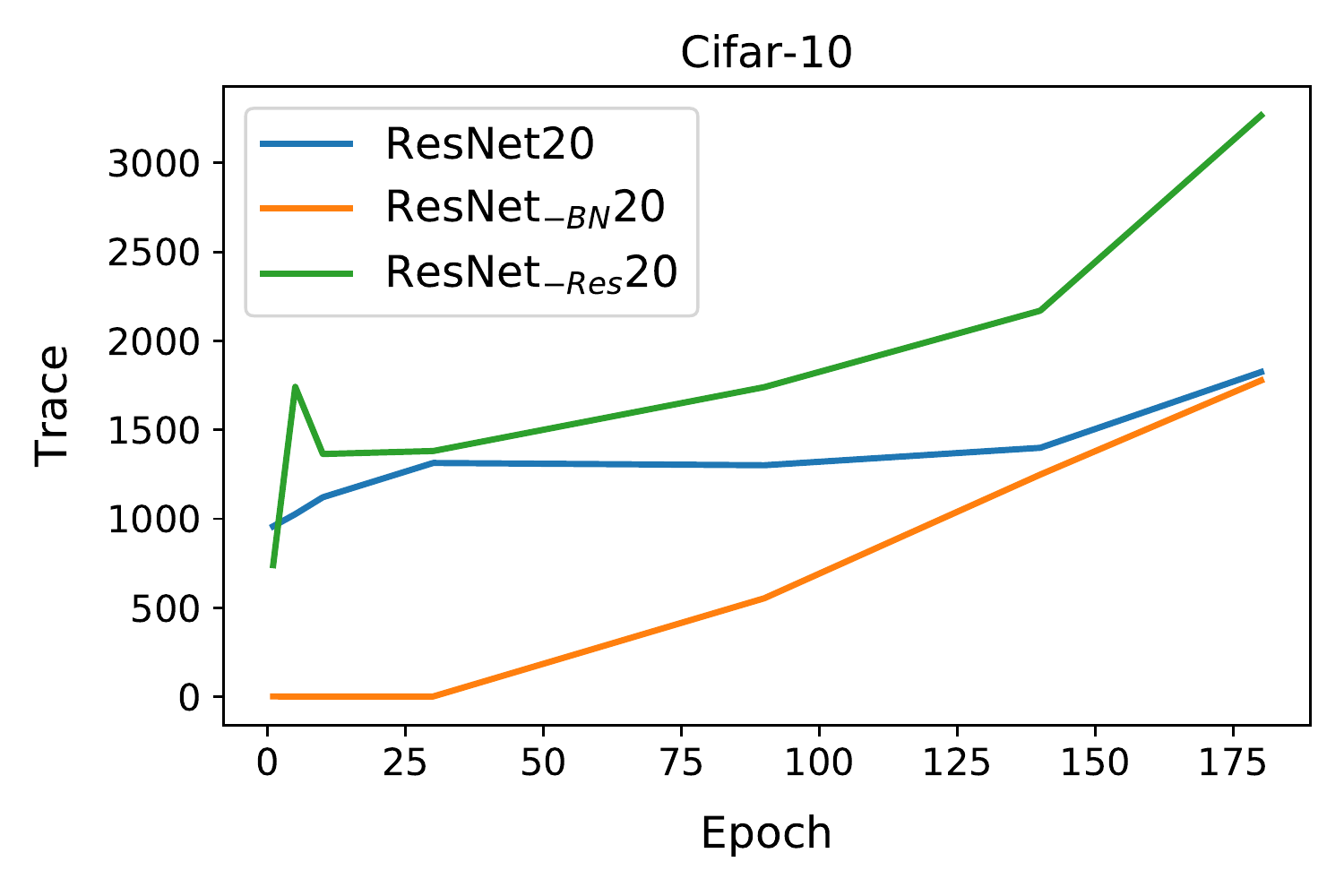}
\includegraphics[width=0.45\textwidth]{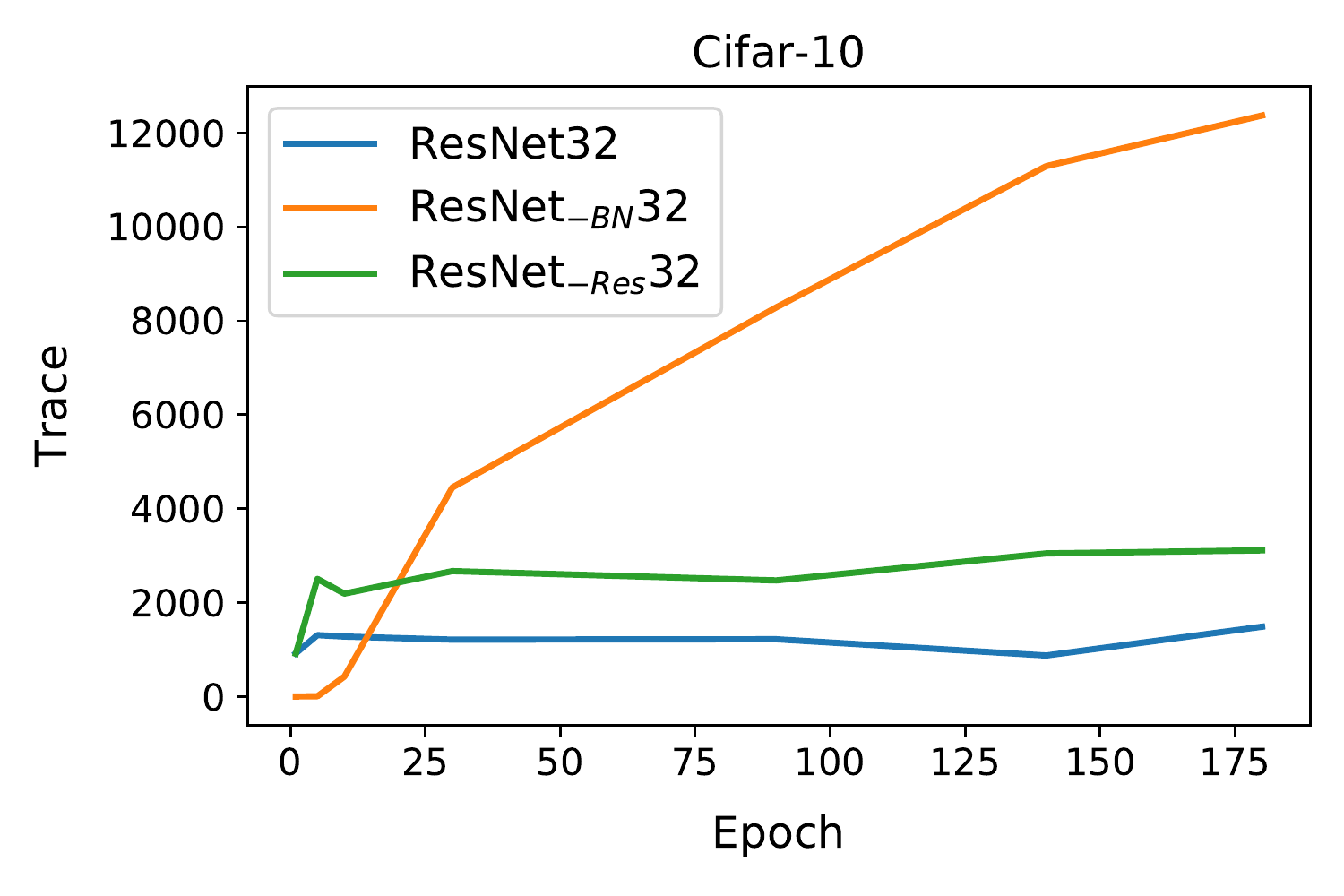}\\
\includegraphics[width=0.45\textwidth]{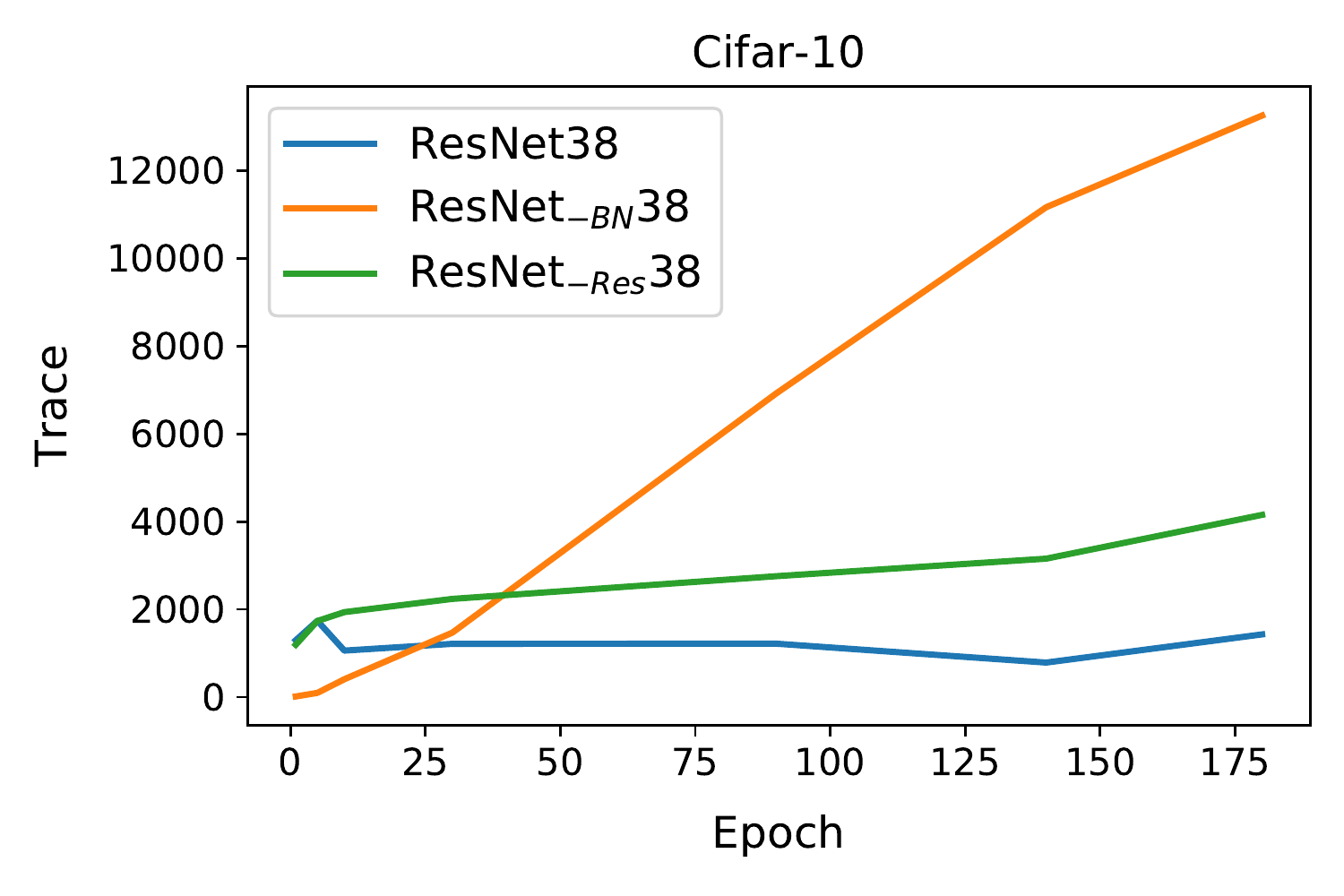}
\includegraphics[width=0.45\textwidth]{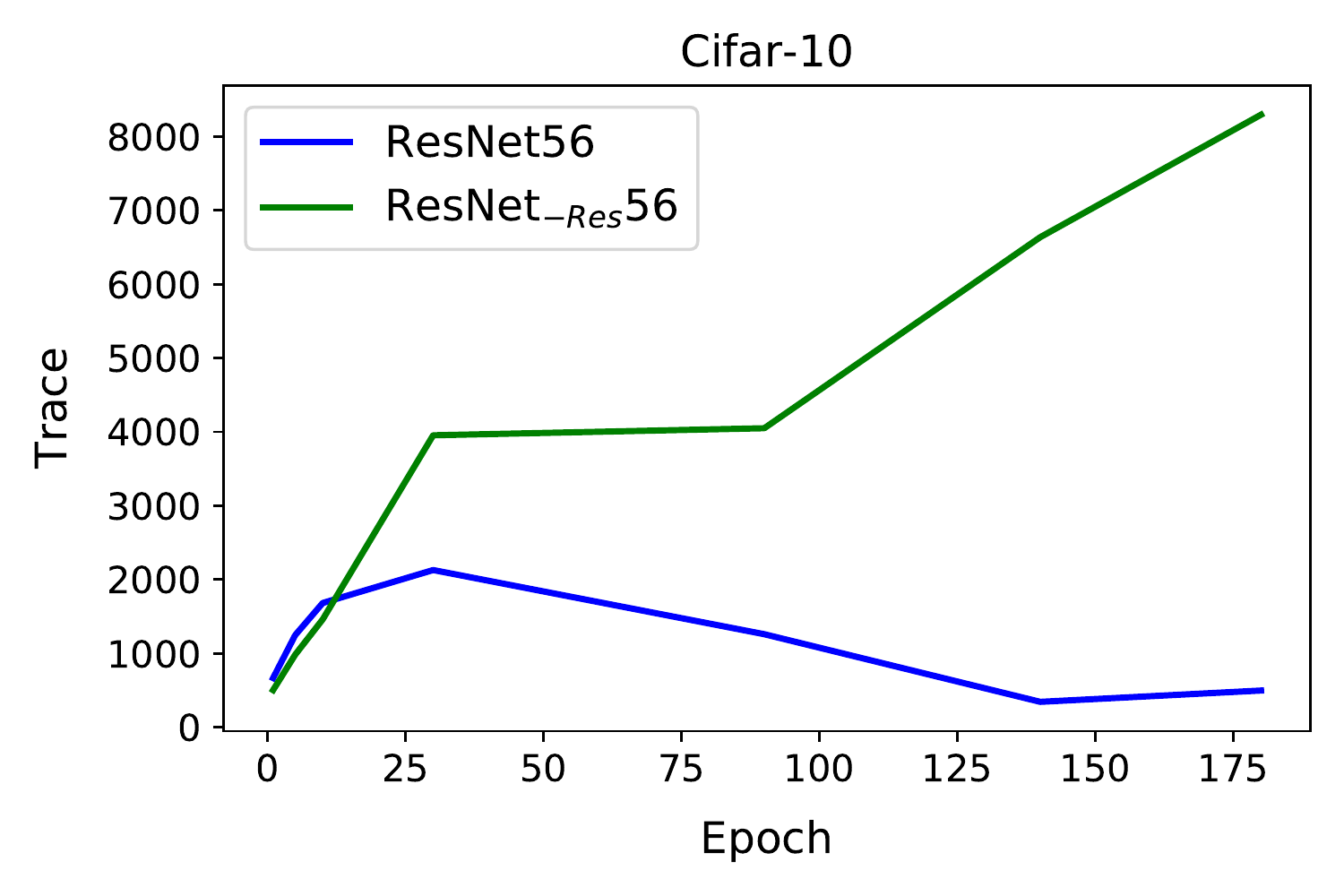}
\caption{
The Hessian trace of the entire network for ResNet/\ResNetBN/\ResNetRes with depth 20/32/38/56 on Cifar-10. For each depth, ResNet (blue) is the baseline.  It can be clearly seen that removing BN from the architecture (orange) generally results in a rapid increase of the Hessian trace. This increase is more pronounced for deeper networks such as ResNet32 and ResNet38. Importantly, the Hessian trace of ResNet20 without BN is lower than the original model (blue). This is in contrast to the claim of~\cite{santurkar2018does}. Also, we generally observe that residual connections smooth the Hessian trace for both shallow and deep networks (compare blue and green lines). 
Results on Cifar-100 also exhibit the same behaviour (as shown in~\fref{fig:resnet20/32/38-hut-full-net-cifar100}). 
}
  \label{fig:resnet20/32/38-hut-full-net}
\end{figure*}

\subsection{Full Eigenvalue Spectral Density}

To provide finer-grained information on
the Hessian spectrum than is provided by the top eigenvalues or the trace, we need
to compute the full empirical spectral density (ESD) of the Hessian eigenvalues, defined as
\small
\begin{equation}
    \phi(t) = \frac1m \sum_{i=1}^m \delta(t-\lambda_i),
    \label{e:phi}
\end{equation}
\normalsize
where $\delta(\cdot)$ is the Dirac distribution and $\lambda_i$ is the $i^{th}$ eigenvalue of $H$, in descending order. 
\begin{algorithm}[t]
\DontPrintSemicolon
\caption{Stochastic Lanczos Quadrature for ESD Computation}
\label{alg:slq}
    \SetAlgoLined
    \KwInput{Parameter: $\theta$, degree m and $n_v$.}
    Compute the gradient of $\theta$ by backpropagation, i.e., compute $g_\theta=\frac{d L}{d \theta}$.
    
    \For(\ \ \quad \quad\quad\quad\quad\tcp*[h]{Different Seeds}){i $=1,2,\ldots n_v$}{
        Draw a random vector $v$ from N(0,1) and normalize it (same dimension as $\theta$).
        
        Get the tridiagonal matrix $T$ through Lanczos algorithm.
        
        Compute $\tau_k^{(i)}$ and $\tilde \lambda_k^{(i)}$ from $T$
        $\phi_\sigma^{z_i} = \sum_{k=1}^q \tau_k f(\tilde \lambda_k; t, \sigma)$
    }
    Return $\phi(t) = \frac{1}{n_v} \sum_{l=1}^{n_v} \left(\sum_{i=1}^q \tau_i^{(l)} f(\tilde\lambda_i^{(l)}; t, \sigma)\right)$
\end{algorithm}

Recent work in NLA/RandNLA has provided efficient matrix-free algorithms to estimate this ESD~\cite{lin2016approximating,golub2009matrices,ubaru2017fast} through Stochastic Lanczos Quadrature (SLQ).
Here, we briefly describe SLQ in simple terms. 
This approach was also  used in~\cite{ghorbani2019investigation} to compute the Hessian ESD.
For more details, see~\cite{lin2016approximating,golub2009matrices,ubaru2017fast}.

Here is a summary of our approach to compute the ESD $\phi(t)$.
First, we approximate $\phi(t)$ (of \eref{e:phi}) by $\phi_\sigma(t)$ (\eref{e:phisigma} below) by applying a Gaussian kernel (\textit{first approximation}), and we express this in the same expectation form as in the Hutchinson algorithm (\eref{e:phisigma_to_expection} below).
Next, since the computation inside the expectation depends directly on $t$ and
the unknown eigenvalues (denoted by $\lambda_i$s), we further simplify the problem by using Gaussian quadrature (\eref{e:riemann_sum_relas} below) (\textit{second approximation}). 
Then, since the weights and $\lambda_i$s in the Gaussian quadrature are still unknown, we use the stochastic Lanczos algorithm to approximate the weights and $\lambda_i$s (\eref{e:slq_relax} below) (\textit{third approximation}). 
Finally, we approximate the expectation of the eigenvalue distribution as a sum (\eref{e:final_slq} below) (\textit{forth~approximation}).

In more detail, for the first approximation, we apply a Gaussian kernel, $f$, with variance $\sigma^2$ to~\eref{e:phi} to obtain
\begin{equation}
    \phi_\sigma(t) = \frac1m \sum_{i=1}^m f(\lambda; t, \sigma),
    \label{e:phisigma}
\end{equation}
where $f(\lambda;t,\sigma) = \frac{1}{\sigma \sqrt{2\pi}}exp(-(t-\lambda)^2/(2\sigma^2))$ is the Gaussian kernel. 
Clearly, $\phi_\sigma(t) \rightarrow \phi(t)$, as $\sigma \rightarrow 0$. 
Thus, if we had an algorithm to approximate~\eref{e:phisigma}, then we could take the limit and reduce the standard deviation of the Gaussian kernel to approximate~\eref{e:phi}.
In our context, the question of how to compute $\phi_\sigma(t)$ amounts to computing the density distribution of the Hessian convolved with a Gaussian kernel. 

To do this, observe that
\begin{equation}
    Tr(f(H)) = Tr(Qf(\Lambda)Q^T) = Tr(f(\Lambda)),
    \label{e:trace7}
\end{equation}
where $Q\Lambda Q^T$ is the eigendecomposition of $H$, and let $f(H)$ be the matrix function defined as
\begin{equation}
    f(H) \triangleq Qf(\Lambda) Q^T \triangleq Q\text{diag}(f(\lambda_1), ..., f(\lambda_m))Q^T.
\end{equation}
We can plug~\eref{e:trace7} into~\eref{e:phisigma} to get
\begin{equation}
    \phi_\sigma(t) = \frac{1}{m} Tr(f(H; t, \sigma)).
\end{equation}
For a given value of $t$, the trace $Tr(f(H; t, \sigma))$ can be efficiently computed using the Hutchinson algorithm (described in \S\ref{sec:trace}).
That is, we draw a random Rademacher vector $v$ and compute the expectation $\E[v^T f(H; t, \sigma) v]$ to get 
\begin{equation}\label{e:phisigma_to_expection}
    \phi_\sigma(t) = \frac{1}{m} \E[v^T f(H; t, \sigma) v]. 
\end{equation}
However, this is still intractable, as the trace computation needs to be repeated for every value of $t$ (which scales with the number of model parameters). 

To get around this, we relax this problem further~\cite{lin2016approximating,ubaru2017fast}.
Define $\phi_\sigma^v(t) = v^T f(H; t, \sigma) v$, in which case we have
\begin{equation}
\begin{aligned}
    \phi_\sigma^v(t) &= v^T f(H;t) v = v^T Q f(\Lambda;t) Q^Tv \\
                  &= \sum_{i=1}^m \mu_i^2 f(\lambda_i;t),
\end{aligned}
\end{equation}
where $\mu_i$ is the magnitude (or dot product) of $v$ along the $i^{th}$ eigenvector of $H$. 
Now let us define a probability distribution w.r.t. $\alpha$ with the cumulative distribution function,  $\pi(\alpha)$, as the following piece-wise function:
\begin{equation}
\pi(\alpha) =
    \begin{cases} 
      0 & \alpha \leq \lambda_m, \\
      \sum_{i=1}^j \mu_i^2 & \lambda_{j}\leq  \alpha \leq \lambda_{j-1}, \\
      \sum_{i=1}^m \mu_i^2 & \lambda_1 \leq \alpha.
   \end{cases}
\end{equation}
Then, by the Riemann-Stieltjes integral, it follows that
\begin{equation}
    \phi_\sigma^v(t) = \int_{\lambda_m}^{\lambda_1} f(\alpha; t) d\pi(\alpha).
\end{equation}
This integral can be estimated by the Gauss quadrature rule~\cite{golub1969calculation}, 
\begin{equation}\label{e:riemann_sum_relas}
    \phi_\sigma^v(t) \approx \sum_{i=1}^q\omega_i f(t_i;t, \sigma),
\end{equation}
where $(\omega_i, t_i)$ is the weight-node pair to estimate the integral.
The stochastic Lanczos algorithm can then be used to estimate accurately this quantity~\cite{ubaru2017fast,golub2009matrices,lin2016approximating}. 
Specifically, for the $q$-step Lanczos algorithm, we have $q$ eigenpairs $(\tilde \lambda_i, \tilde v_i)$. 
Let $\tau_i = (\tilde v_i[1])^2$, where $\tilde v_i[1]$ is the first component of $\tilde v_i$, in which case it follows that
\begin{equation}\label{e:slq_relax}
    \phi_\sigma^v(t) \approx \sum_{i=1}^q\omega_i f(t_i; t, \sigma) \approx \sum_{i=1}^q \tau_i f(\tilde\lambda_i; t, \sigma).
\end{equation}
Therefore, as in the Hutchinson algorithm, with multiple different runs (e.g., $n_v$ times) of Lanczos algorithm, $\phi_\sigma$ can be approximated by
\small 
\begin{equation}\label{e:final_slq}
    \phi_\sigma(t) = Tr(f(H)) \approx\frac{1}{n_v} \sum_{l=1}^{n_v} \left(\sum_{i=1}^q \tau_i^{(l)} f(\tilde\lambda_i^{(l)}; t, \sigma)\right).
\end{equation}
\normalsize

\noindent
See~\aref{alg:slq} for a description of the SLQ algorithm.

\section{Results}
\label{sec:results}

Here, we provide extensive experiments to study the impact of BN and residual connection on
the Hessian spectrum.
We first start by discussing the experimental settings in~\S\ref{sec:experiment_setting_and_convergence}, followed by presenting the Hessian spectrum results for the entire model in~\S\ref{sec:full_net_analysis} as
well different ResNet stages in~\S\ref{sec:stagewise_analysis}.

\subsection{Experimental Setting}
\label{sec:experiment_setting_and_convergence}

Using \OURS, we measure all three Hessian spectrum metrics (i.e., top eigenvalues, trace, and full ESD)
throughout the training process of SGD with momentum.
We consider various ResNet~\cite{he2016deep} architectures, and in particular ResNet20/32/38/56 on the Cifar-10,
and we analyze these models and variants with/without BN and with/without residual connections.
We also experimented with same networks tested on Cifar-100 dataset, and all of the observations were consistent. 
These results are presented in Appendix. 

For clarity, we refer to the networks without BN as \ResNetBN, and we refer to the networks without residual connections as \ResNetRes.
We train each model with various initial learning rates, and we pick the best performing result for analysis.
See Appendix~\ref{sec:training_details} for more details on training settings. 
We analyze the spectrum throughout training at all checkpoints. 
The accuracy of each model is reported in~\tref{tab:model_acc}, and the testing curve is 
shown in~\fref{fig:model_acc}.


\begin{table}[!htbp]
\caption{
Accuracy of ResNet, \ResNetBN, and \ResNetRes, with different depths, on Cifar-10. The accuracy drops if the BN layer is removed (denoted by \ResNetBN), and this degradation is more pronounced for deeper models. In fact, \ResNetBN56 cannot be trained at all. Removing
the residual connections (denoted as \ResNetRes) also results in slight performance degradation.
Accuracy of models on Cifar-100 is reported in~\tref{tab:model_acc_cifar100}. 
}
\small
\setlength\tabcolsep{2.35pt}
\label{tab:model_acc}
\centering
\begin{tabular}{lcccccccccccccc} \toprule
 Model\textbackslash Depth  & 20       & 32       & 38       & 56 \\ 
\midrule 
\hc ResNet       & 92.01\%  & 92.05\%  & 92.37\%  & 93.59\%\\
\ha \ResNetBN    & 87.27\%  & 66.57\%  & 53.65\%  & N/A\\ 
\hc \ResNetRes   & 90.66\%  & 89.8\%   & 88.92\%  & 87.38\%\\
     \bottomrule 
\end{tabular}
\end{table}

\subsection{Full Network Hessian Analysis}\label{sec:full_net_analysis}

\begin{figure*}[t]
\centering
\includegraphics[width=0.32\textwidth]{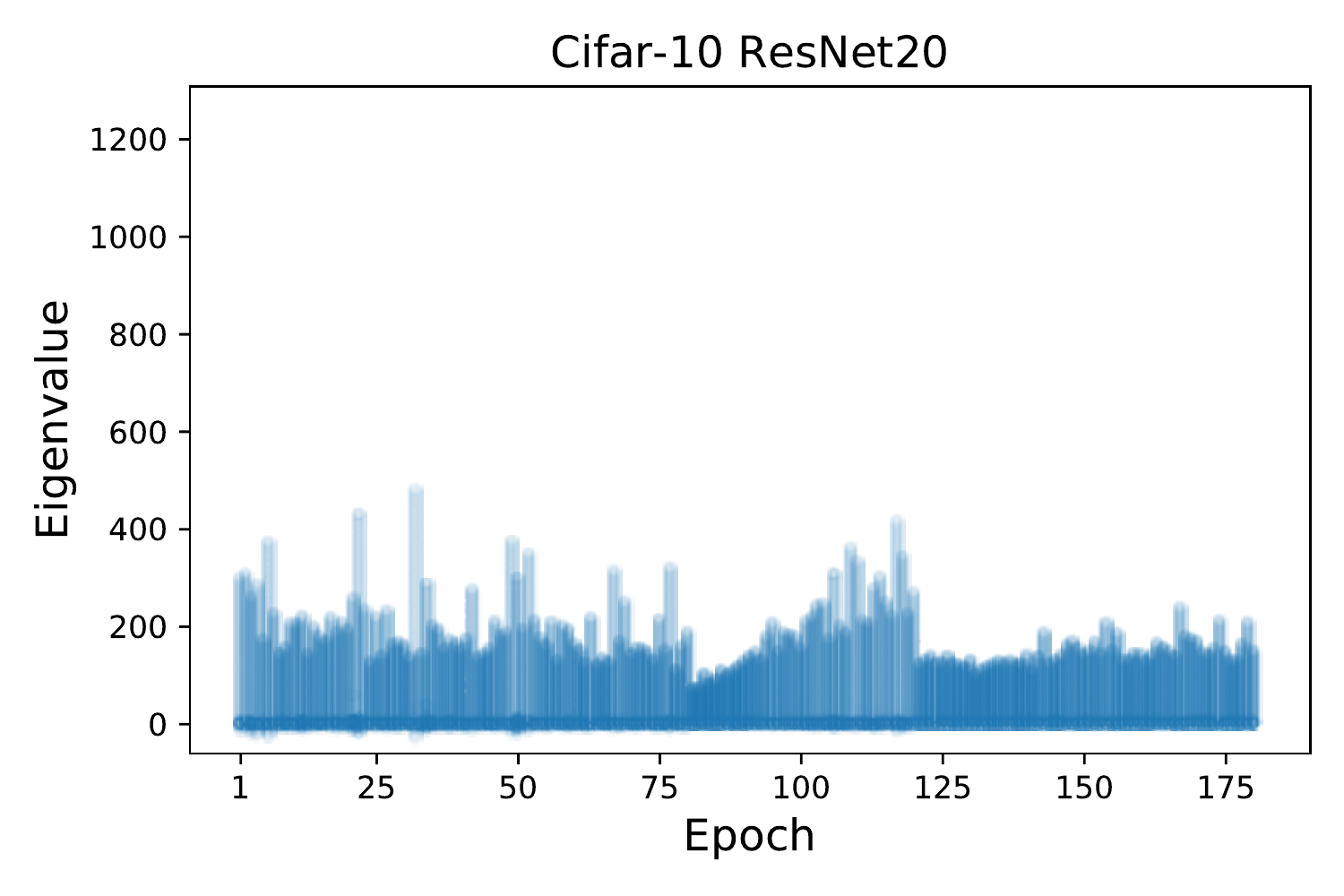}
\includegraphics[width=0.32\textwidth]{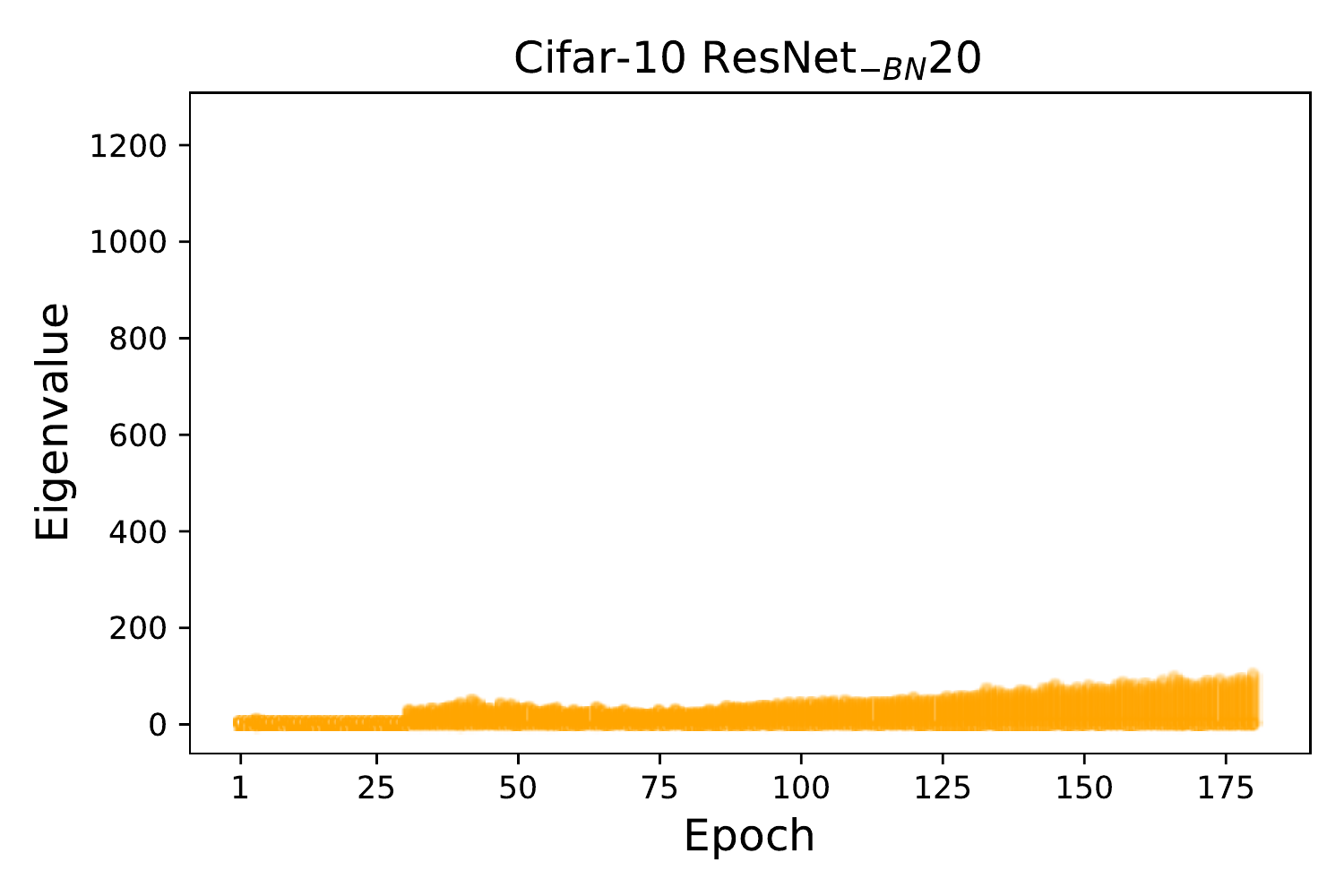}
\includegraphics[width=0.32\textwidth]{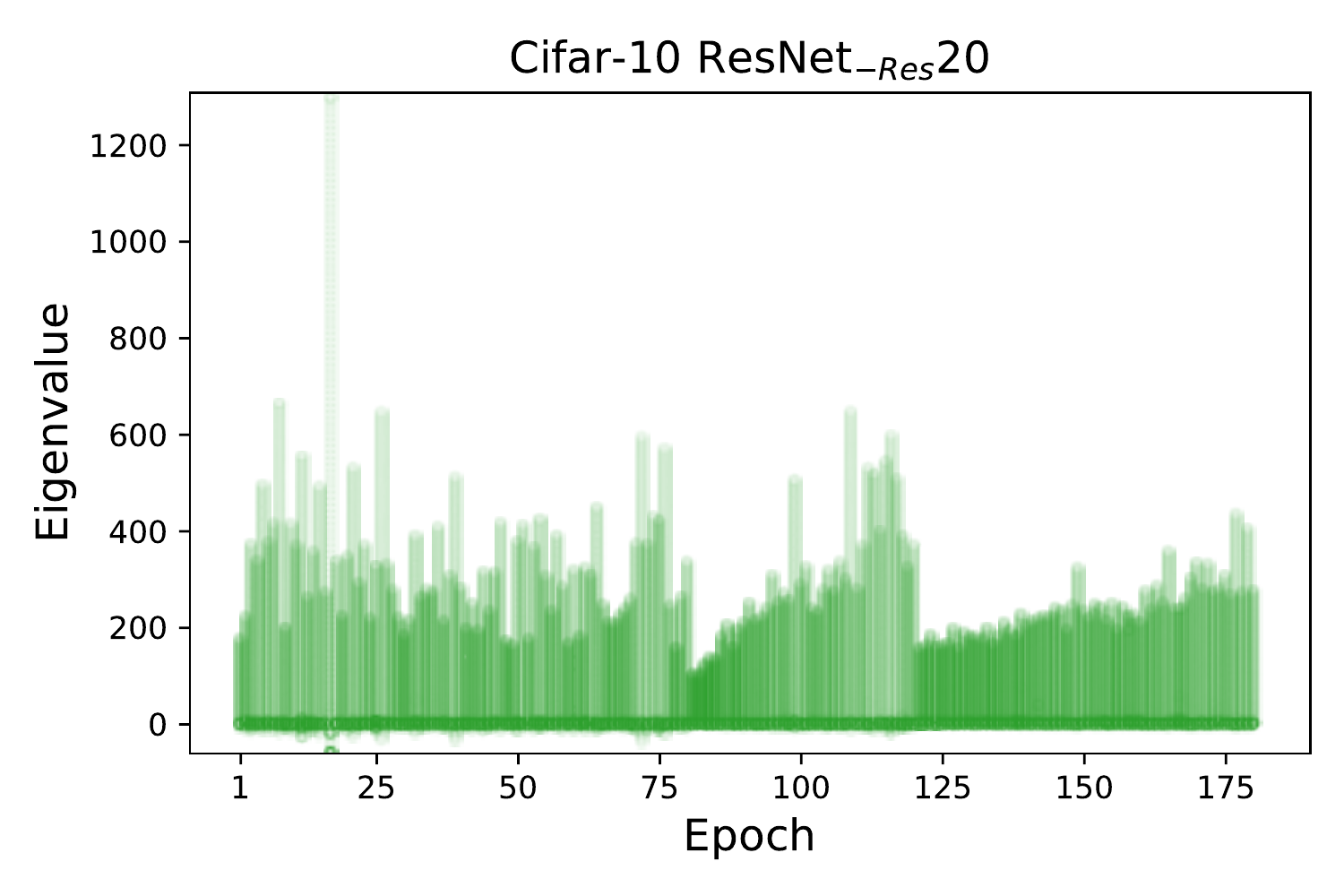}\\
\includegraphics[width=0.32\textwidth]{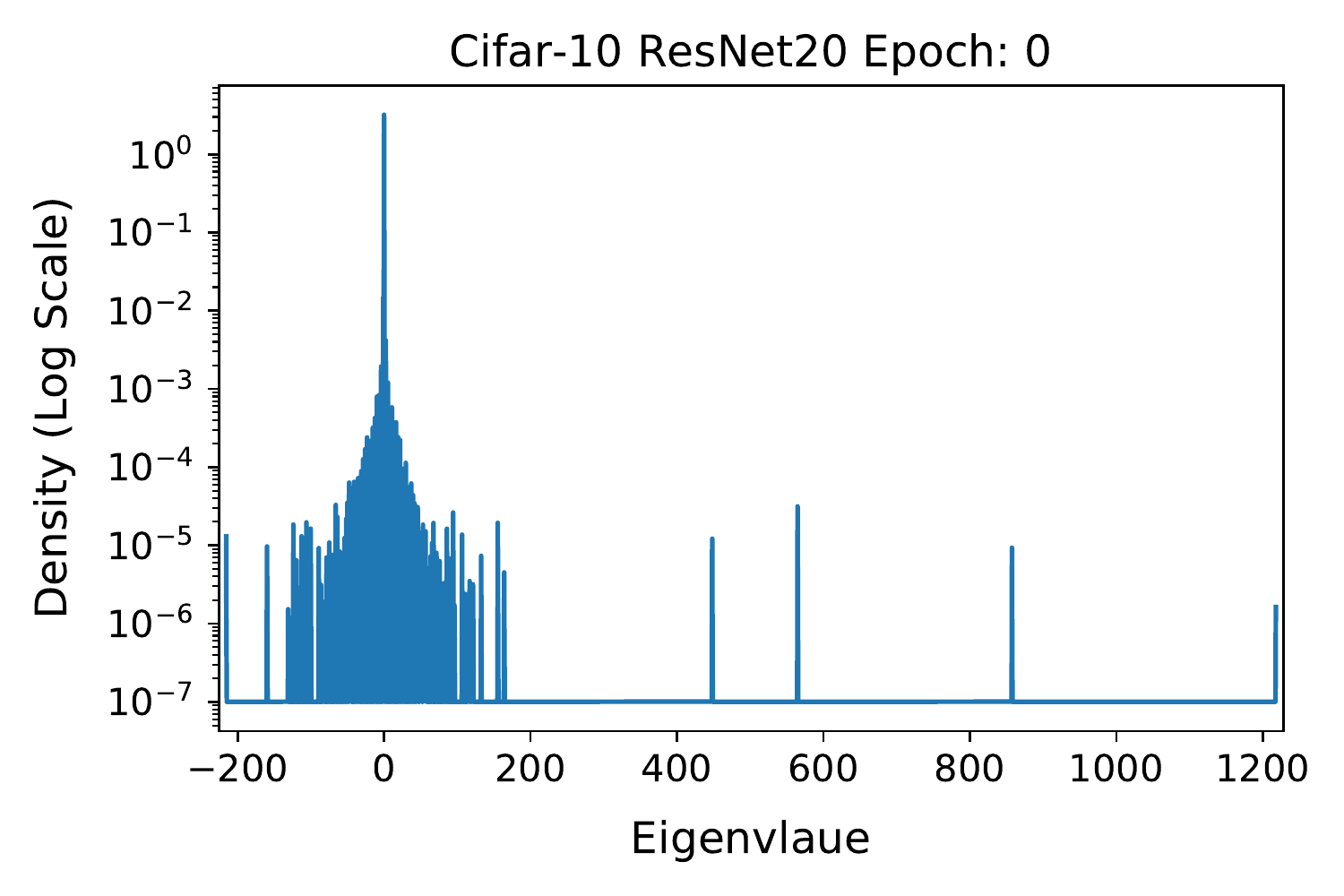}
\includegraphics[width=0.32\textwidth]{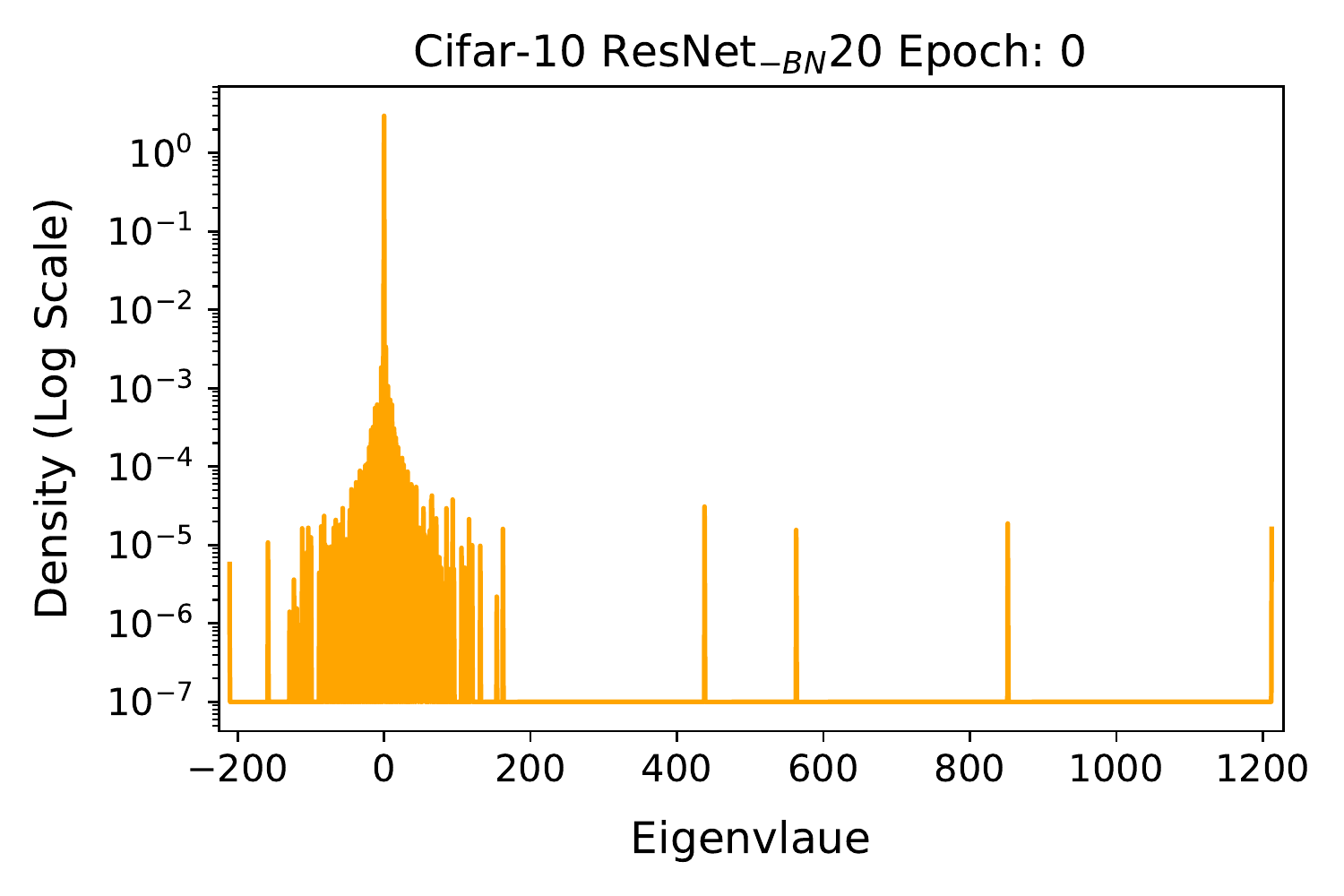}
\includegraphics[width=0.32\textwidth]{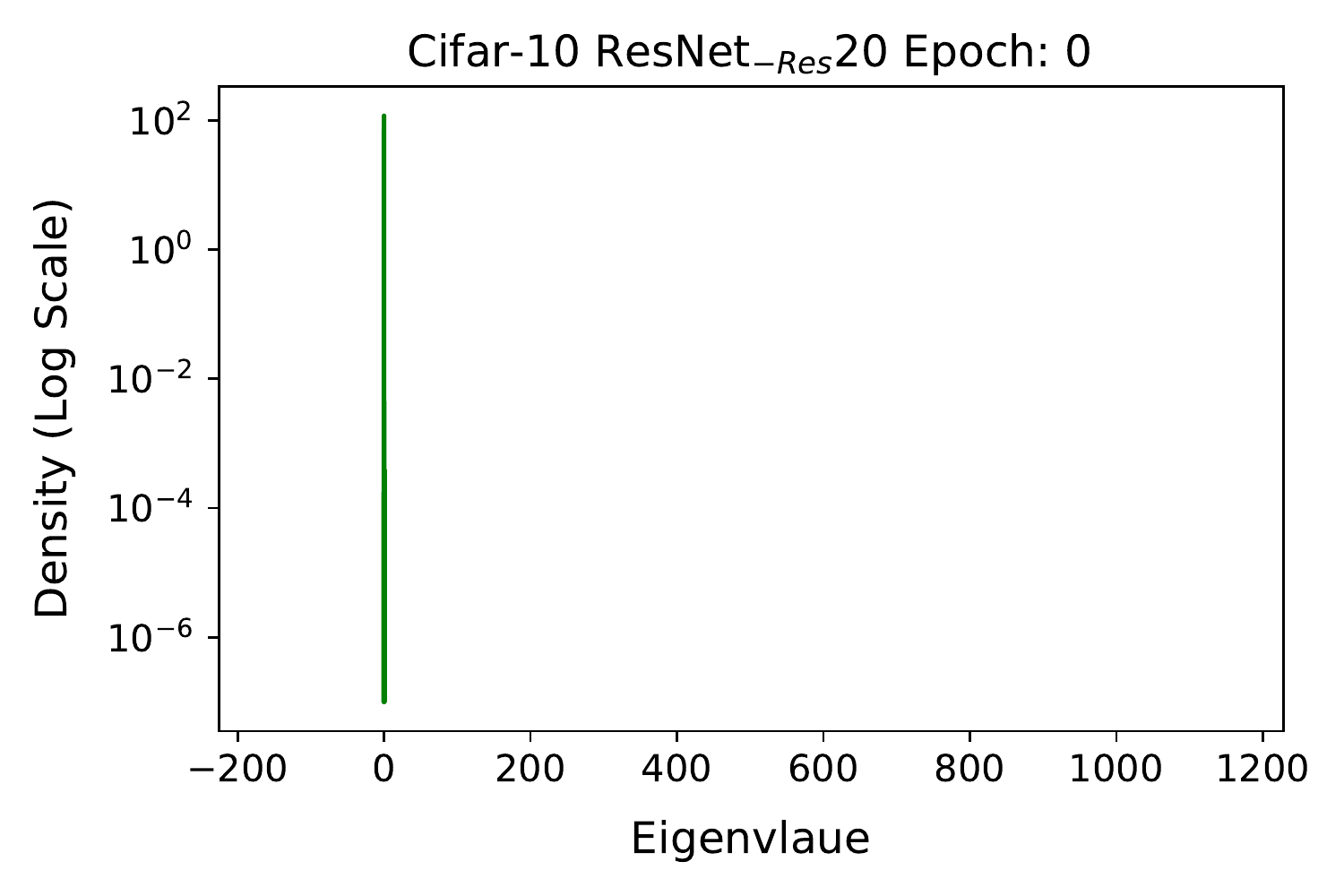}\\
\includegraphics[width=0.32\textwidth]{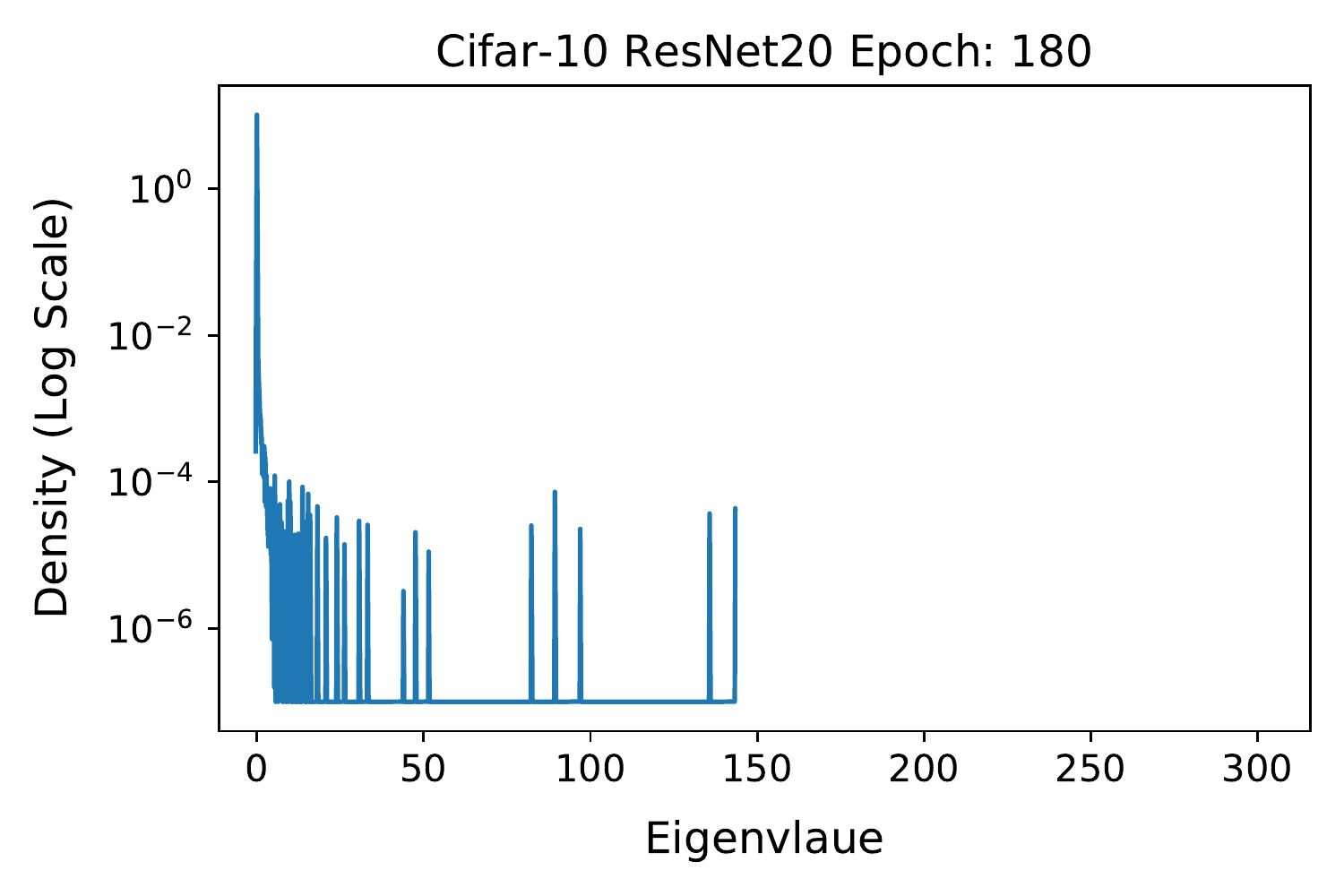}
\includegraphics[width=0.32\textwidth]{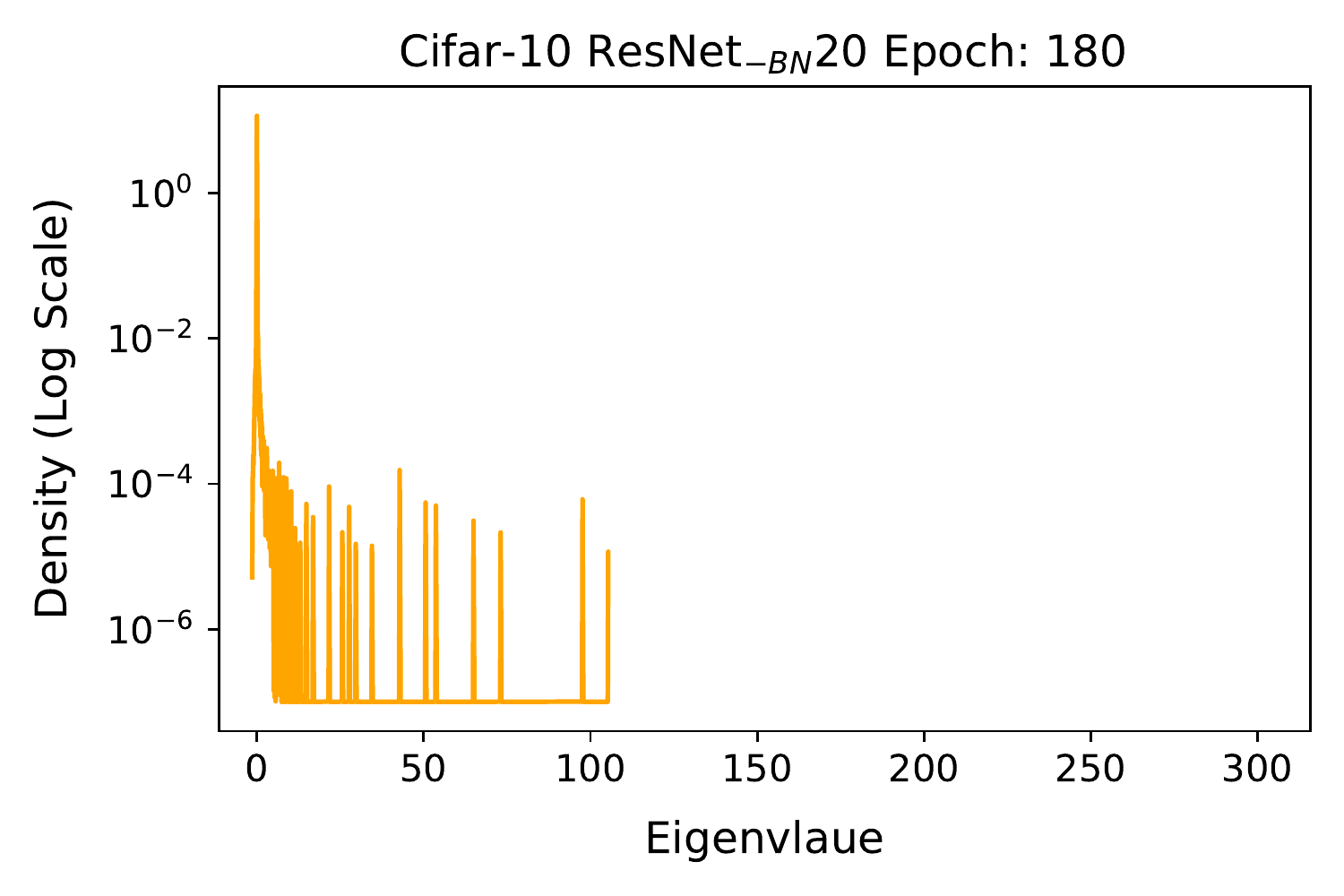}
\includegraphics[width=0.32\textwidth]{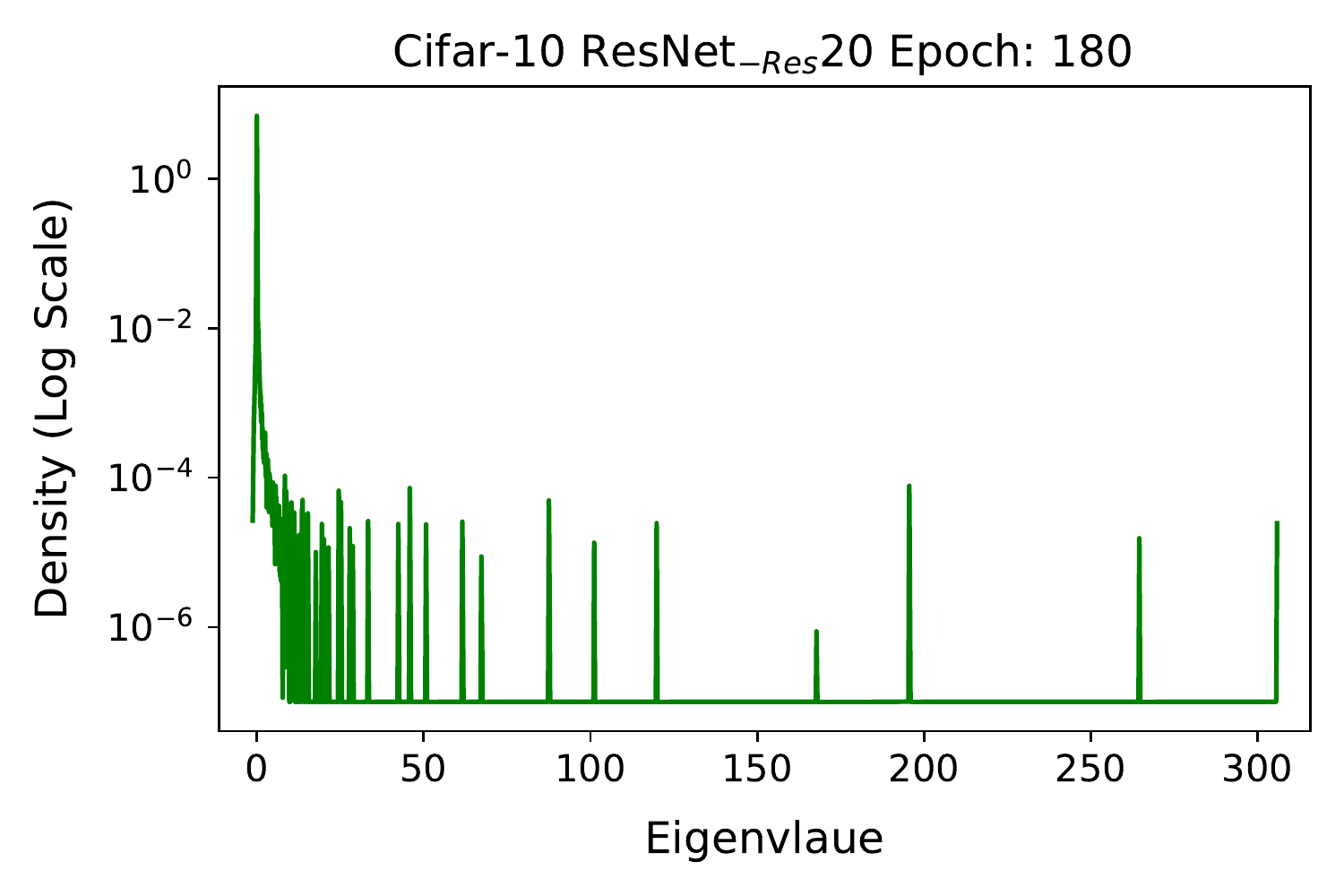}\\
\caption{
(first row) We show the Hessian ESD throughout training for ResNet/\ResNetBN/\ResNetRes 
(each shown in a different column) with depth 20 on Cifar-10.
For a fixed epoch, every point corresponds to a Hessian eigenvalue.
These plots show several important phenomena. First note that removing BN (middle column) does not lead to a non-smooth loss landscape as was claimed
by~\cite{santurkar2018does}. We can clearly see that this is true throughout training. However, removing the residual connection leads
makes the loss landscape non-smooth throughout training 
(middle/last row). 
We show the Hessian ESD at epoch 0 and epoch 180. This clearly shows that removing BN leads to a maximum eigenvalue of ~100, whereas the baseline
has a maximum Hessian eigenvalue of ~150.
See~\fref{fig:resnet20-slq-full-net-all}, where we plot the Hessian ESD for several other epochs throughout training. 
We observed the same behaviour on Cifar-100 dataset (as shown in \fref{fig:resnet20-slq-full-net-all-cifar100}). 
}
  \label{fig:resnet20-slq-full-net-part}
\end{figure*}

We start with the original ResNet model with BN and residual connections. 
Hereafter we refer to this as \ResNet. 
The behaviour of the Hessian trace throughout training is shown in~\fref{fig:resnet20/32/38-hut-full-net}. 
Furthermore, we show the evolution of the Hessian ESD throughout training
in~\fref{fig:resnet20-slq-full-net-part} for Cifar-10. 


\paragraph{\textbf{Batch Normalization}}
As discussed before, a BN layer is crucial for training NN models,
and removing this component can adversely affect the generalization performance, as is shown in~\tref{tab:model_acc}. 
The drop in performance is very significant for deeper models.
For example, we could not even train ResNet56 on Cifar-10 without a BN layer, even with hyperparameter~tuning.

The first interesting observation is that removing BN layer (denoted by \ResNetBN)
exhibits different behaviour for shallower versus deeper models. 
For example, for ResNet20 we see that removing BN results in smaller trace and Hessian ESD values, as compared to baseline, as shown in~\fref{fig:resnet20/32/38-hut-full-net}
 (orange curve versus blue curve),  and \ref{fig:resnet20-slq-full-net-part} (second versus first column).
In more detail, from the evolution plot of~\fref{fig:resnet20-slq-full-net-part} throughout training, it can be seen that the ESD of \ResNetBN20
initially reduces significantly and centers around zero. 
That is, the model gets attracted to areas with a significantly large number of small/degenerate Hessian directions.
This continues until epoch 30, at which point the training gets attracted to regions of the loss landscape with several non-degenerate Hessian~directions.

This clearly shows that training without BN makes training harder, but it does not necessarily mean that the Hessian spectrum is going to be larger than the baseline model,
despite the claim made by~\cite{santurkar2018does}. 
In fact, we only observe the smoothing behaviour proposed by~\cite{santurkar2018does} for deeper NN models.
For example, observe the Hessian trace plot of ResNet32/38, shown in~\fref{fig:resnet20/32/38-hut-full-net}.  
Here, the Hessian trace of \ResNetBN32 increases to 10000 from zero, as compared to 2000 for \ResNet.
The Hessian ESD also exhibits the same behaviour, as shown in~\fref{fig:resnet32-slq-full-net-all} and \ref{fig:resnet38-slq-full-net-all}. 
We can clearly see that the range of eigenvalues of \ResNetBN is significantly larger, as compared to \ResNet.

The Hessian ESD of ResNet32 and ResNet38 \emph{throughout the training process} is shown in~\fref{fig:resnet32-slq-full-net-all},~\ref{fig:resnet38-slq-full-net-all}. 
Again, we see the interesting behaviour that
without the BN layer, the spectrum initially converges to degenerate Hessian directions, before finding non-degenerate directions in later epochs of training. 
The Hessian trace and the range of the Hessian ESD significantly increases as the model gets deeper.

These plots show the numerical values of the Hessian spectrum.
However, the results could be more intuitively presented via parametric plots of the loss landscape.
We plot the parametric 3D loss landscapes of ResNet20/38 on Cifar-10 with/without BN in~\fref{fig:resnet38-loss-landscape-part} (compare left and middle columns).
These plots are computed by perturbing the model parameters across the first and second eigenvectors of the Hessian. 
For ResNet20, it can be clearly seen that removing the BN layer (middle plot) results in convergence to a flatter local minimum, as compared to ResNet20 with BN. 
This observation is the opposite of the common belief that adding BN layer makes the loss landscape smoother~\cite{santurkar2018does}.
However, for ResNet38, we can also see that removing the BN layer results in convergence to a point with a higher value of loss. 
The visualizations corroborate our finding that initially \ResNetBN finds points with degenerate Hessian directions, before converging to a point with non-degenerate directions. 
We provide more visualizations for ResNet20 (\fref{fig:resnet20-loss-landscape-all}), ResNet32 (\fref{fig:resnet32-loss-landscape-all}), and ResNet38 (\fref{fig:resnet38-loss-landscape-all}), which show the same behaviour.

In summary, our empirical results highlight two points. 
First, our findings show several fine-scale behaviours when the BN layer is removed. 
Importantly, we find that the observation made in~\cite{santurkar2018does} only holds for deeper models, and are not necessarily true for shallow networks.
Second, using the scalable Hessian-based techniques implemented in \OURS, one can test the hypotheses that these or other claims hold more~generally.
For example, we observed exactly similar behaviour for Cifar-100 dataset, as shown in Appendix~\ref{sec:extra_results}.

\begin{figure*}[!ht]
\centering
\includegraphics[width=0.995\textwidth]{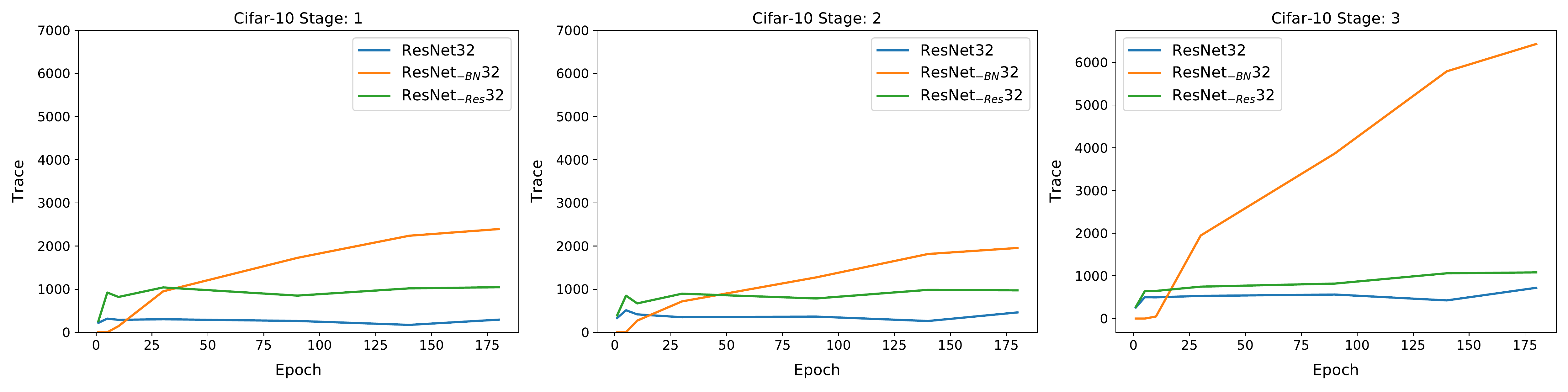}
\caption{Stage-wise Hessian trace of ResNet/\ResNetBN/\ResNetRes32 on Cifar-10. (See~\fref{fig:resnet20/38_stagewise_hut} for depth 20/32; and see~\fref{fig:resnet20_illustration} for stage illustration.) 
Removing BN layer from the third stage significantly increases the trace, compared to removing BN layer from the first/second stage. 
This has a direct correlation with the final generalization performance, as shown in~\tref{tab:model_acc_rm_bn}. 
ResNet/\ResNetBN/\ResNetRes on Cifar-100 has the similar trend as shown in~\fref{fig:resnet20/38_stagewise_hut-cifar100}. 
}
  \label{fig:resnet32_stagewise_hut}
\end{figure*}

\begin{table}[!ht]
\caption{
Accuracy of ResNet models on Cifar-10 with different depths is shown in the first row. 
Accuracy of the corresponding architectures, but with BN removed from one of the stages, is shown in the next three rows, respectively. 
(See~\fref{fig:resnet20_illustration} for stage definition.)
For instance, the last row is a ResNet model with no BN layer in the third stage.
We observe a general correlation between the accuracy drop and stage based Hessian analysis, shown in~\fref{fig:resnet32_stagewise_hut}. 
In particular, we see that stages which significantly affect accuracy also exhibit a significant increase
in the Hessian trace.
Models on Cifar-100 have the similar trend, as shown in~\tref{tab:model_acc_rm_bn-cifar100}. 
}
\small
\setlength\tabcolsep{2.35pt}
\label{tab:model_acc_rm_bn}
\centering
\begin{tabular}{lcccccccccccccc} \toprule
 Model\textbackslash Depth  & 20       & 32       & 38       & 56 \\ 
\midrule 
\hc ResNet                 & 92.01\%  & 92.05\%  & 92.37\%  & 93.59\%\\
\ha RM BN stage 1          & 91.28\%  & 91.98\%  & 92.20\%  & 92.19\%\\
\hc RM BN stage 2          & 91.49\%  & 91.94\%  & 91.70\%  & 92.20\%\\ 
\ha RM BN stage 3          & 90.59\%  & 88.57\%  & 86.96\%  & 73.77\%\\
     \bottomrule 
\end{tabular}
\end{table}

\paragraph{\textbf{Residual Connection}}
We next study the impact of residual connections on the smoothness of the loss landscape. 
Removing residual connections leads to slightly poorer generalization, as shown in~\tref{tab:model_acc},
although the degradation is much smaller than removing the BN layer. 

We report the behaviour of the Hessian trace for \ResNetRes in~\fref{fig:resnet20/32/38-hut-full-net} for ResNet20/32/38/56 on Cifar-10.
It can clearly be seen that the trace of \ResNetRes is consistently higher than that of ResNet, for both shallow and deep models on different datasets. 

In addition, from the Hessian ESD 
in~\fref{fig:resnet20-slq-full-net-part},~\ref{fig:resnet20-slq-full-net-all},~\ref{fig:resnet32-slq-full-net-all},~\ref{fig:resnet38-slq-full-net-all}, and~\ref{fig:resnet56-slq-full-net-all},
we can see that the top eigenvalues as well as the support range of ESD of \ResNetRes increases for deeper models.
These results are in line with the findings of~\cite{li2018visualizing}.

We also visualize the loss landscape of these models in~\fref{fig:resnet38-loss-landscape-part}, \ref{fig:resnet20-loss-landscape-all}, \ref{fig:resnet32-loss-landscape-all},~\ref{fig:resnet38-loss-landscape-all}, and~\ref{fig:resnet56-loss-landscape-all}. 
It can clearly be seen that the converging point for \ResNetRes becomes sharper, as compared with ResNet,
as the depth grows.

Again, our empirical results highlight two points.
First, we make observations that provide a finer-scale understanding of seemingly-contradictory claims in the previous literature.
Second,  using the scalable Hessian-based techniques that are implemented in \OURS, one can ask these questions and test the hypotheses that these or other claims hold more generally. 
Similar to the previous section, we saw exactly similar behaviour for Cifar-100, as reported in
Appendix~\ref{sec:extra_results}.

\subsection{Stage-wise Hessian Analysis}\label{sec:stagewise_analysis}

We also analyzed the impact of removing BN and residual connection from different
stages of the model.
We define each stage of ResNet as blocks with the same activation resolution, as
schematically shown in~\fref{fig:resnet20_illustration}. 

We plot the Hessian trace for the three stages of ResNet32 on Cifar-10 in~\fref{fig:resnet32_stagewise_hut} (similar plots for ResNet20/38/56 on Cifar-10 is shown in~\fref{fig:resnet20/38_stagewise_hut}). 
We can clearly see that removing the BN from the last stage of ResNet32 results in a more rapid increase in the Hessian trace, as compared to removing BN from the first or second stage. 
Interestingly, this has a direct correlation with the final generalization performance reported in~\tref{tab:model_acc_rm_bn}.
We can see that removing BN in the third stage results in higher accuracy drop, as compared to removing it from other stages. 
A similar trend exists for other models (ResNet20/38); and we generally observe the same behaviour on Cifar-100, as reported
in~\fref{fig:resnet20/38_stagewise_hut-cifar100} and~\tref{tab:model_acc_rm_bn-cifar100}.

As for the residual connections, we can see that removing them results in a relatively smaller
increase in the Hessian trace, and correspondingly the impact of removing the residual connections on accuracy is 
smaller, as compared to removing BN.
See~\tref{tab:model_acc_rm_res} for Cifar-10. 


\begin{table}[!htbp]
\caption{
Accuracy of ResNet on Cifar-10 is reported for baseline (first row), along with architectures where the residual connection is removed at different stages.
}
\small
\setlength\tabcolsep{2.35pt}
\label{tab:model_acc_rm_res}
\centering
\begin{tabular}{lcccccccccccccc} \toprule
 Model\textbackslash Depth  & 20       & 32       & 38       & 56 \\ 
\midrule 
\hc ResNet                  & 92.01\%  & 92.05\%  & 92.37\%  & 93.59\%\\
\ha RM Res stage 1          & 91.52\%  & 92.27\%  & 91.74\%  & 91.79\%\\
\hc RM Res stage 2          & 91.06\%  & 91.07\%  & 91.08\%  & 91.28\%\\ 
\ha RM Res stage 3          & 91.54\%  & 92.09\%  & 92.14\%  & 92.34\%\\
     \bottomrule 
\end{tabular}
\end{table}

\subsection{Summary of Results}
\tref{tab:navigation} presents a summary of the tables and figures used in this work and their corresponding properties, i.e., Accuracy, Trace, ESD, and Loss Landscape. 

\begin{table}[!ht]
\caption{
Navigation summary for all figures and tables used throughout this paper.
}
\small
\setlength\tabcolsep{2.35pt}
\label{tab:navigation}
\centering
\begin{tabular}{l|c|ccccccccccccc} \toprule
                      & Cifar-10                           & Cifar-100 \\
\midrule
\hc Accuracy              & \tref{tab:model_acc}, \fref{fig:model_acc}  & \tref{tab:model_acc_cifar100}, \fref{fig:model_acc_cifar100}           \\
\ha RM BN Acc.  & \tref{tab:model_acc_rm_bn}                                   & \tref{tab:model_acc_rm_bn-cifar100}          \\
\hc RM Res Acc & \tref{tab:model_acc_rm_res}                                   & \tref{tab:model_acc_rm_res_cifar100}          \\
\ha Trace                 & \fref{fig:resnet20/32/38-hut-full-net} &  \fref{fig:resnet20/32/38-hut-full-net-cifar100}          \\
\hc Stage-wise Trace      & \fref{fig:resnet32_stagewise_hut}, \ref{fig:resnet20/38_stagewise_hut}  &\fref{fig:resnet20/38_stagewise_hut-cifar100}  \\
\ha \multirow{2}{*}{}    
ESD
 &\fref{fig:resnet20-slq-full-net-part}, \ref{fig:resnet20-slq-full-net-all}, \ref{fig:resnet32-slq-full-net-all}, 
& \fref{fig:resnet20-slq-full-net-all-cifar100}, \ref{fig:resnet32-slq-full-net-all-cifar100},  \\
&\ref{fig:resnet38-slq-full-net-all}, \ref{fig:resnet56-slq-full-net-all} 
& \ref{fig:resnet38-slq-full-net-all-cifar100}  \\
\hc \multirow{2}{*}{}
Loss Landscape
& \fref{fig:resnet38-loss-landscape-part}, \ref{fig:resnet20-loss-landscape-all}, \ref{fig:resnet32-loss-landscape-all}, 
& \fref{fig:resnet20-loss-landscape-all-cifar100}, \ref{fig:resnet32-loss-landscape-all-cifar100},  \\
\hc & \ref{fig:resnet38-loss-landscape-all}, \ref{fig:resnet56-loss-landscape-all} 
& \ref{fig:resnet38-loss-landscape-all-cifar100}
\end{tabular}
\end{table}
\section{Conclusions}
\label{sec:conclusions}

We have developed \OURS, an open-source framework for analyzing NN behaviour through the lens of the Hessian~\cite{pyhessian}.
\OURS enables direct and efficient computation of Hessian-based statistics, including the top eigenvalues, the trace, and the full ESD, with support for distributed-memory execution on cloud/supercomputer systems.
Importantly, since it uses matrix-free techniques, \OURS accomplishes this without the need to form the full Hessian.
This means that we can compute second-order statistics for state-of-the-art NNs in times that
are only marginally longer than the time used by popular stochastic gradient-based techniques.

As a typical application, we have also shown how \OURS can be used to study in detail the impact of popular NN 
architectural changes (such as adding/modifying BN and residual connections) on the NN loss landscape. 
Importantly, we found that adding BN layers does not necessarily result in a smoother loss landscape, as claimed by~\cite{santurkar2018does}.
We have observed this phenomenon only for deeper models, where removing the BN layer results in convergence to 
``sharp'' local minima that have high training loss and poor generalization, but it does not seem to hold for shallower models. 
We also showed that removing residual connections resulted in a slightly coarser loss landscape, a finding which we illustrated with parametric 3D visualizations, and which all three Hessian spectrum metrics confirmed.
We have open-sourced \OURS to encourage reproducibility and as a scalable framework for research on second-order optimization methods, on practical diagnostics for NN learning/generalization, as well as on analytics tools for NNs more generally.

\section*{Acknowledgments}
This work was supported by a gracious fund from Intel and Samsung.
We are also grateful for a gracious fund from Google Cloud, Google TFTC team, as well as  support from the Amazon AWS.
We would also like to acknowledge ARO, DARPA, NSF, and ONR for providing partial support of this~work.

{
\bibliographystyle{plain}

\bibliography{ref.bib}

}
\clearpage
\onecolumn
\appendix

In this appendix, we present additional results to complement and extend the results presented in the main text.

\counterwithin{figure}{section}
\counterwithin{table}{section}

\subsection{Illustration of ResNet Stages}
\begin{figure}[H]
\centering
\includegraphics[width=0.98\textwidth]{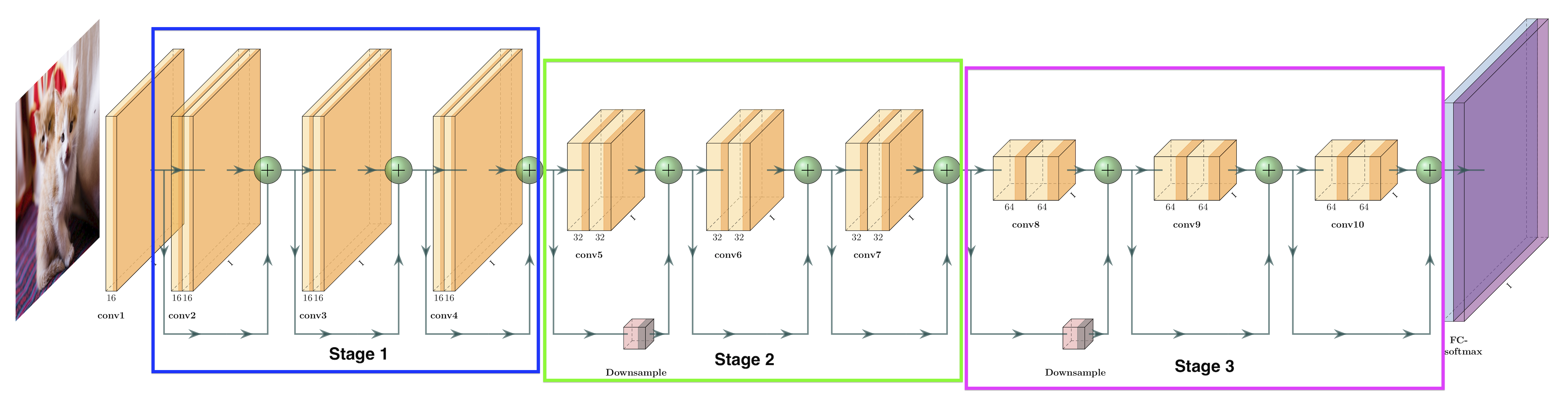}
\caption{
Illustration of ResNet20 on Cifar-10/100 and its three stages. Blue, green, and purple boxes shows the first, second and third stages, respectively.
}
  \label{fig:resnet20_illustration}
\end{figure}

In \fref{fig:resnet20_illustration}, we show the illustration of ResNet20 on Cifar-10/100 and its three stages. 

\subsection{Algorithms}
We provide the pseudo-code for power iteration, Hutchinson algorithm, and stochastic Lanczos Quadrature in this section. 
See \aref{alg:power_iteration} and \aref{alg:hutchinson}.
(\aref{alg:slq} is presented in the main text.)
\begin{algorithm}[h] 
\DontPrintSemicolon
\caption{Power Iteration for Top Eigenvalue Computation}
\label{alg:power_iteration}
    \SetAlgoLined
    \KwInput{Parameter: $\theta$.
    }
    
    Compute the gradient of $\theta$ by backpropagation, i.e., compute $g_\theta=\frac{d L}{d \theta}$.
    
    Draw a random vector $v$ from $N(0,1)$  (same dimension as $\theta$).
    
    Normalize $v$, $v=\frac{v}{\|v\|_2}$
    
    \For(\ \ \quad \quad\quad\quad\quad\tcp*[h]{Power Iteration}){i $=1,2,\ldots$}{
        Compute $gv = g_\theta^Tv$ \tcp*{Inner product}
        
        Compute $Hv$ by backpropagation, $Hv = \frac{d(gv)}{d\theta}$ \tcp*{Get Hessian vector product}
        
        Normalize and reset $v$, $v = \frac{H v}{\|H v\|_2}$
    }
\end{algorithm}
\begin{algorithm}[h] 
\DontPrintSemicolon
\caption{Hutchinson Method for Trace Computation}
\label{alg:hutchinson}
    \SetAlgoLined
    \KwInput{Parameter: $\theta$.
    }
    
    Compute the gradient of $\theta$ by backpropagation, i.e., compute $g_\theta=\frac{d L}{d \theta}$.

    \For(\ \ \quad \quad\quad\quad\quad\tcp*[h]{Hutchinson Steps}){i $=1,2,\ldots$}{
        Draw a random vector $v$ from Rademacher distribution (same dimension as $\theta$).
        
        Compute $gv = g_\theta^Tv$ \tcp*{Inner product}
        Compute $H v$ by backpropagation, $H v = \frac{d(gv)}{d\theta}$ \tcp*{Get Hessian vector product}
        Compute and record $v^TH v$
    }
    Return the average of all computed $v^TH v$.
\end{algorithm}

\subsection{Training Details}
\label{sec:training_details}

We train each model (ResNet, \ResNetBN, and \ResNetRes) for 180 epochs, with five different initial learning rates (0.1, 0.05, 0.01, 0.005, 0.001) on Cifar-10, and ten different initial learning rates (0.1, 0.05, 0.01, 0.005, 0.001, 0.0005, 0.0004, 0.0003, 0.0002, 0.00001) on Cifar-100.
The optimizer is SGD with momentum (0.9). 
The learning rate decays by a factor of 10 at epoch 80, 120.

\begin{table}[H]
\caption{
Accuracy of ResNet, \ResNetBN, and \ResNetRes with different depths, on Cifar-100. Results are similar to those shown in~\tref{tab:model_acc}, i.e., removing BN (\ResNetBN) or residual connections (\ResNetRes) results in performance degradation.
}
\small
\setlength\tabcolsep{2.35pt}
\label{tab:model_acc_cifar100}
\centering
\begin{tabular}{lcccccccccccccc} \toprule
 Model\textbackslash Depth  & 20       & 32       & 38       \\ 
\midrule 
\hc ResNet       & 66.47\%  & 68.26\%  & 69.06\%  \\
\ha \ResNetBN    & 62.82\%  & 25.89\%  & 11.25\%  \\ 
\hc \ResNetRes   & 64.59\%  & 62.08\%  & 62.75\%  \\
     \bottomrule 
\end{tabular}
\end{table}

\begin{figure}[H]
\centering
\includegraphics[width=0.245\textwidth]{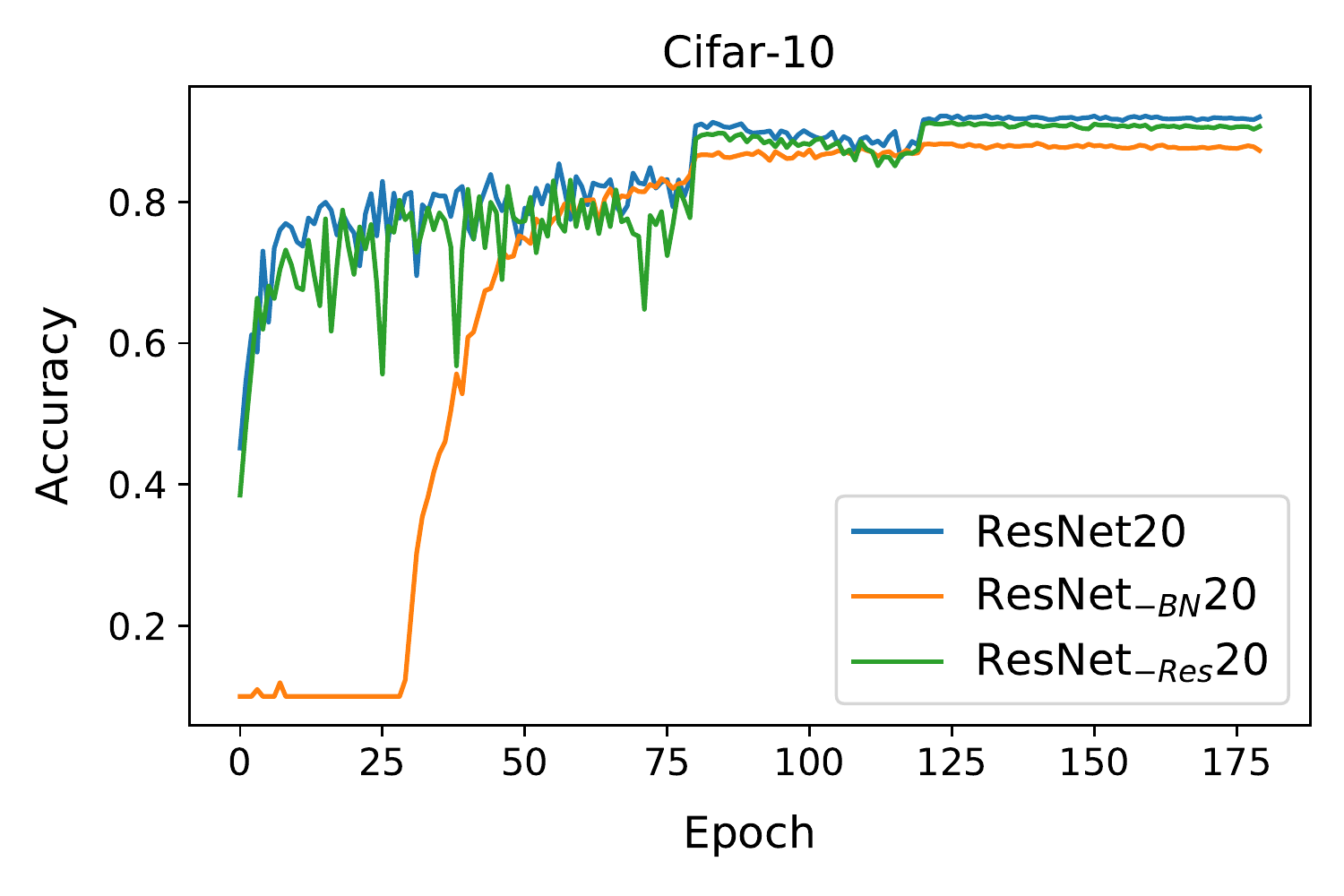}
\includegraphics[width=0.245\textwidth]{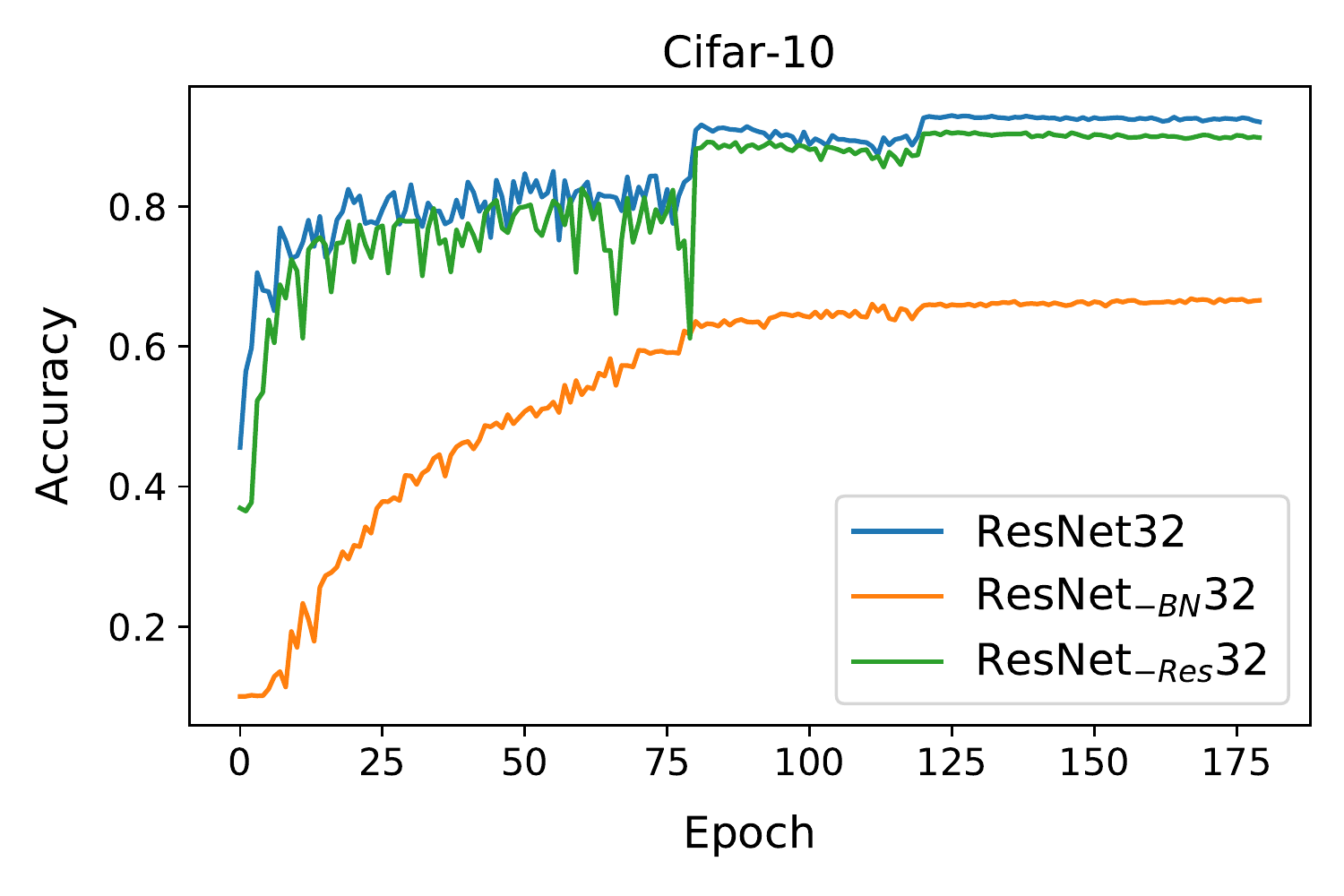}
\includegraphics[width=0.245\textwidth]{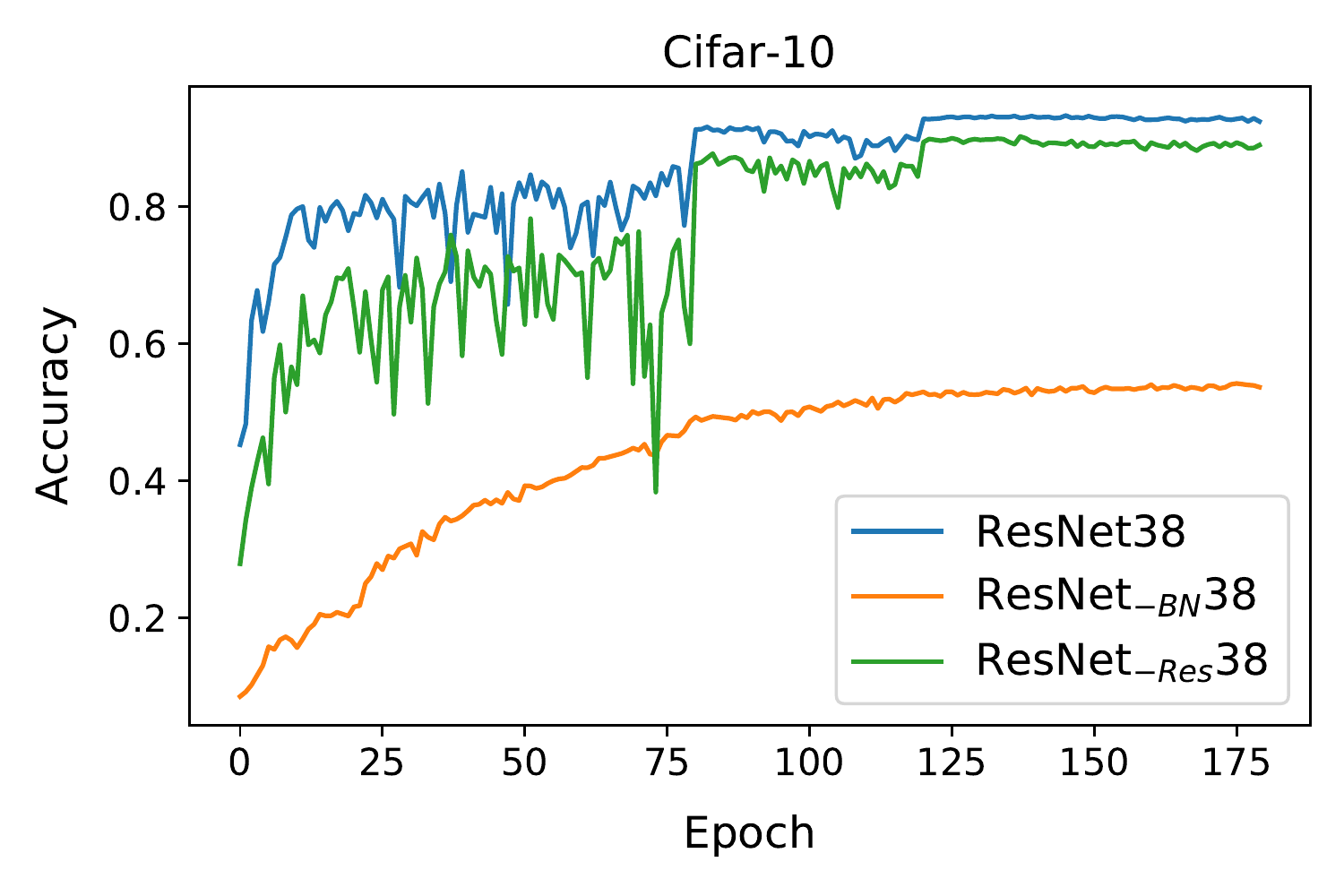}
\includegraphics[width=0.245\textwidth]{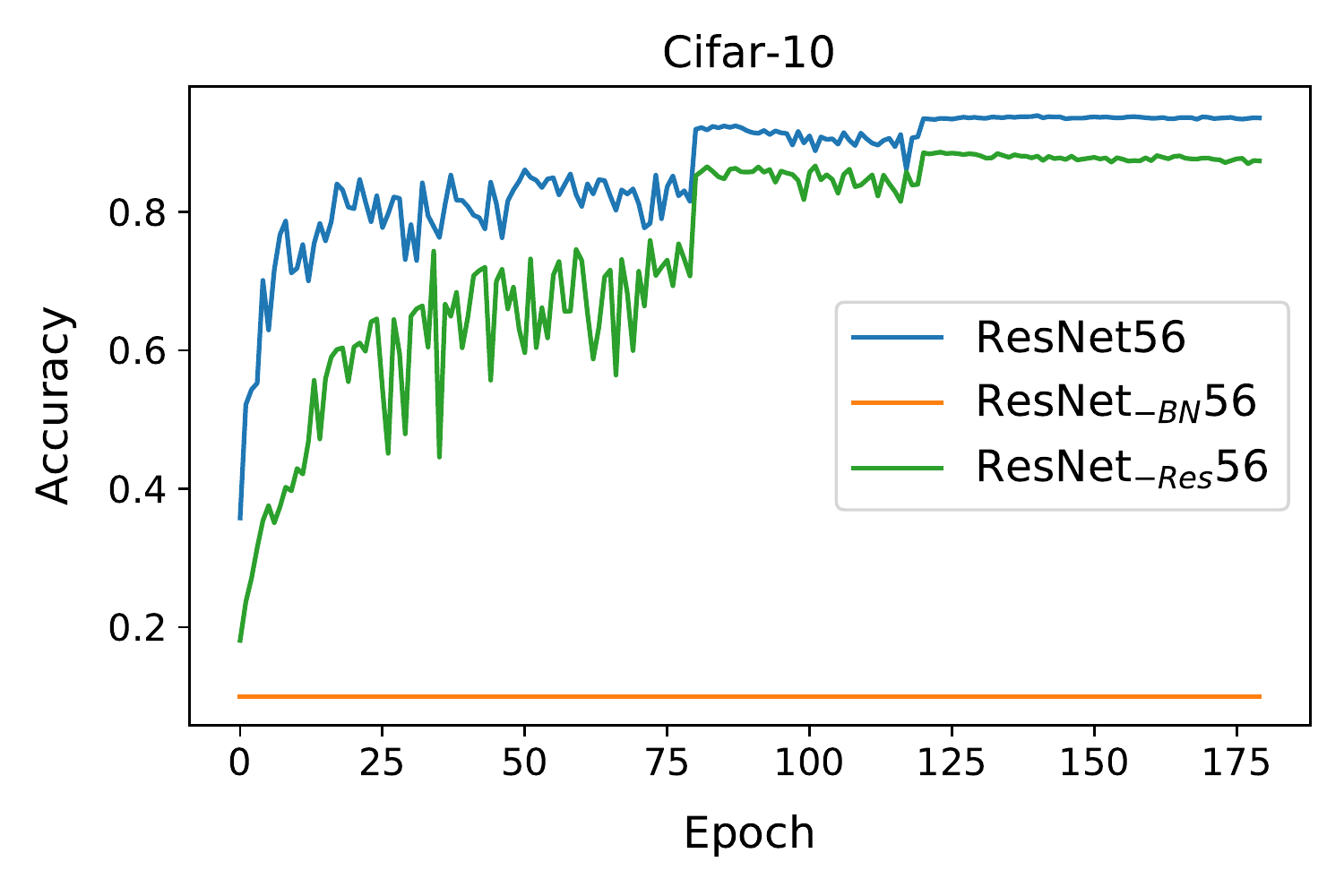}
\caption{
Testing curve of all models reported in~\tref{tab:model_acc}. 
The generalization performance of models without BN (denoted as \ResNetBN) is much worse than the baseline
(denoted as \ResNet). We see a similar but much smaller generalization loss when
the residual connection is removed (denoted as \ResNetRes).
}
  \label{fig:model_acc}
\end{figure}

\begin{figure}[H]
\centering
\includegraphics[width=0.329\textwidth]{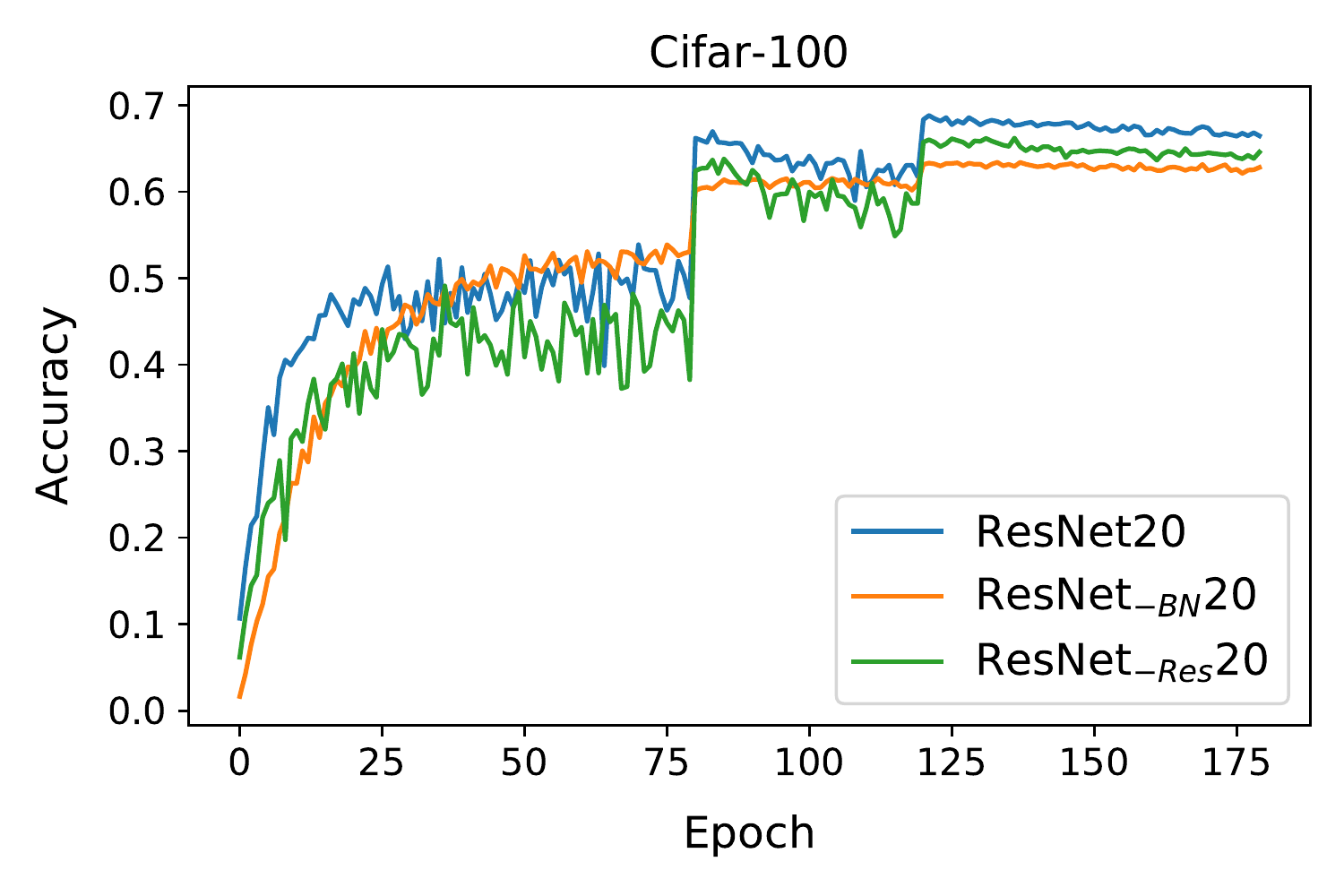}
\includegraphics[width=0.329\textwidth]{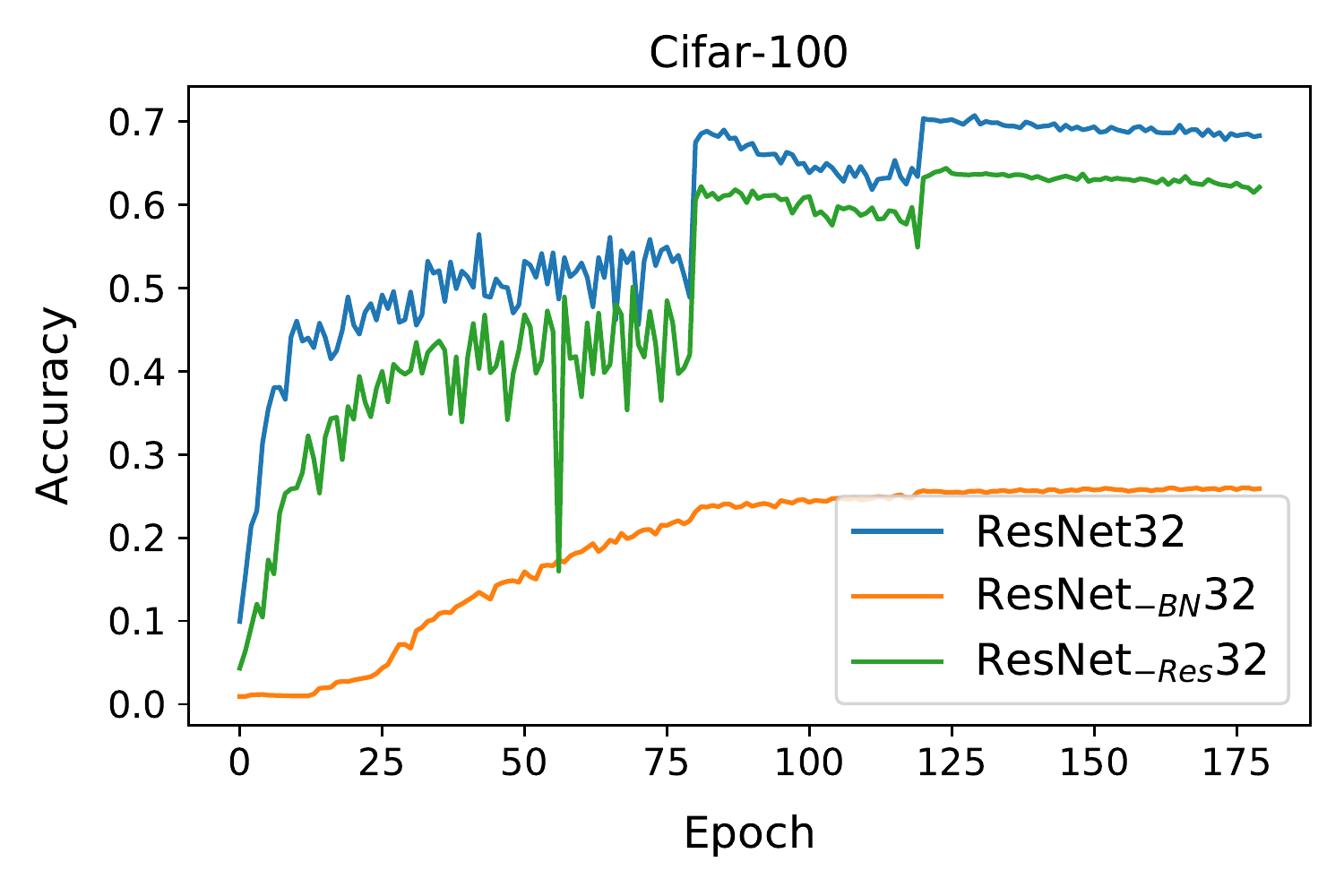}
\includegraphics[width=0.329\textwidth]{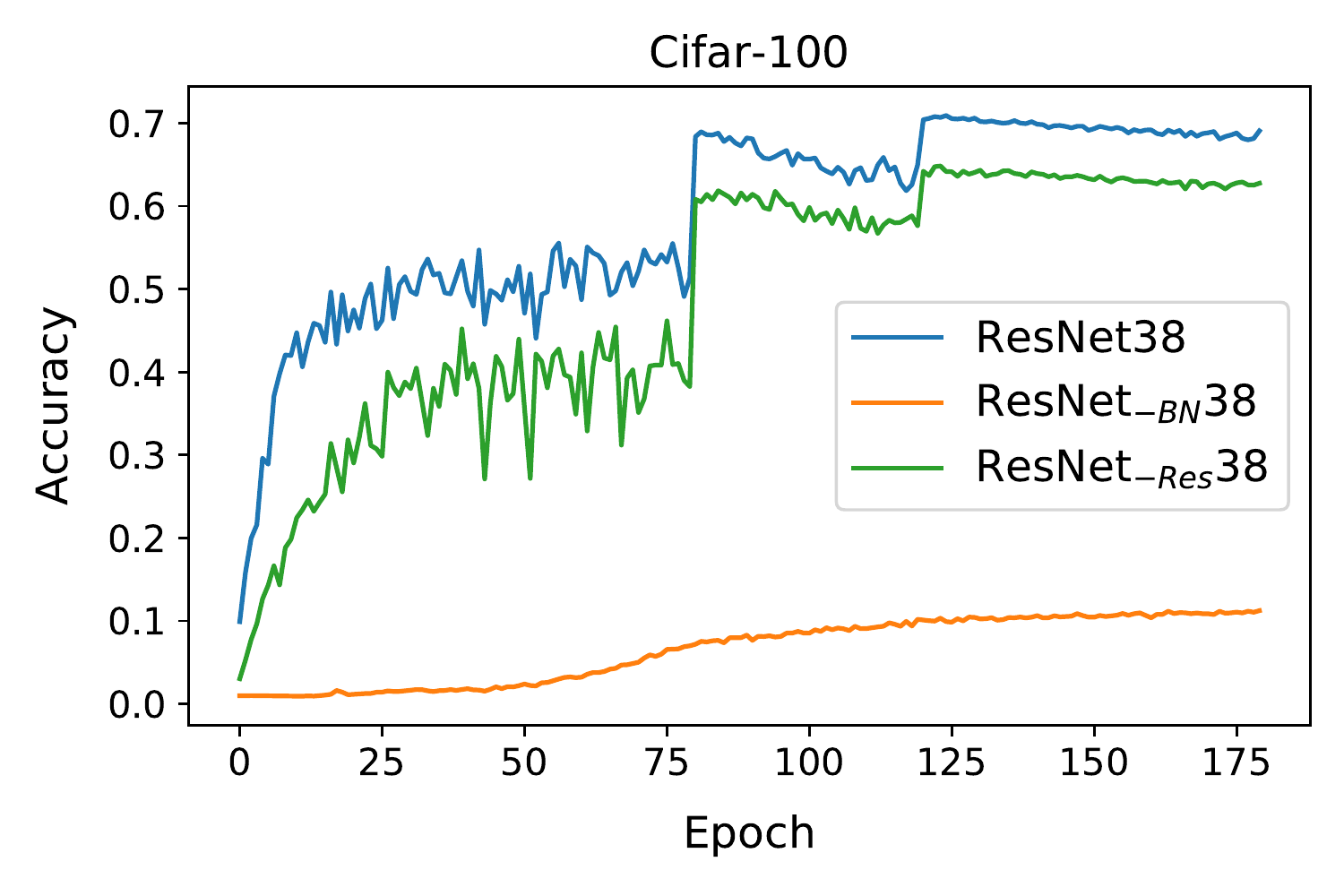}
\caption{
Testing curve of all models reported in~\tref{tab:model_acc_cifar100}. 
The generalization performance of models without BN (denoted as \ResNetBN) is much worse than the baseline
(denoted as \ResNet). We see a similar but much smaller generalization loss when
the residual connection is removed (denoted as \ResNetRes).
}
  \label{fig:model_acc_cifar100}
\end{figure}

\subsection{Loss Landscape Details}
\label{sec:loss_landscape_explanation}
The parametric loss landscape plots are plotted by perturbing the model parameters, $\theta$, along the first and second top eigenvectors of the
Hessian, denoted as $v_1$ and $v_2$.
Then, we compute the loss of K (in our case, $K=4096$) data points with the following formula,
\[
loss = \tilde L(\theta + \epsilon_1v_1 + \epsilon_2v_2) = \frac{1}{K}\sum_{i=1}^{K} l(M(x_i), y_i; \theta + \epsilon_1v_1 + \epsilon_2v_2).
\]


\subsection{Extra Results}
\label{sec:extra_results}
In the remainder of this appendix, we present additional results that we described in the main text.
See \tref{tab:navigation} for a~summary.

\begin{figure*}[!htbp]
\centering
\includegraphics[width=0.98\textwidth]{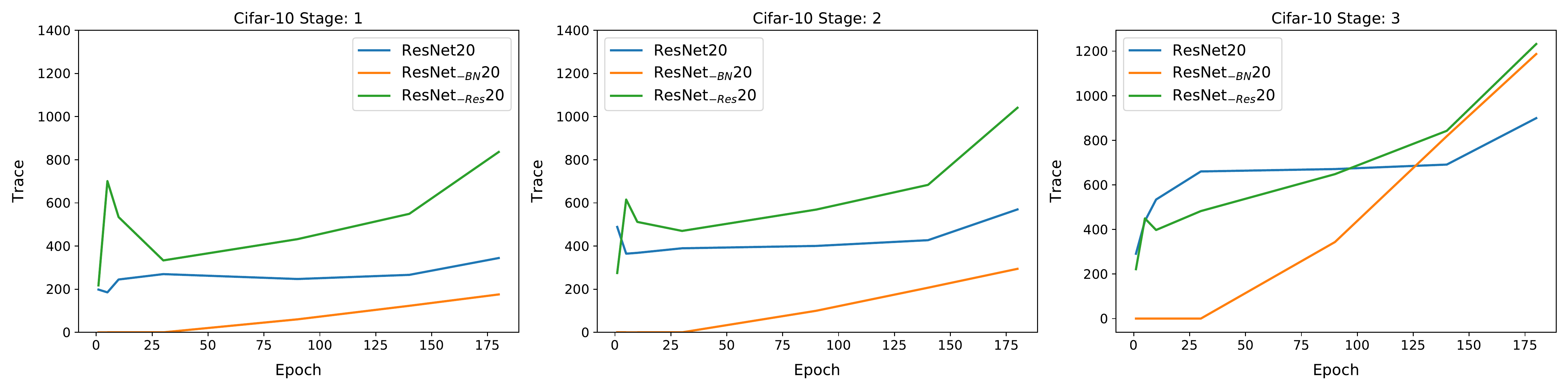}
\includegraphics[width=0.98\textwidth]{figures/resnet32/stagewise/resnet32_hut_50000.pdf}
\includegraphics[width=0.98\textwidth]{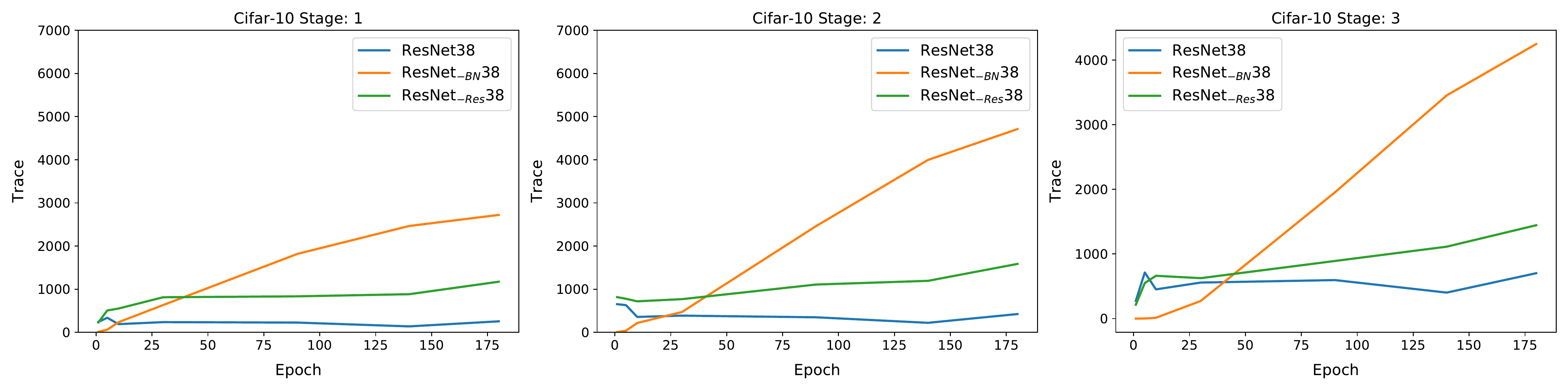}
\includegraphics[width=0.98\textwidth]{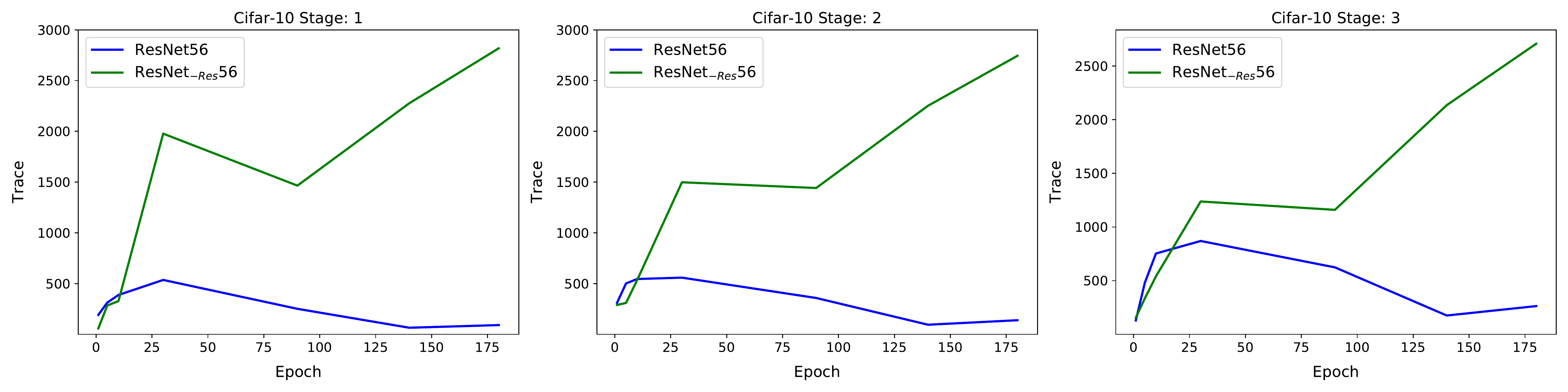}
\caption{Stage-wise Hessian trace of ResNet/\ResNetBN/\ResNetRes with depth 20/32/38/56 on Cifar-10. 
See~\fref{fig:resnet20_illustration} for stage illustration. 
Removing BN layer from the third stage significantly increases the trace, compared to removing BN layer from the first/second stage. 
This has a direct correlation with the final generalization performance, as shown in~\tref{tab:model_acc_rm_bn}. 
}
  \label{fig:resnet20/38_stagewise_hut}
\end{figure*}

\begin{figure*}[!ht]
\centering
\includegraphics[width=0.329\textwidth]{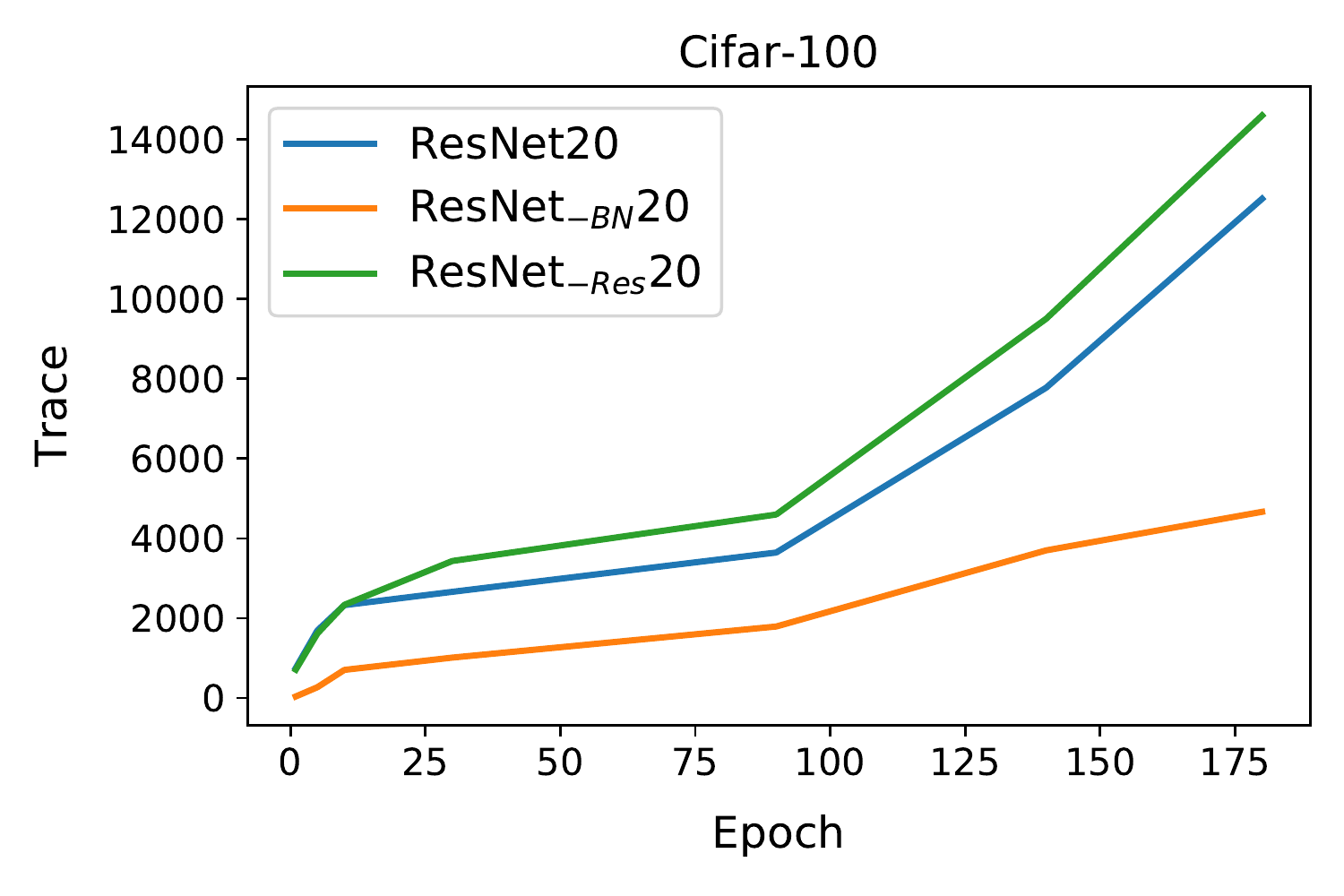}
\includegraphics[width=0.329\textwidth]{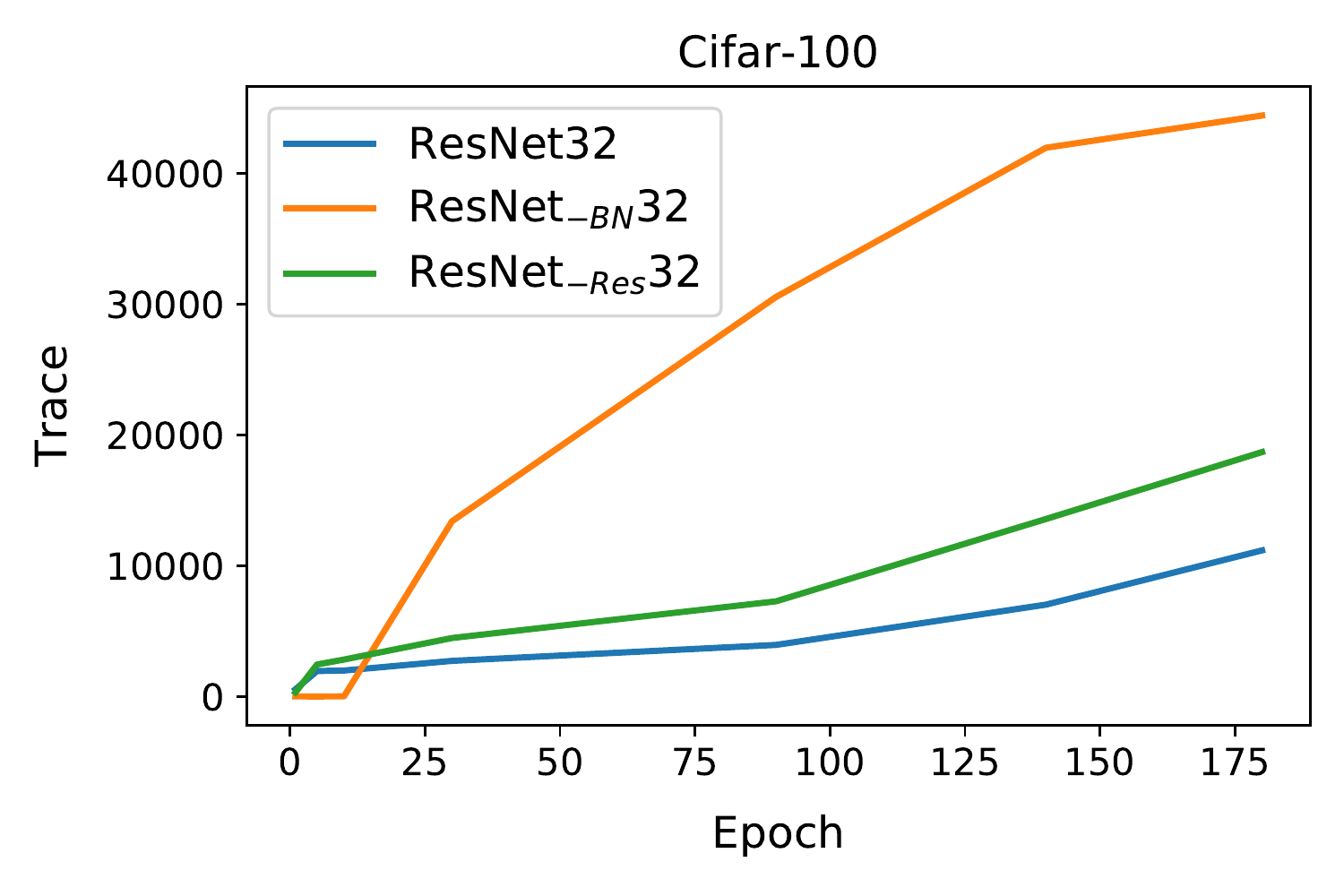}
\includegraphics[width=0.329\textwidth]{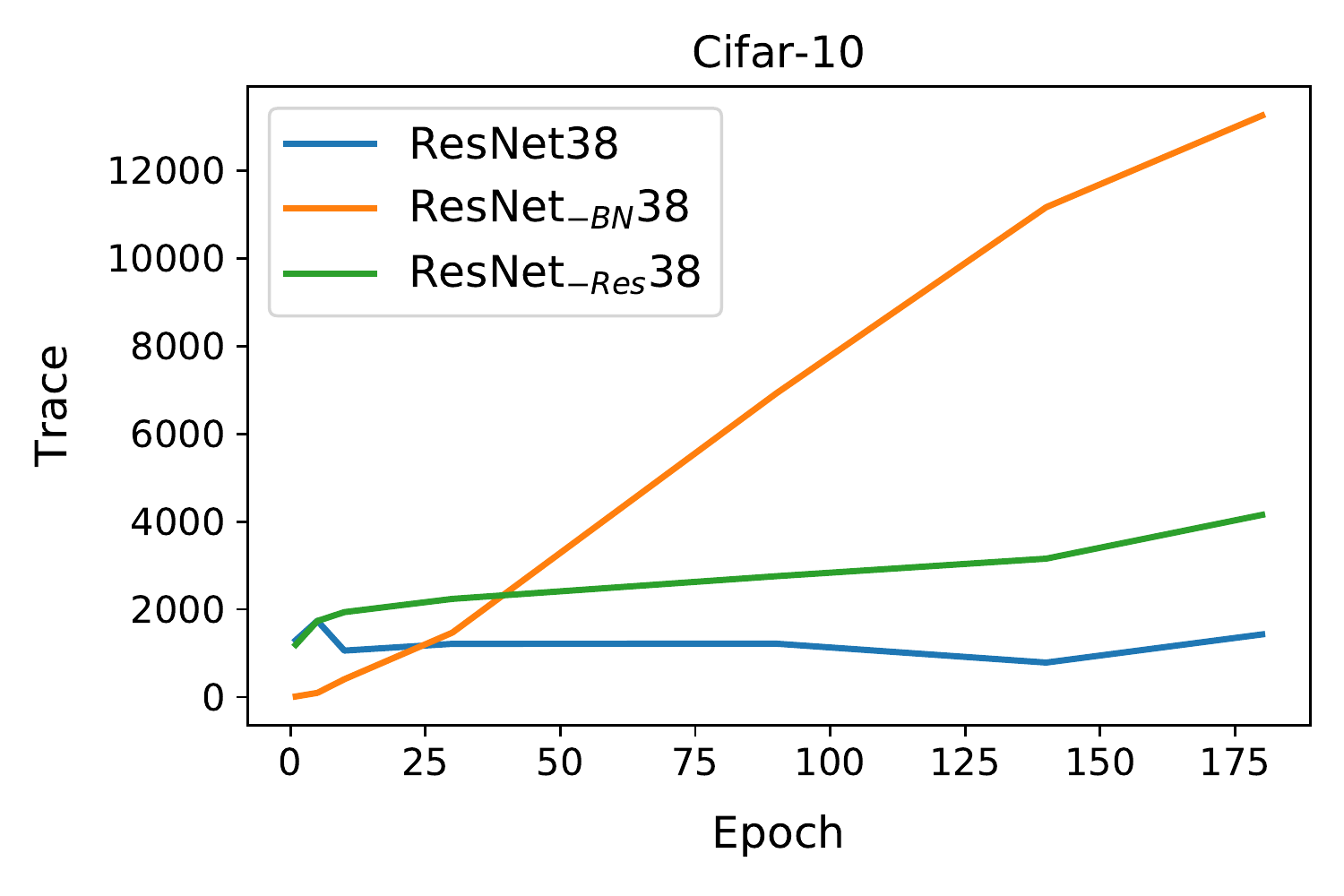}
\caption{
The Hessian trace of the entire network for ResNet/\ResNetBN/\ResNetRes with depth 20/32/38 on Cifar-100. Similar to the results for Cifar-10, shown in~\fref{fig:resnet20/32/38-hut-full-net}, we see that removing the BN layer results in a rapid increase of the Hessian trace, and that removing the residual connection leads to sharper loss landscape throughout training.
}
  \label{fig:resnet20/32/38-hut-full-net-cifar100}
\end{figure*}

\begin{figure*}[!htbp]
\centering
\includegraphics[width=0.98\textwidth]{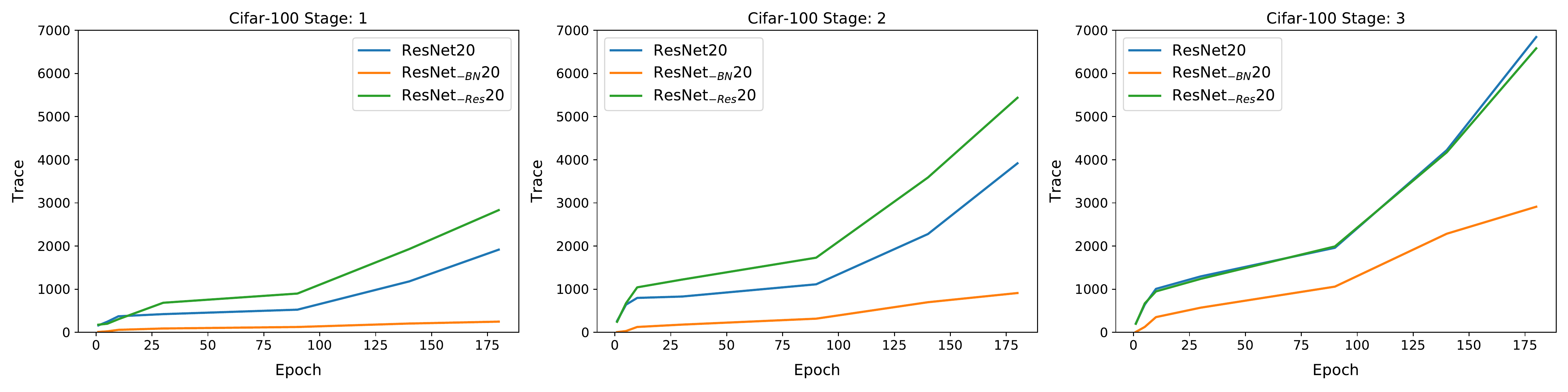}
\includegraphics[width=0.98\textwidth]{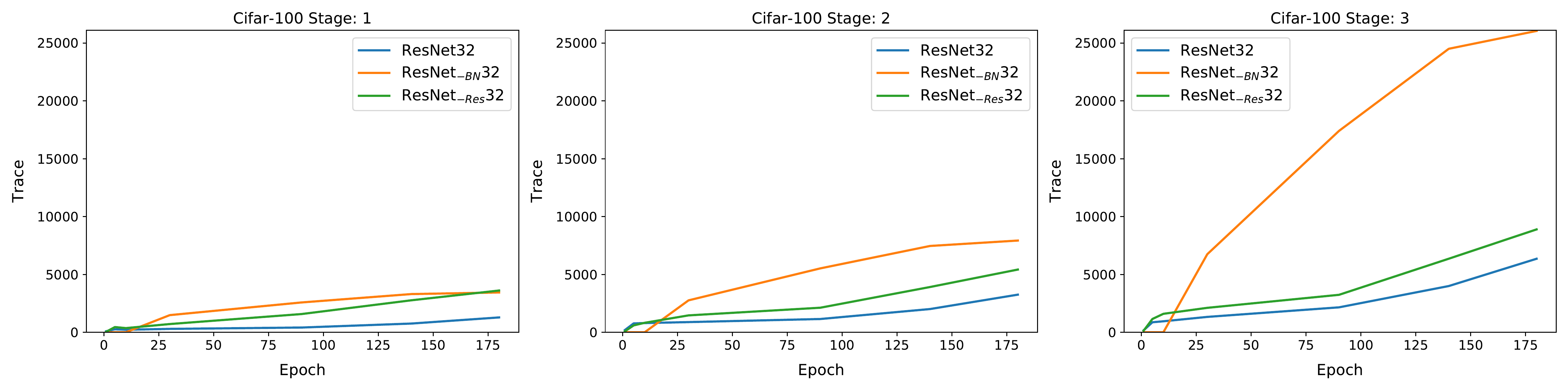}
\includegraphics[width=0.98\textwidth]{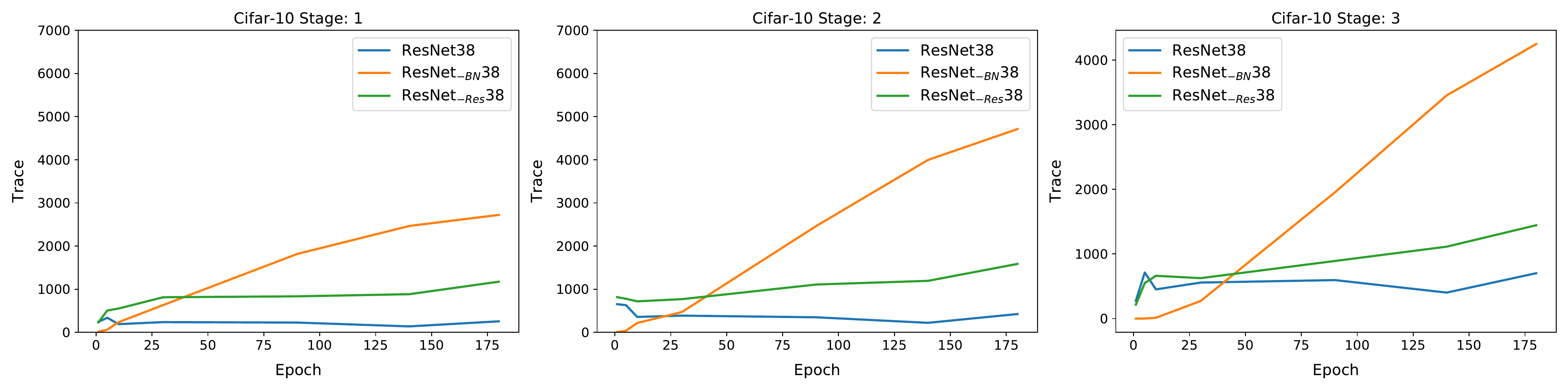}
\caption{Stage-wise Hessian trace of ResNet/\ResNetBN/\ResNetRes with depth 20/32/38 on Cifar-100. 
See~\fref{fig:resnet20_illustration} for stage illustration. 
Removing BN layer from the third stage significantly increases the trace, compared to removing BN layer from the first/second stage. 
This has a direct correlation with the final generalization performance, as shown in~\tref{tab:model_acc_rm_bn-cifar100}. 
}
  \label{fig:resnet20/38_stagewise_hut-cifar100}
\end{figure*}

\begin{figure*}[!htbp]
\centering
\includegraphics[width=0.295\textwidth]{figures/resnet20/full_net/resnet_slq_50000_0.pdf}
\includegraphics[width=0.295\textwidth]{figures/resnet20/full_net/resnetbn_slq_50000_0.pdf}
\includegraphics[width=0.295\textwidth]{figures/resnet20/full_net/resnetres_slq_50000_0.pdf}\\
\includegraphics[width=0.295\textwidth]{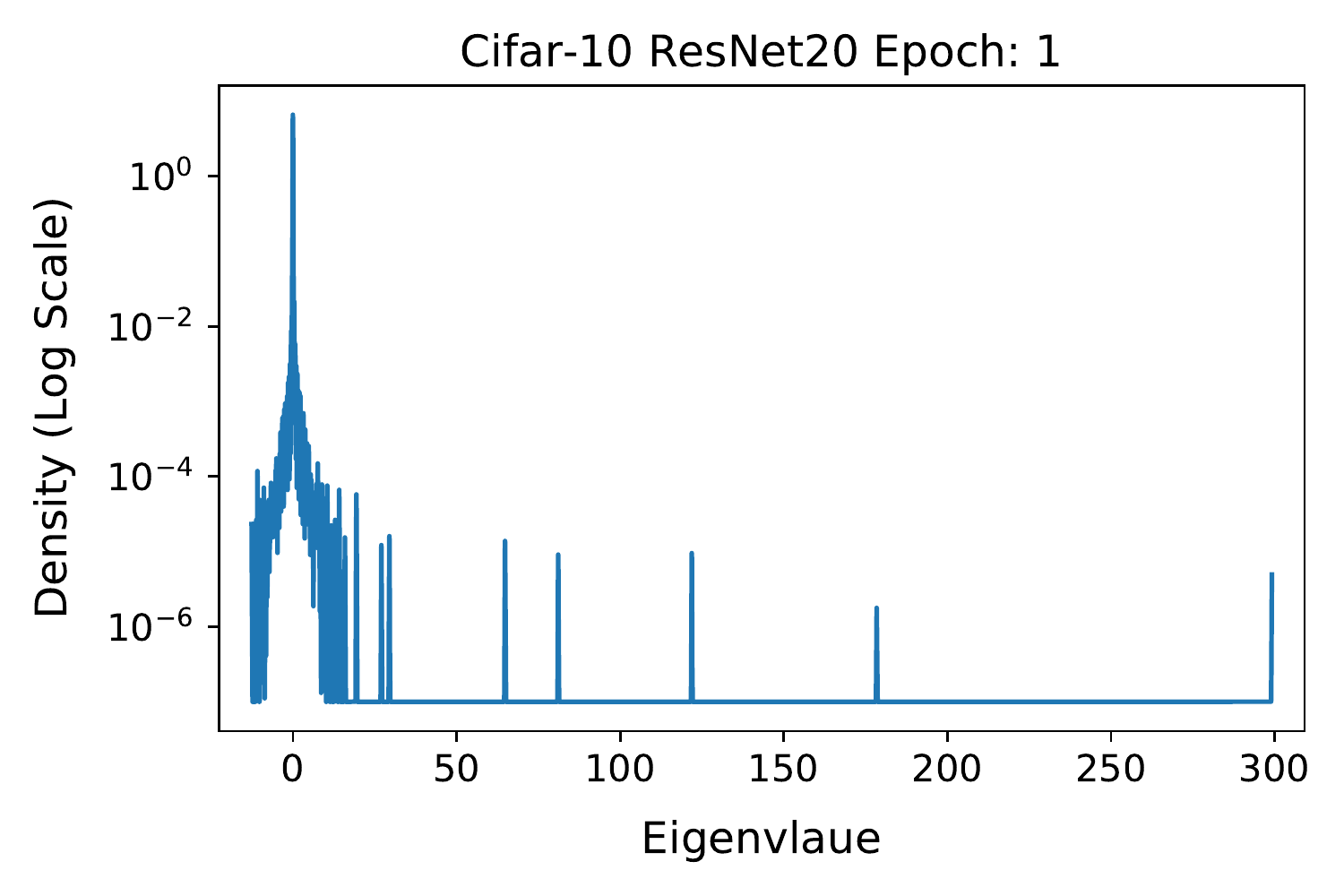}
\includegraphics[width=0.295\textwidth]{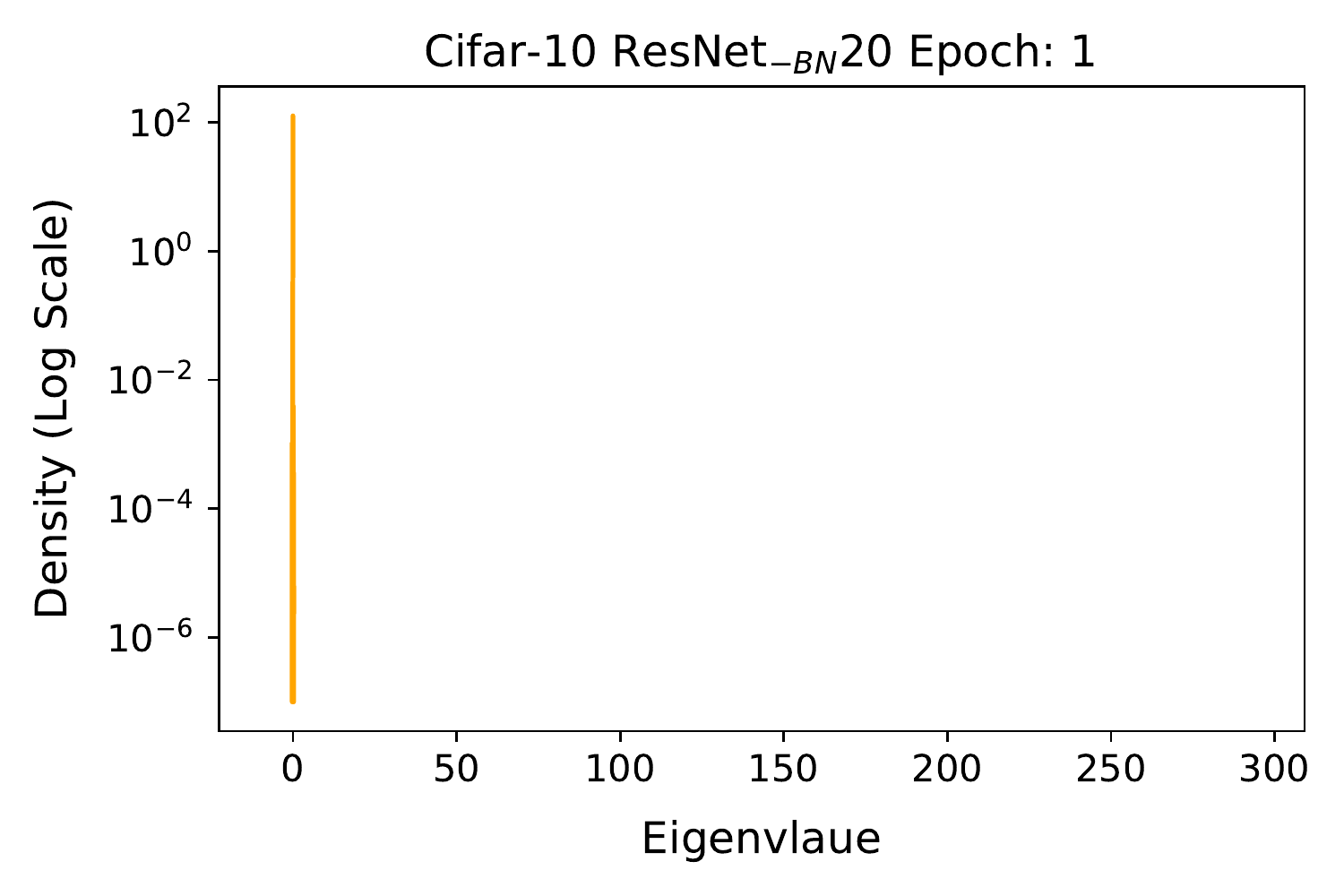}
\includegraphics[width=0.295\textwidth]{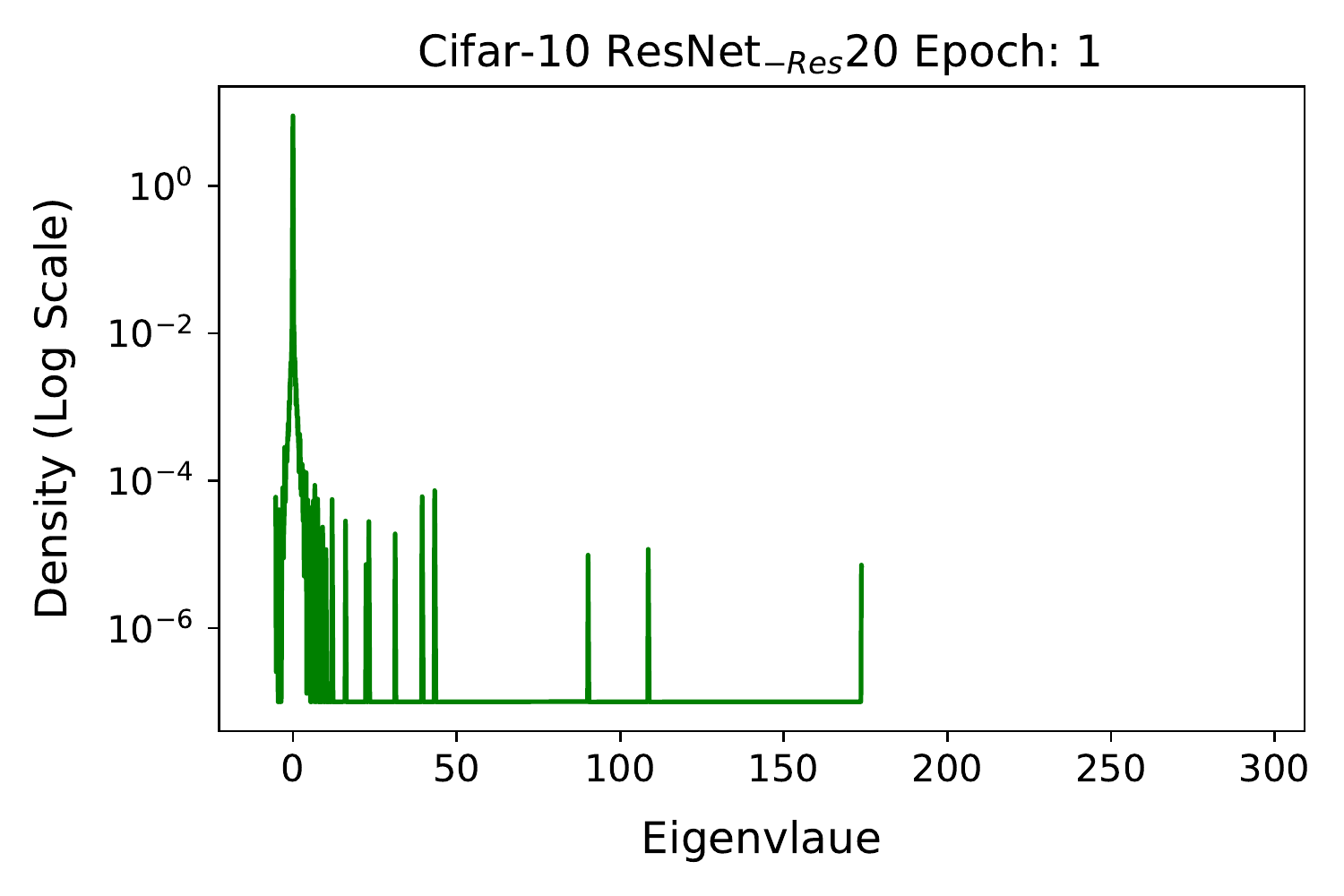}\\
\includegraphics[width=0.295\textwidth]{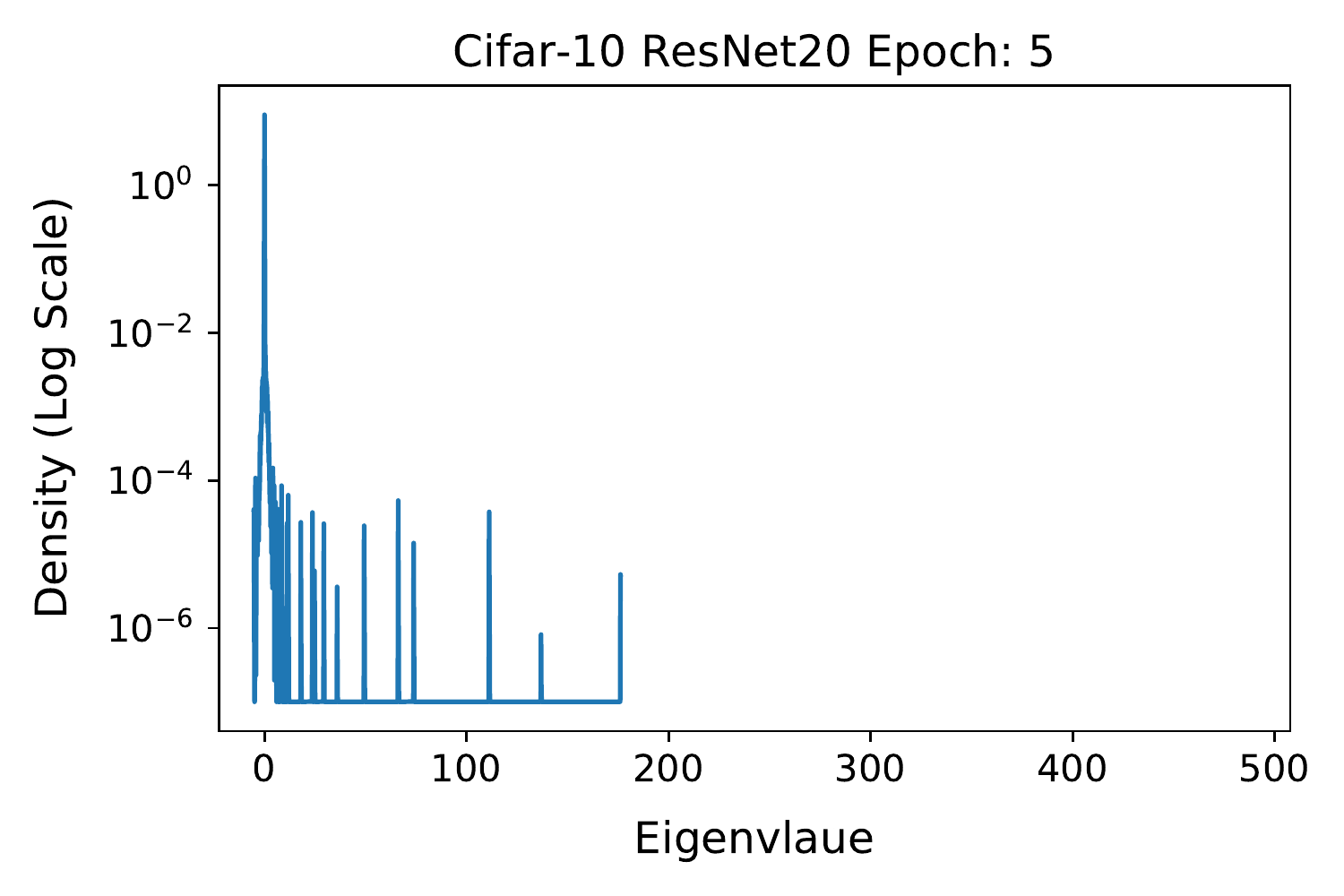}
\includegraphics[width=0.295\textwidth]{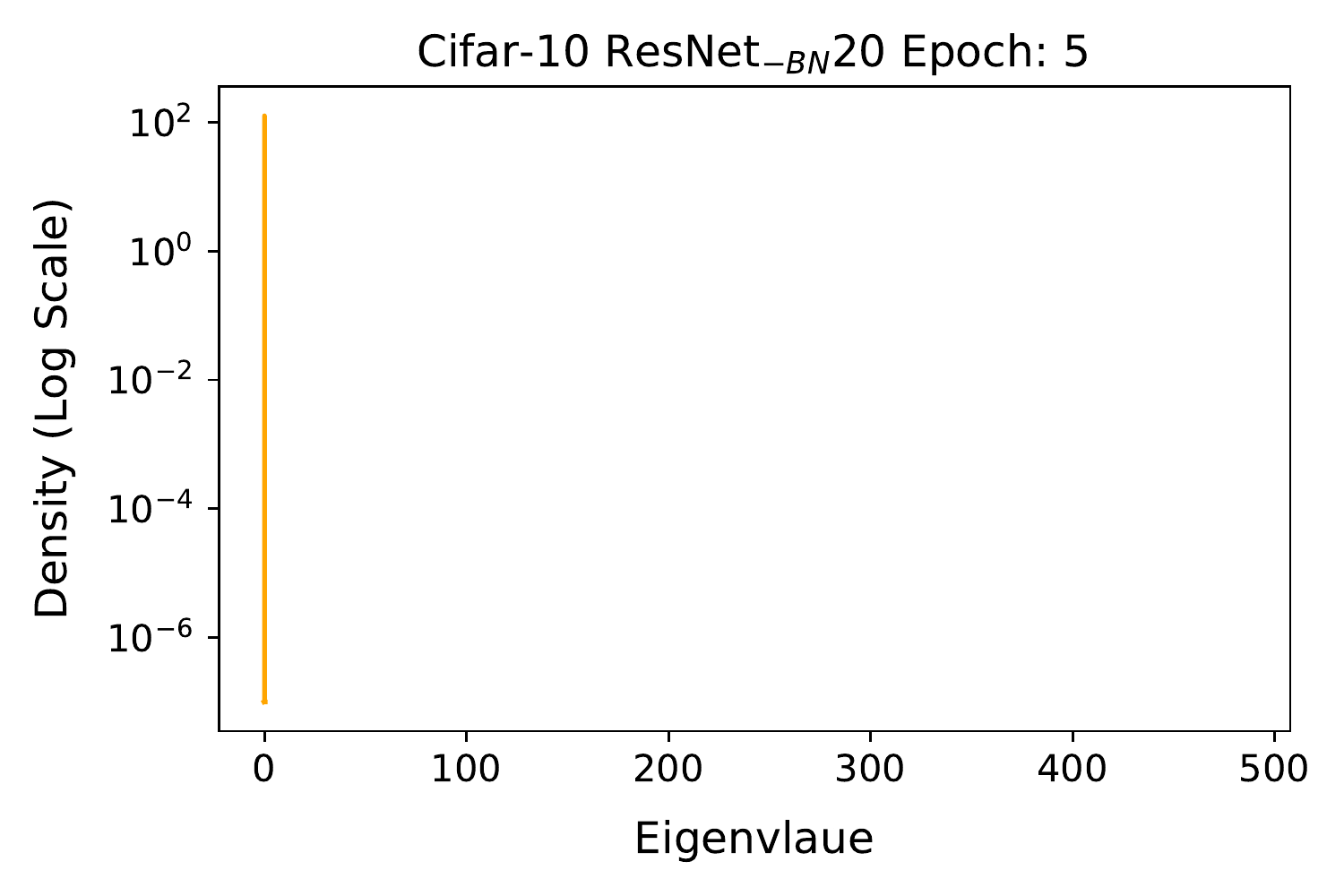}
\includegraphics[width=0.295\textwidth]{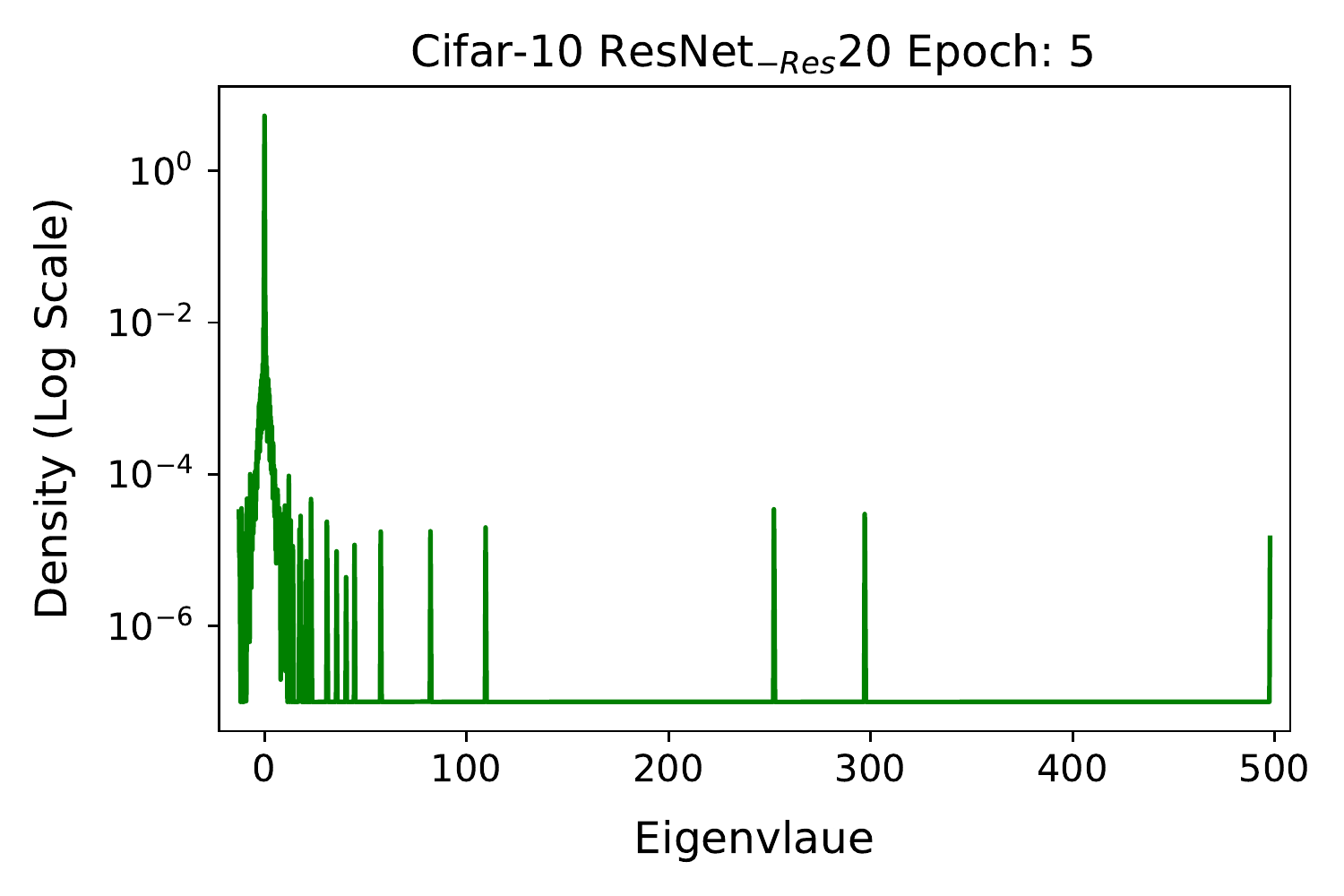}\\
\includegraphics[width=0.295\textwidth]{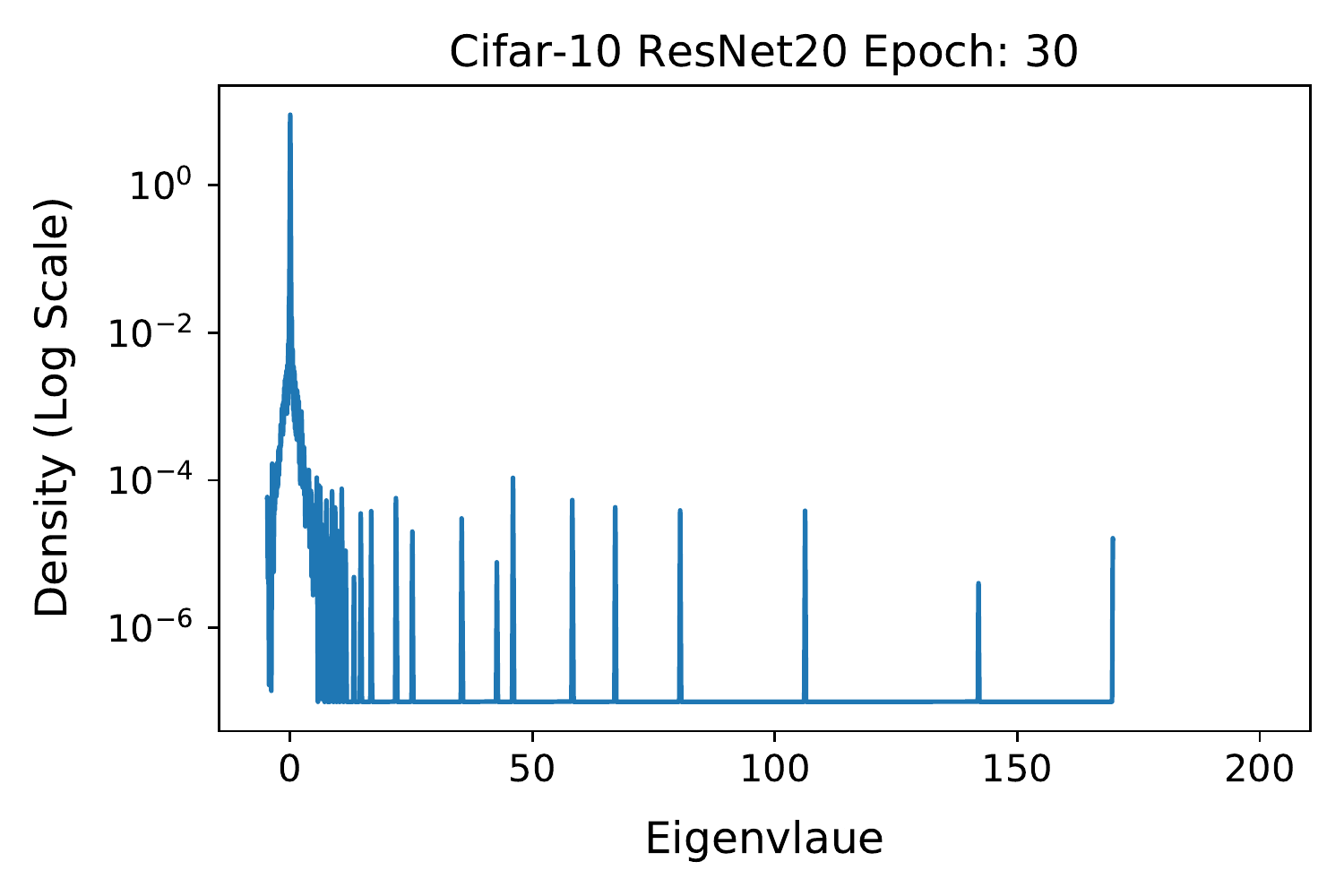}
\includegraphics[width=0.295\textwidth]{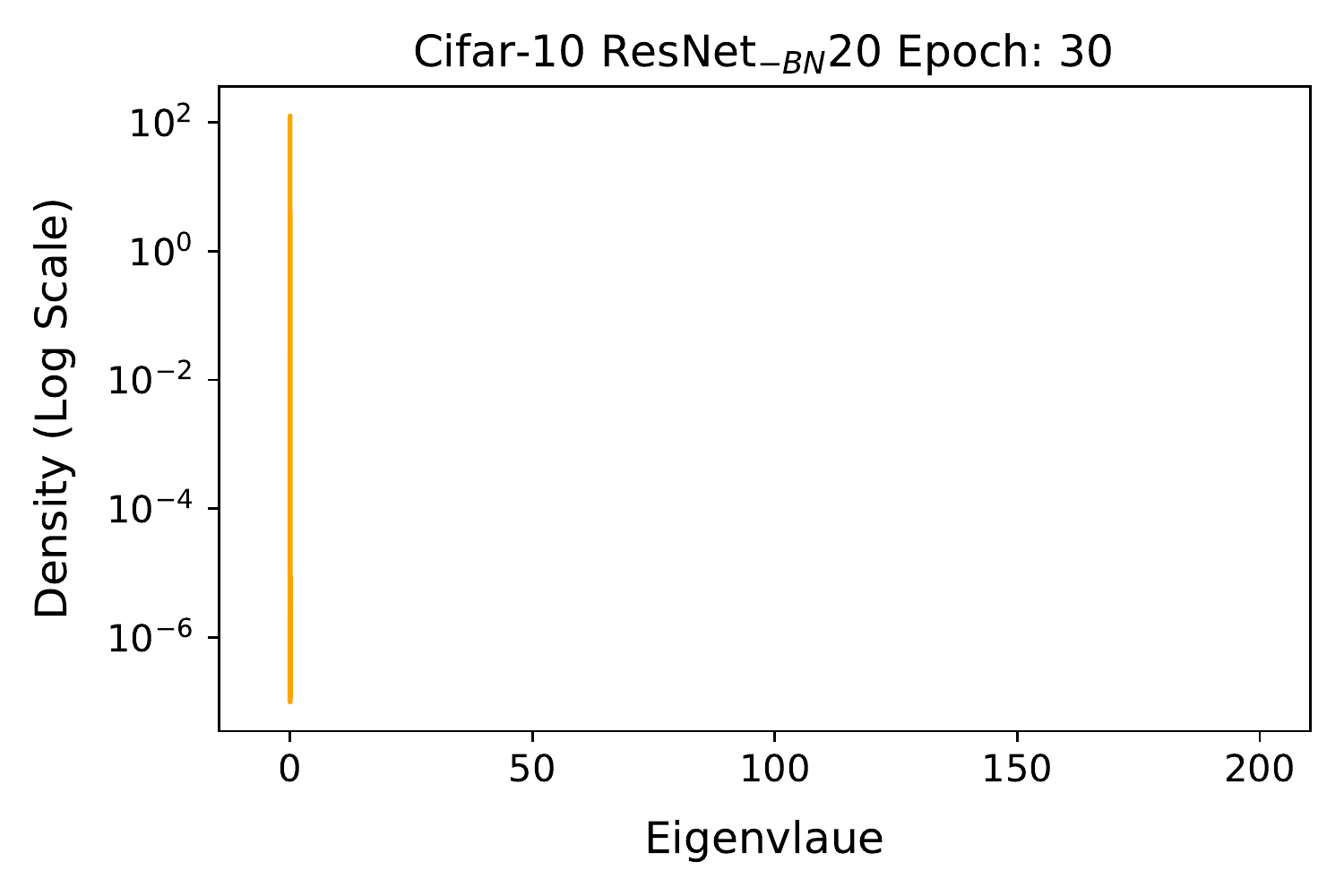}
\includegraphics[width=0.295\textwidth]{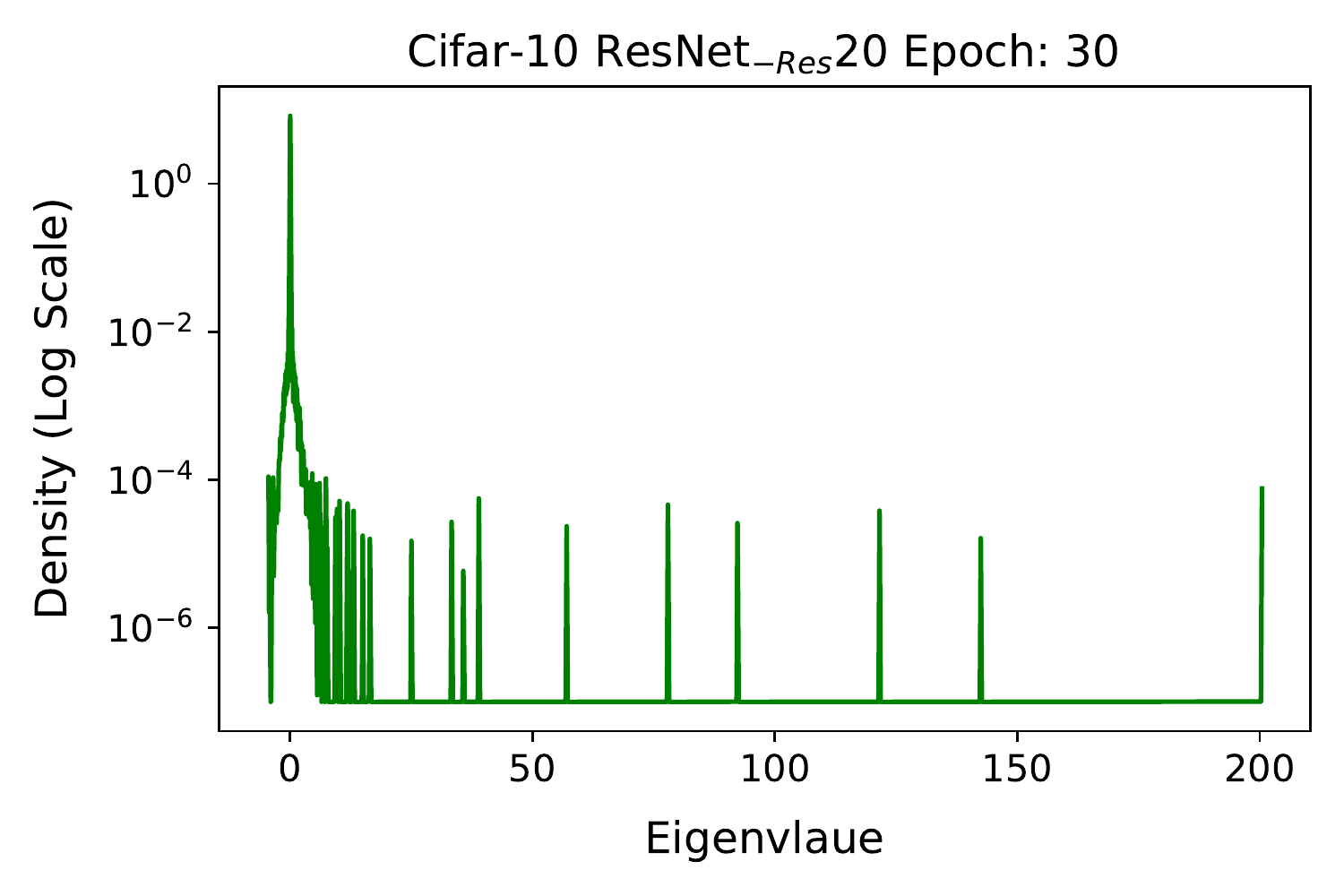}\\
\includegraphics[width=0.295\textwidth]{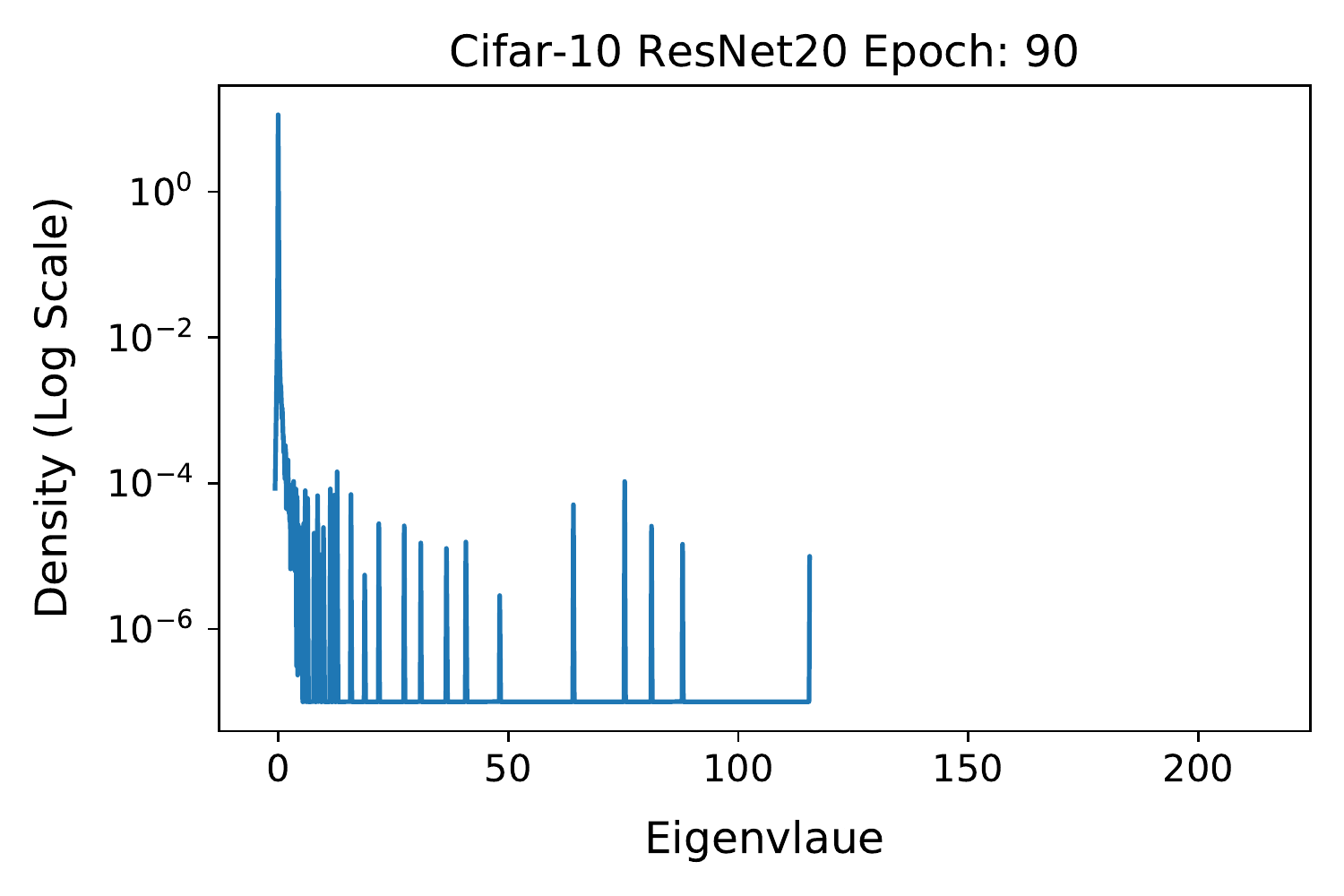}
\includegraphics[width=0.295\textwidth]{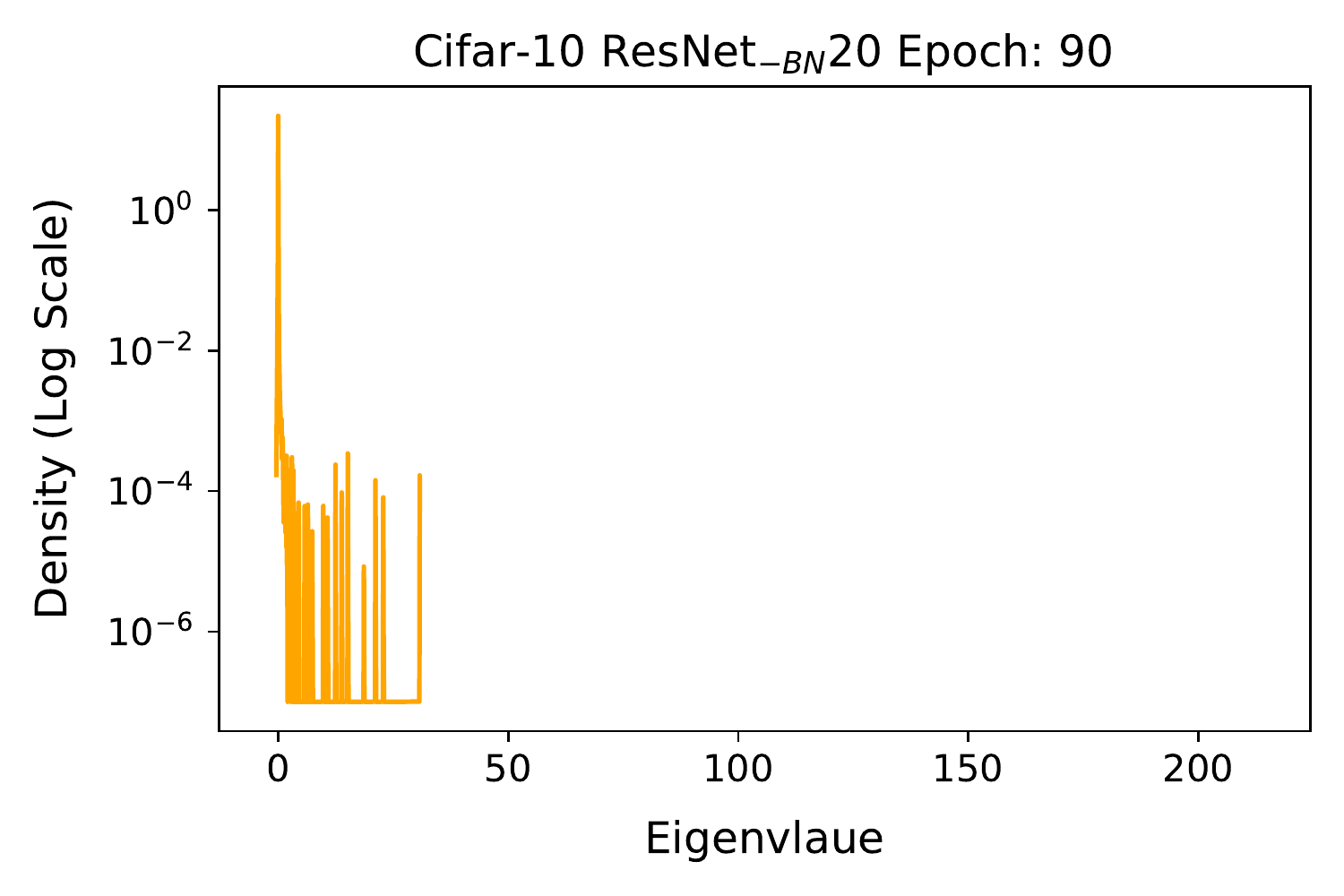}
\includegraphics[width=0.295\textwidth]{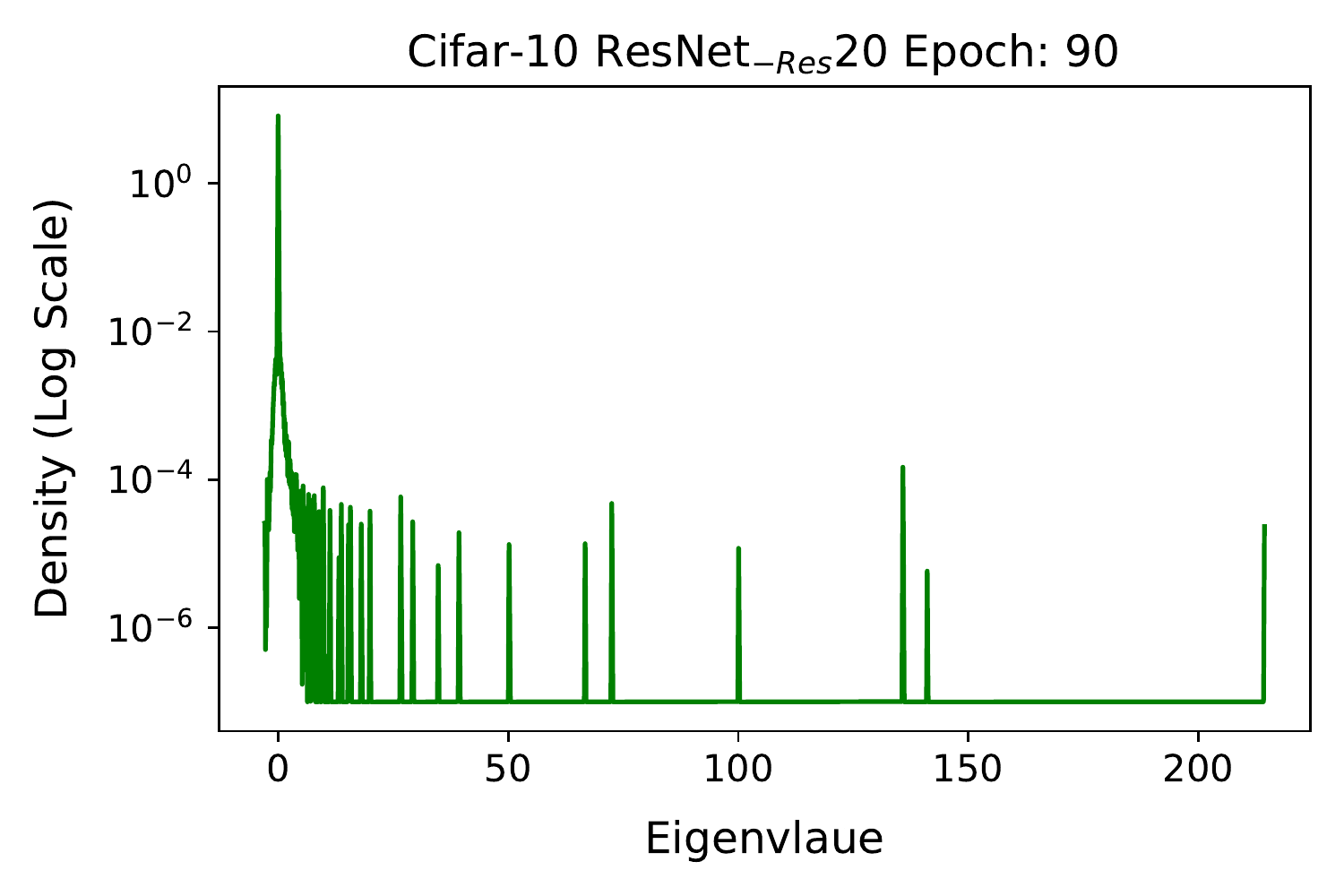}\\
\includegraphics[width=0.295\textwidth]{figures/resnet20/full_net/resnet_slq_50000_180.pdf}
\includegraphics[width=0.295\textwidth]{figures/resnet20/full_net/resnetbn_slq_50000_180.pdf}
\includegraphics[width=0.295\textwidth]{figures/resnet20/full_net/resnetres_slq_50000_180.pdf}\\
\caption{
Hessian ESD of the entire network for ResNet/\ResNetBN/\ResNetRes with depth 20 on Cifar-10 with Hessian batch size 50000. 
This figure shows the Hessian ESD throughout the training process, which is an full version of~\fref{fig:resnet20-slq-full-net-part}. 
One notable thing here is that although \ResNetBN20 has smaller Hessian ESD support range than \ResNet20 does, the Hessian ESD of \ResNetBN20 centers around zero (at least) until epoch 90. 
This clearly shows that training without BN is indeed~harder. 
}
  \label{fig:resnet20-slq-full-net-all}
\end{figure*}

\begin{figure*}[!htbp]
\centering
\includegraphics[width=0.295\textwidth]{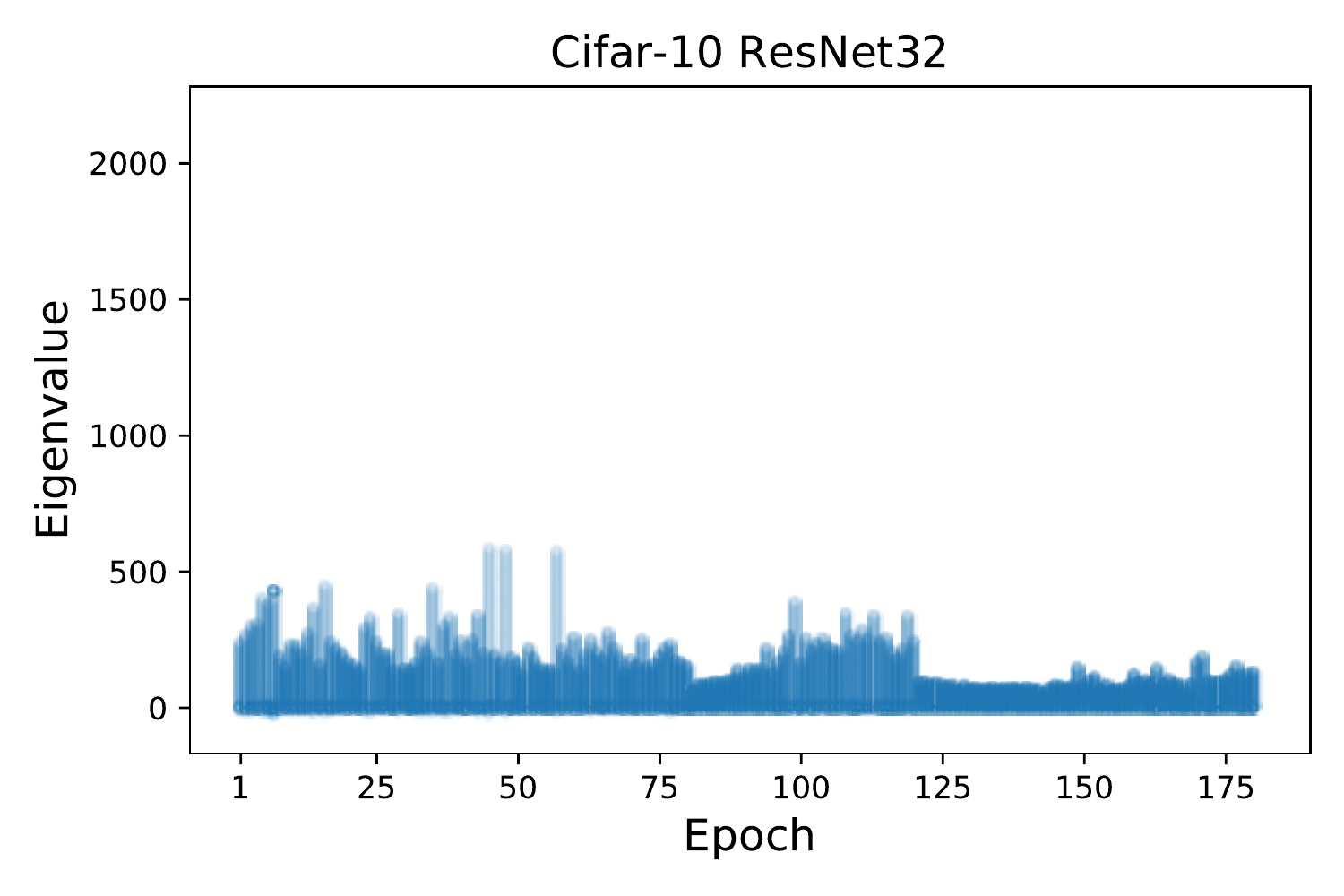}
\includegraphics[width=0.295\textwidth]{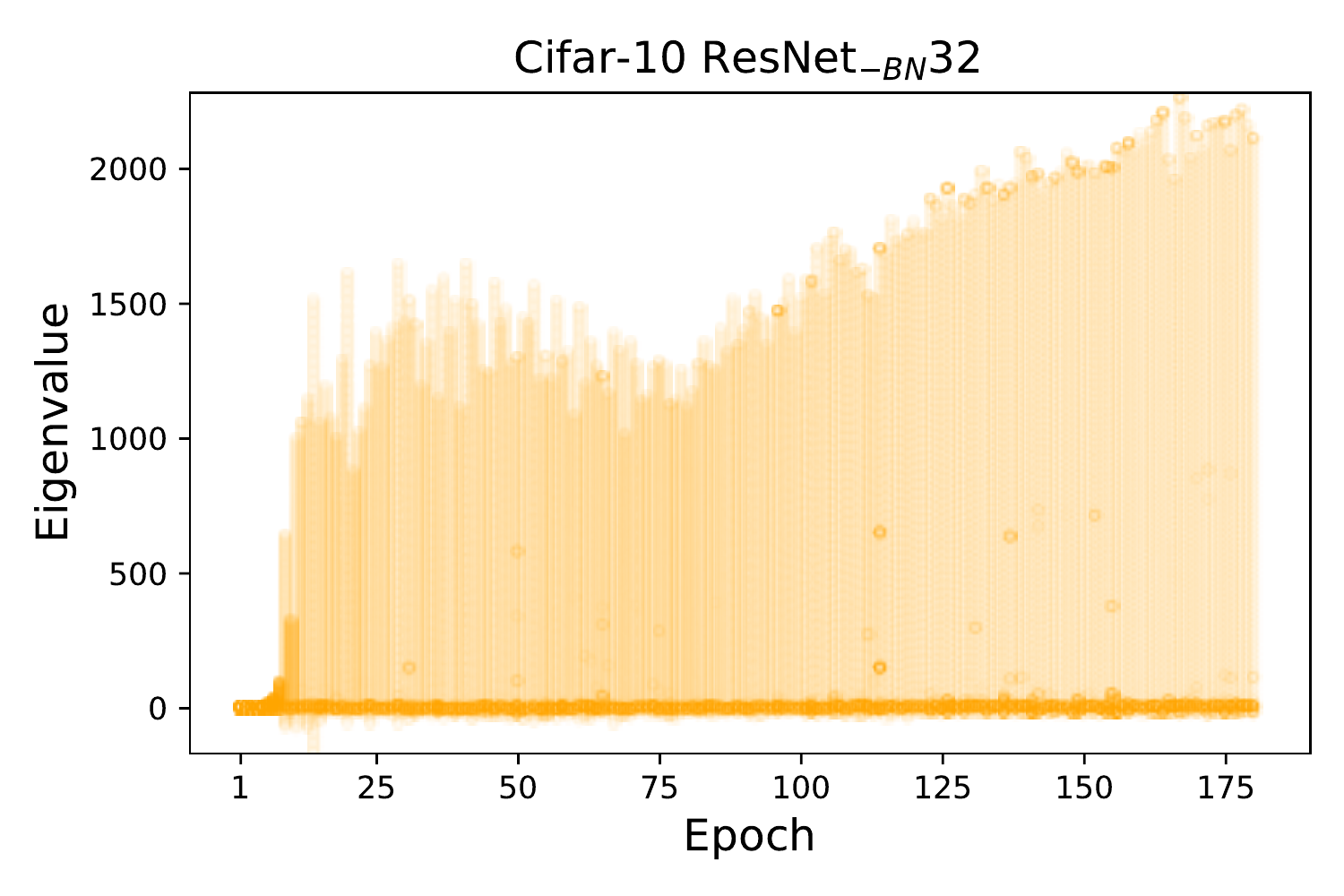}
\includegraphics[width=0.295\textwidth]{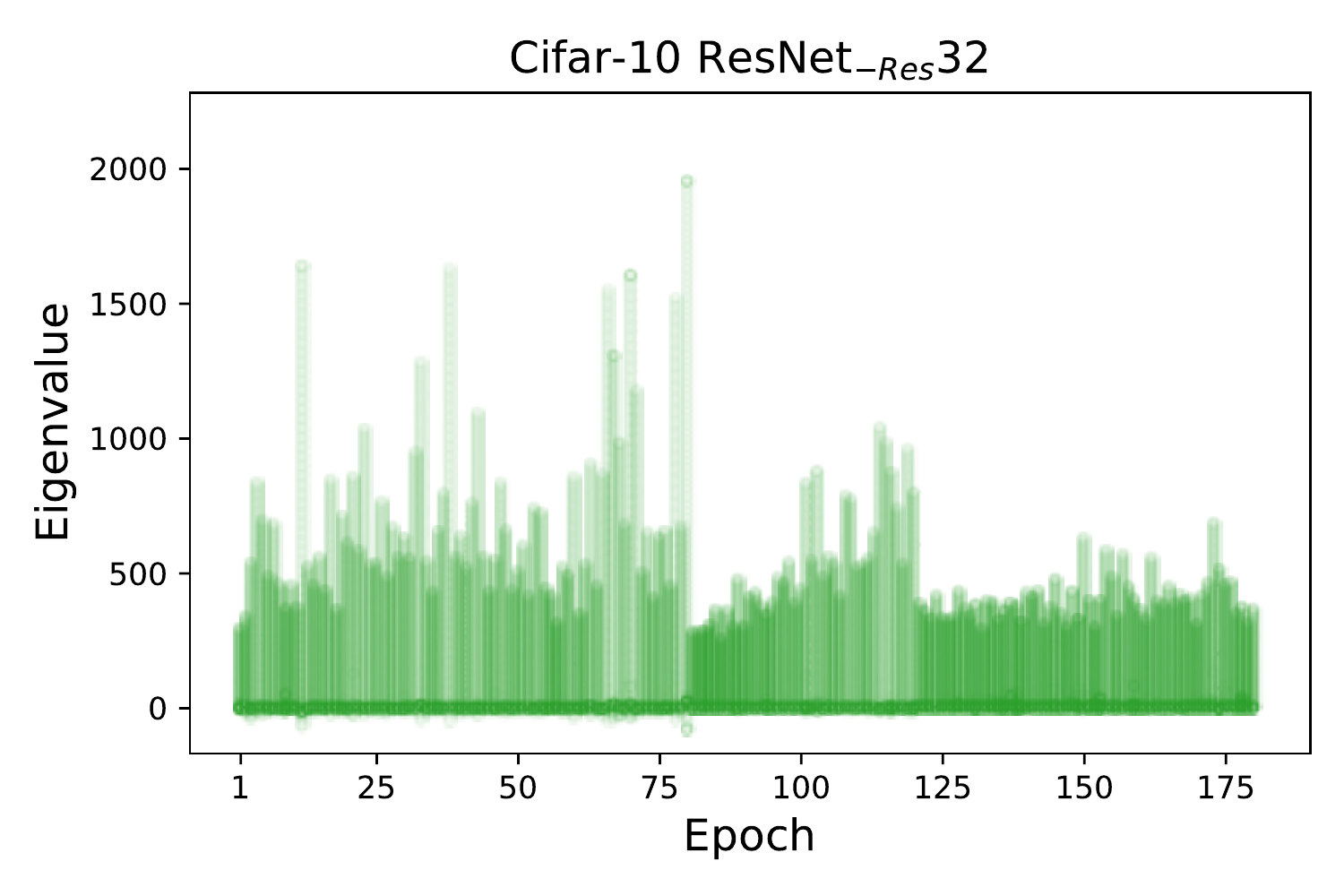}\\
\includegraphics[width=0.295\textwidth]{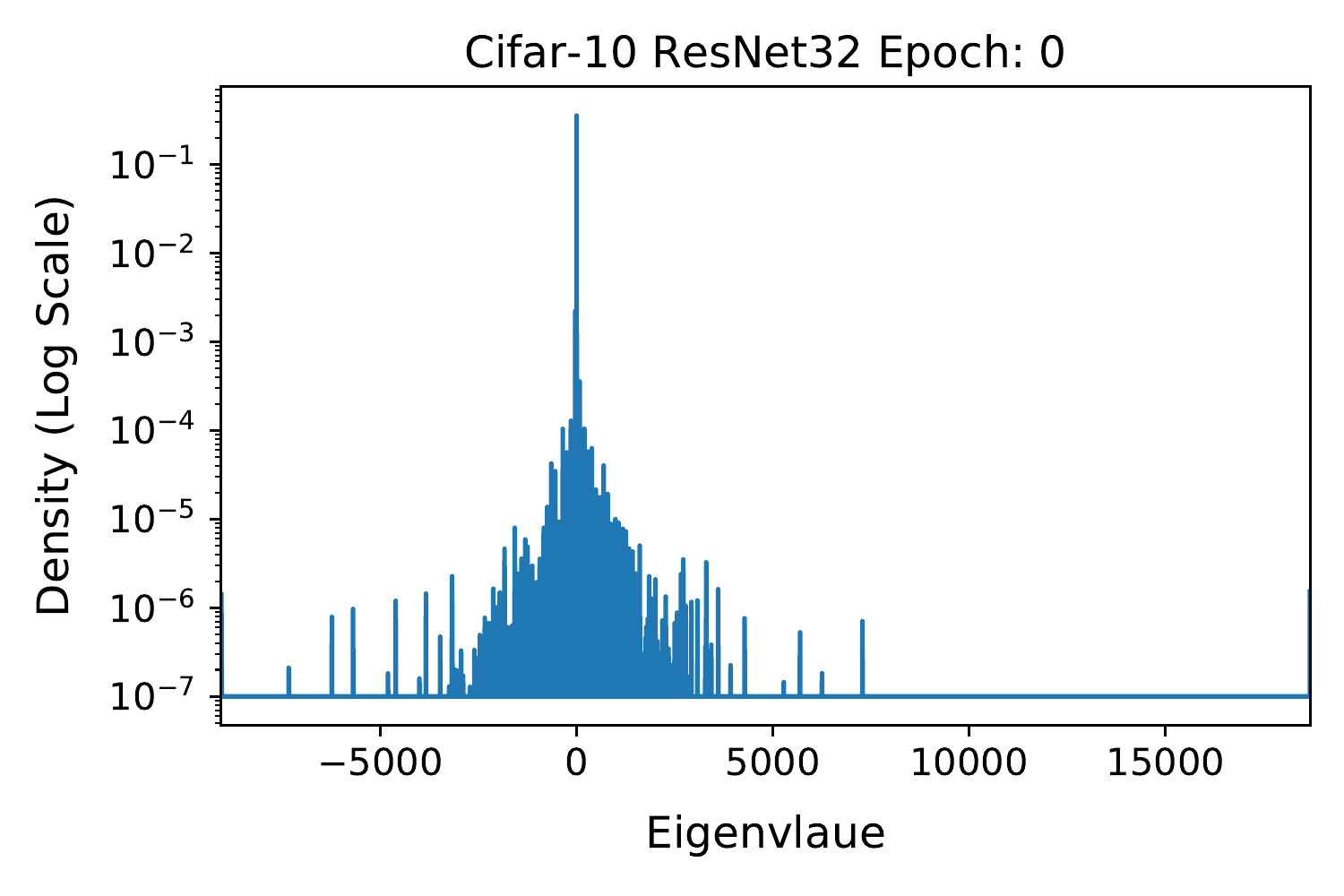}
\includegraphics[width=0.295\textwidth]{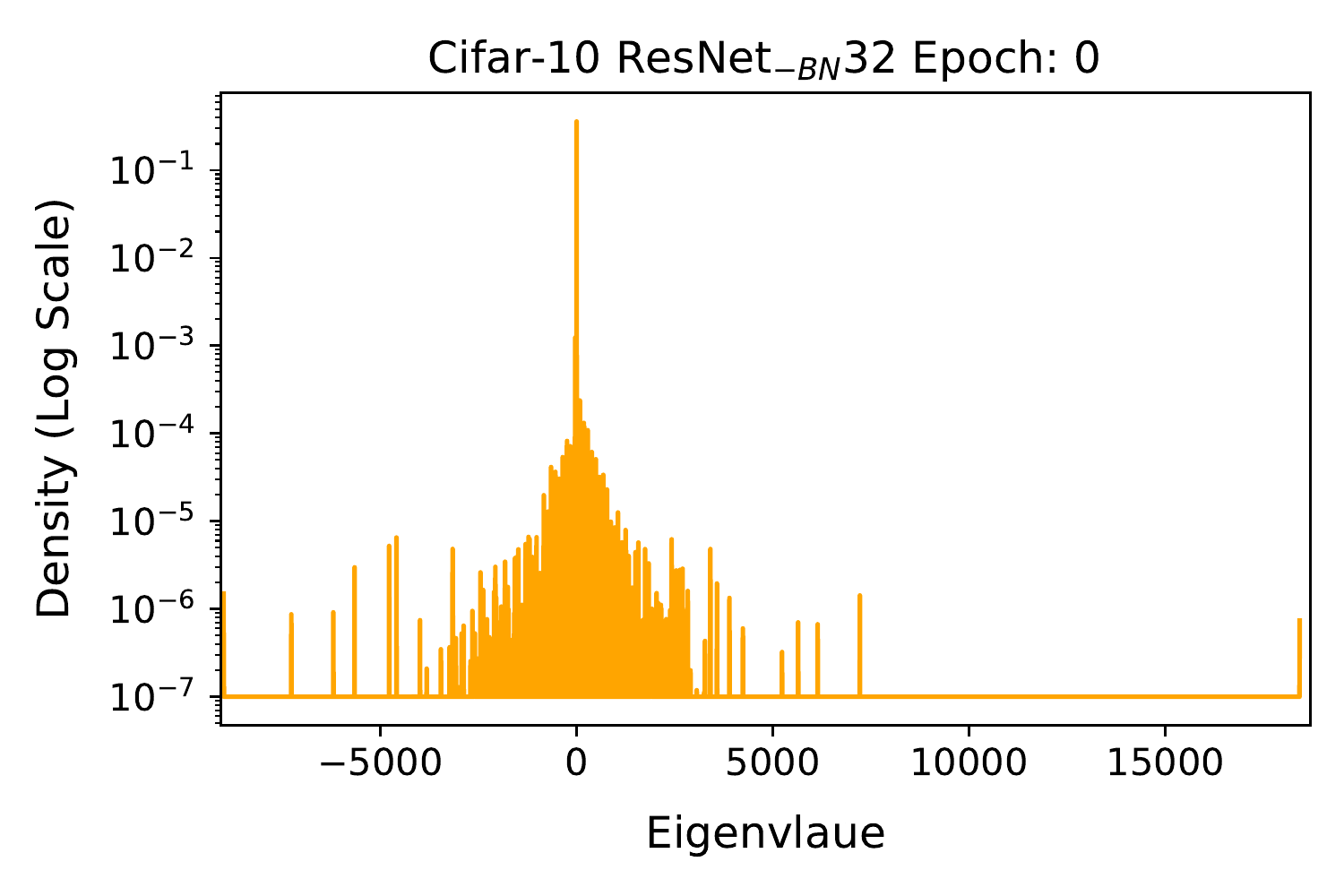}
\includegraphics[width=0.295\textwidth]{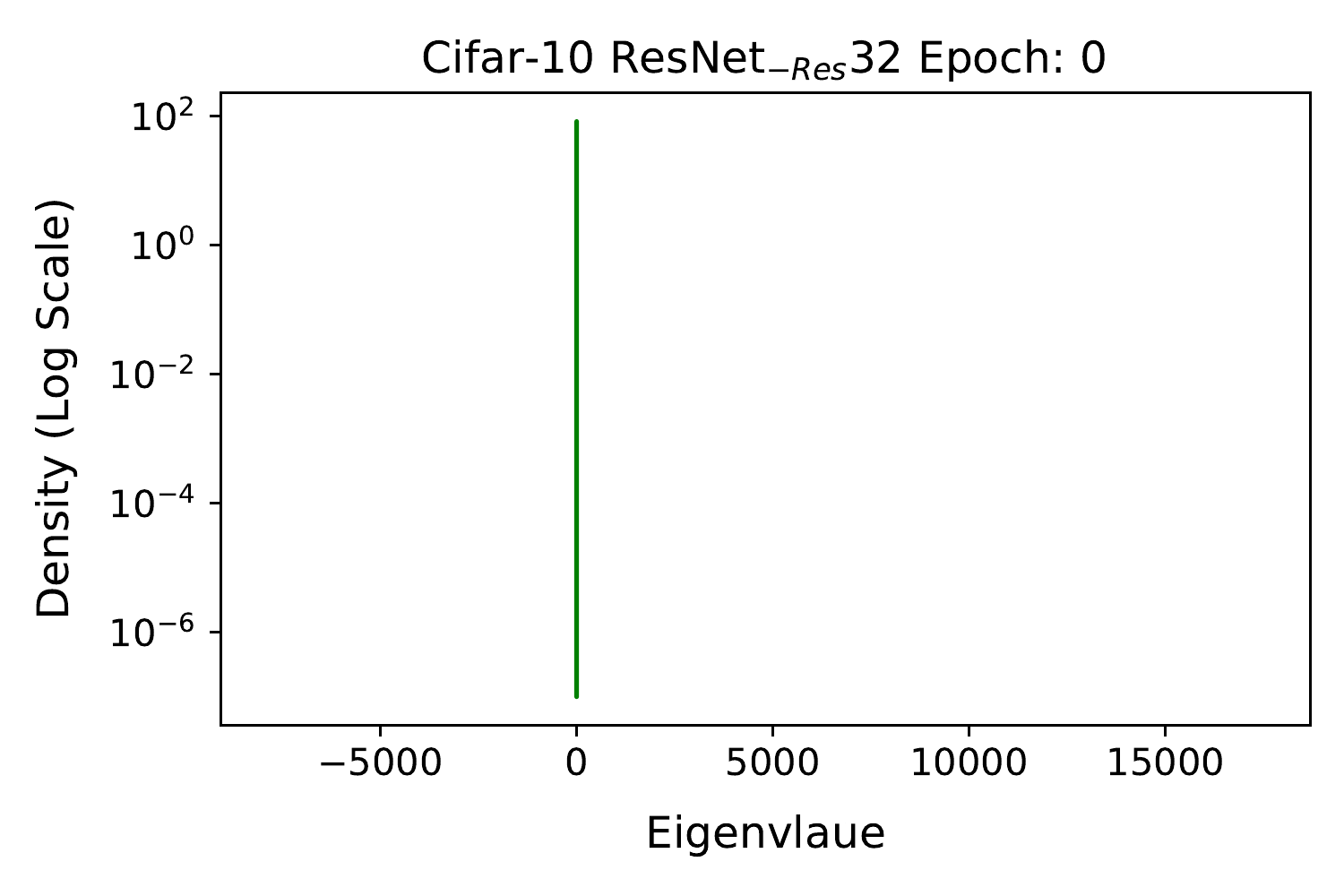}\\
\includegraphics[width=0.295\textwidth]{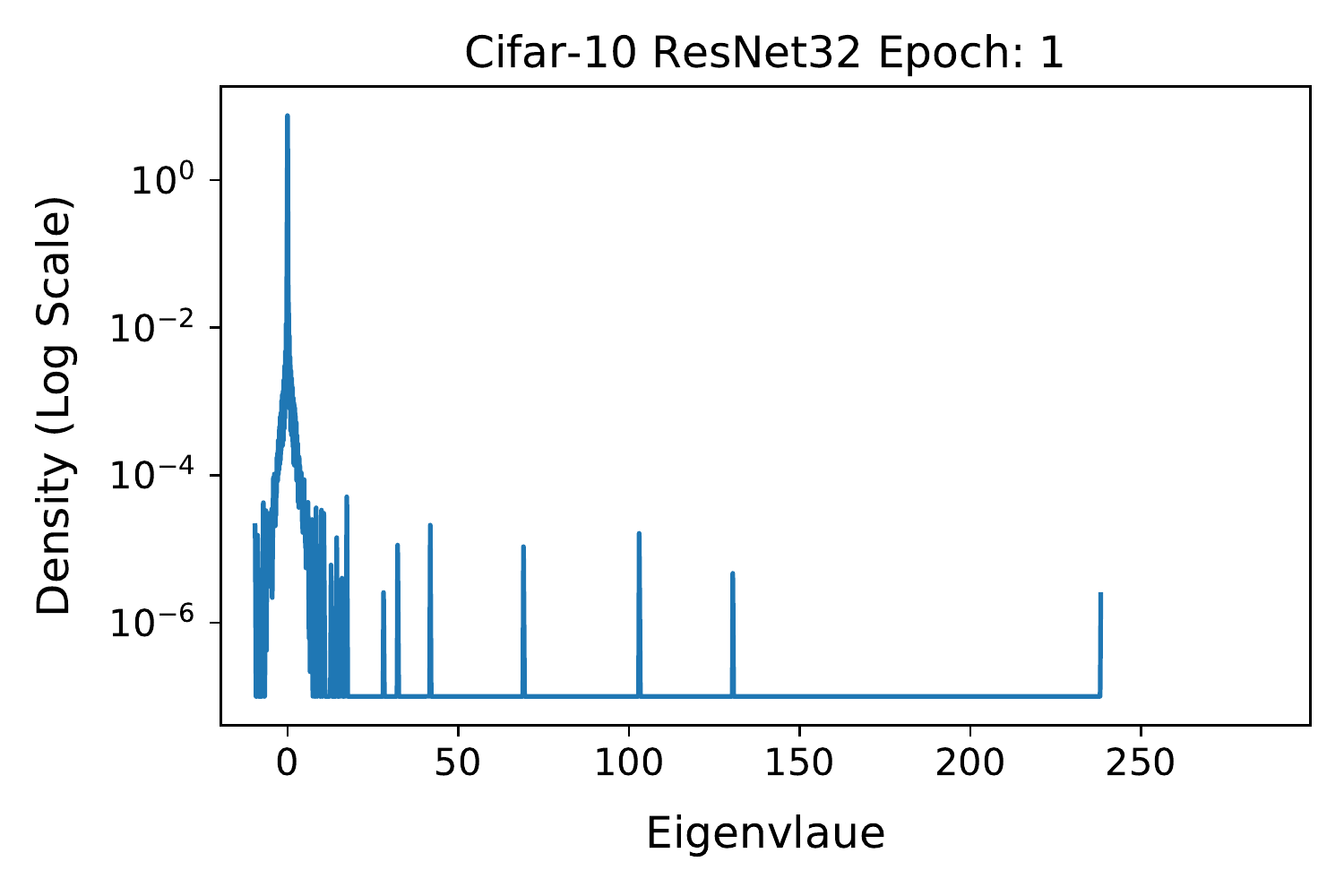}
\includegraphics[width=0.295\textwidth]{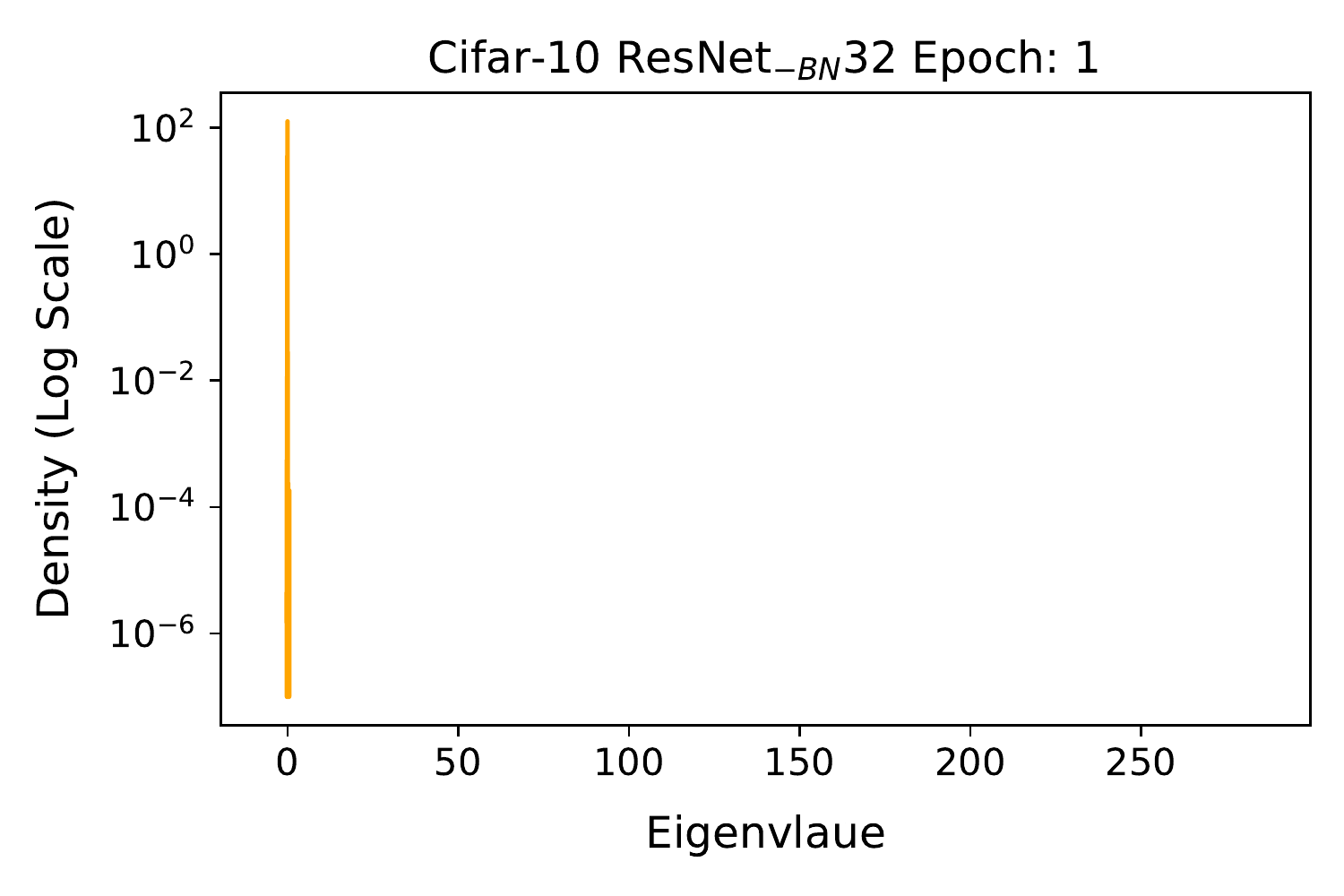}
\includegraphics[width=0.295\textwidth]{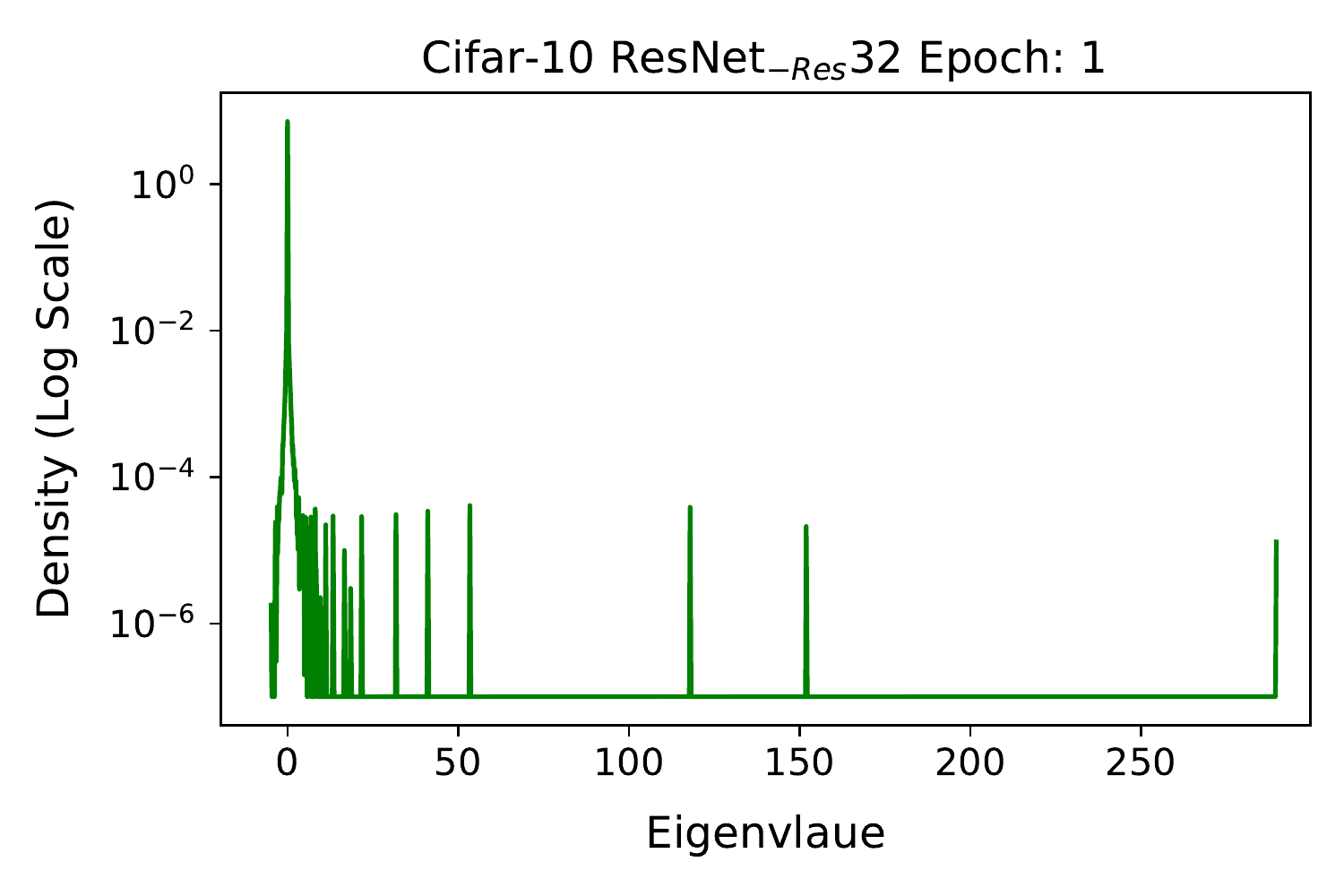}\\
\includegraphics[width=0.295\textwidth]{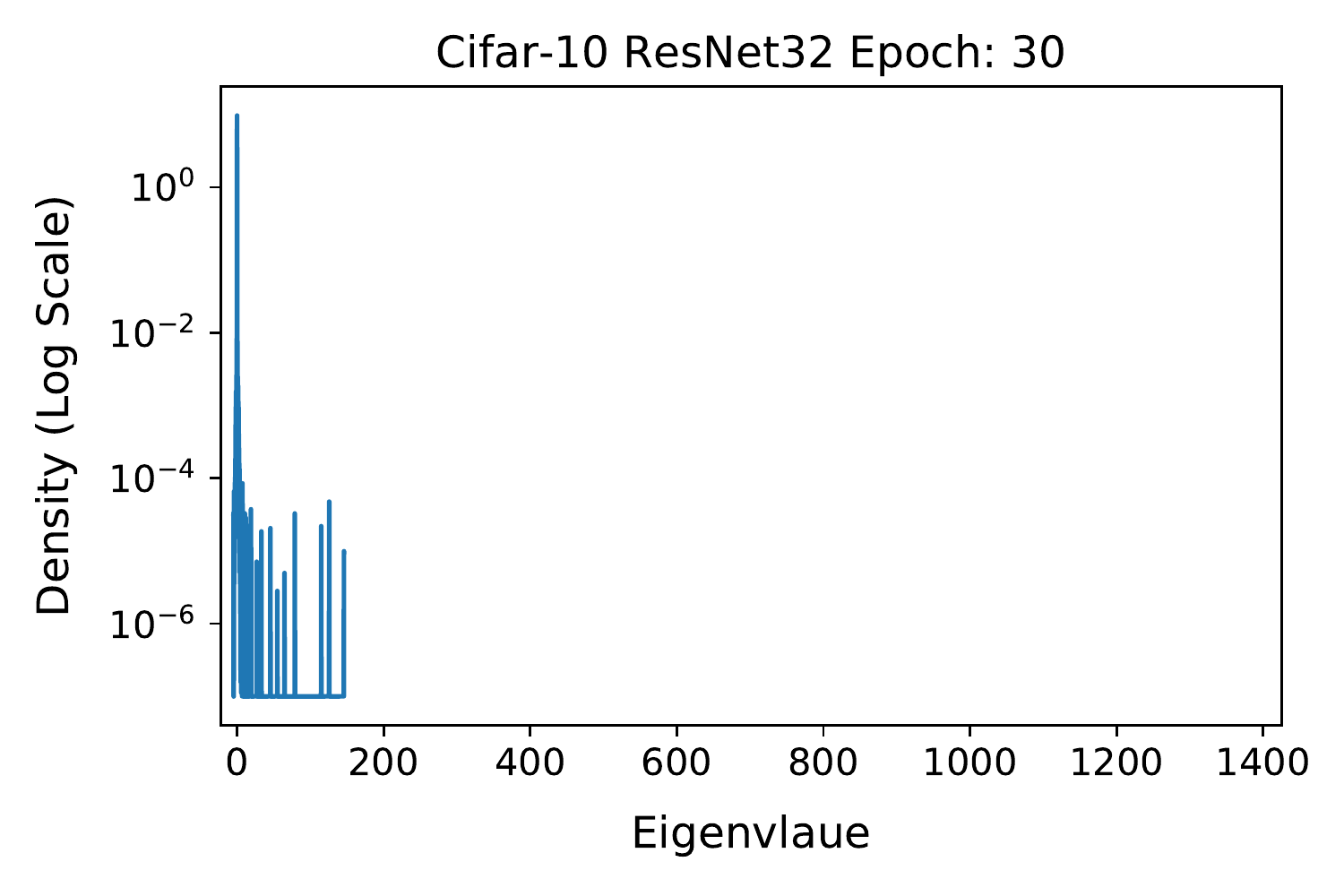}
\includegraphics[width=0.295\textwidth]{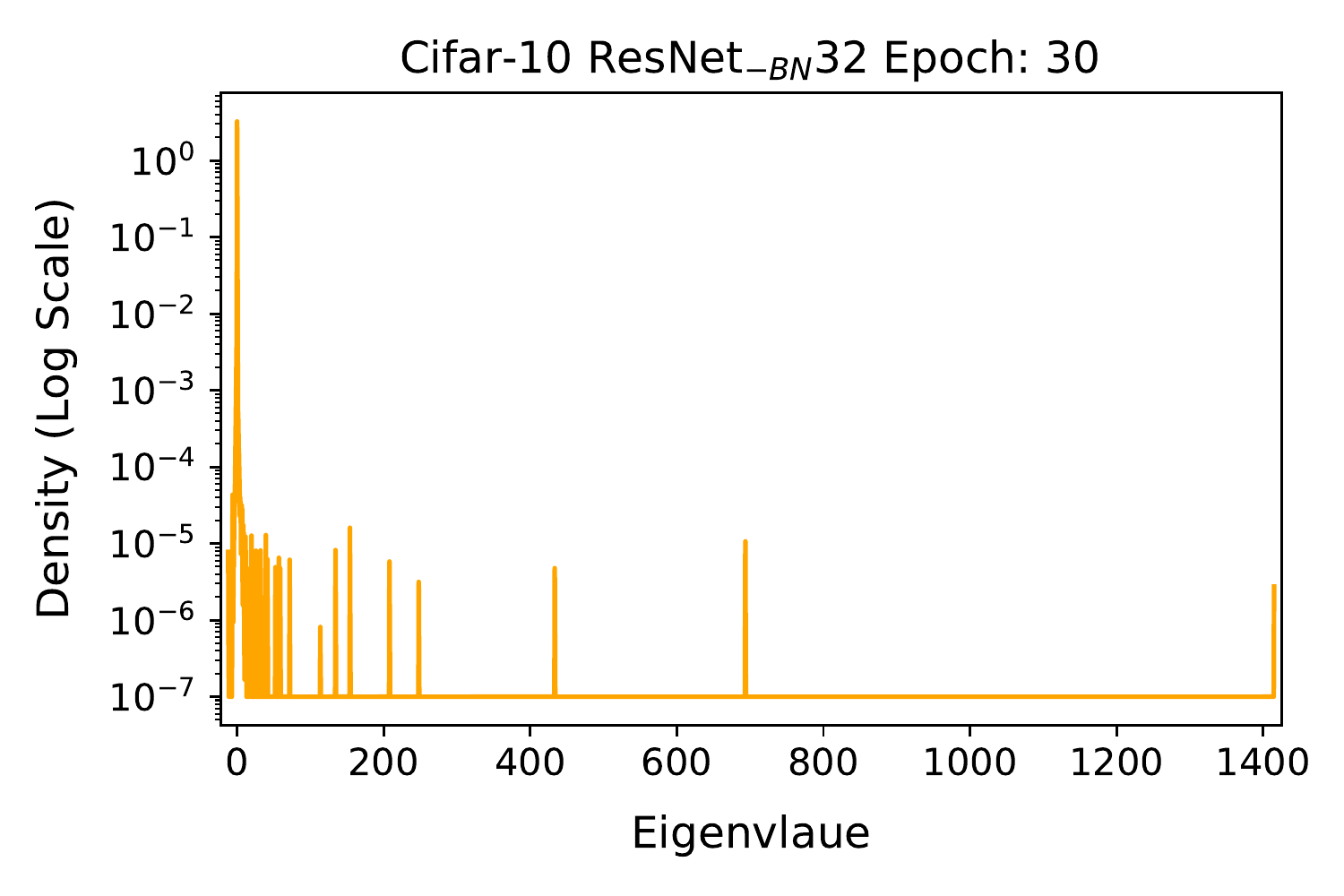}
\includegraphics[width=0.295\textwidth]{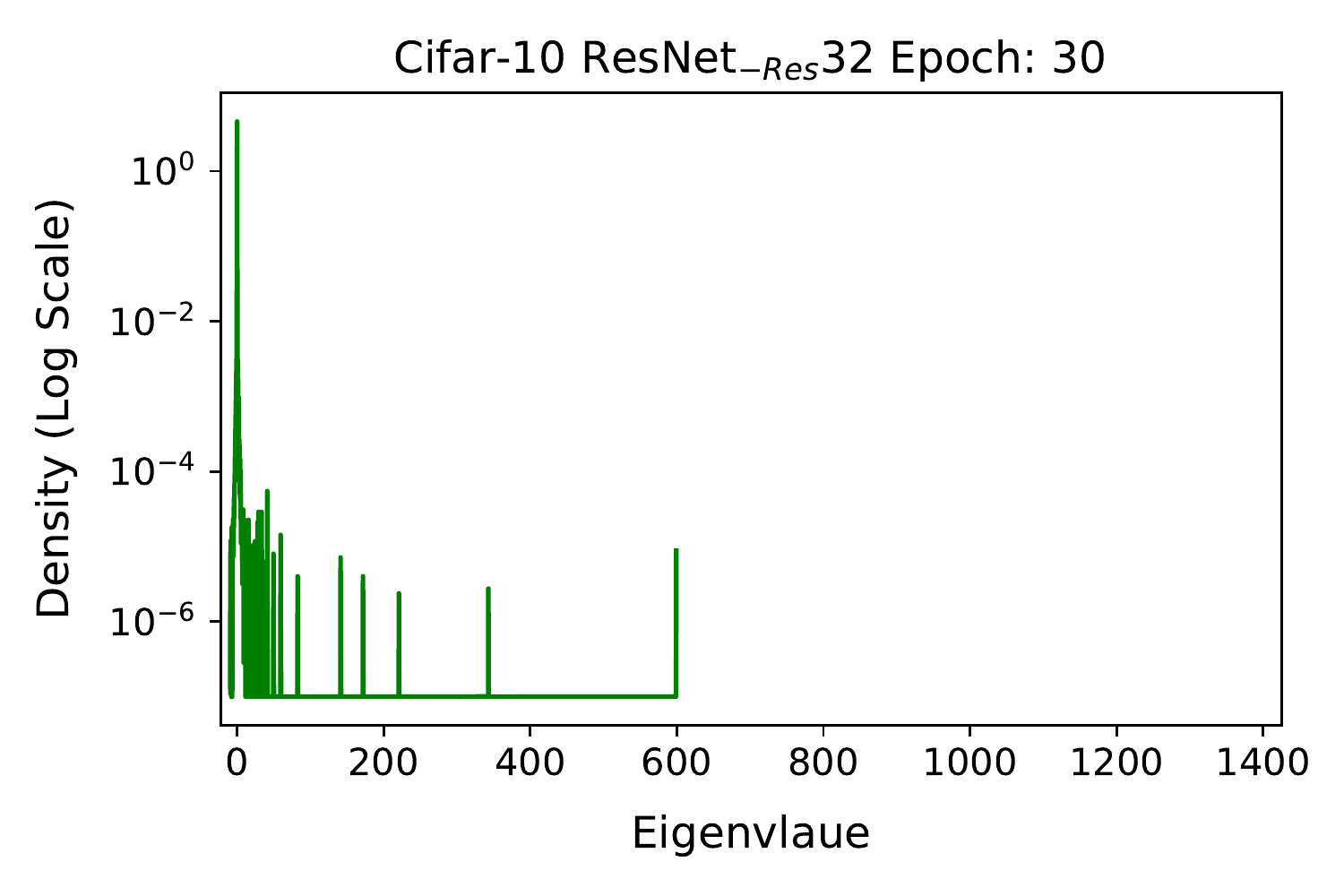}\\
\includegraphics[width=0.295\textwidth]{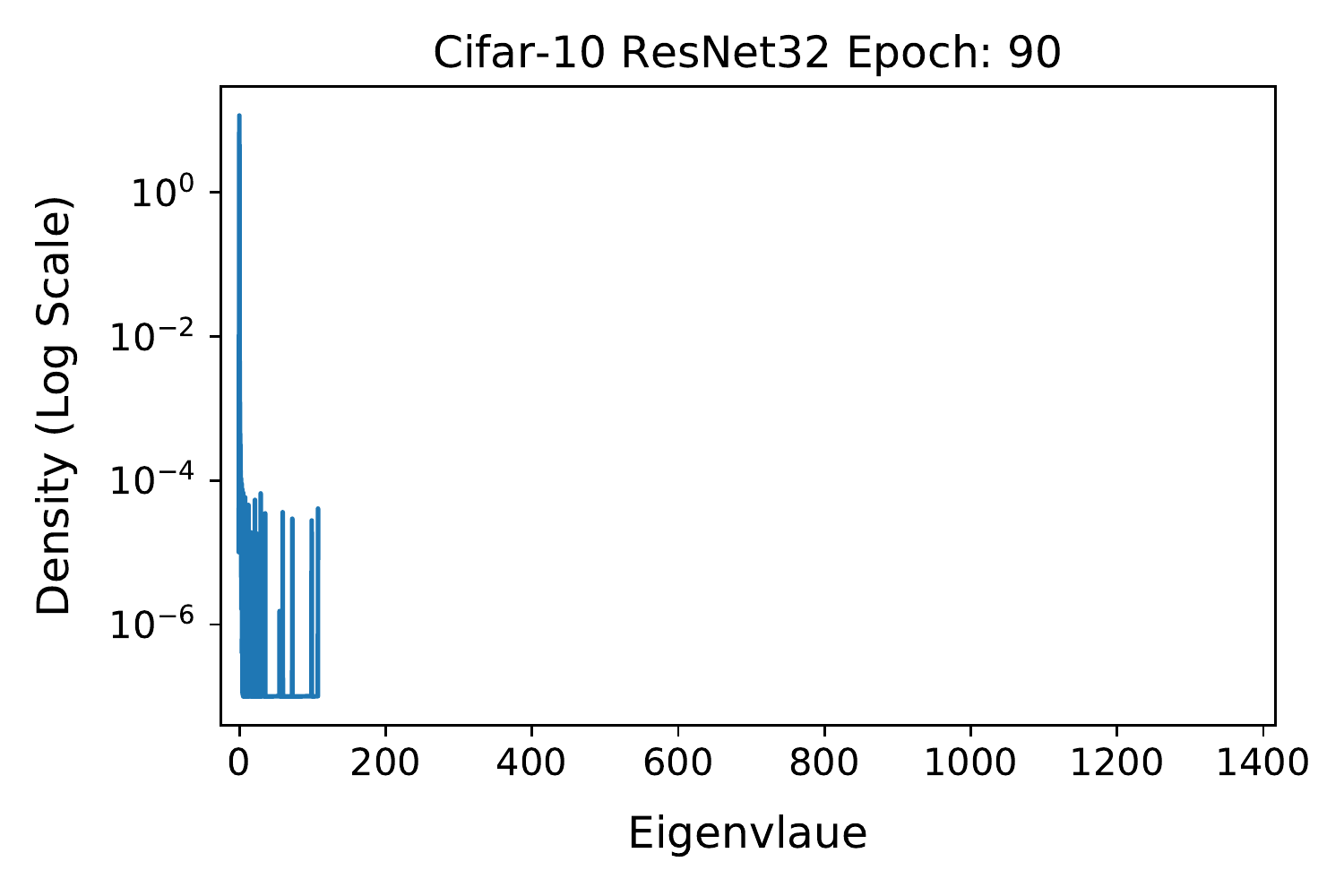}
\includegraphics[width=0.295\textwidth]{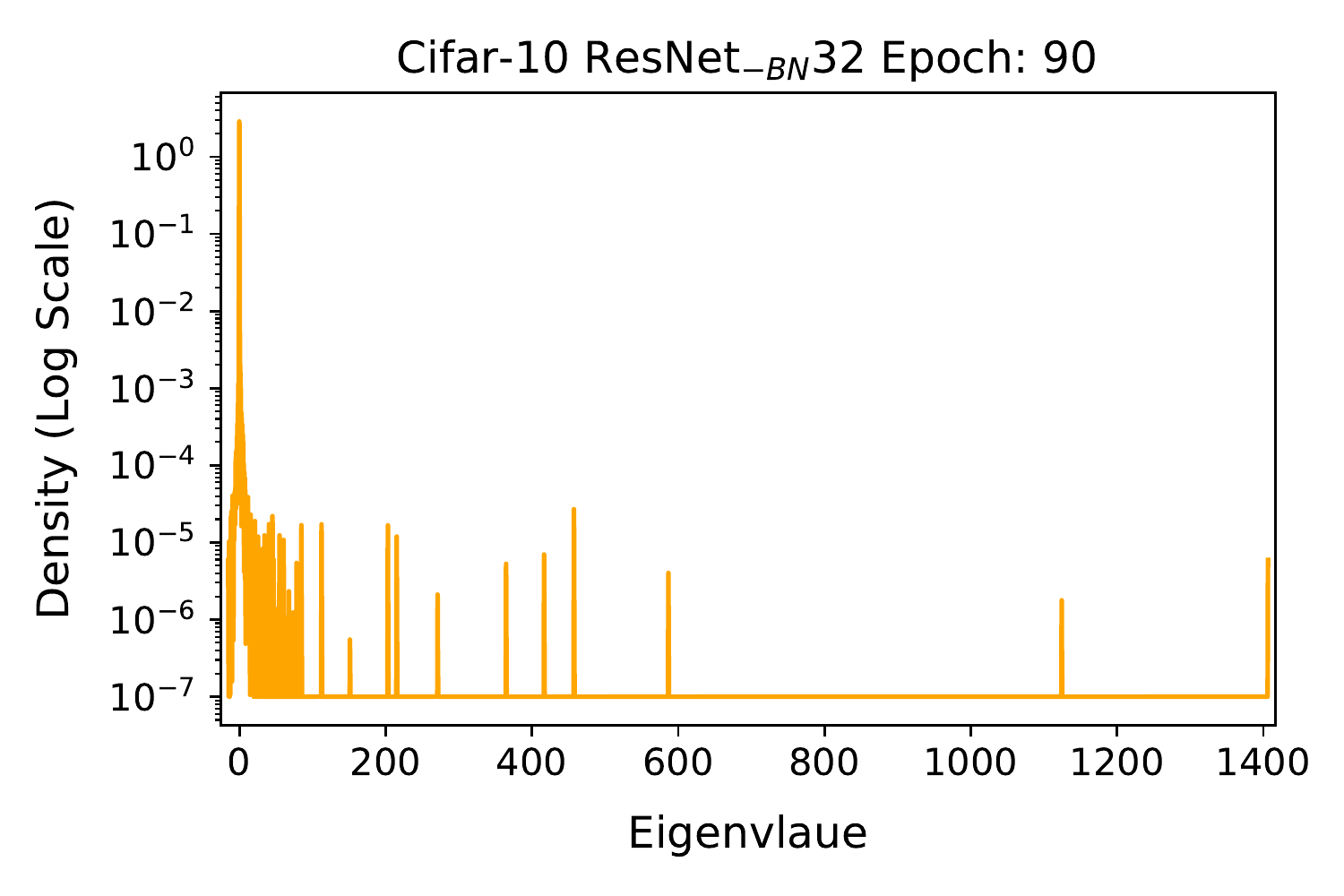}
\includegraphics[width=0.295\textwidth]{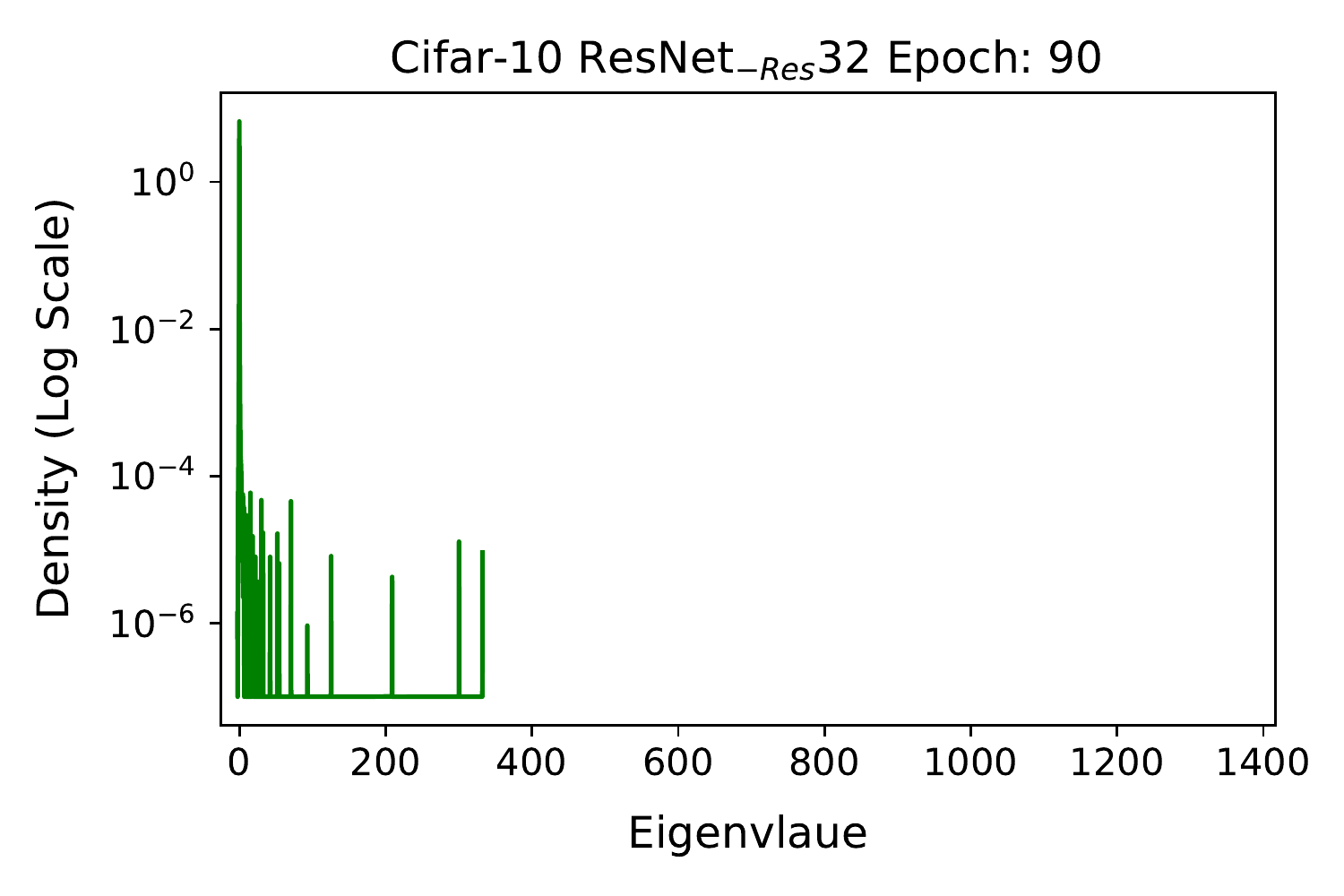}\\
\includegraphics[width=0.295\textwidth]{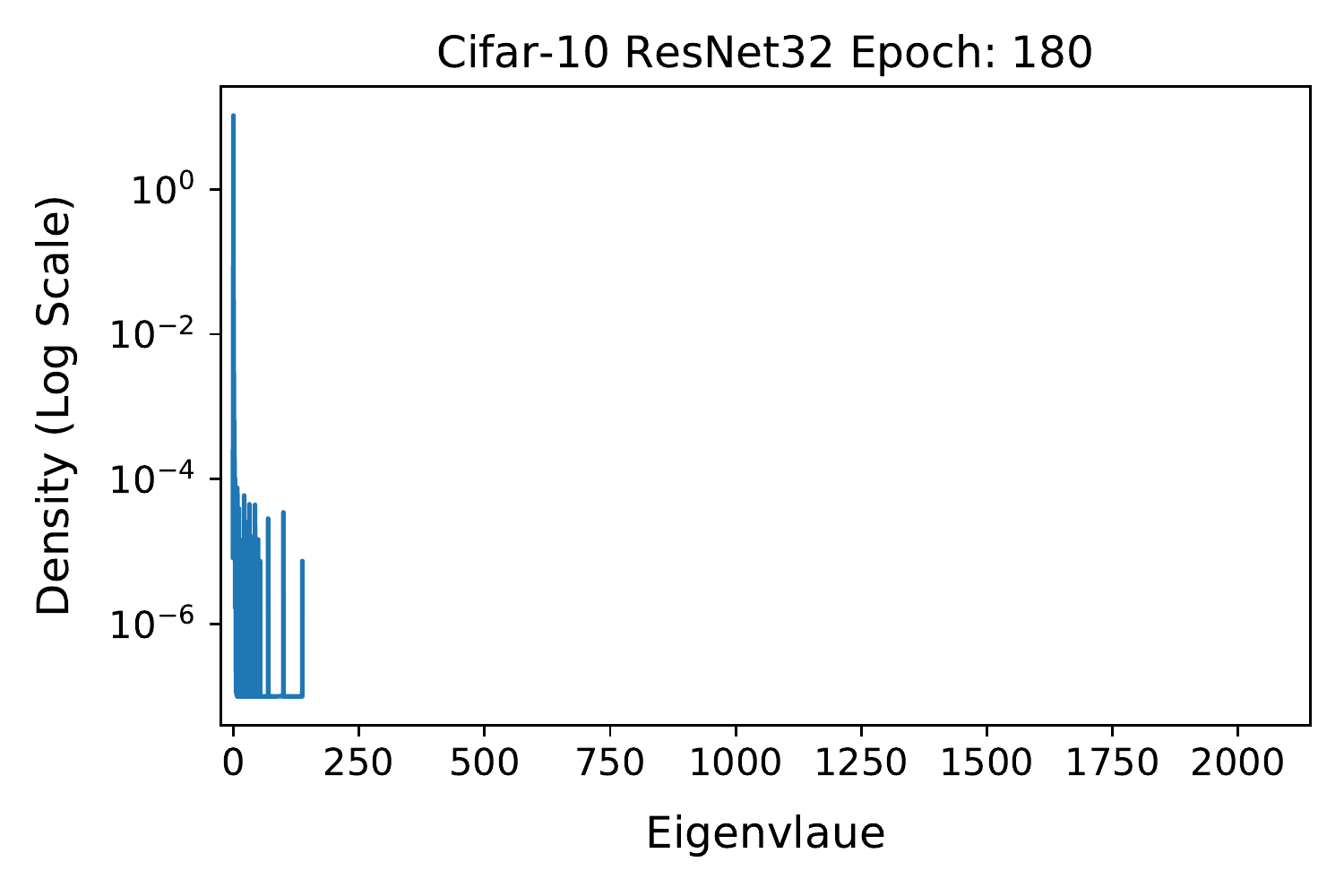}
\includegraphics[width=0.295\textwidth]{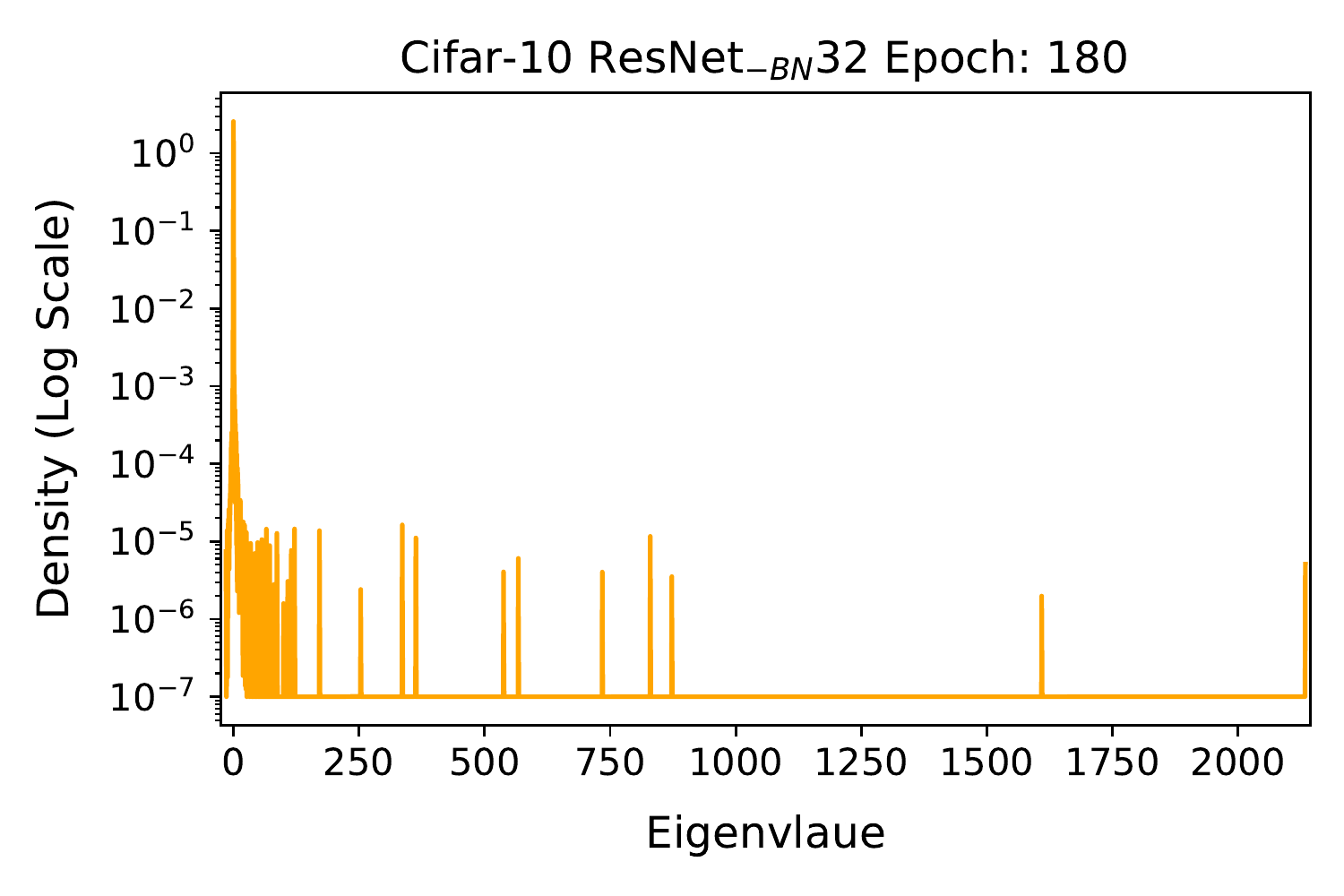}
\includegraphics[width=0.295\textwidth]{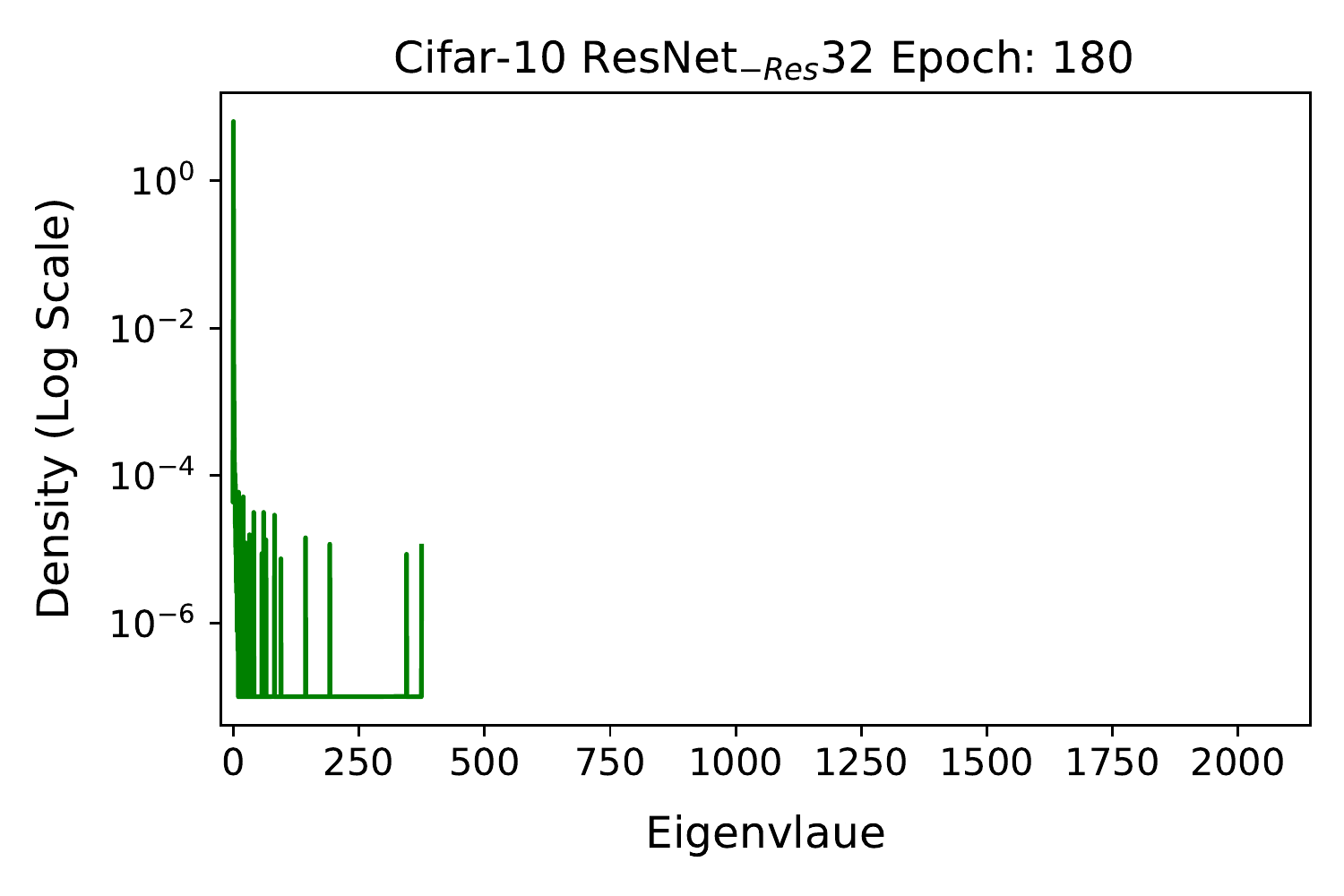}\\
\caption{
Hessian ESD of the entire network for ResNet/\ResNetBN/\ResNetRes with depth 32 on Cifar-10 with Hessian batch size 50000. 
This figure shows the Hessian ESD throughout the training process.
One notable thing here is that the Hessian ESD of \ResNetBN32 centers around zero (at least) until epoch 5. 
This clearly shows that training without BN is indeed harder. 
}
  \label{fig:resnet32-slq-full-net-all}
\end{figure*}

\begin{figure*}[!htbp]
\centering
\includegraphics[width=0.295\textwidth]{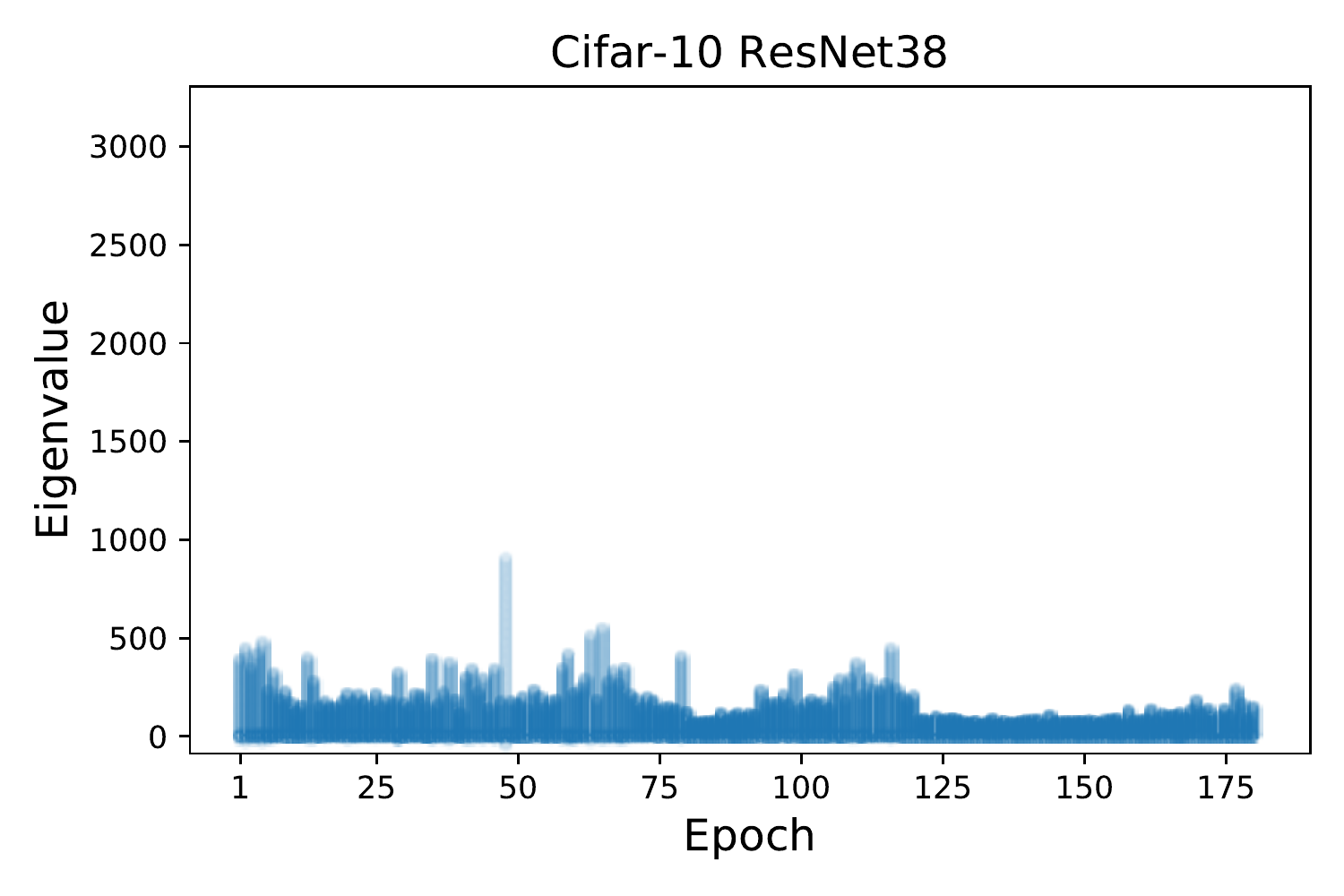}
\includegraphics[width=0.295\textwidth]{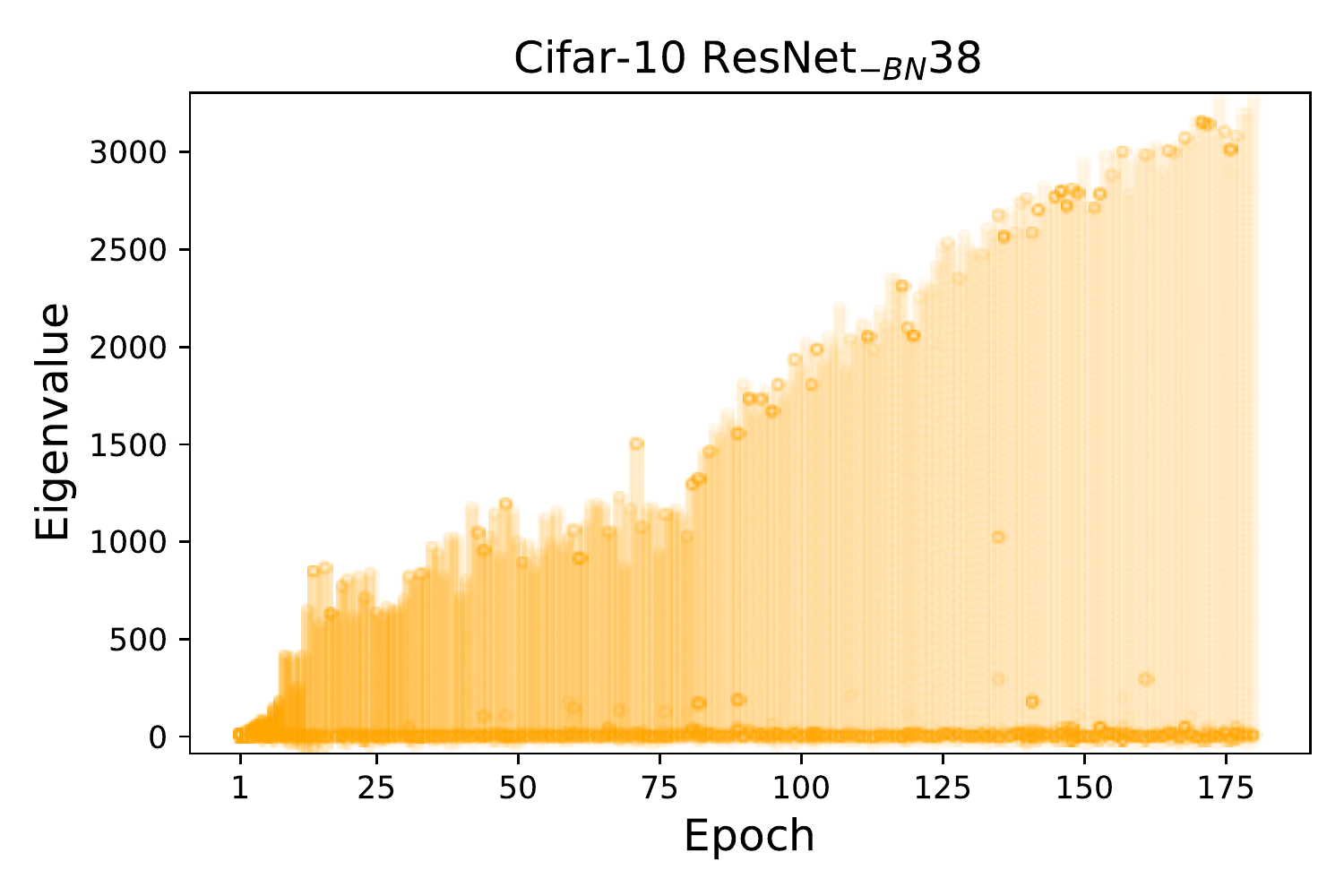}
\includegraphics[width=0.295\textwidth]{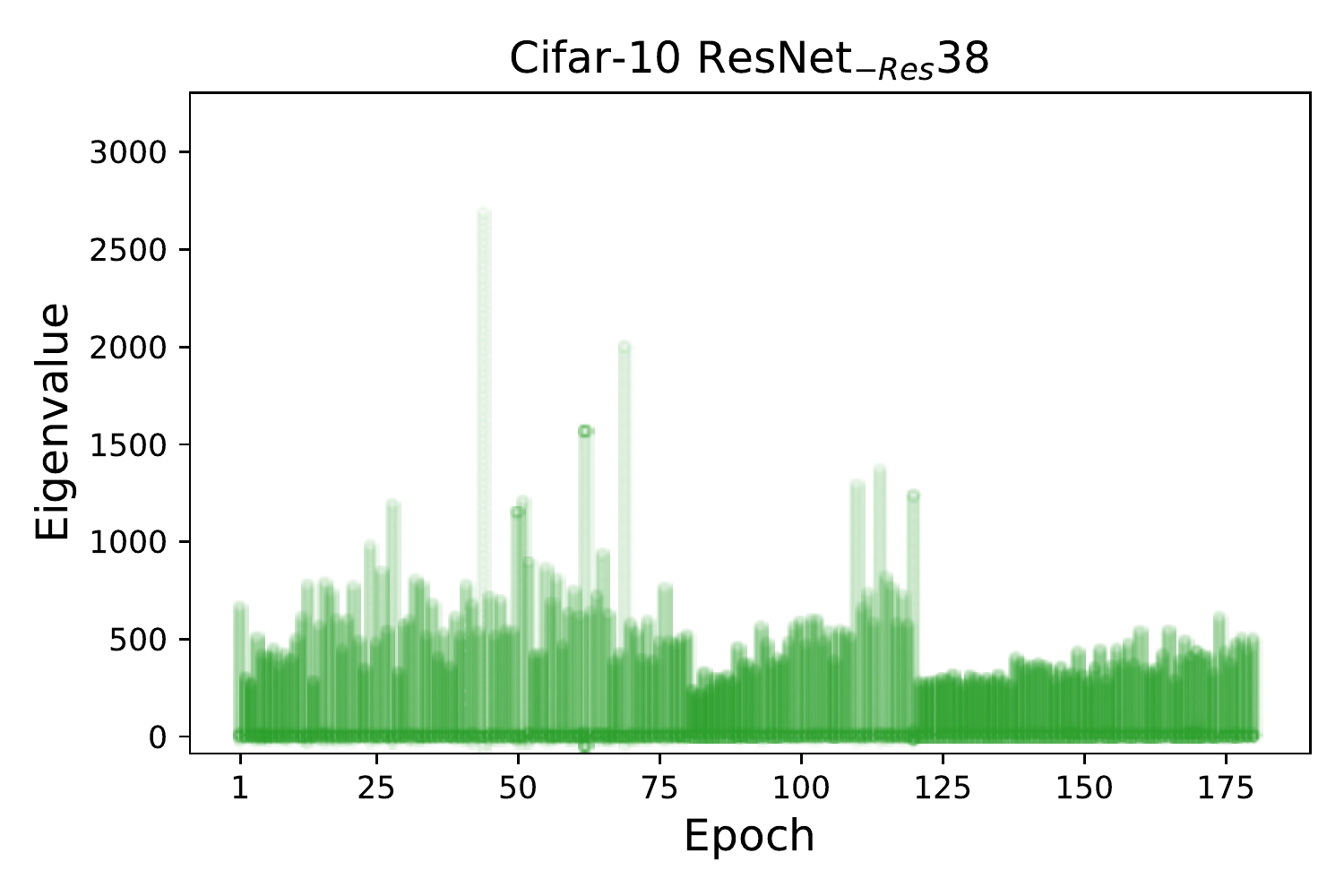}\\
\includegraphics[width=0.295\textwidth]{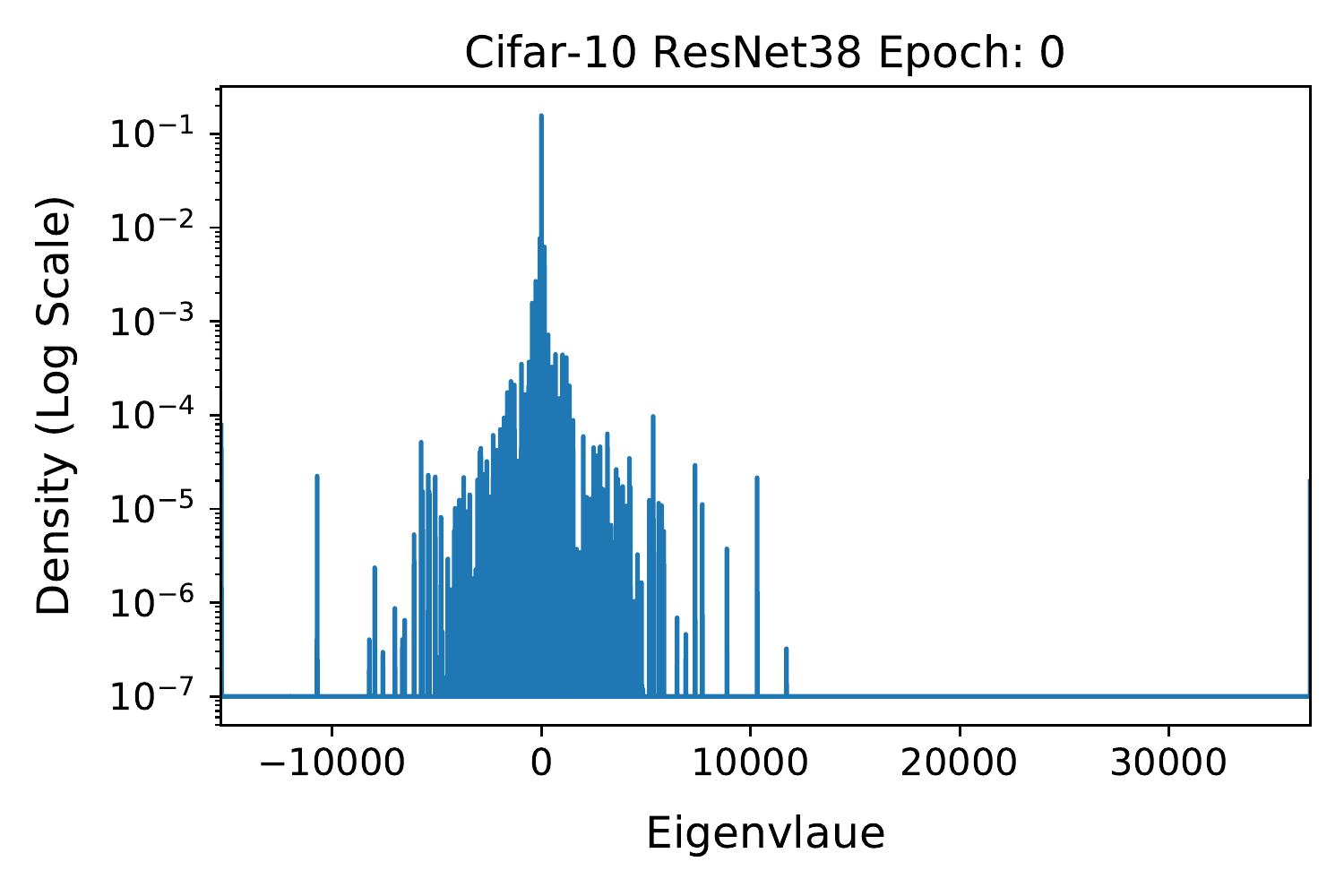}
\includegraphics[width=0.295\textwidth]{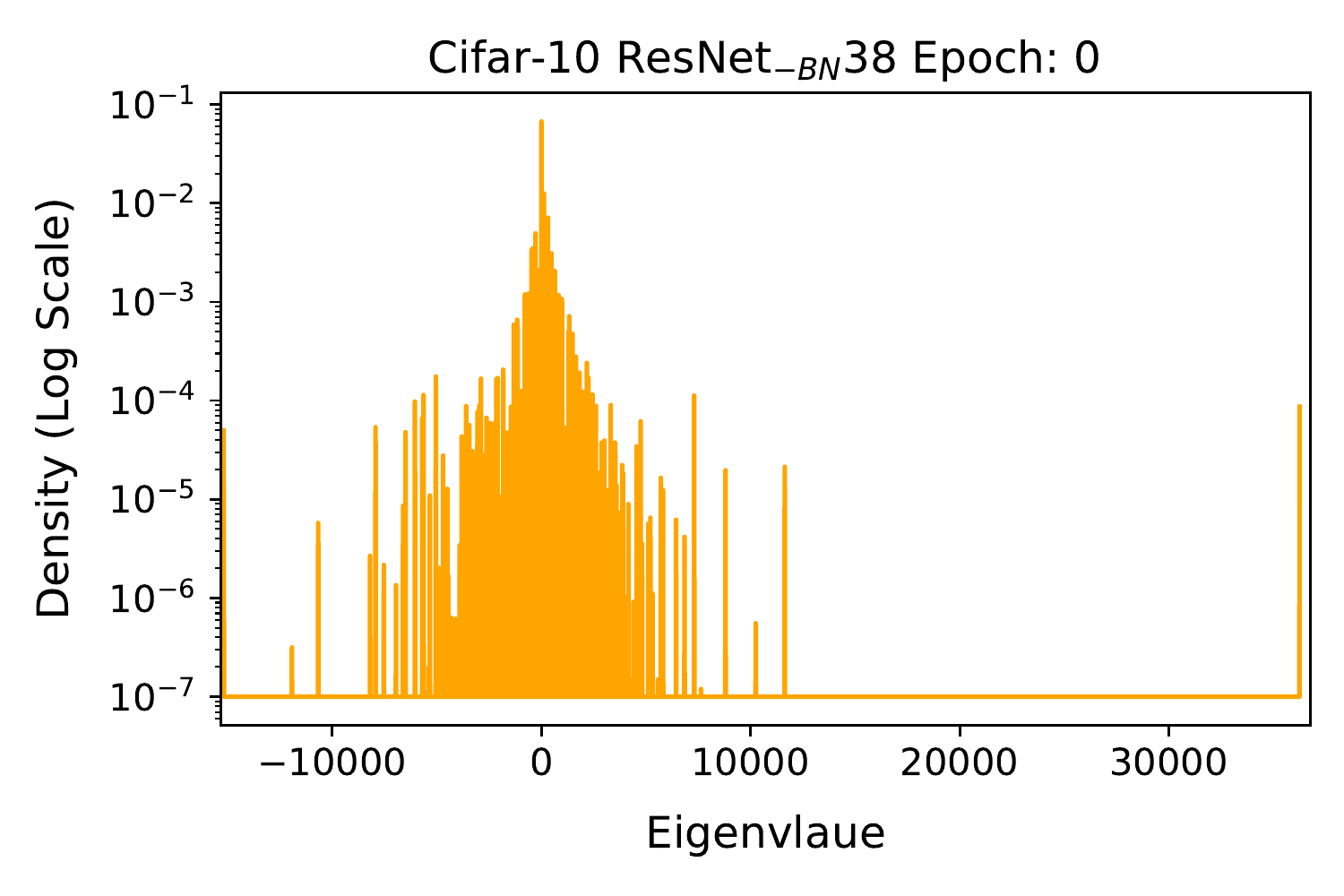}
\includegraphics[width=0.295\textwidth]{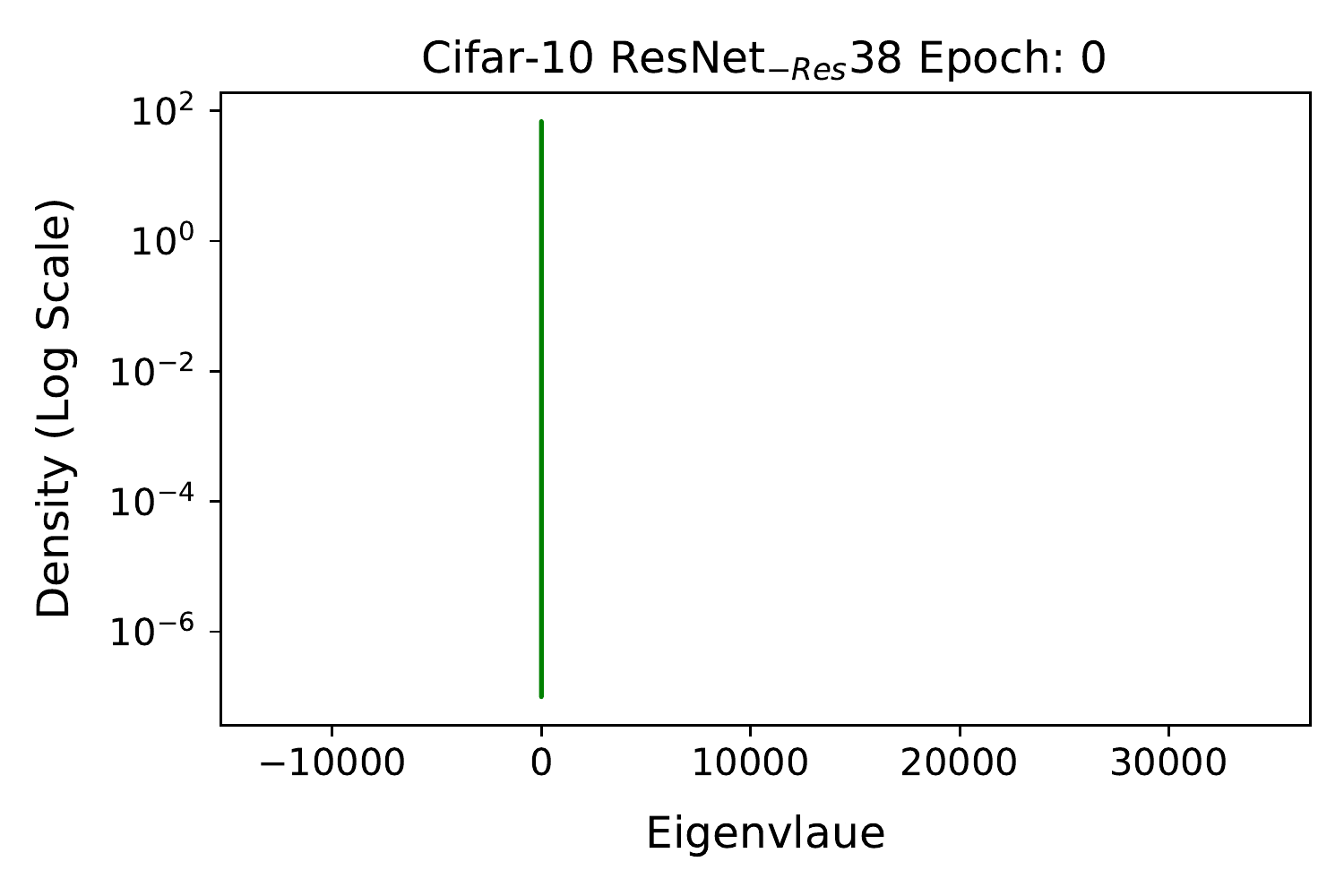}\\
\includegraphics[width=0.295\textwidth]{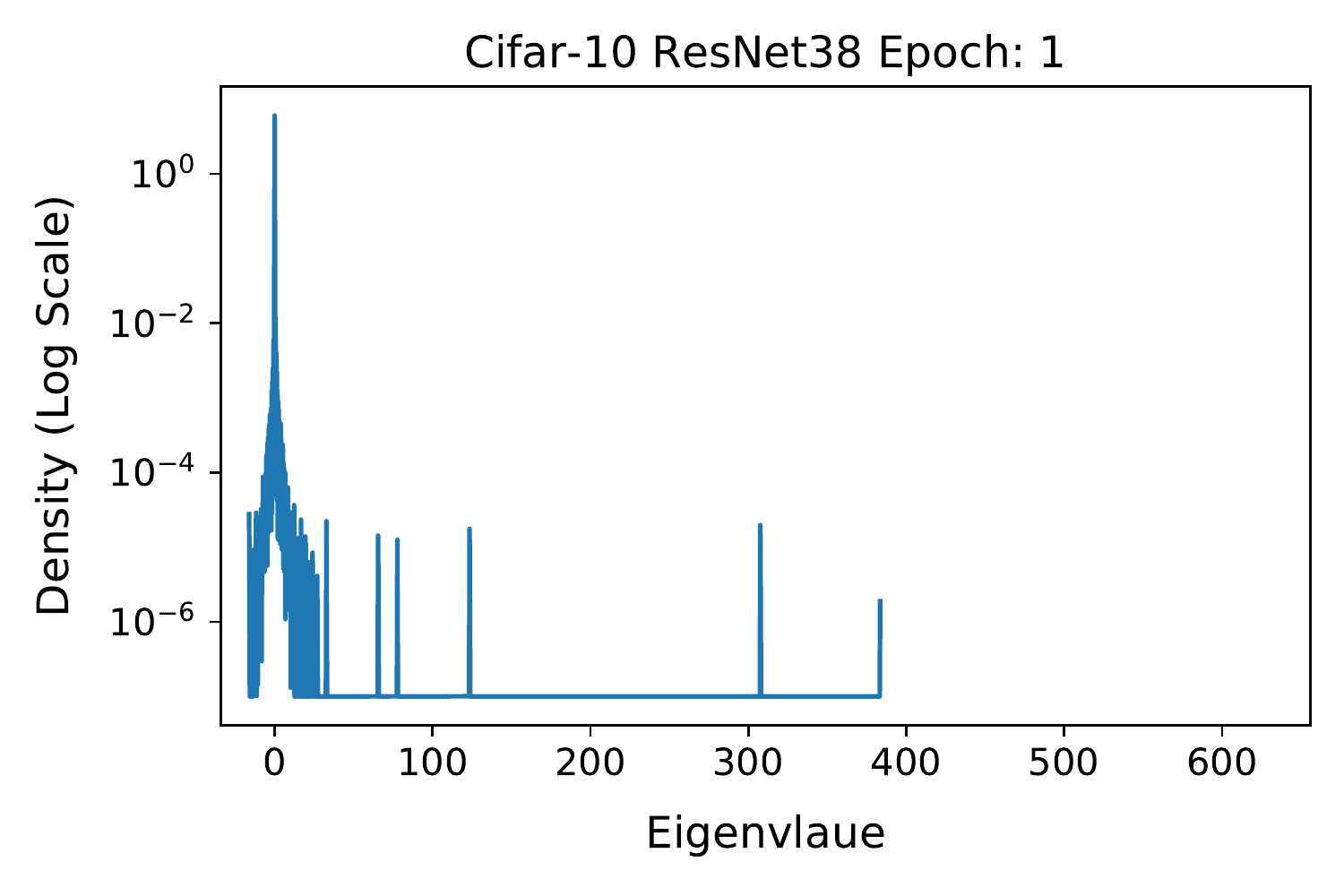}
\includegraphics[width=0.295\textwidth]{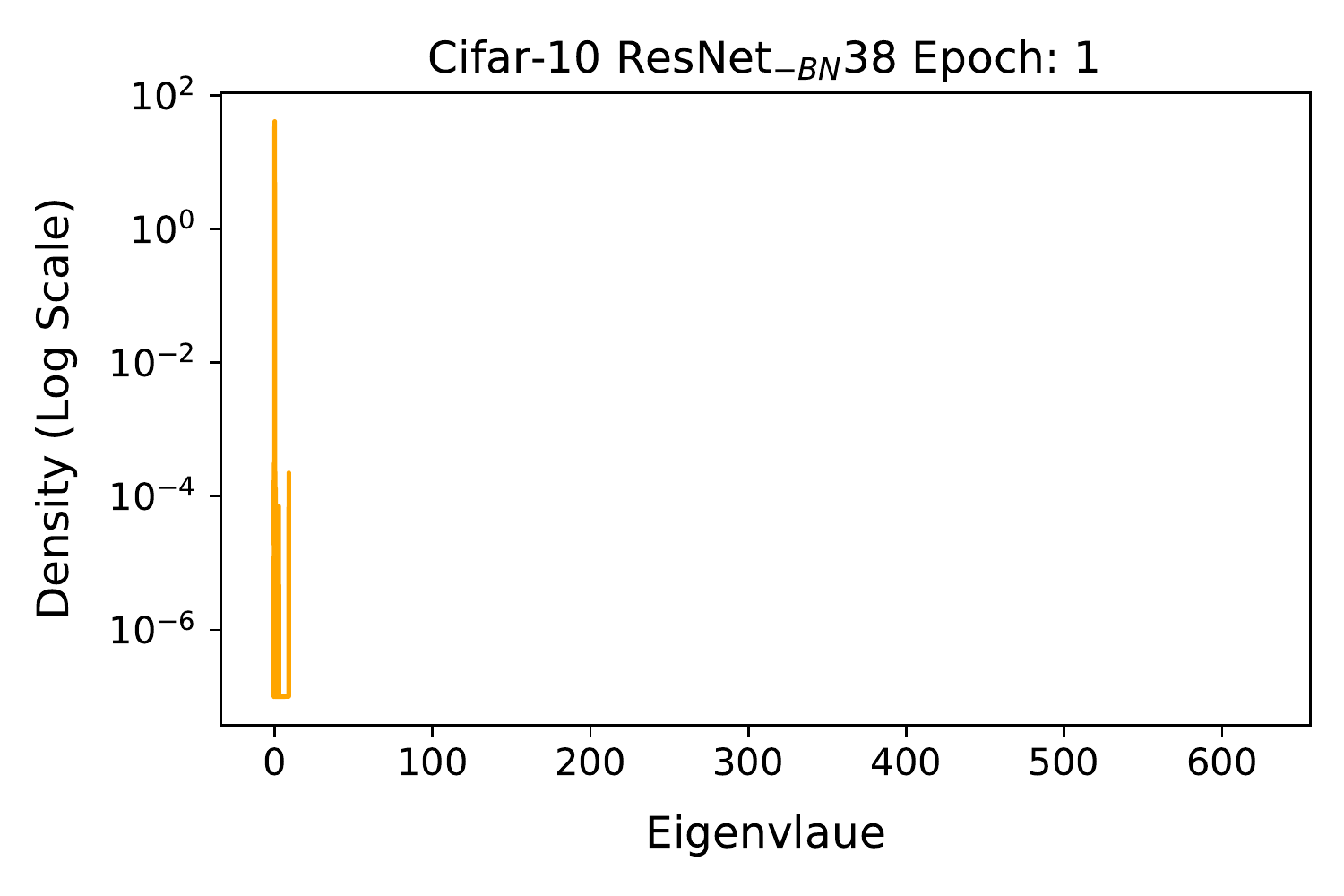}
\includegraphics[width=0.295\textwidth]{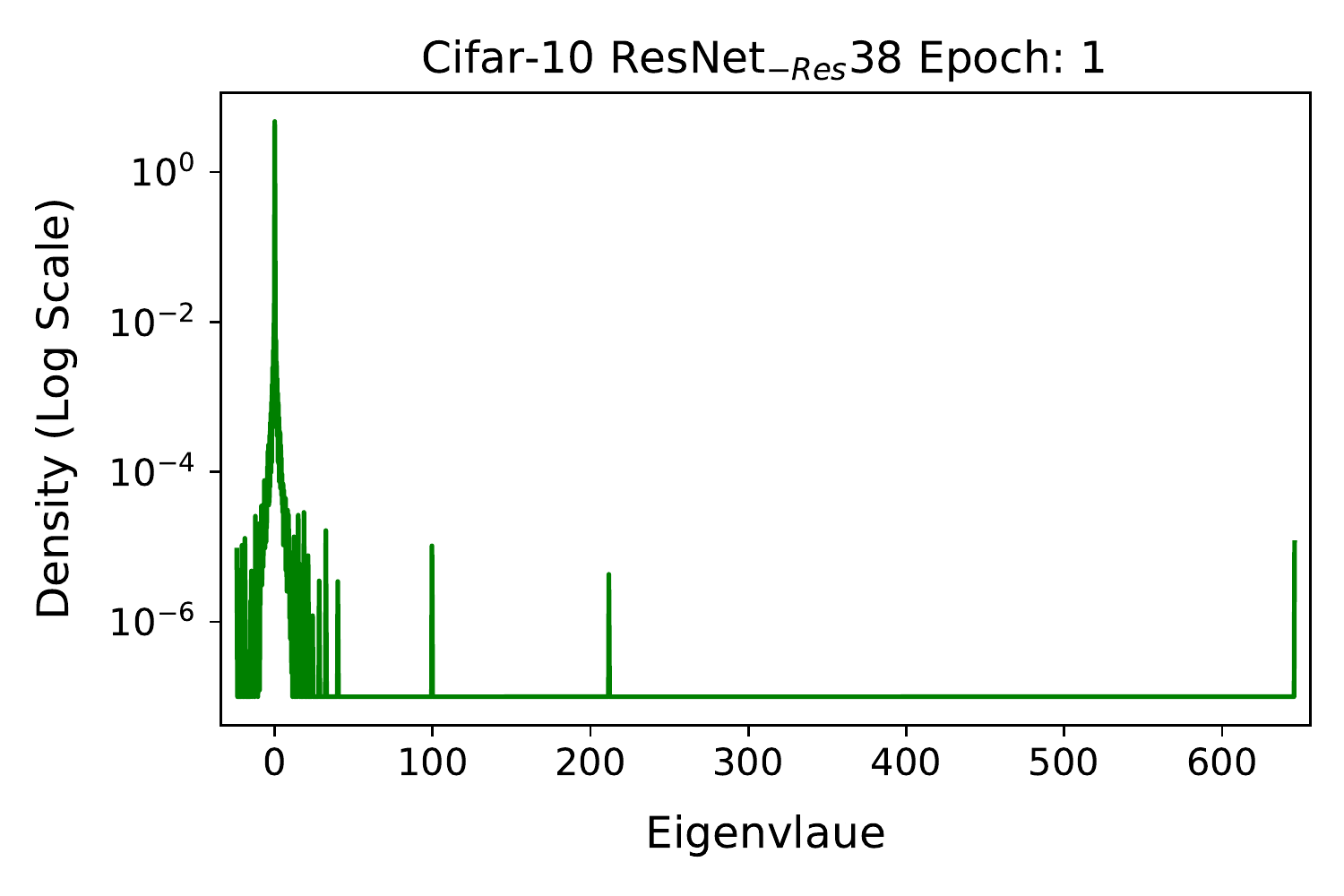}\\
\includegraphics[width=0.295\textwidth]{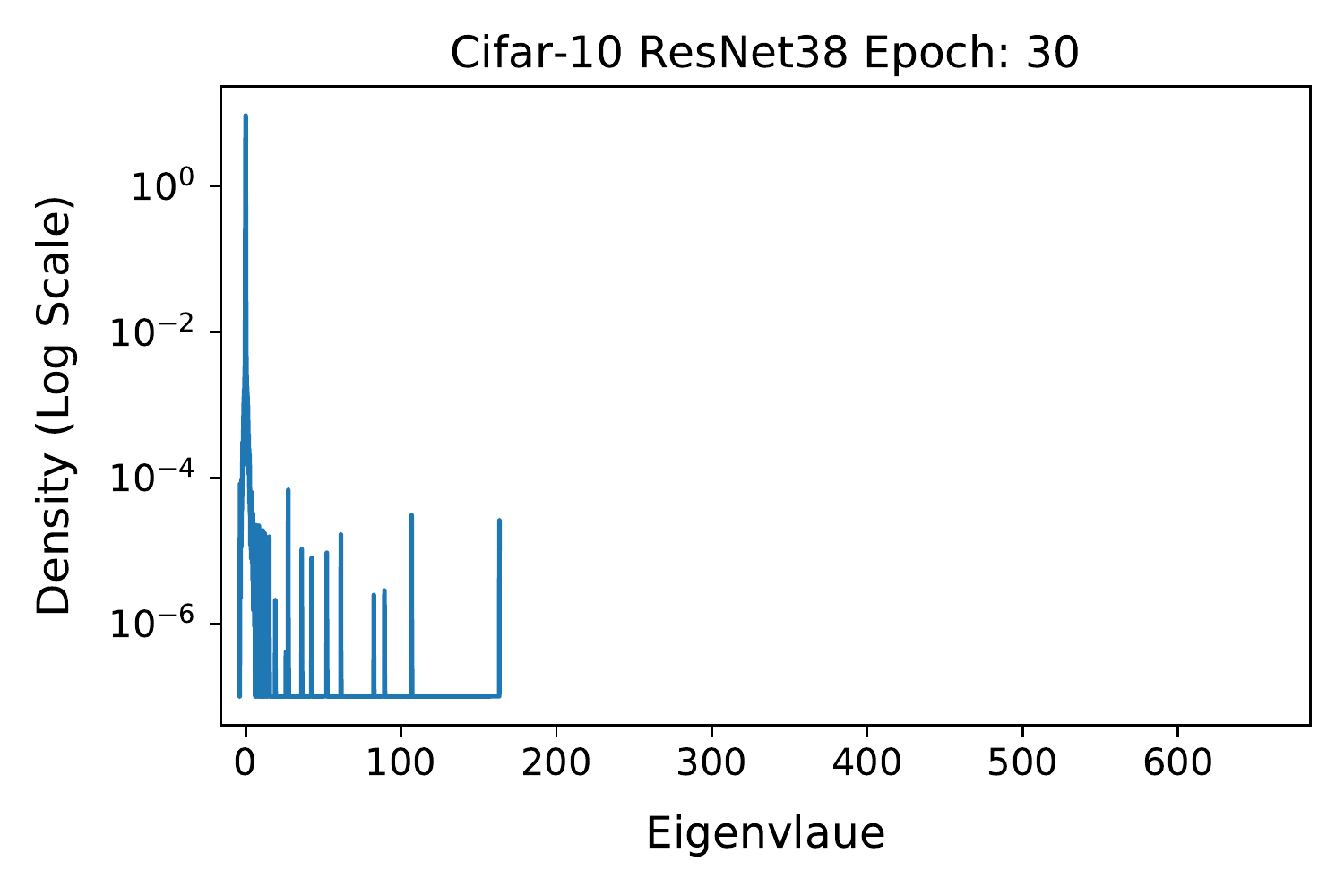}
\includegraphics[width=0.295\textwidth]{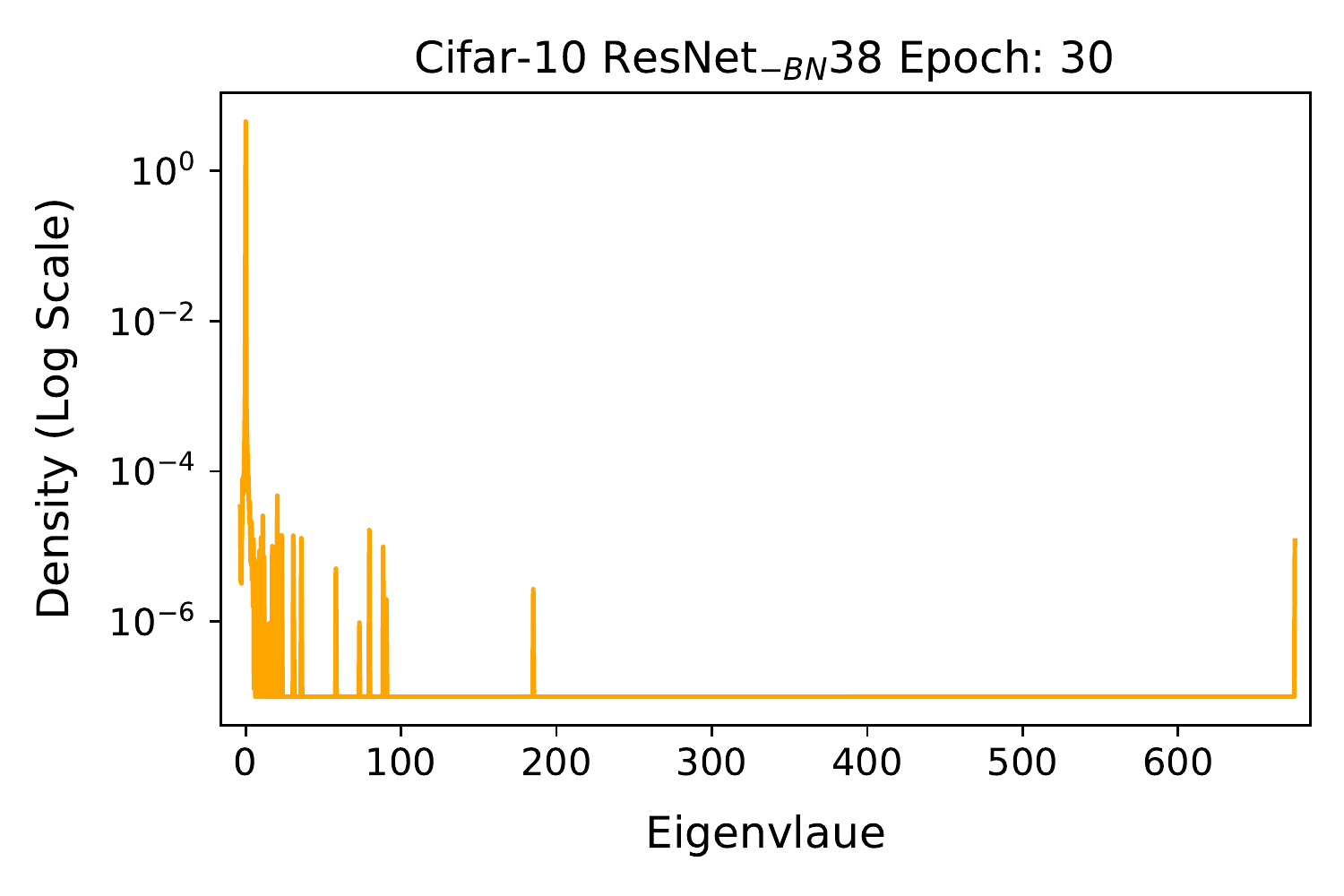}
\includegraphics[width=0.295\textwidth]{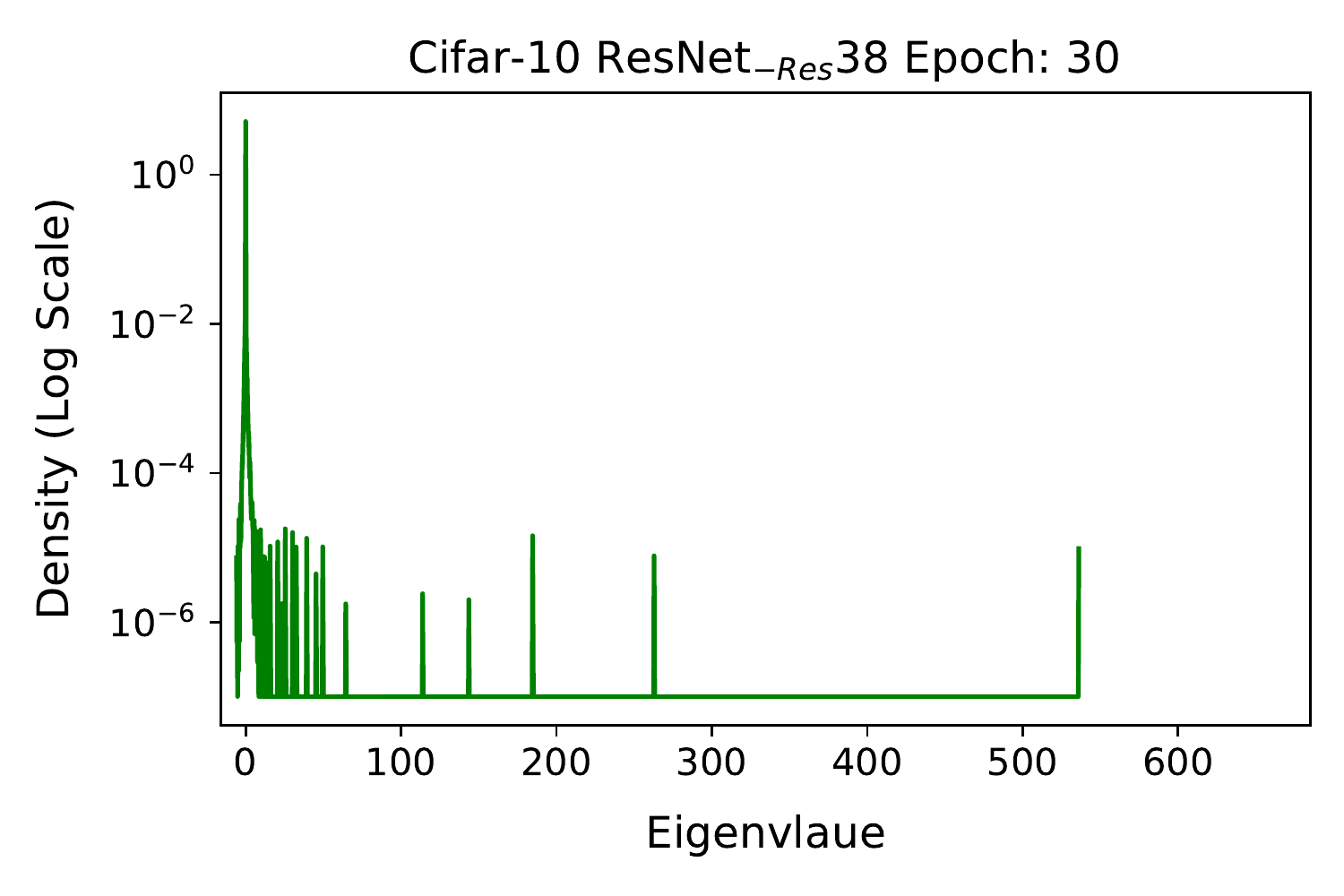}\\
\includegraphics[width=0.295\textwidth]{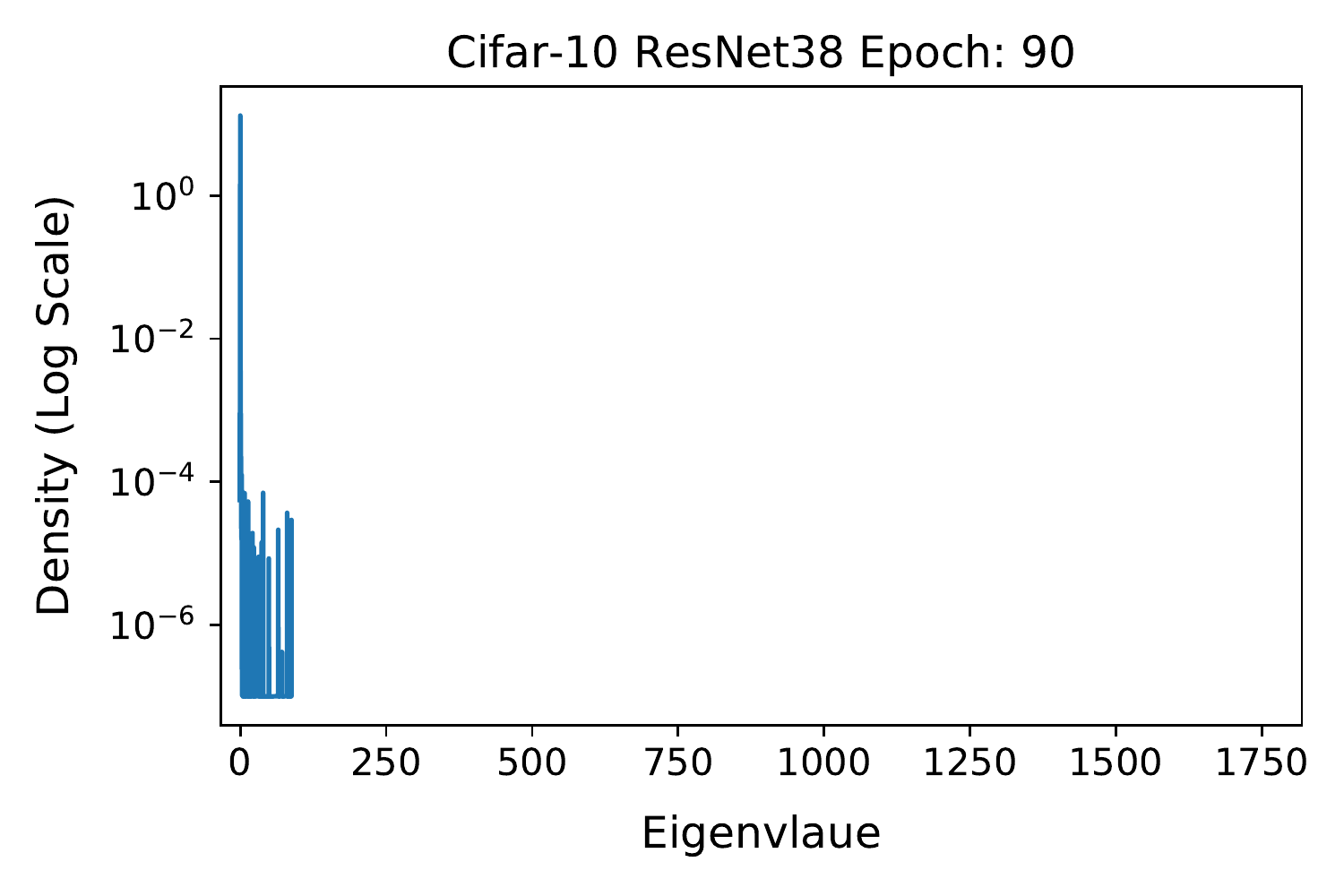}
\includegraphics[width=0.295\textwidth]{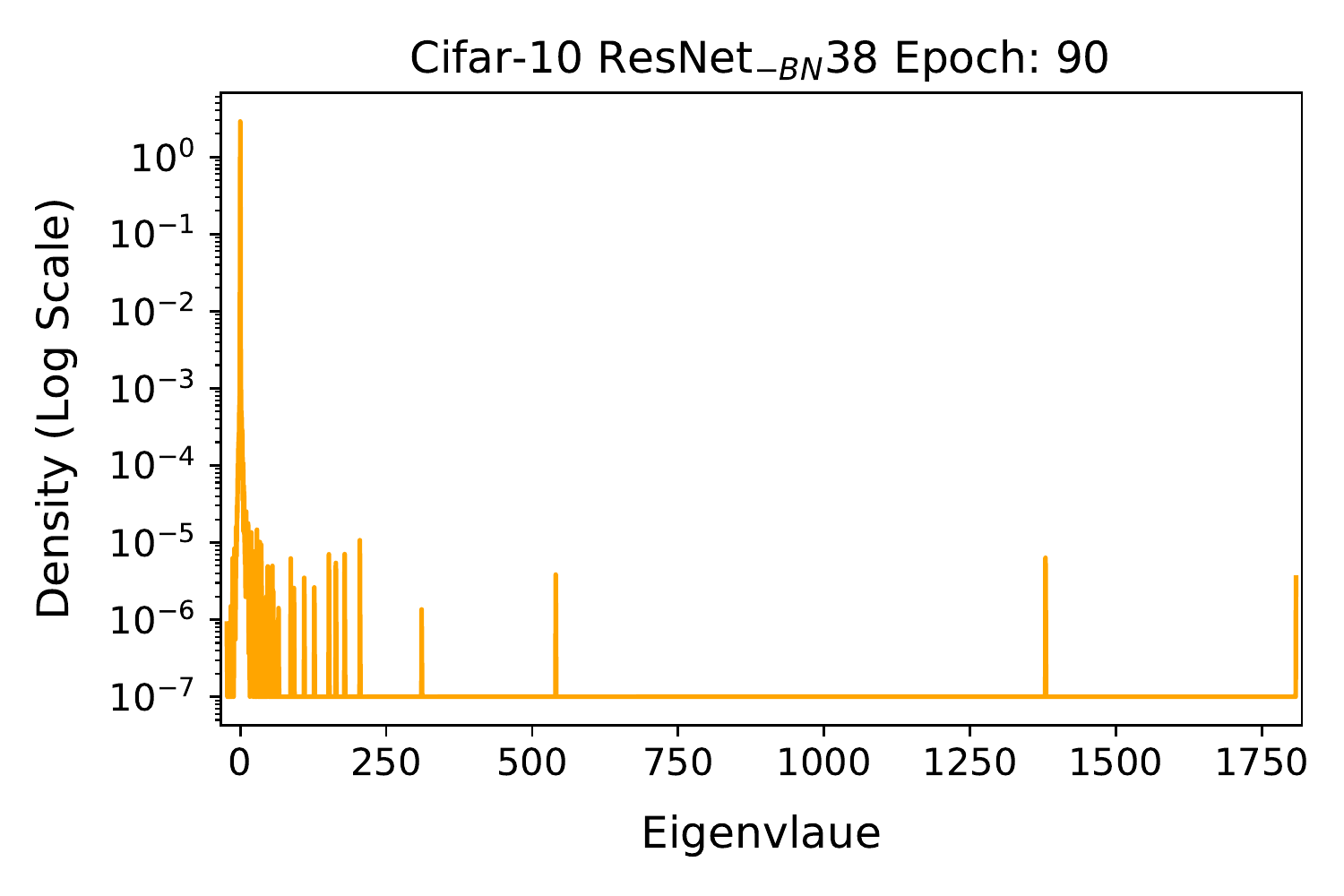}
\includegraphics[width=0.295\textwidth]{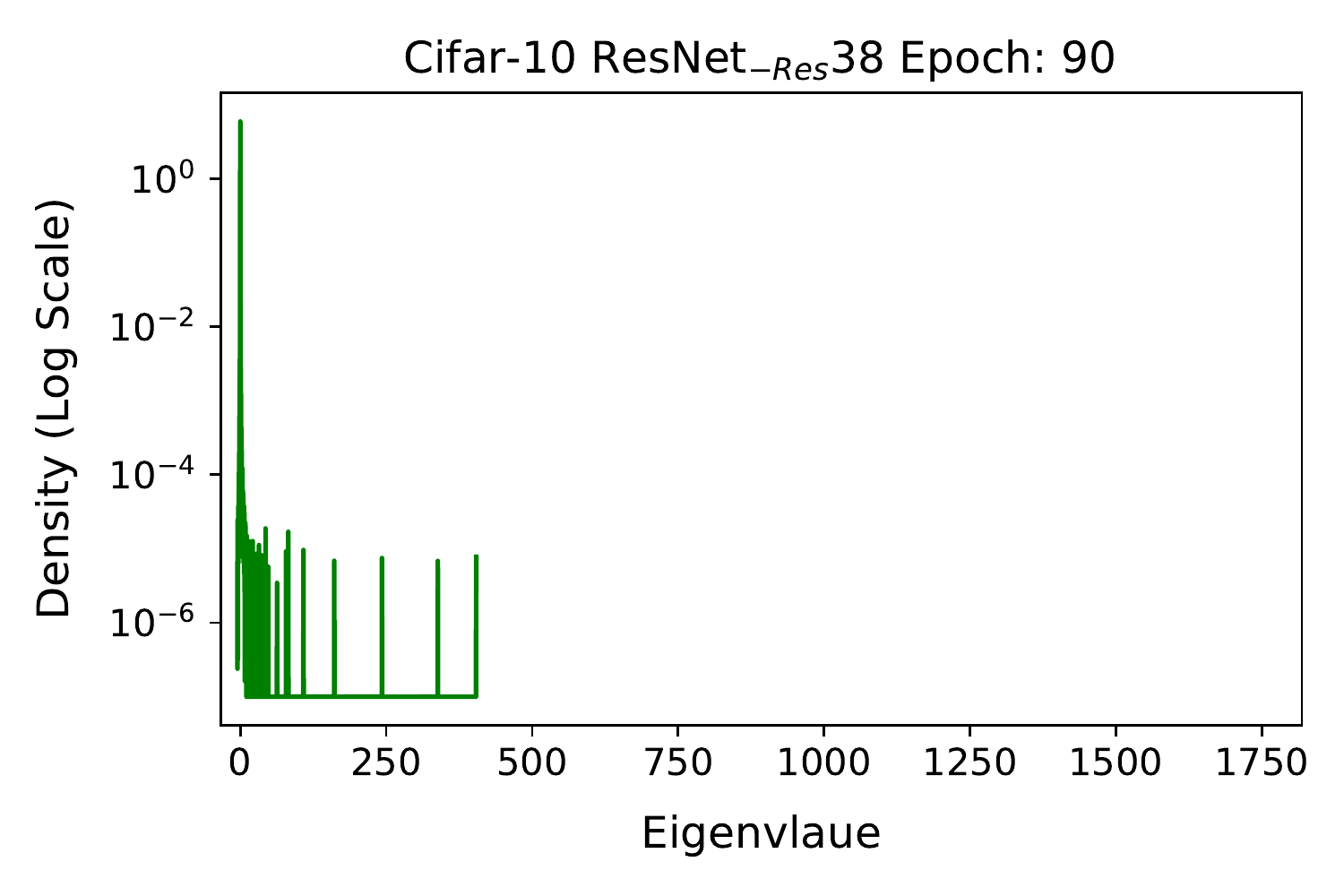}\\
\includegraphics[width=0.295\textwidth]{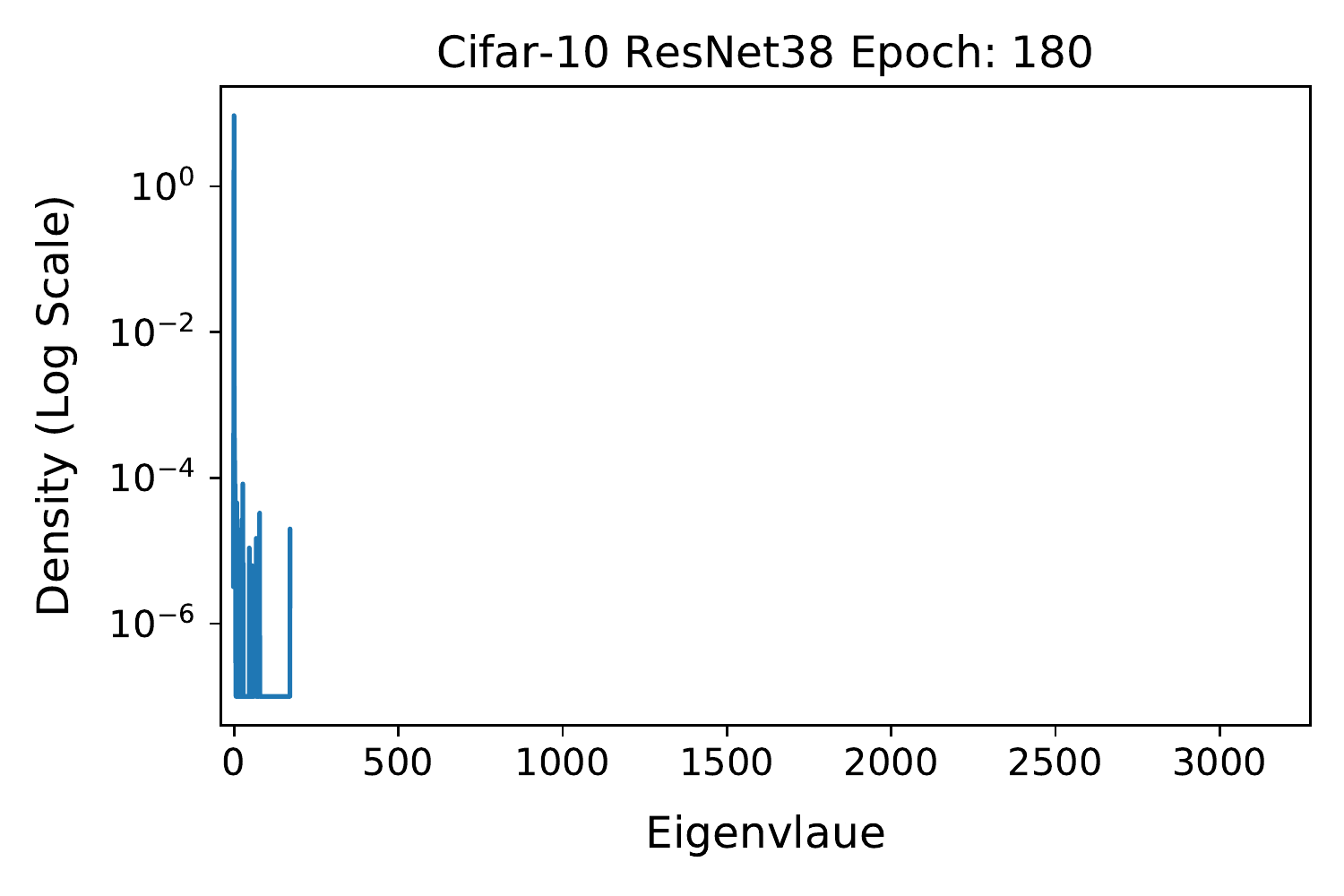}
\includegraphics[width=0.295\textwidth]{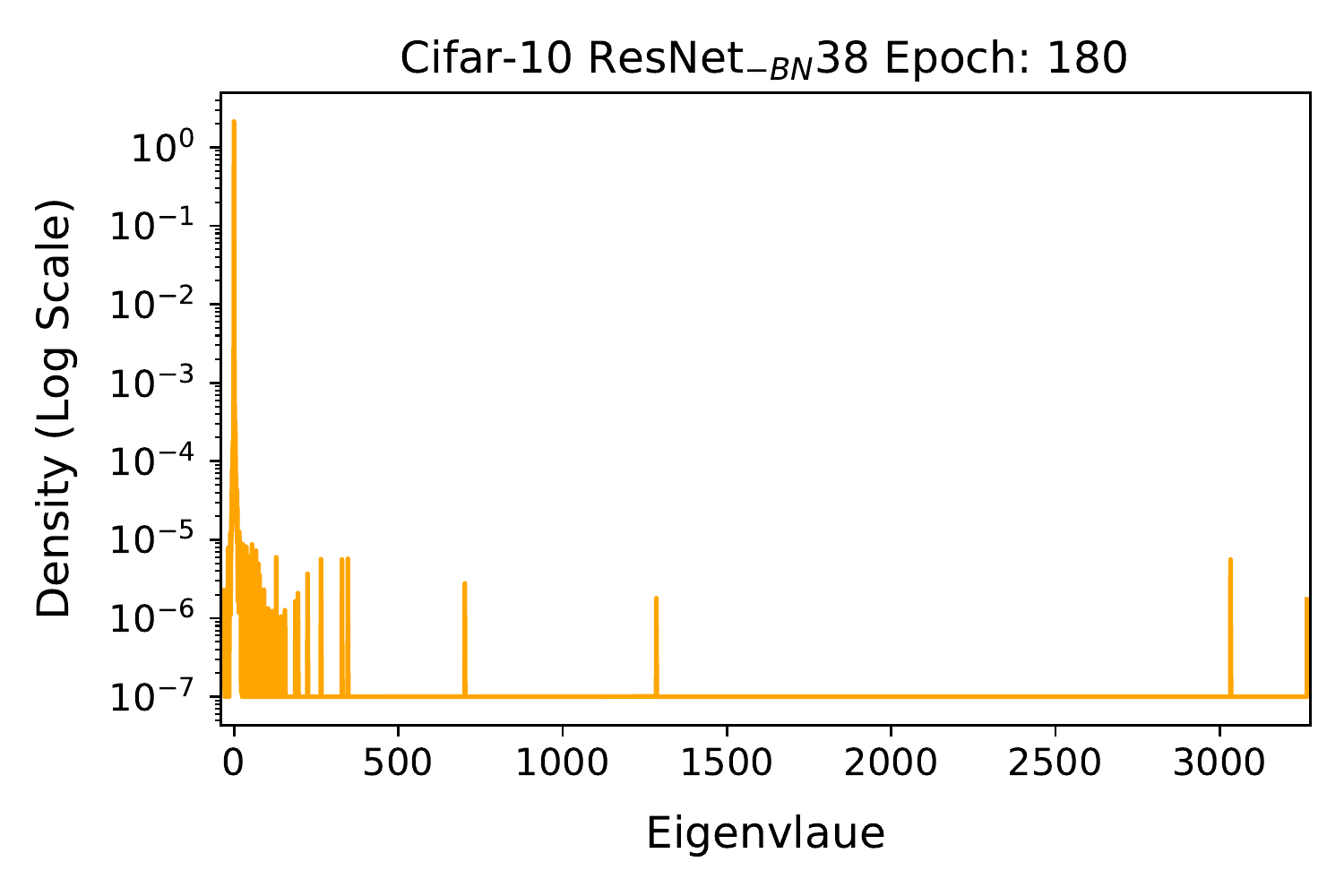}
\includegraphics[width=0.295\textwidth]{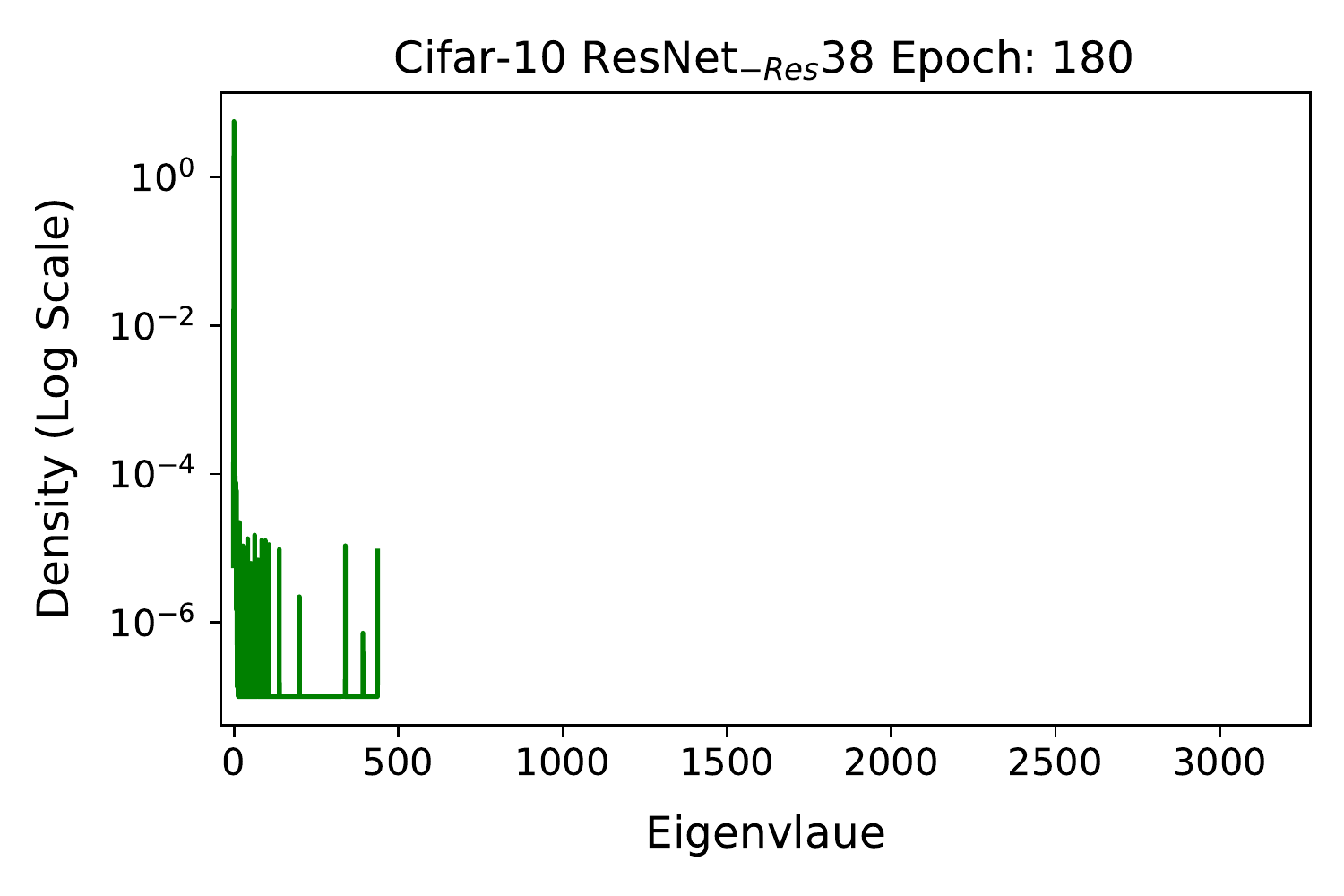}\\
\caption{
Hessian ESD of the entire network for ResNet/\ResNetBN/\ResNetRes with depth 38 on Cifar-10 with Hessian batch size 50000. 
This figure shows the Hessian ESD throughout the training process.
One notable thing here is that the Hessian ESD of \ResNetBN38 centers around zero at very beginning phase. 
This clearly shows that training without BN is indeed harder. 
}
  \label{fig:resnet38-slq-full-net-all}
\end{figure*}

\begin{figure*}[!htbp]
\centering
\includegraphics[width=0.295\textwidth]{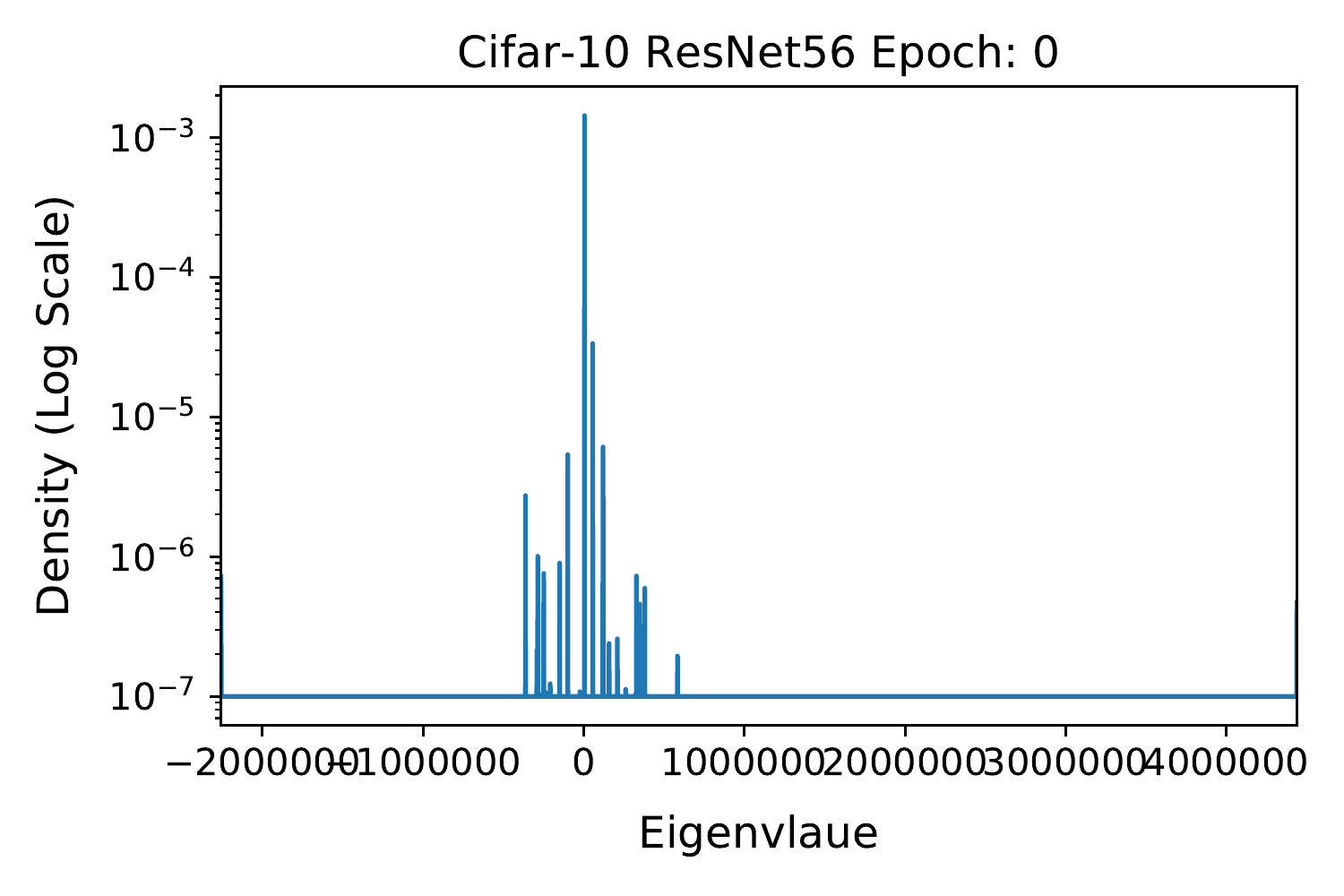}
\includegraphics[width=0.295\textwidth]{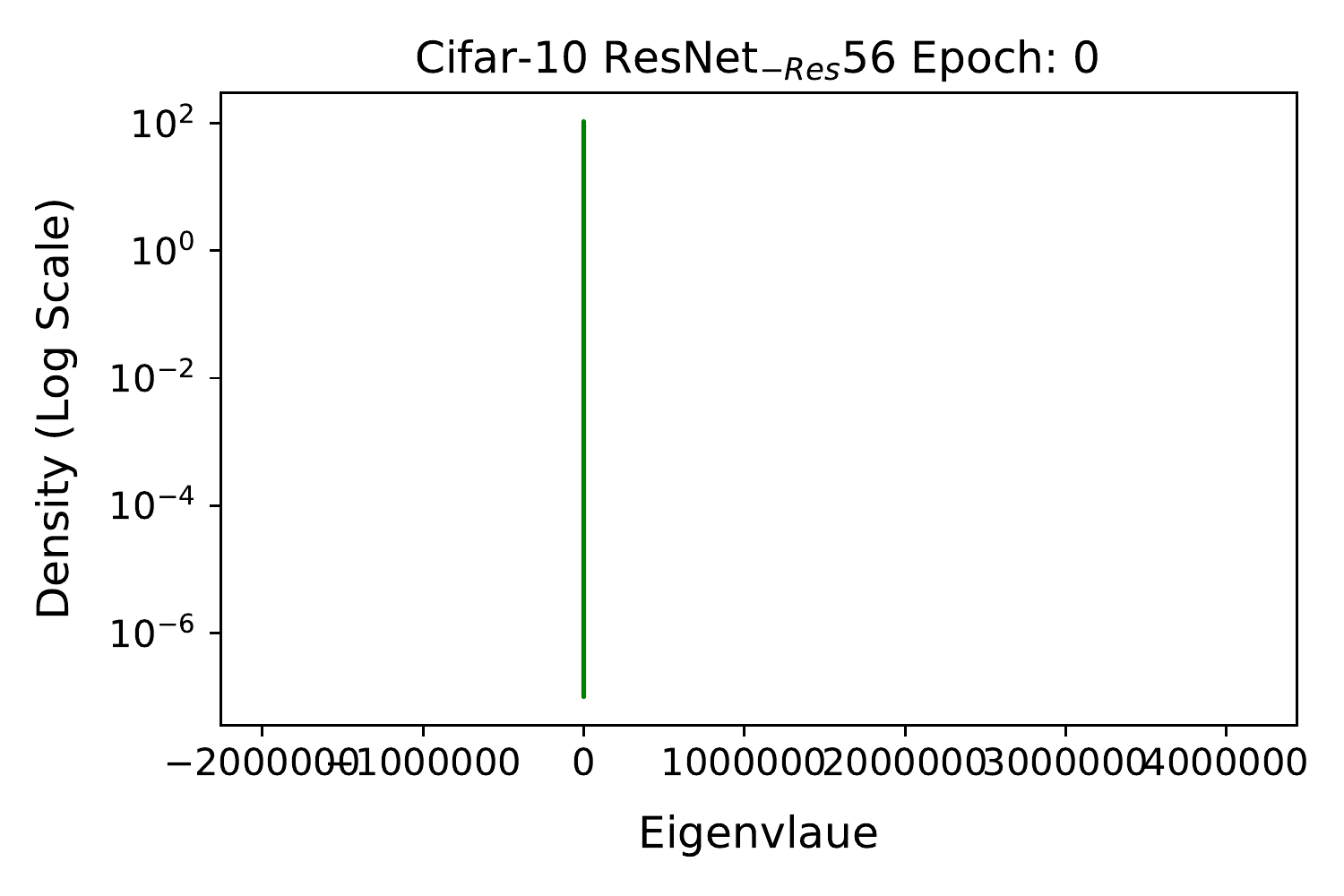}\\
\includegraphics[width=0.295\textwidth]{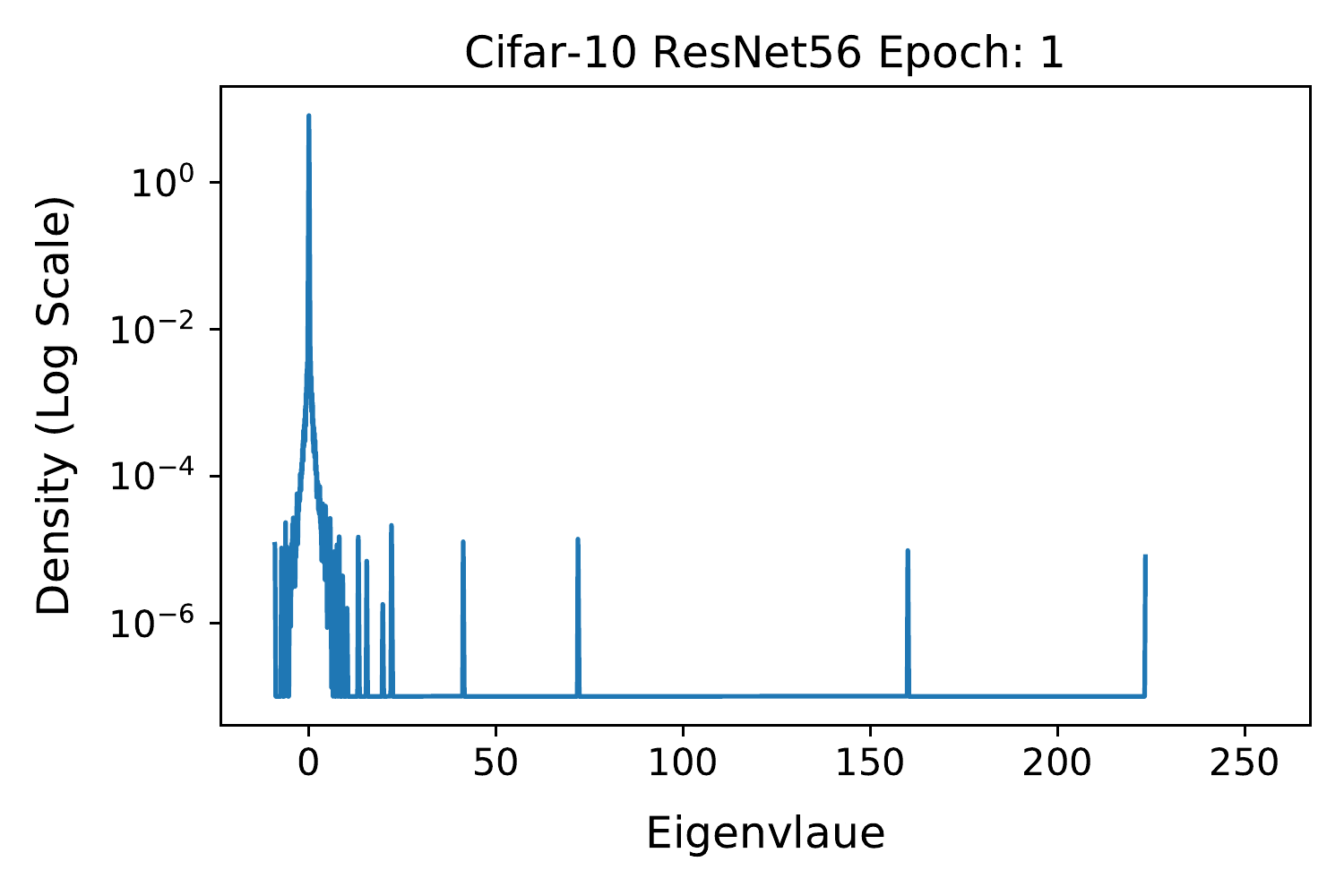}
\includegraphics[width=0.295\textwidth]{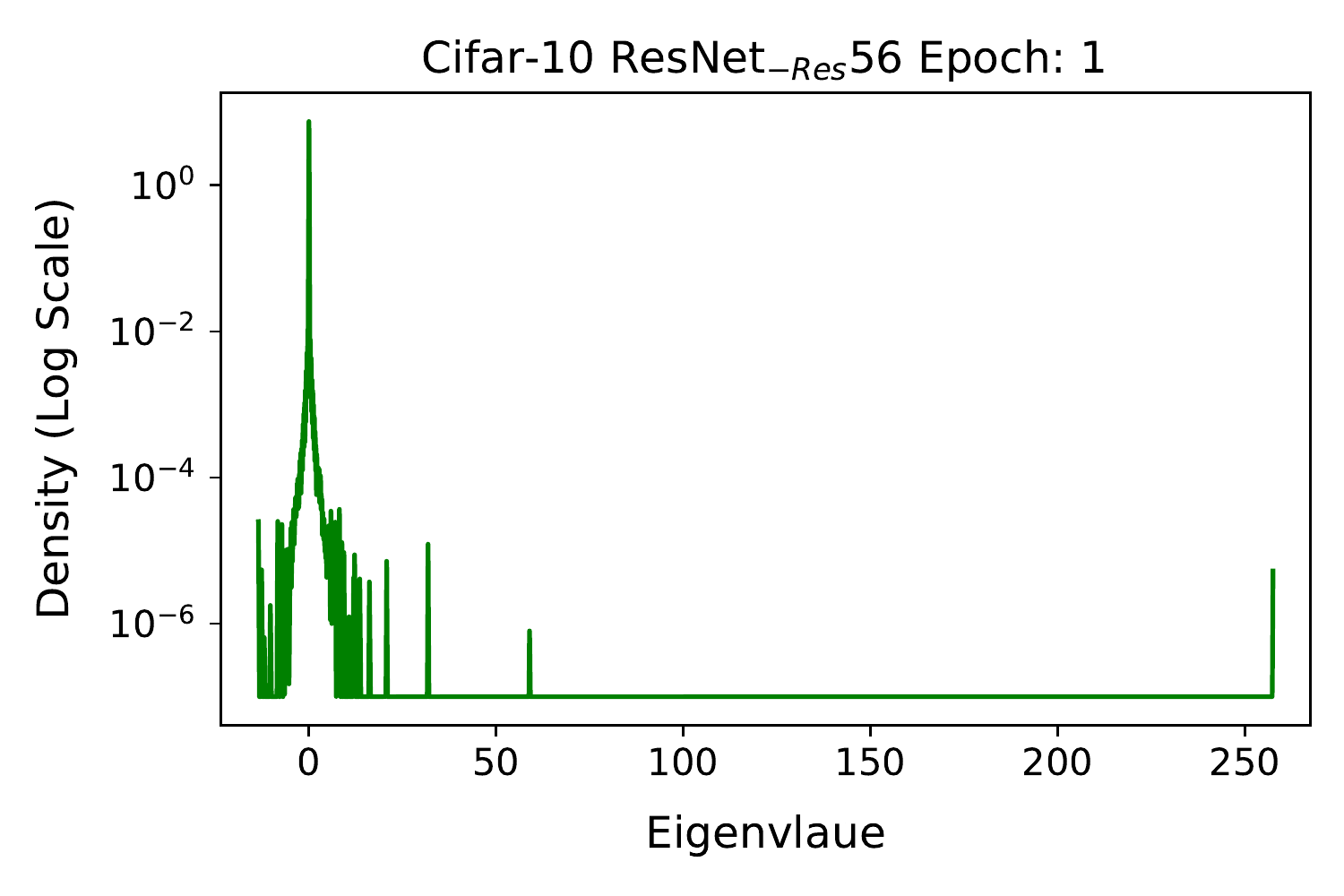}\\
\includegraphics[width=0.295\textwidth]{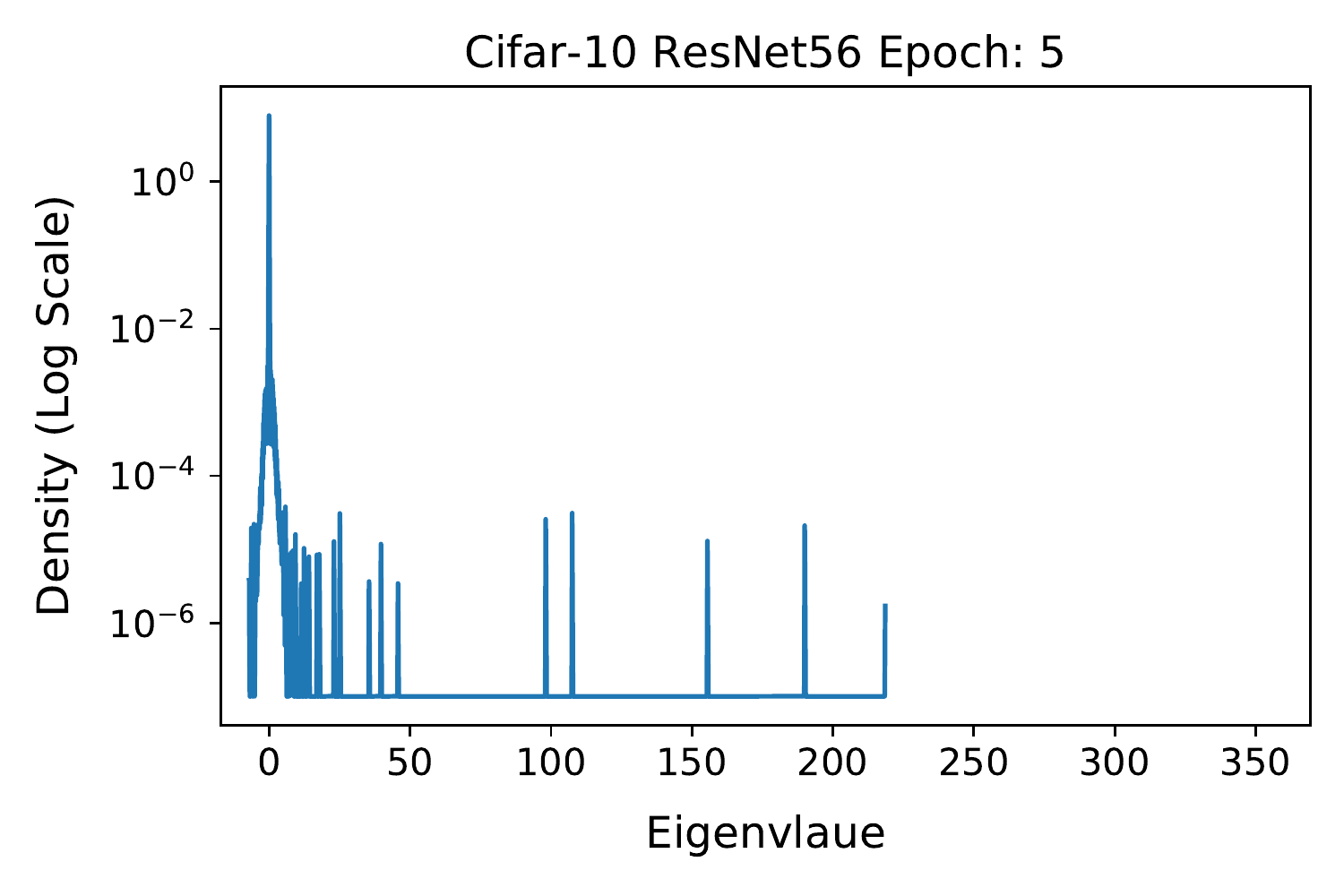}
\includegraphics[width=0.295\textwidth]{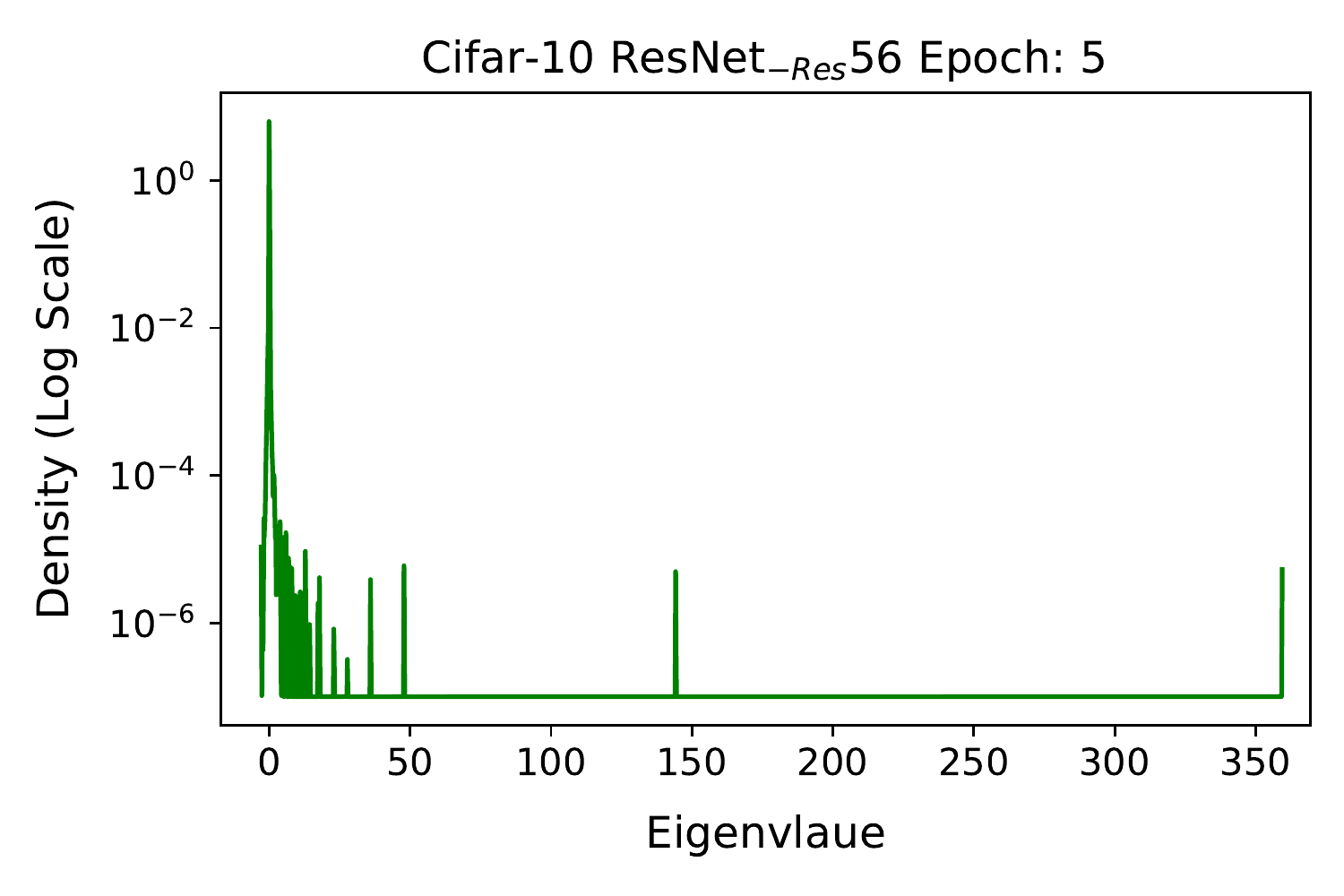}\\
\includegraphics[width=0.295\textwidth]{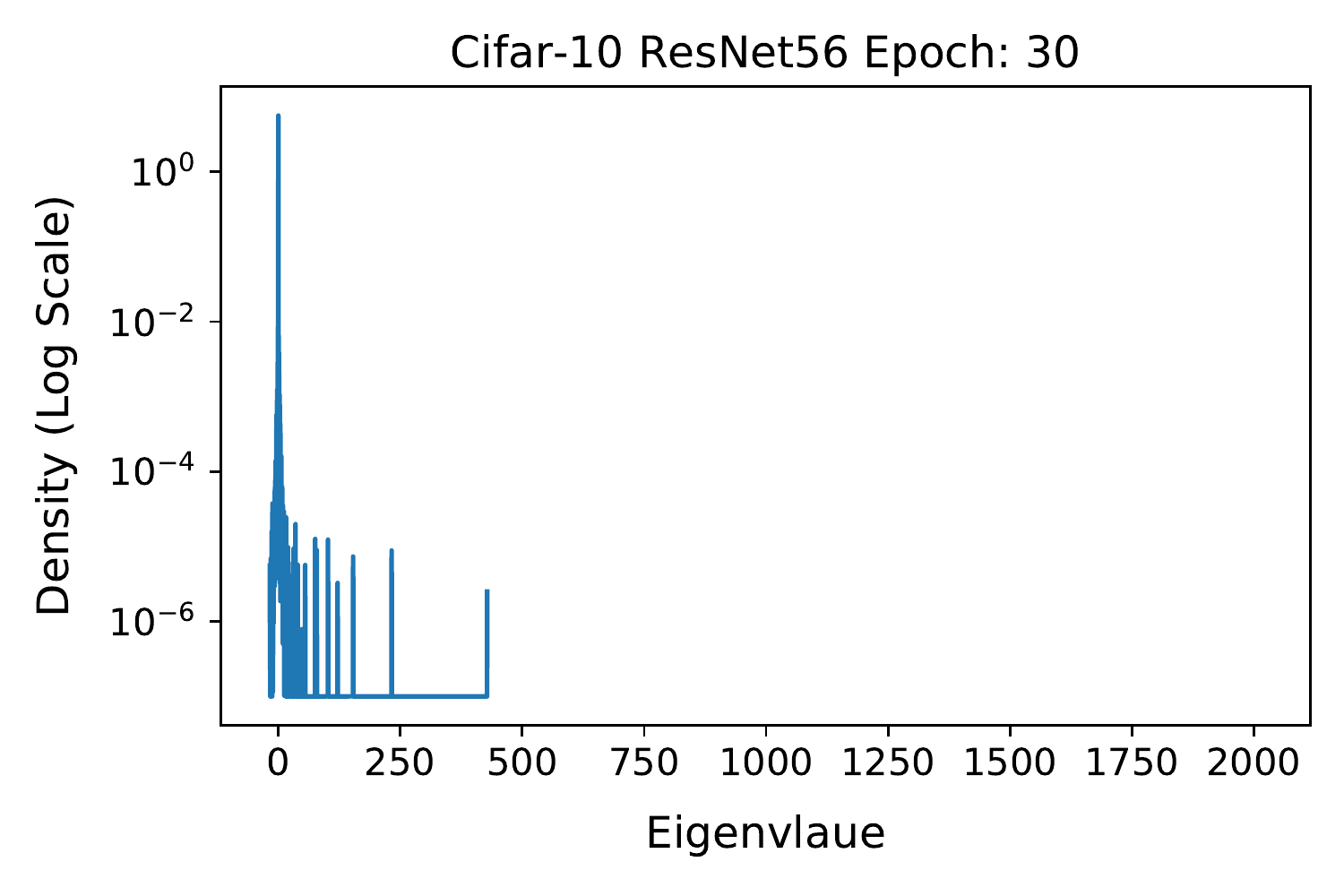}
\includegraphics[width=0.295\textwidth]{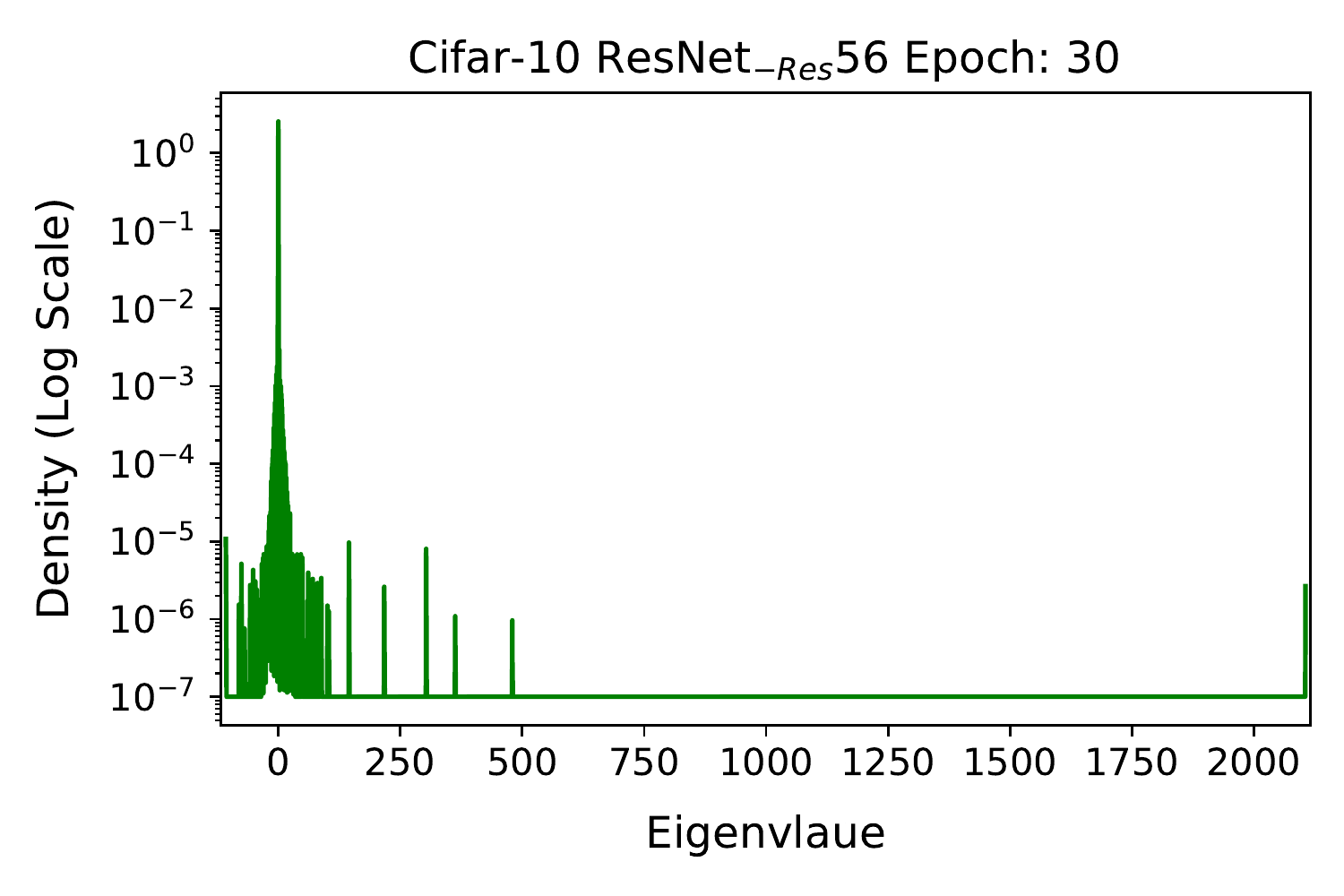}\\
\includegraphics[width=0.295\textwidth]{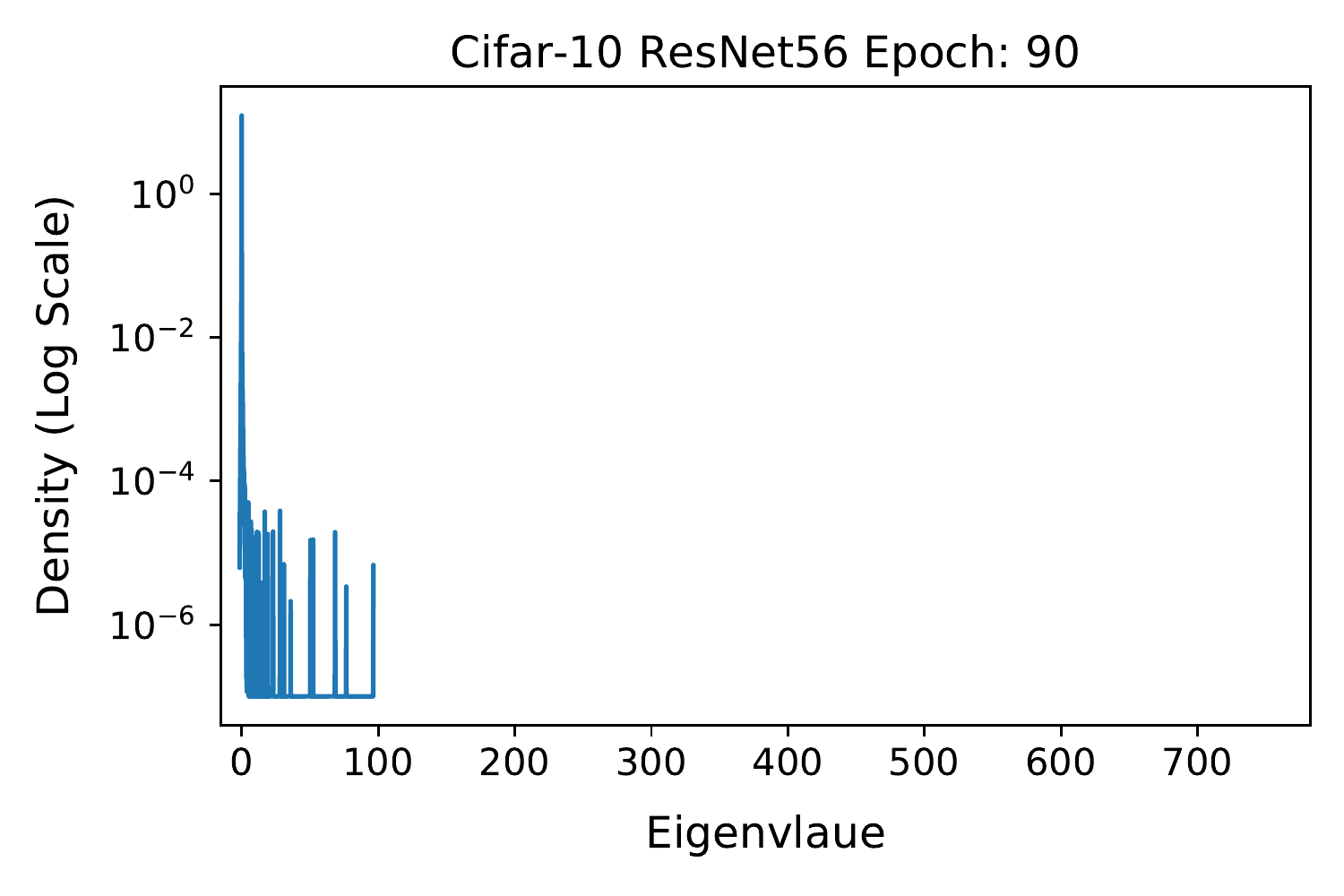}
\includegraphics[width=0.295\textwidth]{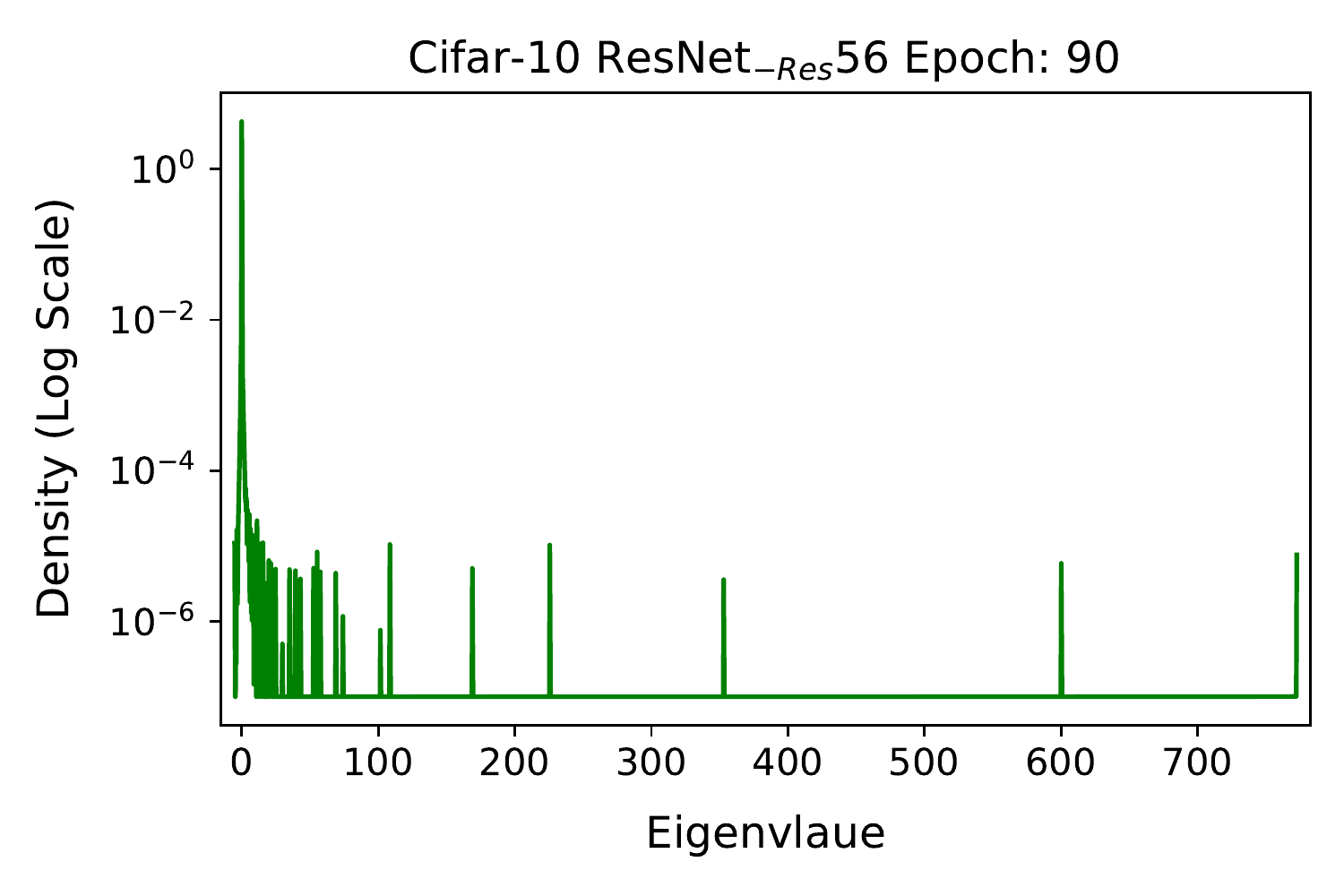}\\
\includegraphics[width=0.295\textwidth]{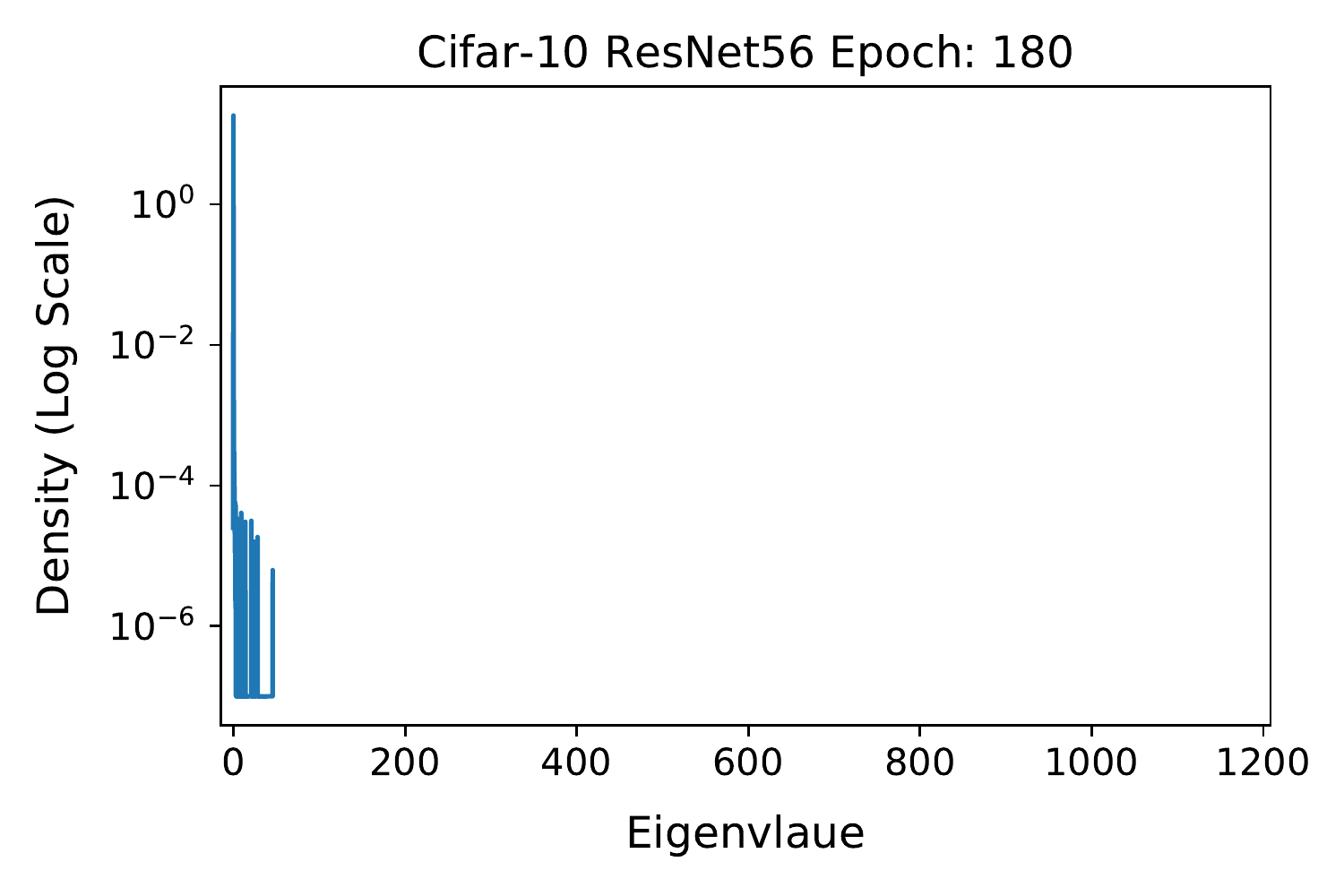}
\includegraphics[width=0.295\textwidth]{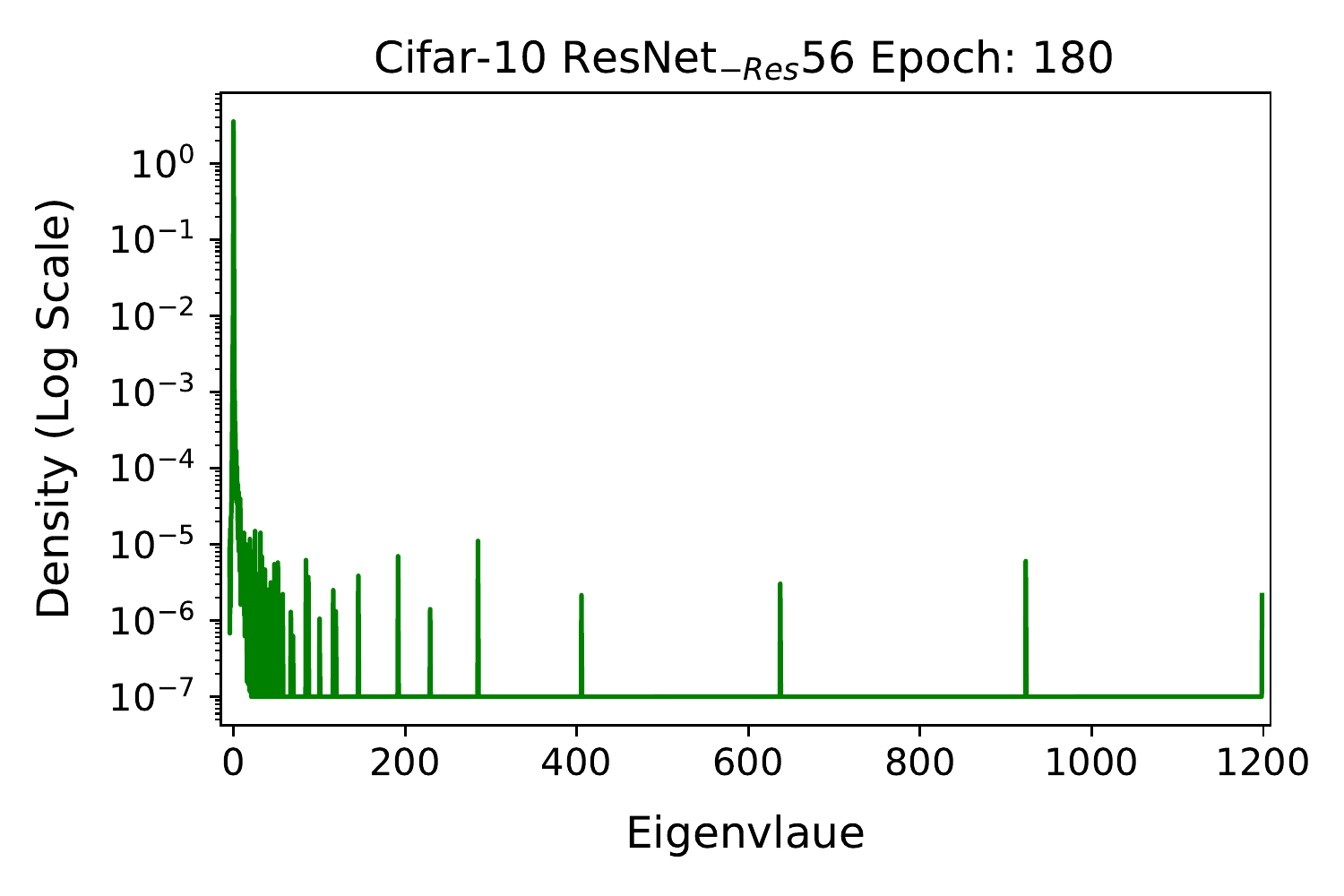}\\
\caption{
Hessian ESD of the entire network for ResNet/\ResNetRes with depth 56 with Hessian batch size 50000. 
Residual connection can help smooth the loss landscape. 
}
  \label{fig:resnet56-slq-full-net-all}
\end{figure*}

\begin{figure*}[!htbp]
\centering
\includegraphics[width=0.295\textwidth]{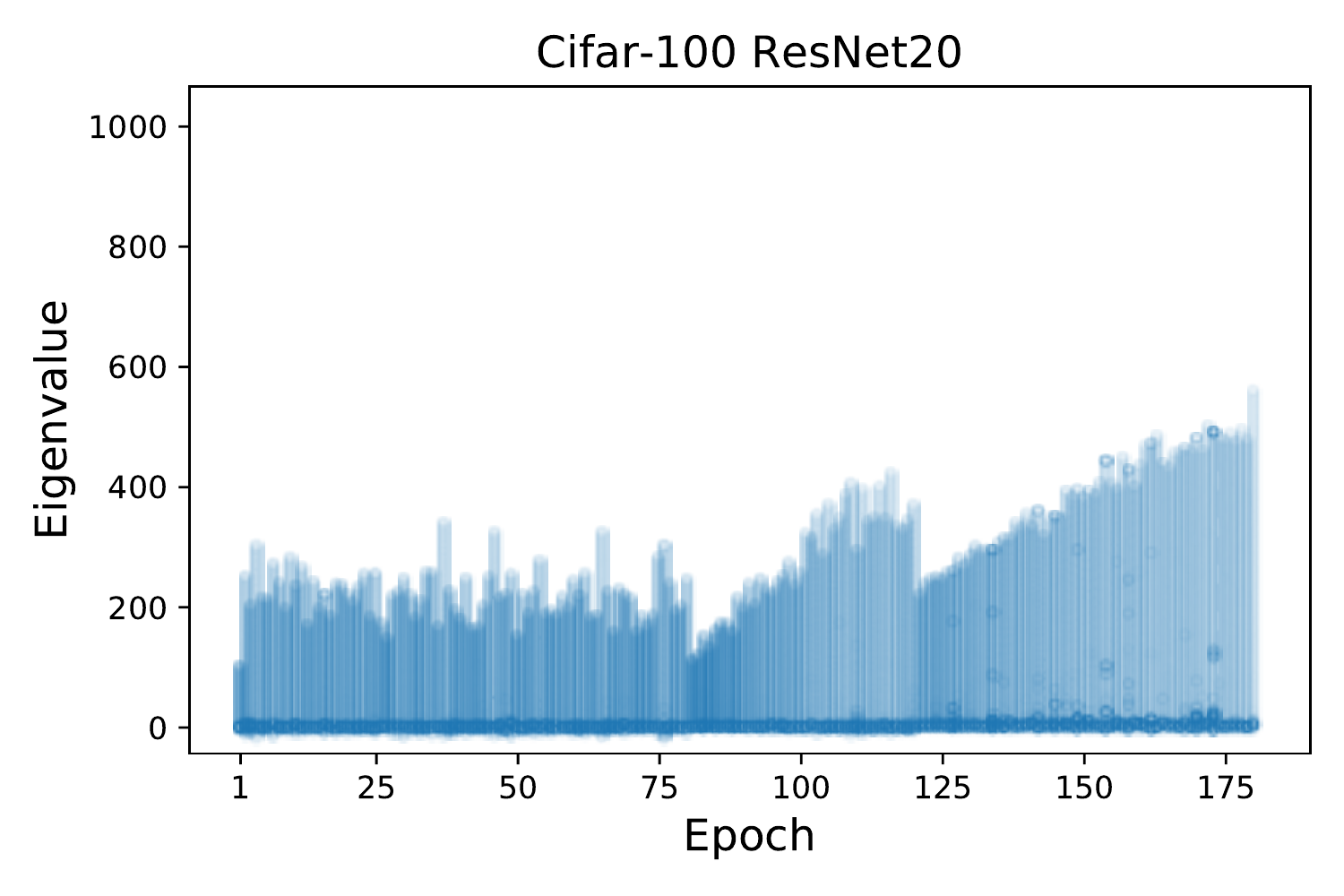}
\includegraphics[width=0.295\textwidth]{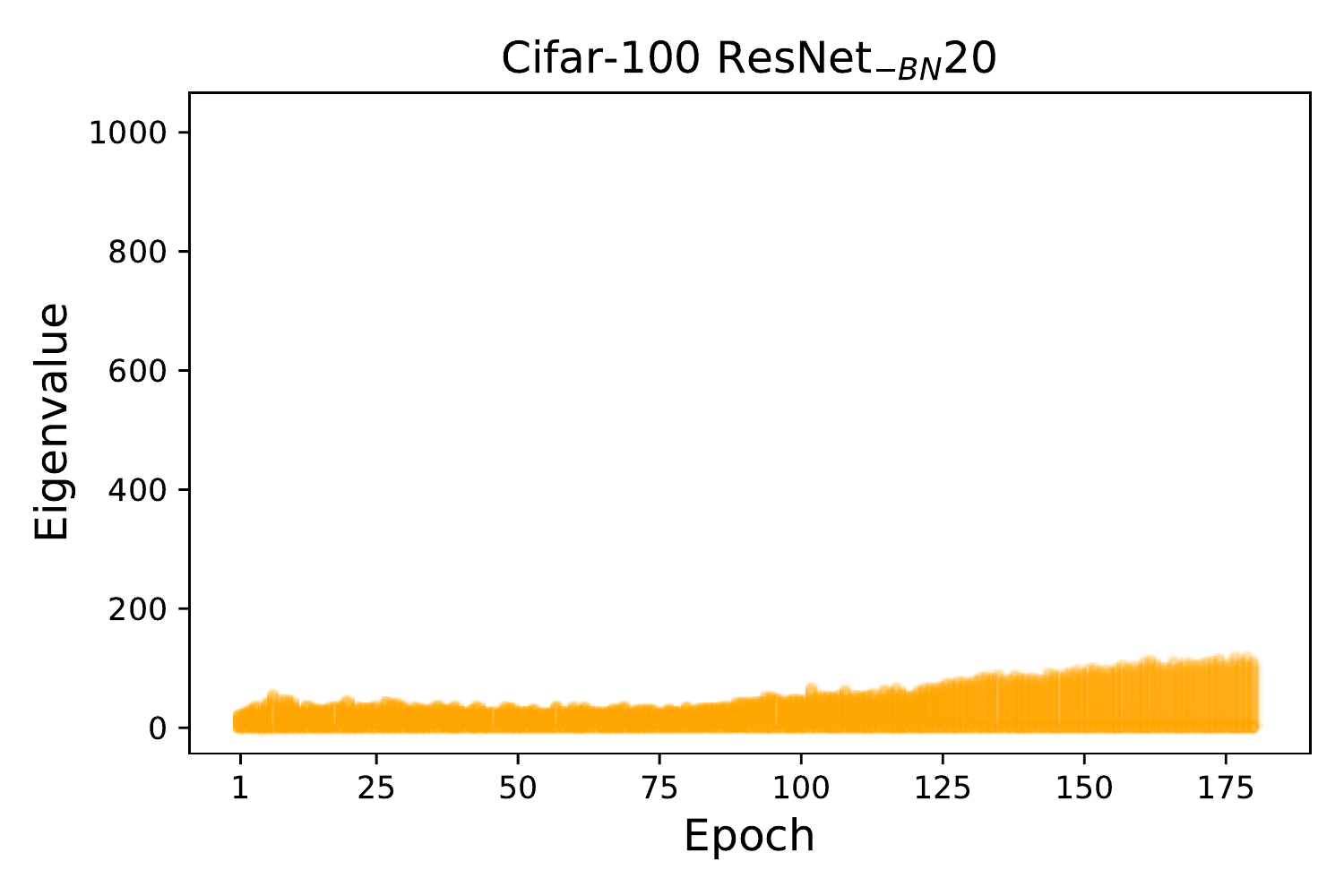}
\includegraphics[width=0.295\textwidth]{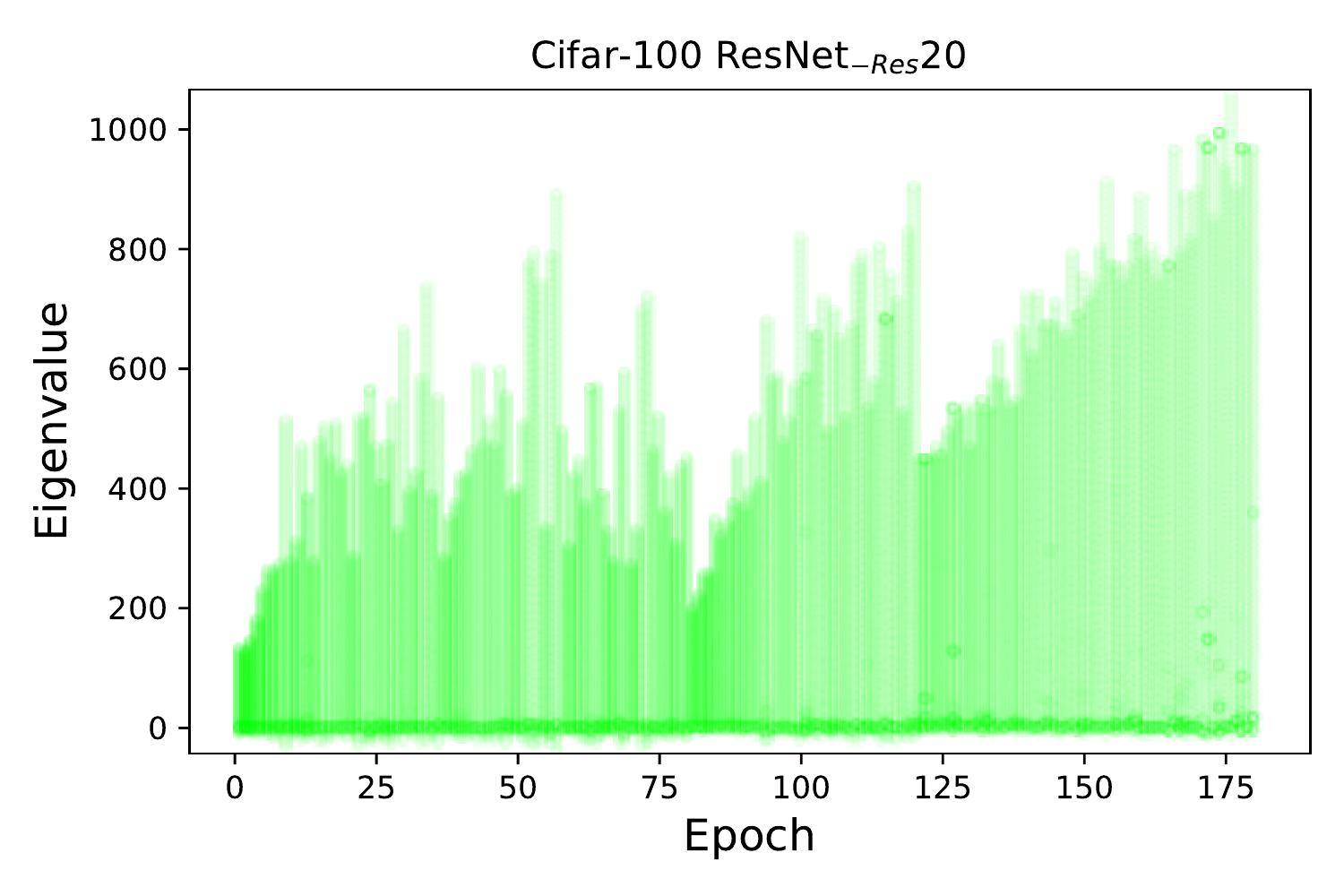}\\
\includegraphics[width=0.295\textwidth]{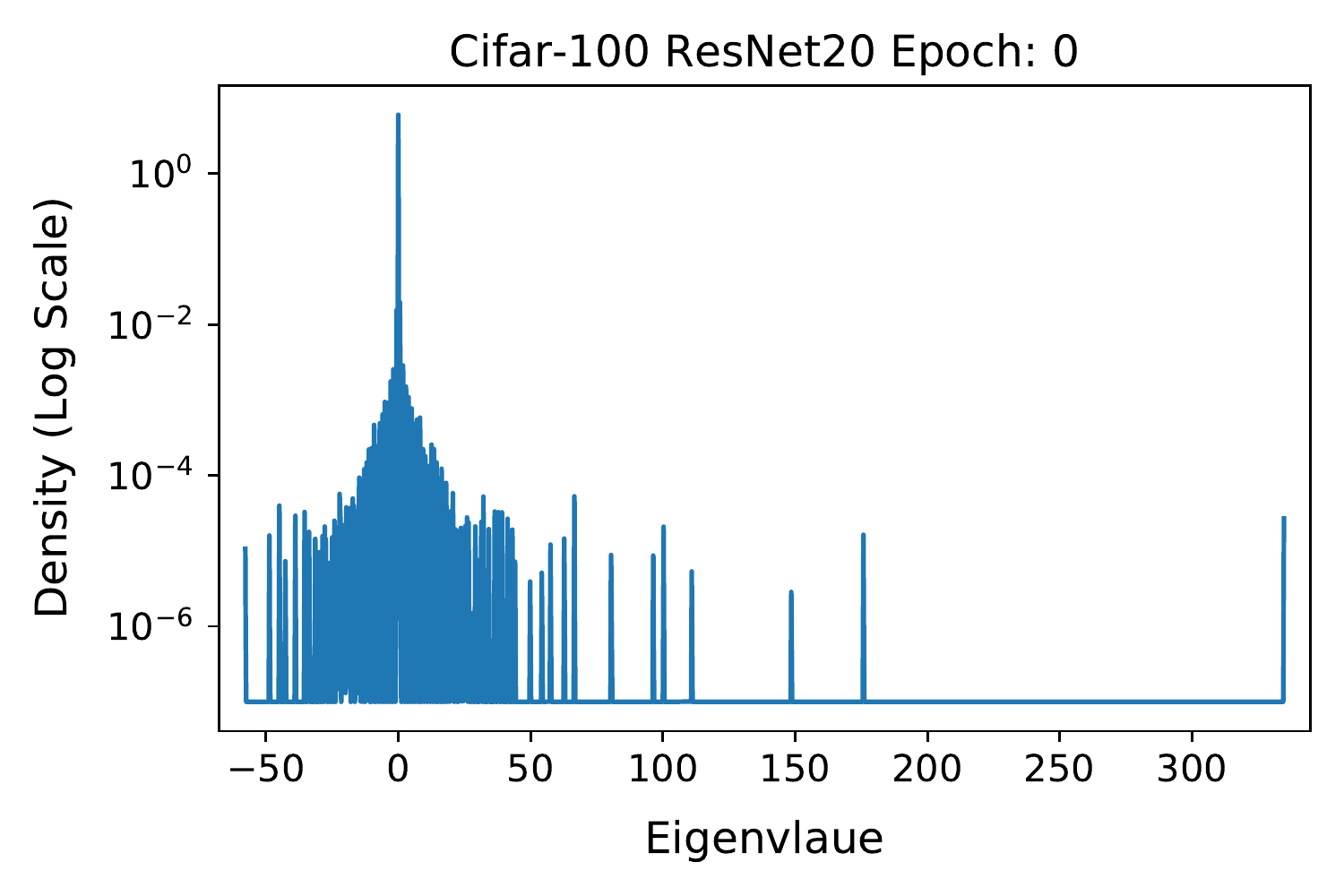}
\includegraphics[width=0.295\textwidth]{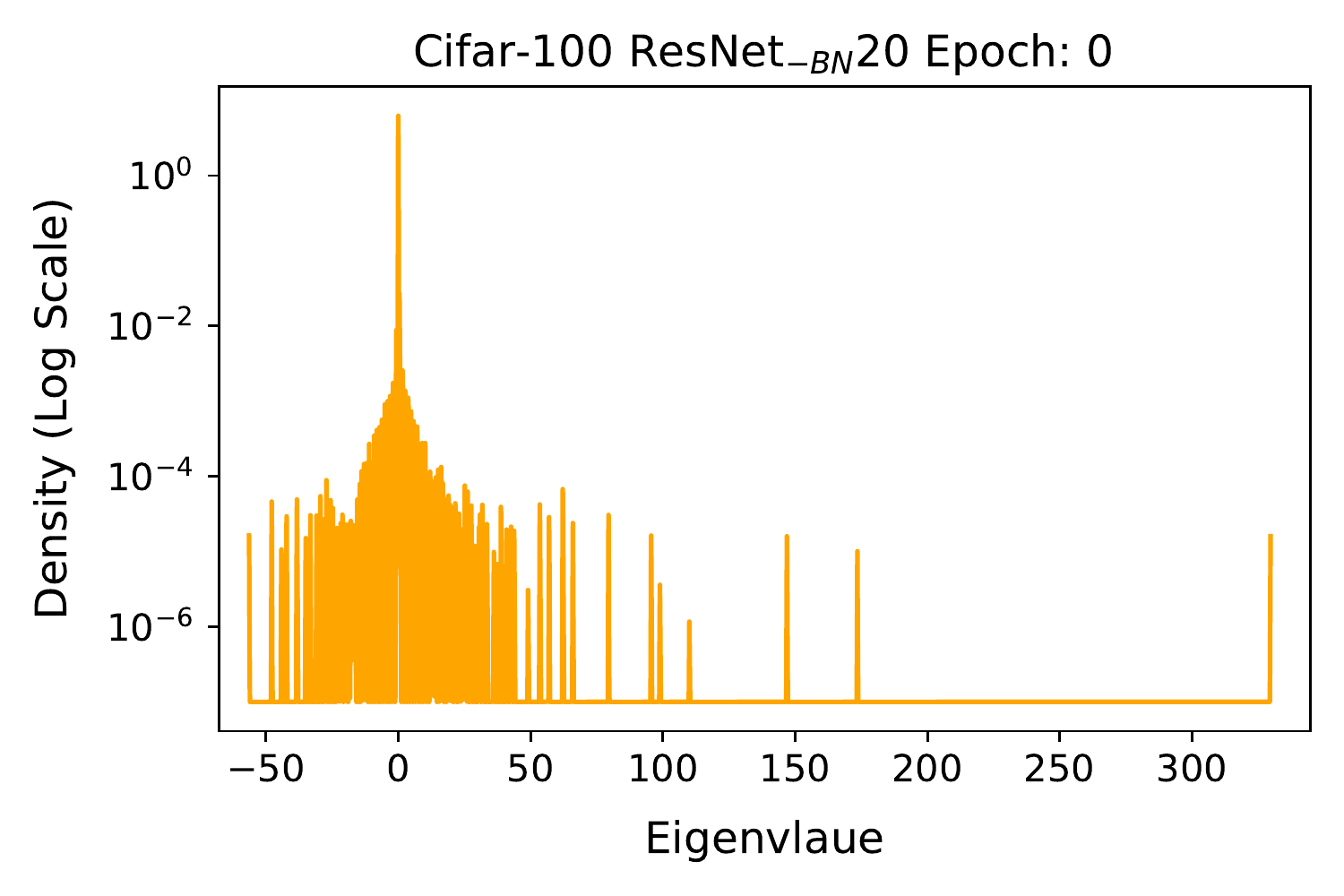}
\includegraphics[width=0.295\textwidth]{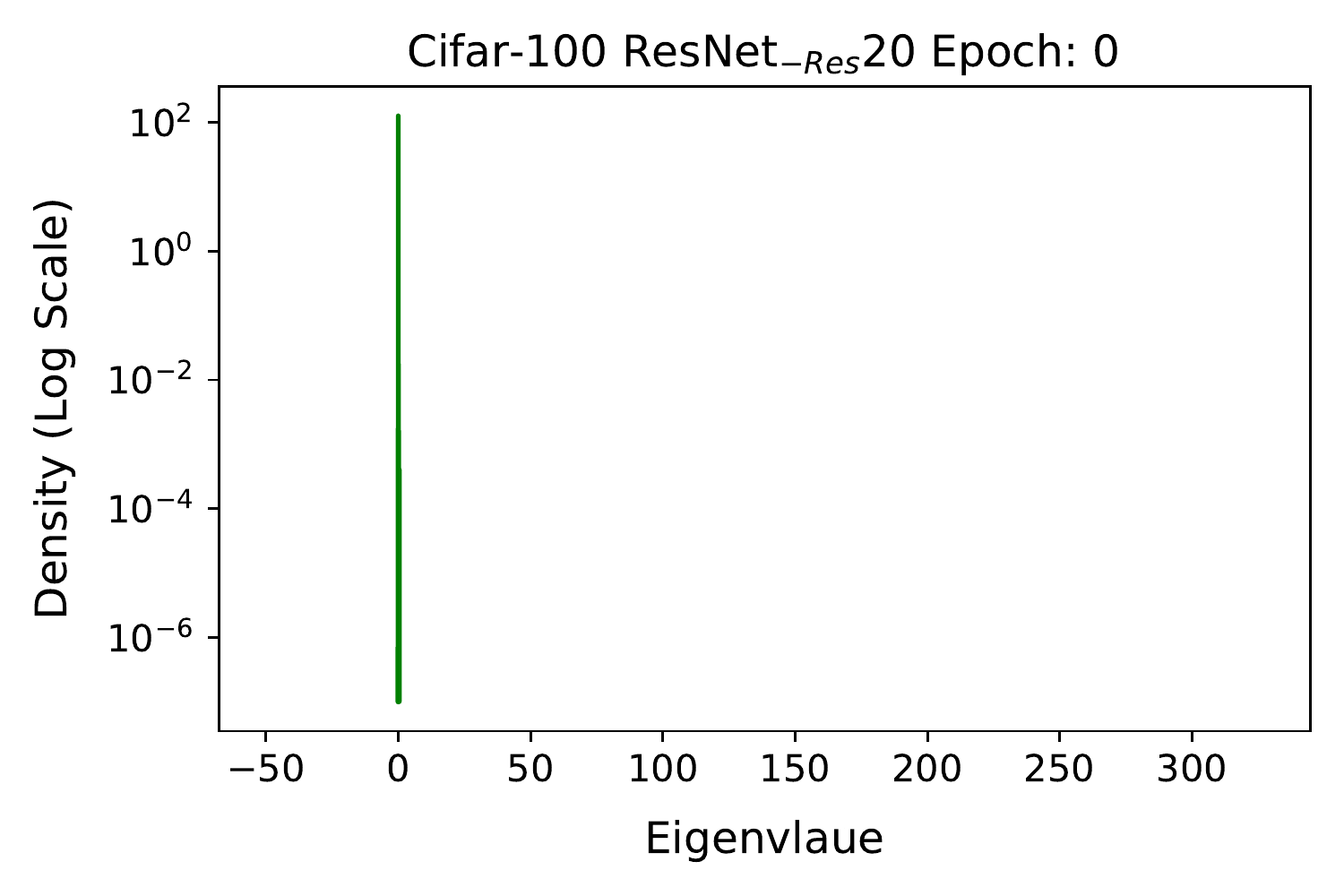}\\
\includegraphics[width=0.295\textwidth]{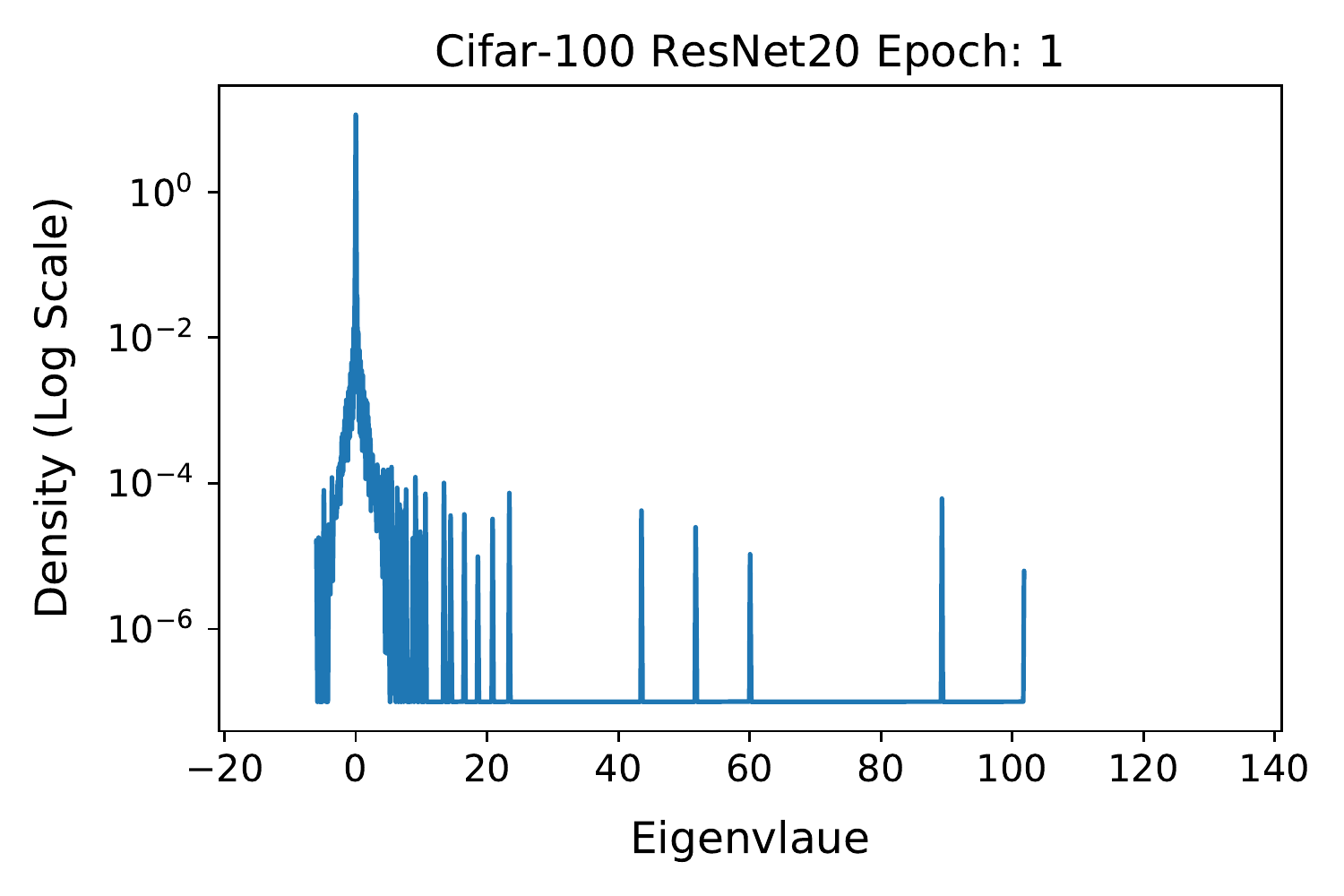}
\includegraphics[width=0.295\textwidth]{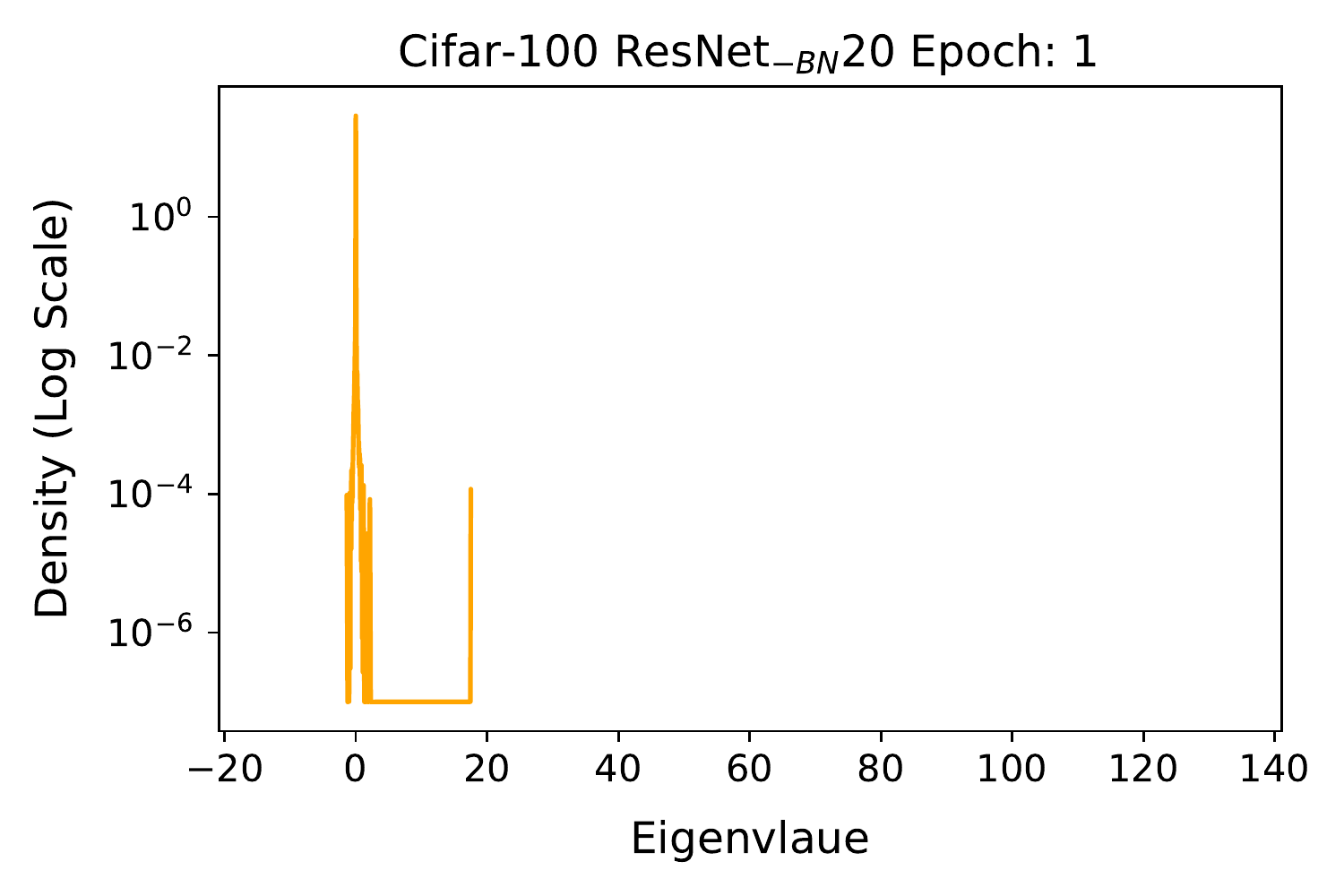}
\includegraphics[width=0.295\textwidth]{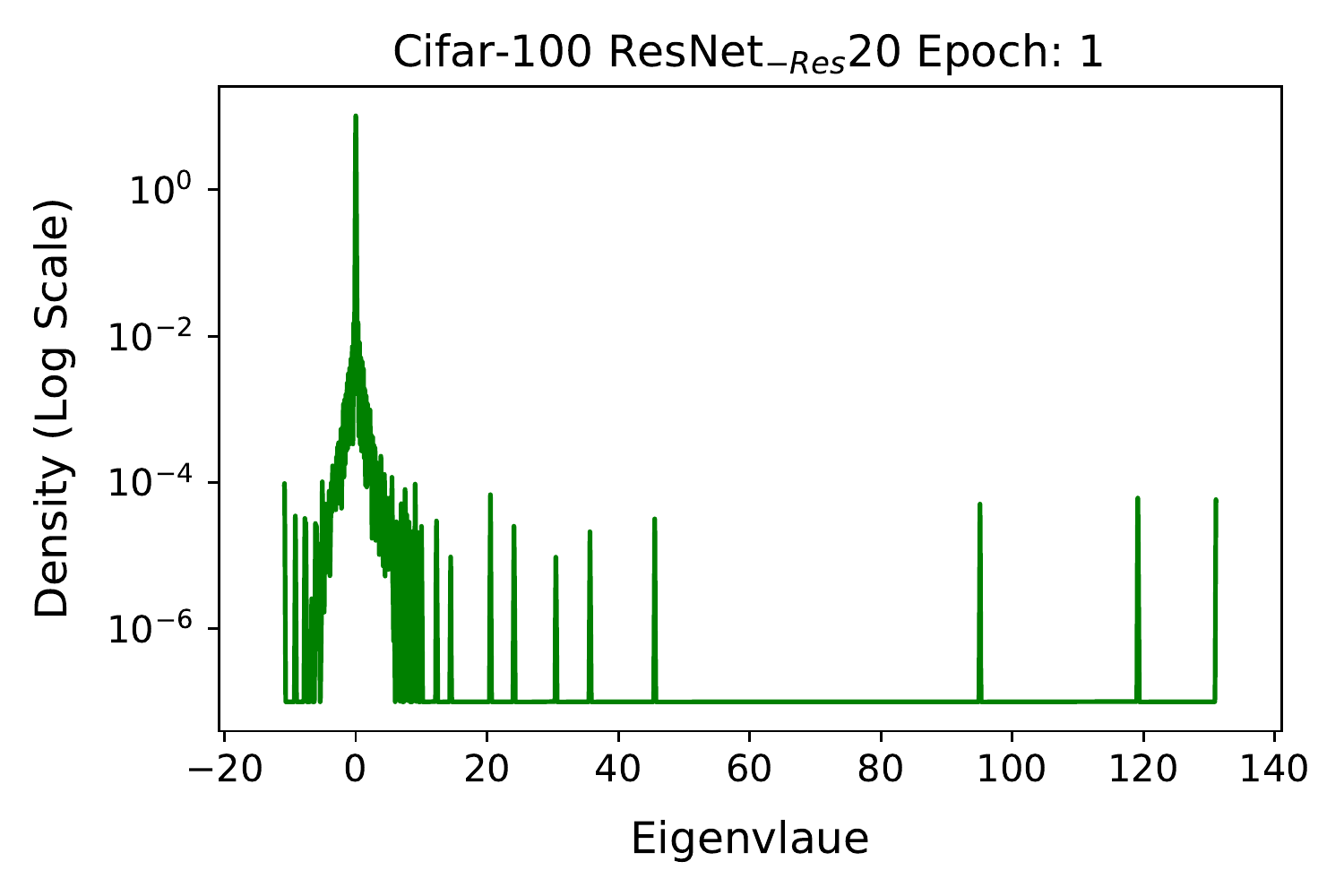}\\
\includegraphics[width=0.295\textwidth]{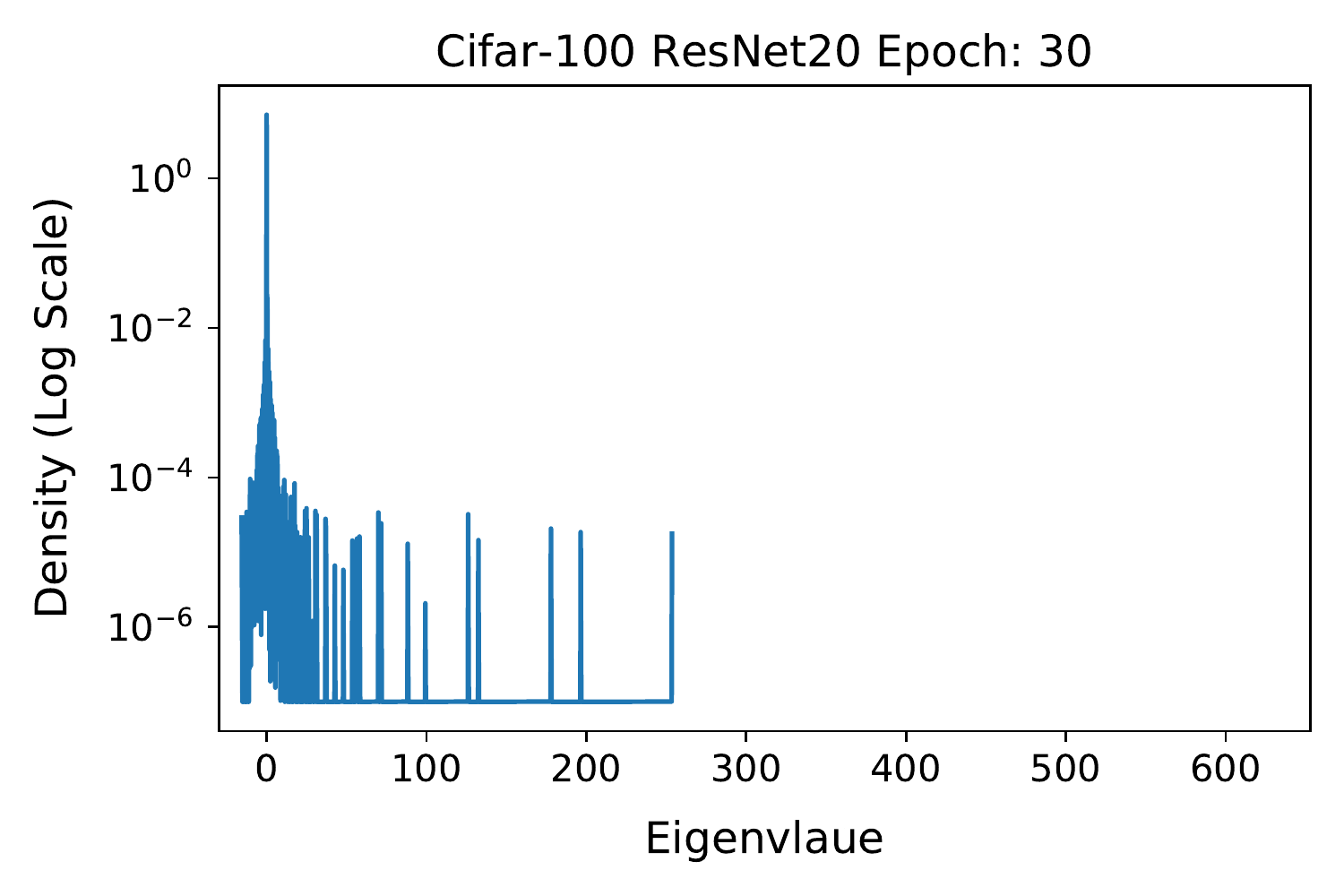}
\includegraphics[width=0.295\textwidth]{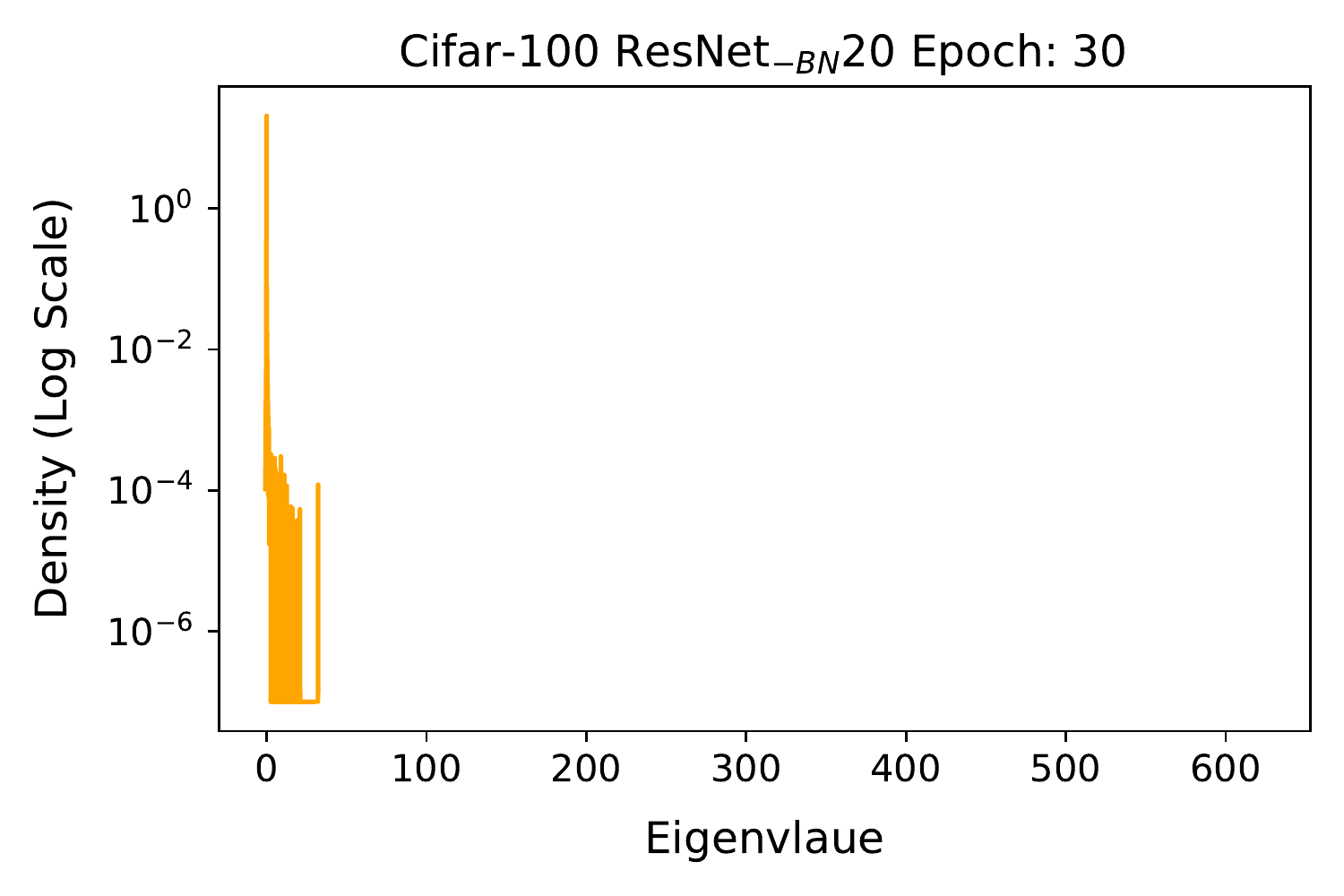}
\includegraphics[width=0.295\textwidth]{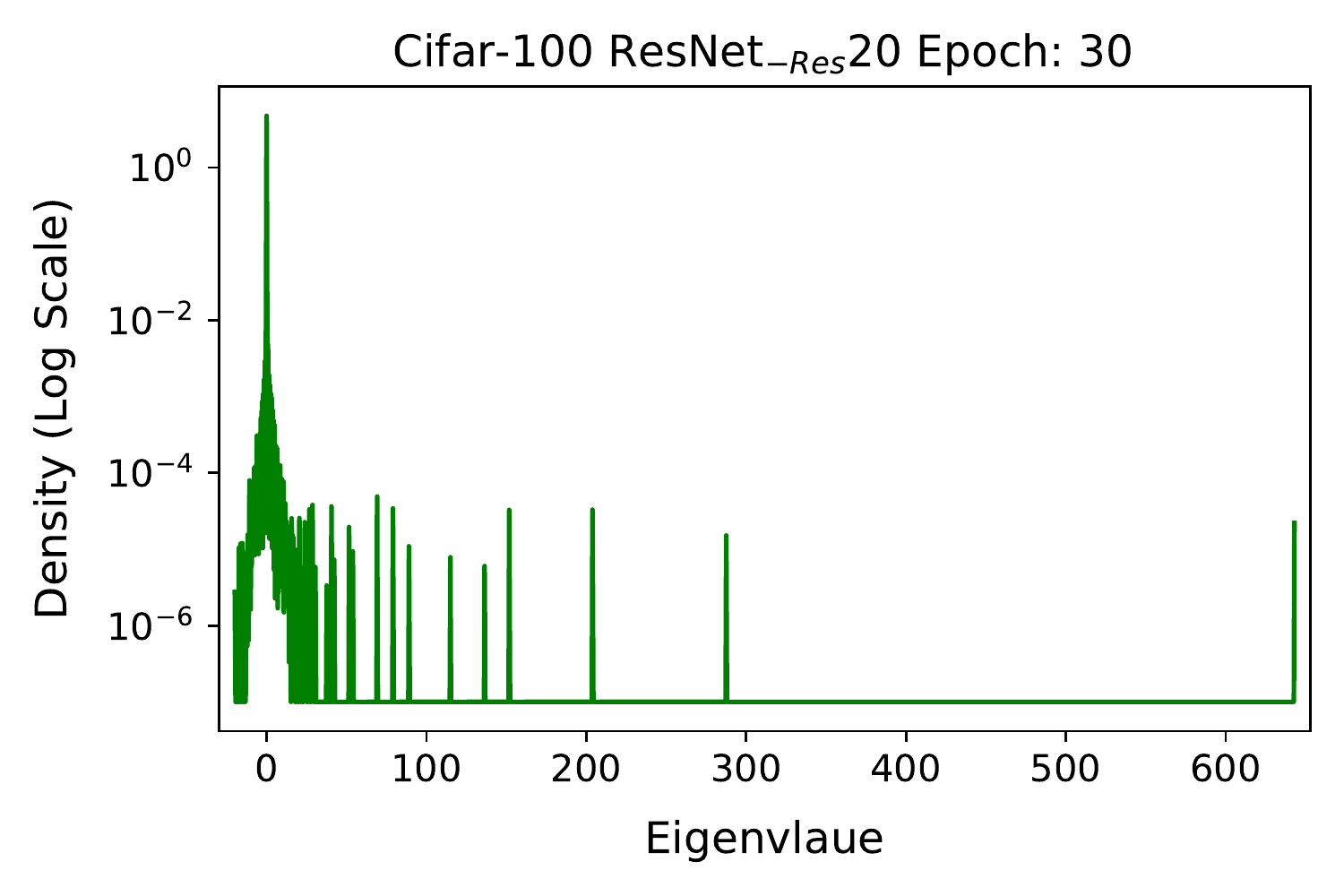}\\
\includegraphics[width=0.295\textwidth]{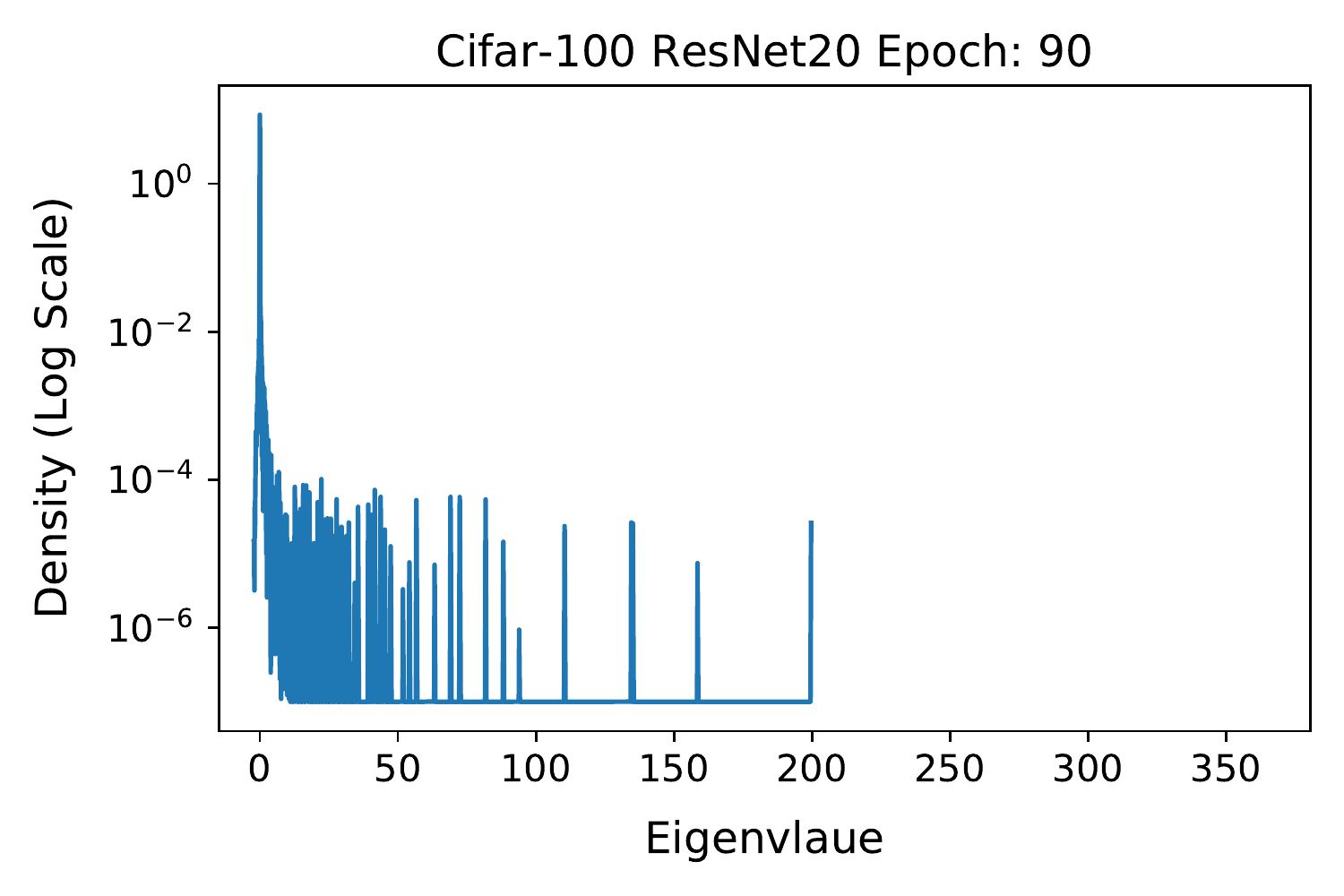}
\includegraphics[width=0.295\textwidth]{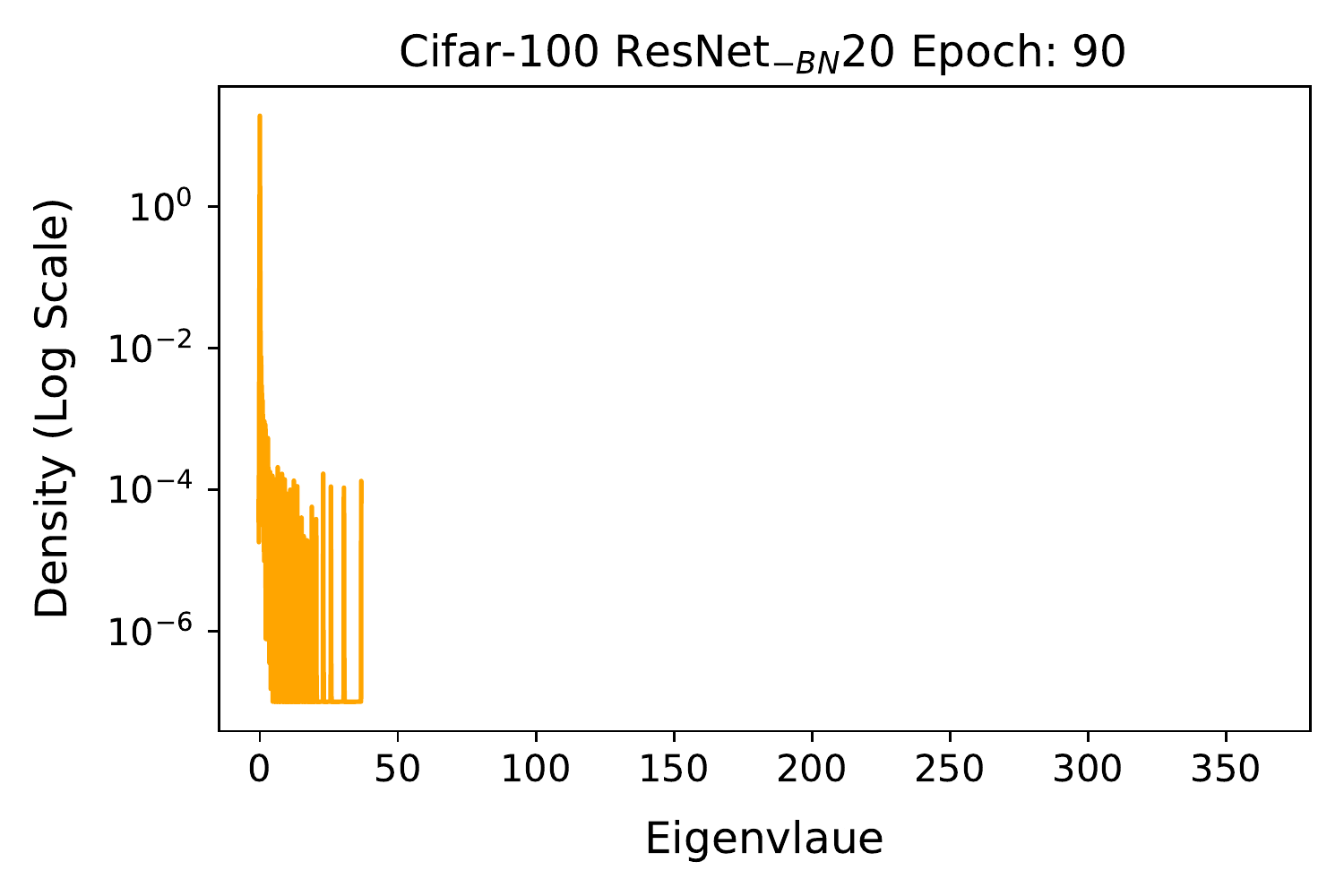}
\includegraphics[width=0.295\textwidth]{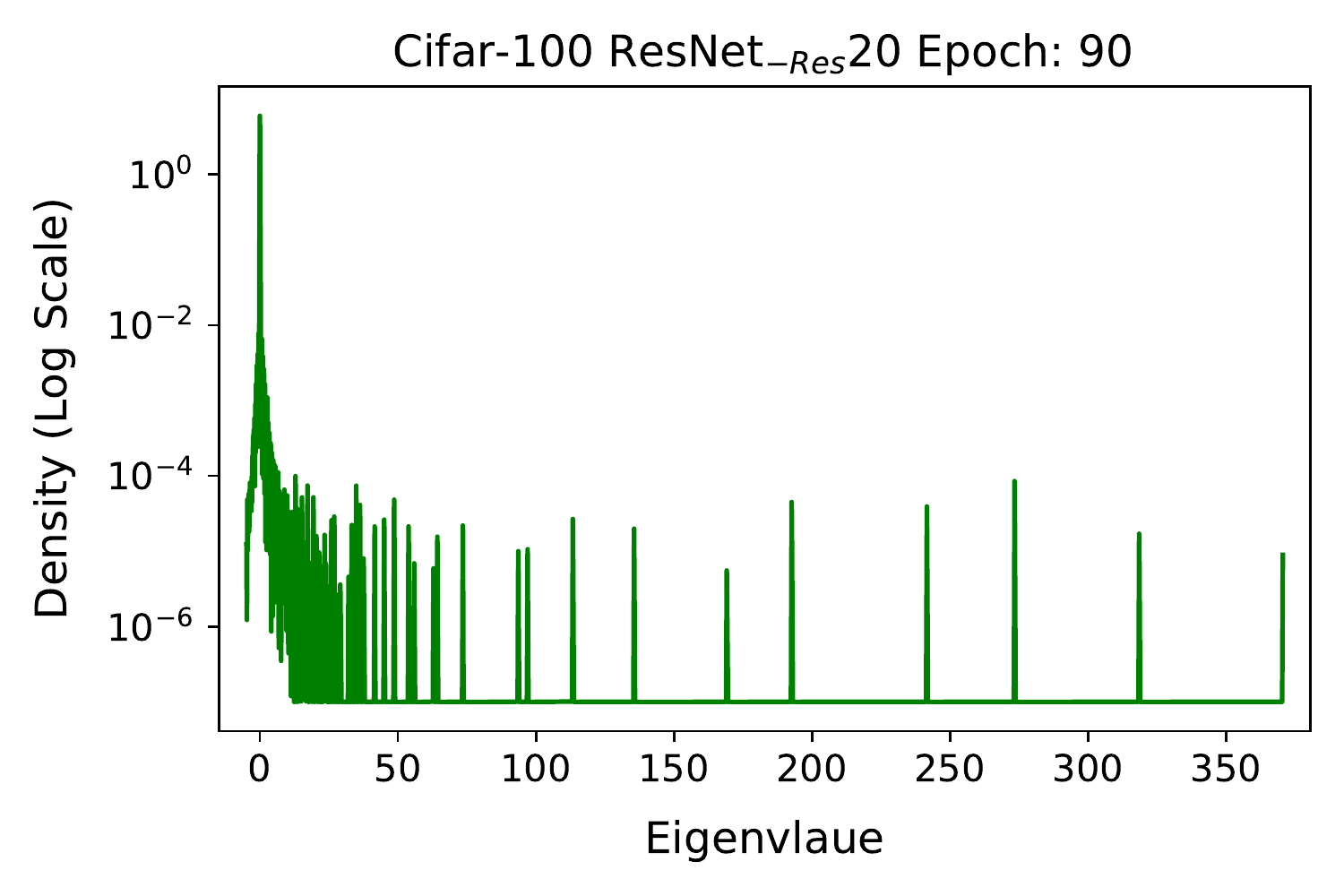}\\
\includegraphics[width=0.295\textwidth]{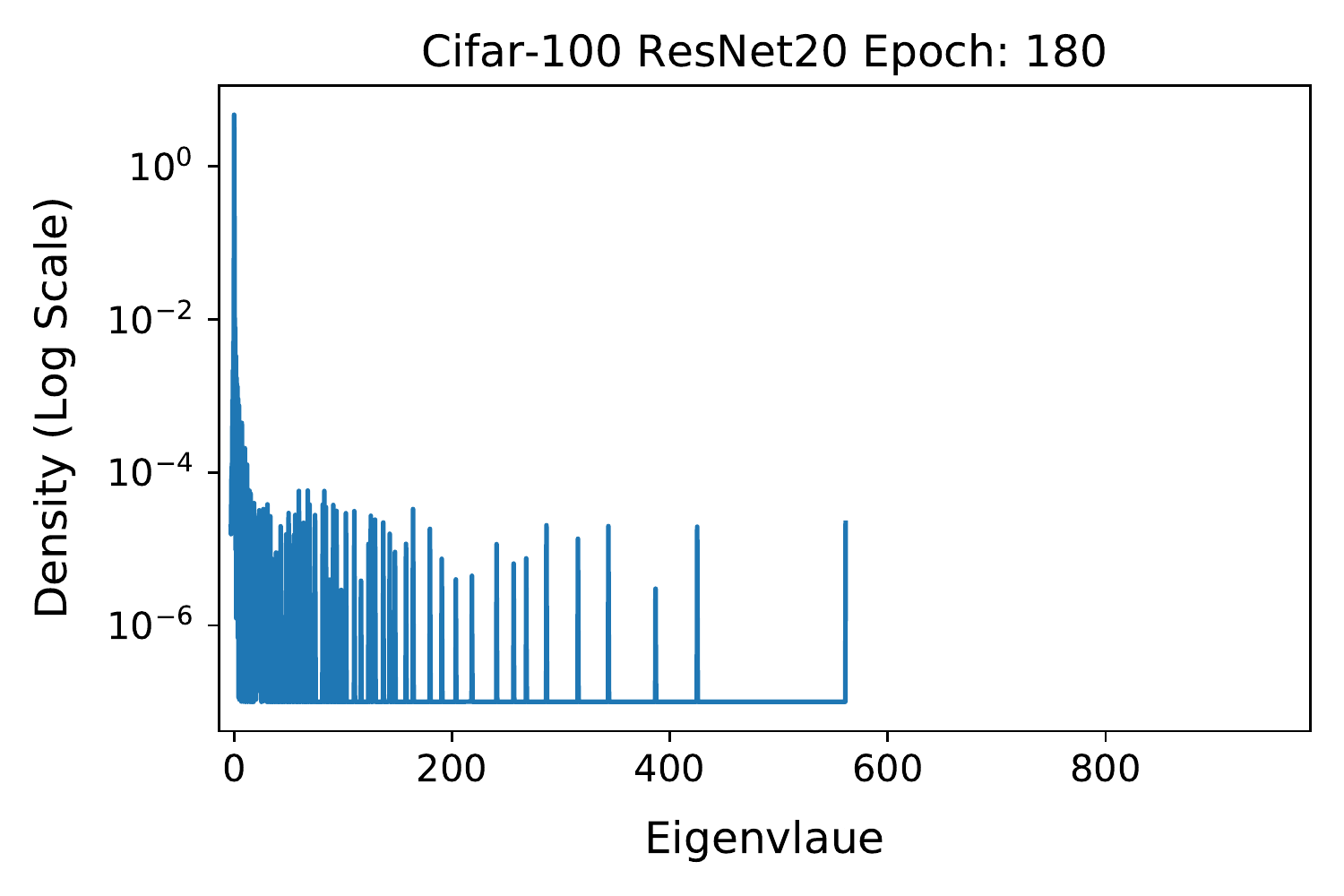}
\includegraphics[width=0.295\textwidth]{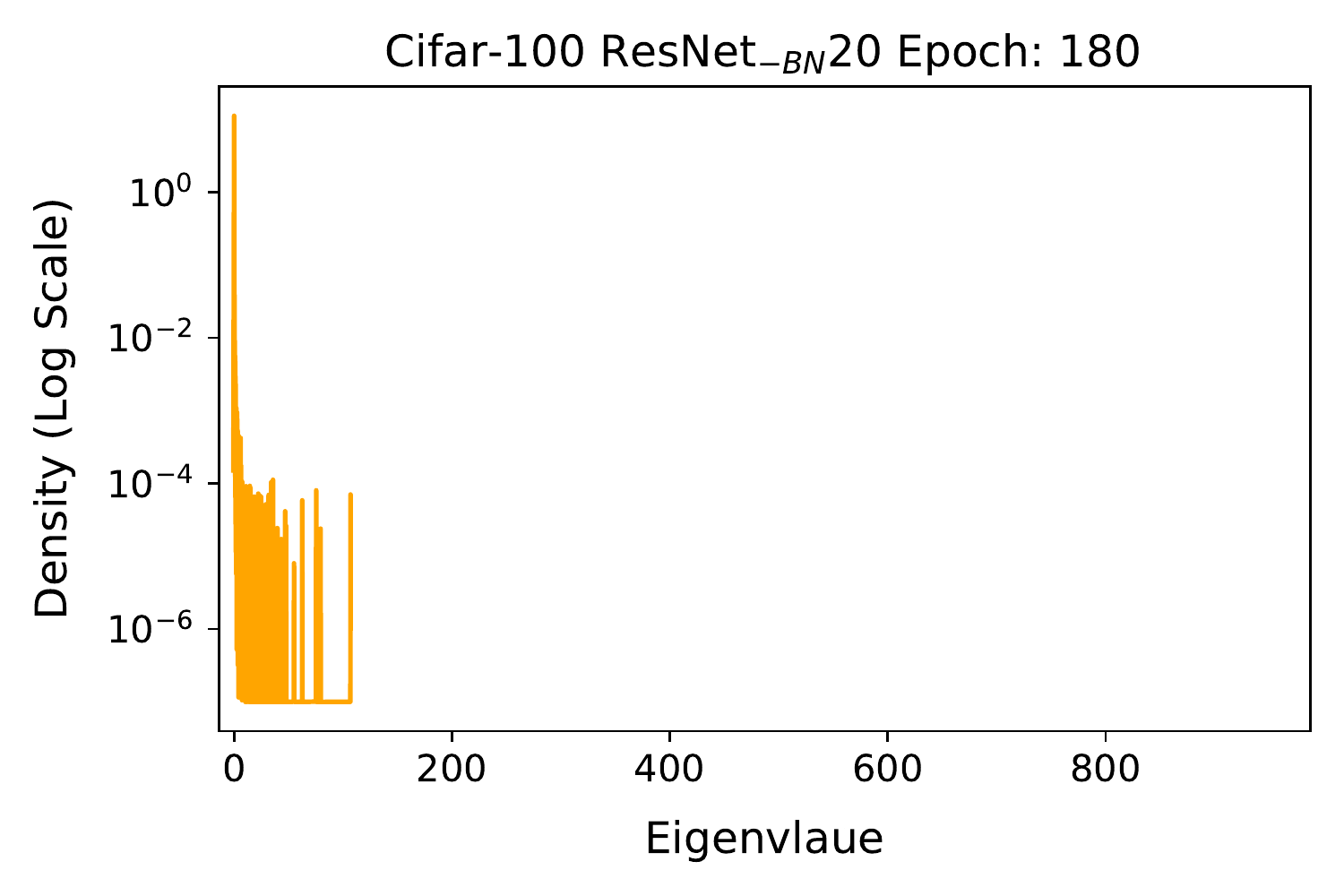}
\includegraphics[width=0.295\textwidth]{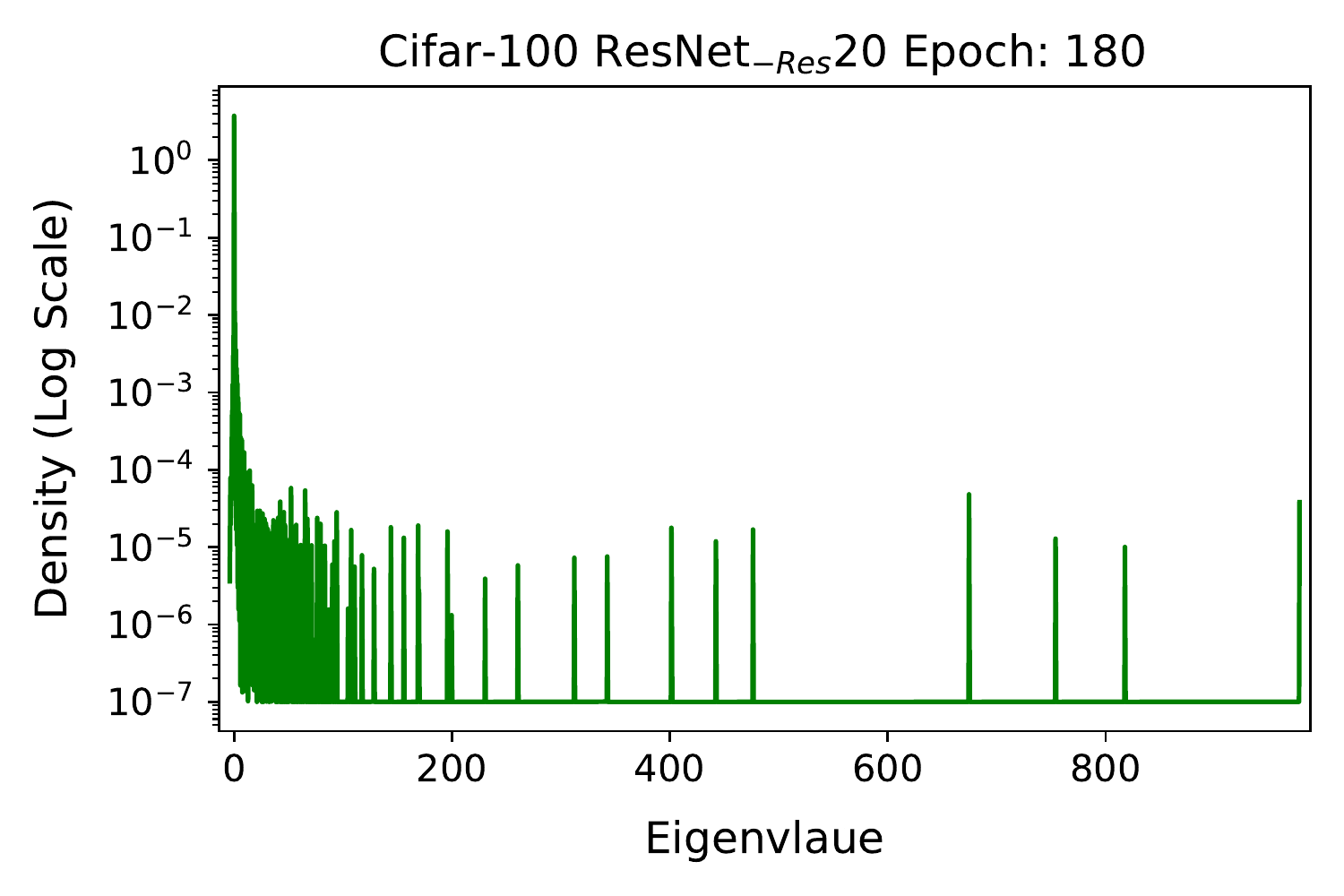}\\
\caption{
Hessian ESD of the entire network for ResNet/\ResNetBN/\ResNetRes with depth 20 on Cifar-100 with Hessian batch size 50000. 
This figure shows the Hessian ESD throughout the training process.
One notable thing here is that although \ResNetBN20 has smaller Hessian ESD support range than \ResNet20 does, the Hessian ESD of \ResNetBN20 centers around zero at very beginning (epoch 1). 
This clearly shows that training without BN is indeed~harder. 
}
  \label{fig:resnet20-slq-full-net-all-cifar100}
\end{figure*}

\begin{figure*}[!htbp]
\centering
\includegraphics[width=0.295\textwidth]{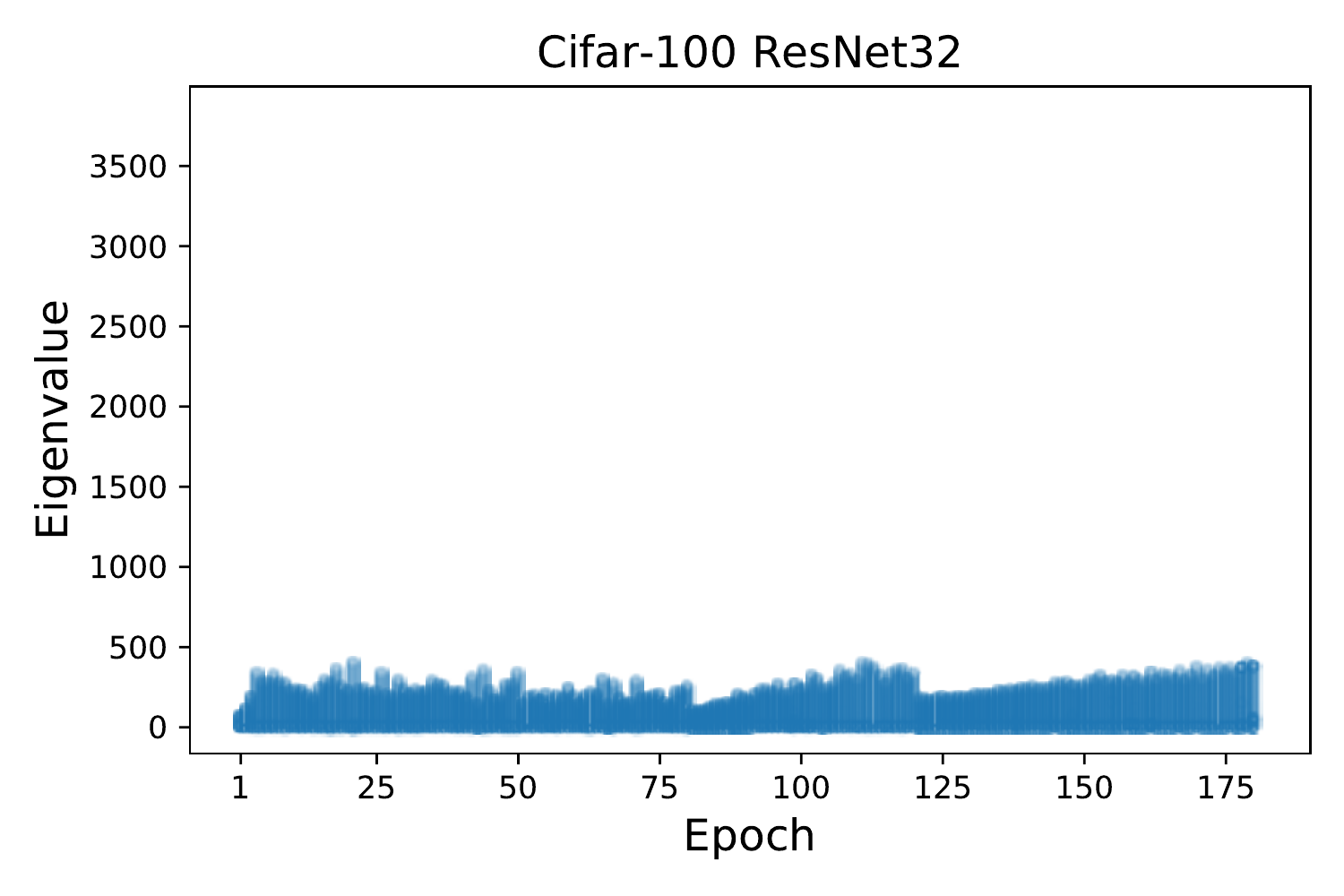}
\includegraphics[width=0.295\textwidth]{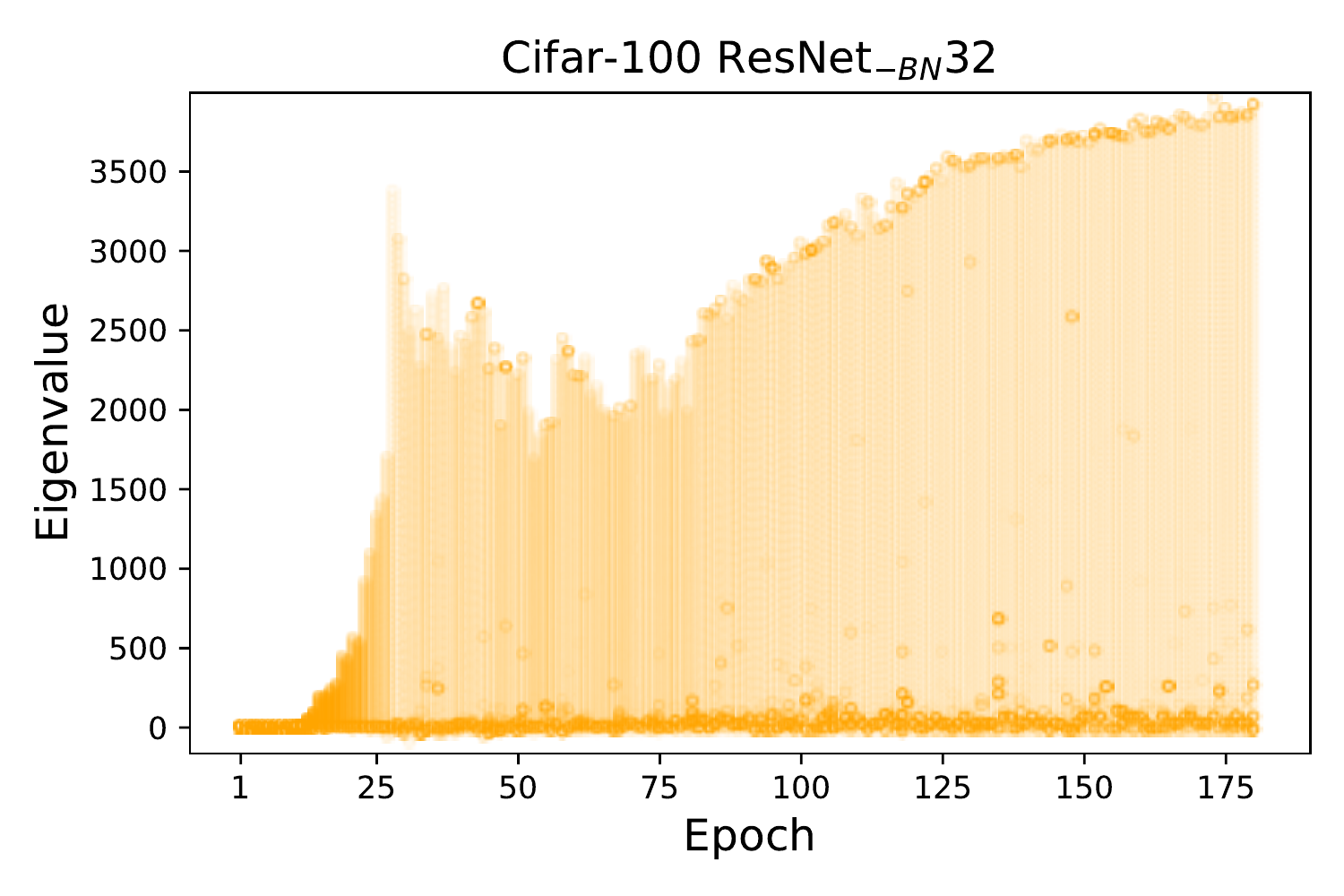}
\includegraphics[width=0.295\textwidth]{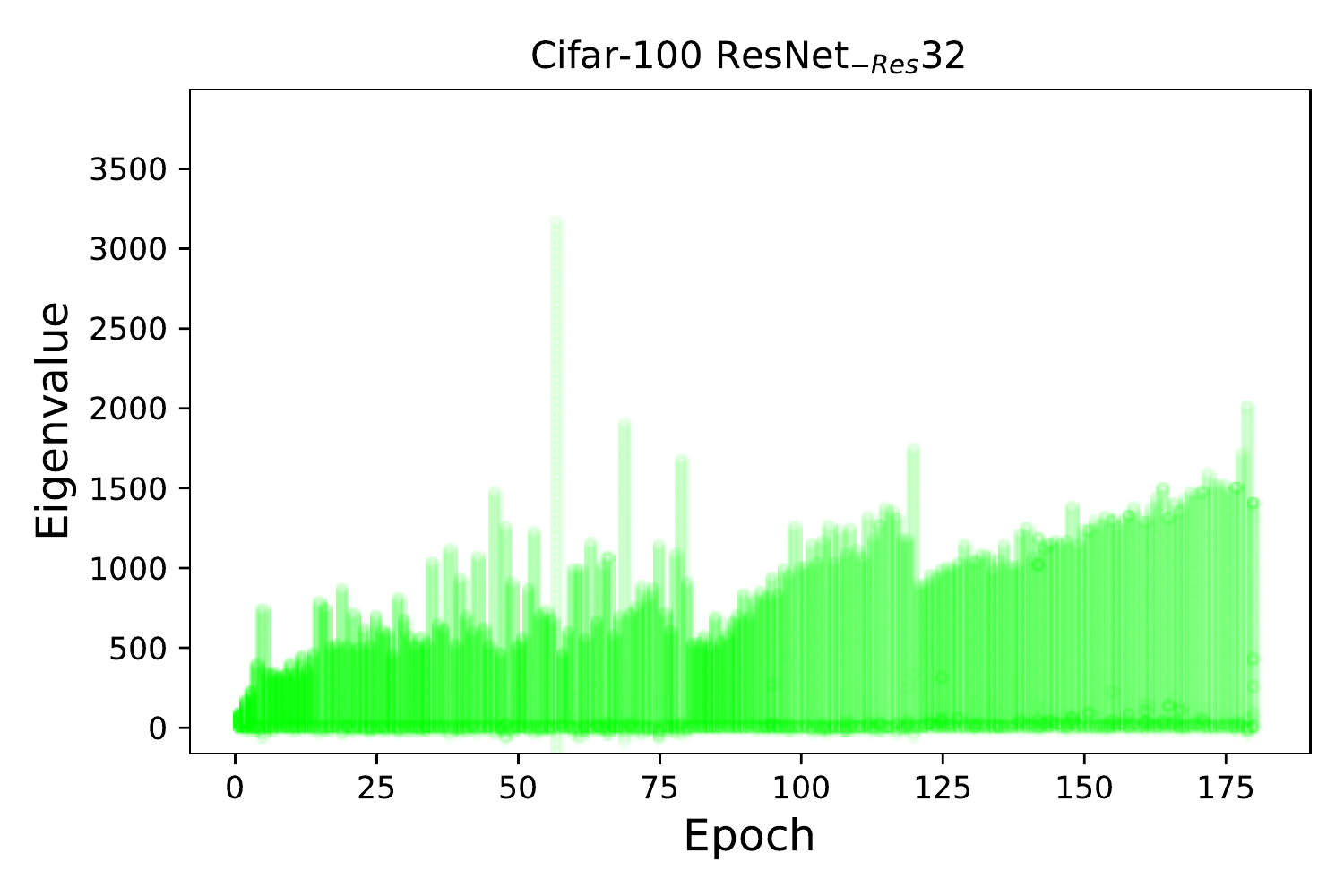}\\
\includegraphics[width=0.295\textwidth]{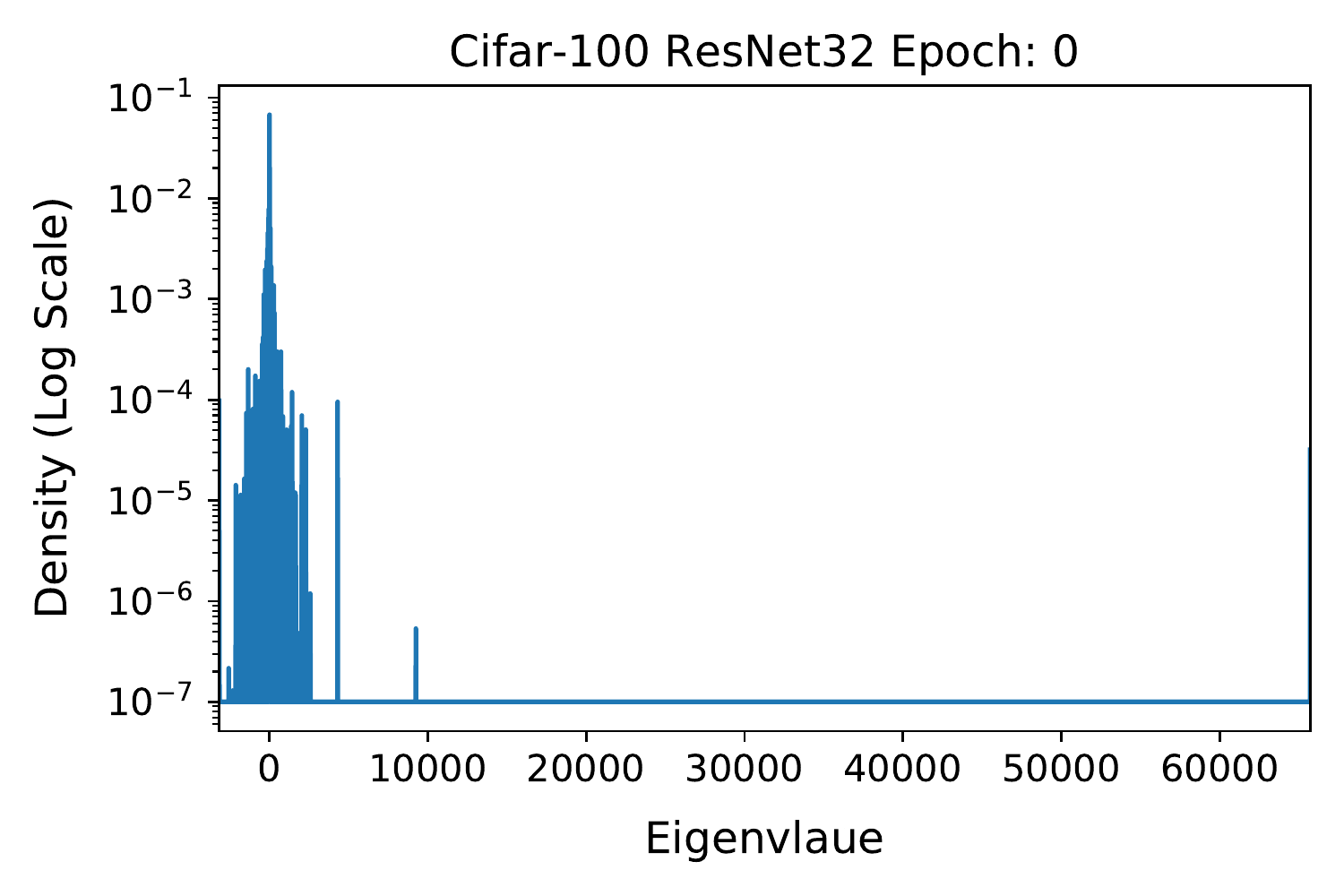}
\includegraphics[width=0.295\textwidth]{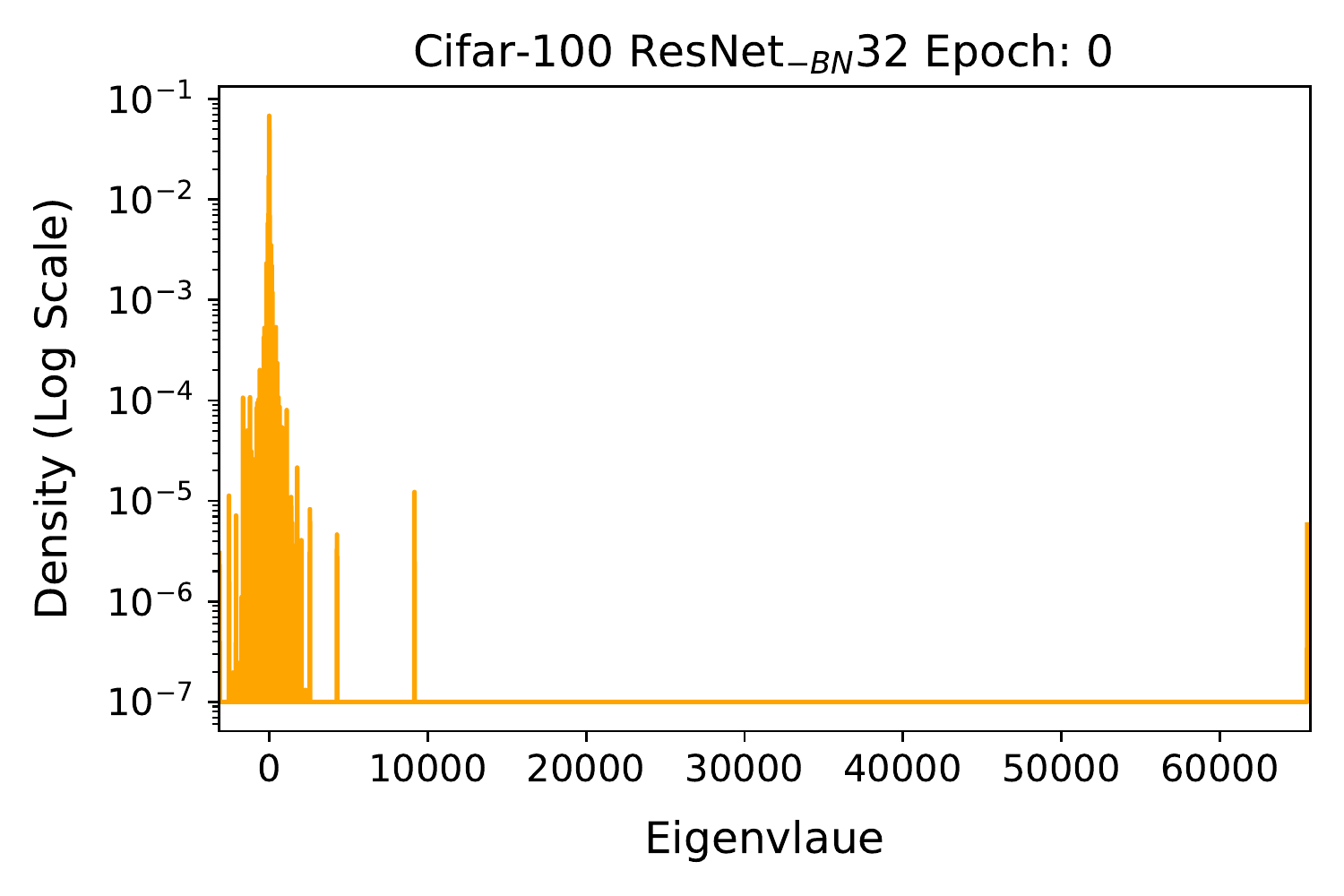}
\includegraphics[width=0.295\textwidth]{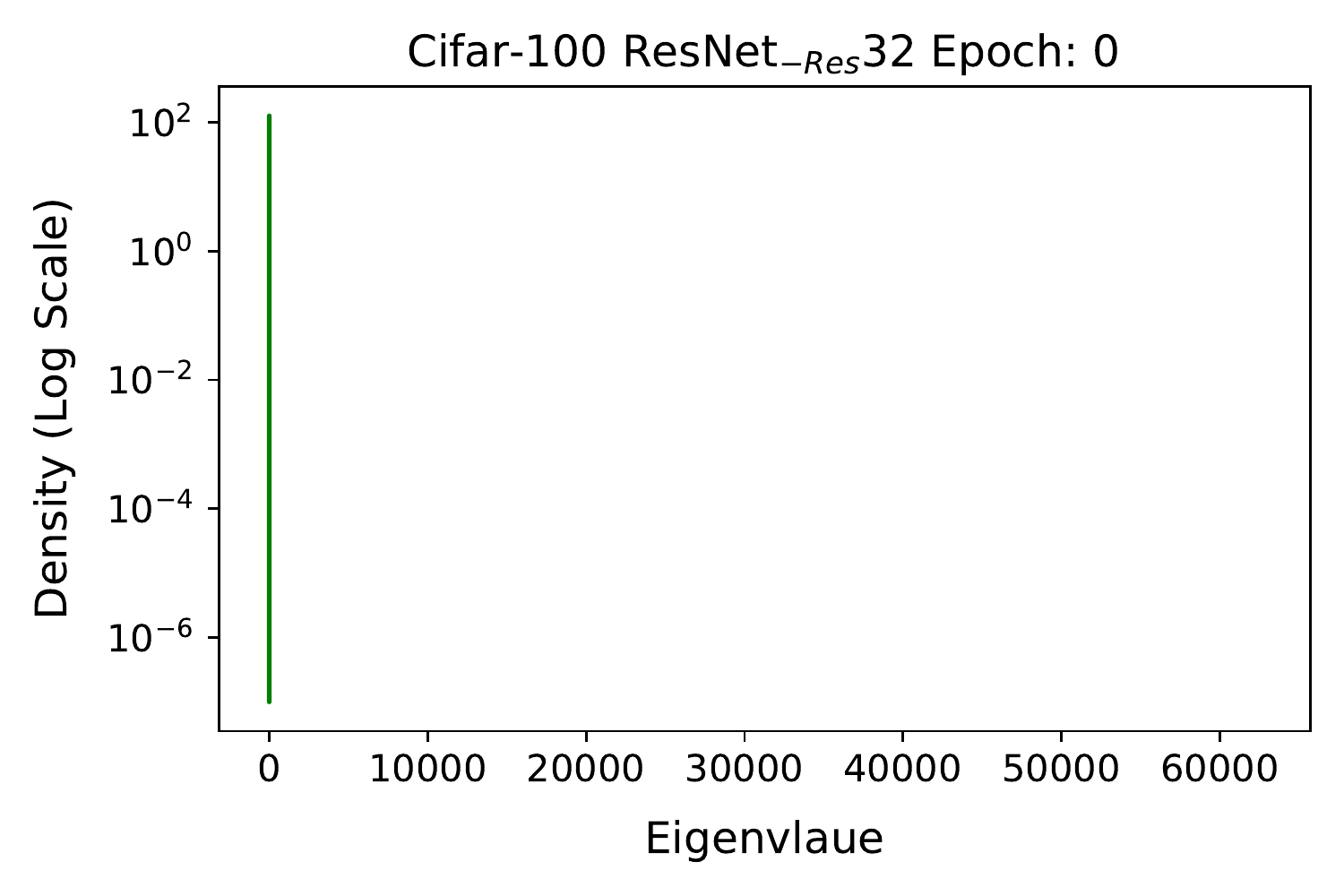}\\
\includegraphics[width=0.295\textwidth]{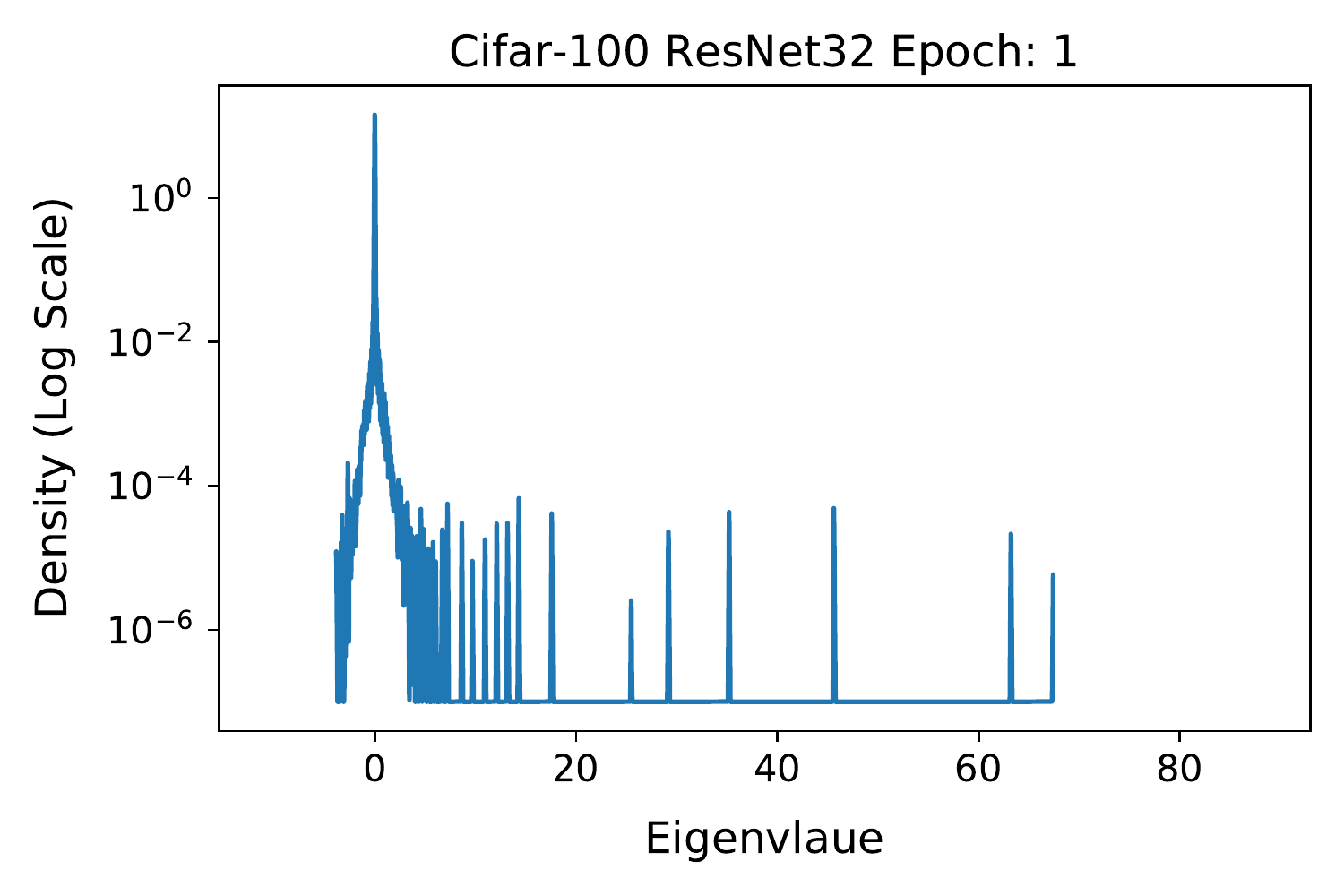}
\includegraphics[width=0.295\textwidth]{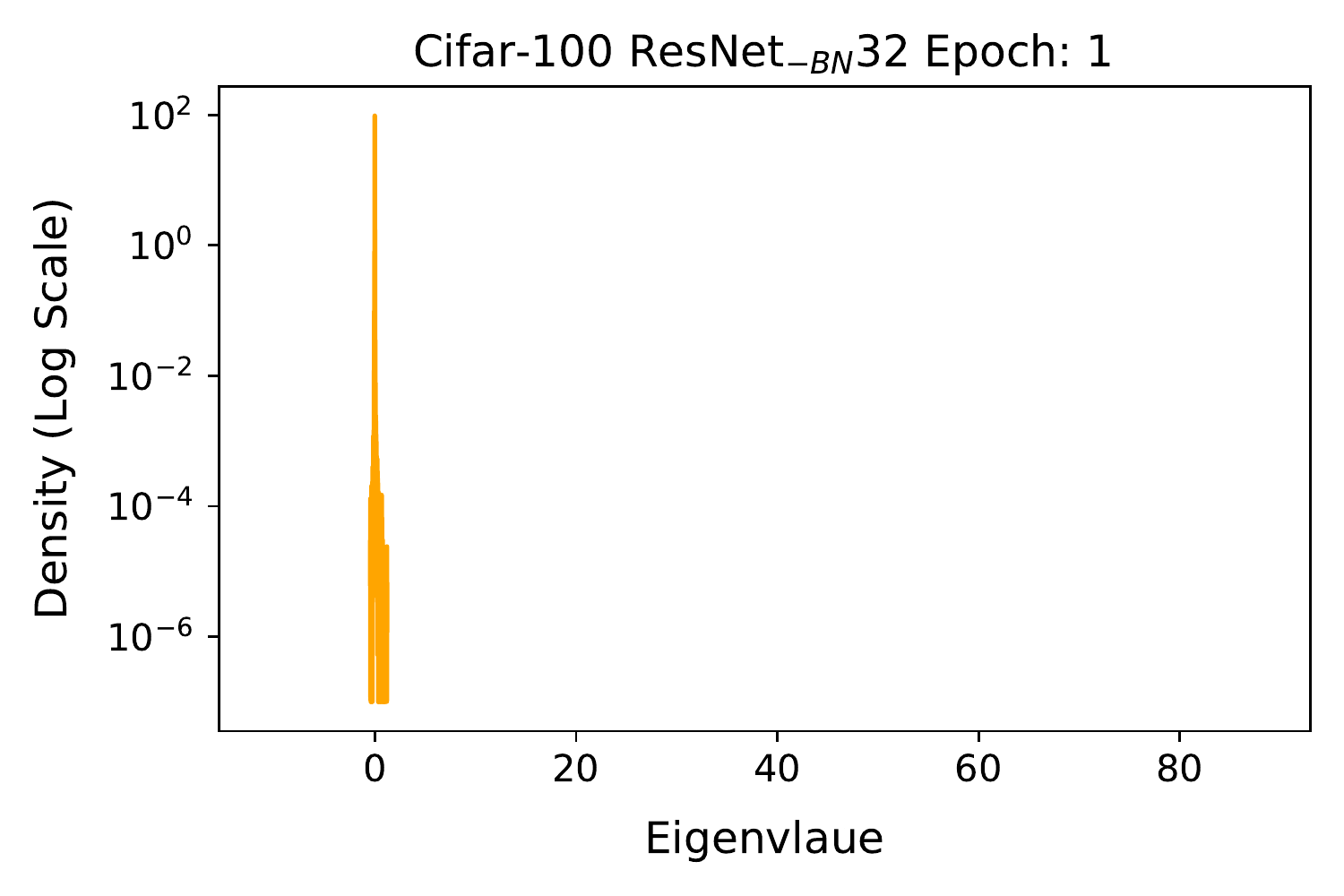}
\includegraphics[width=0.295\textwidth]{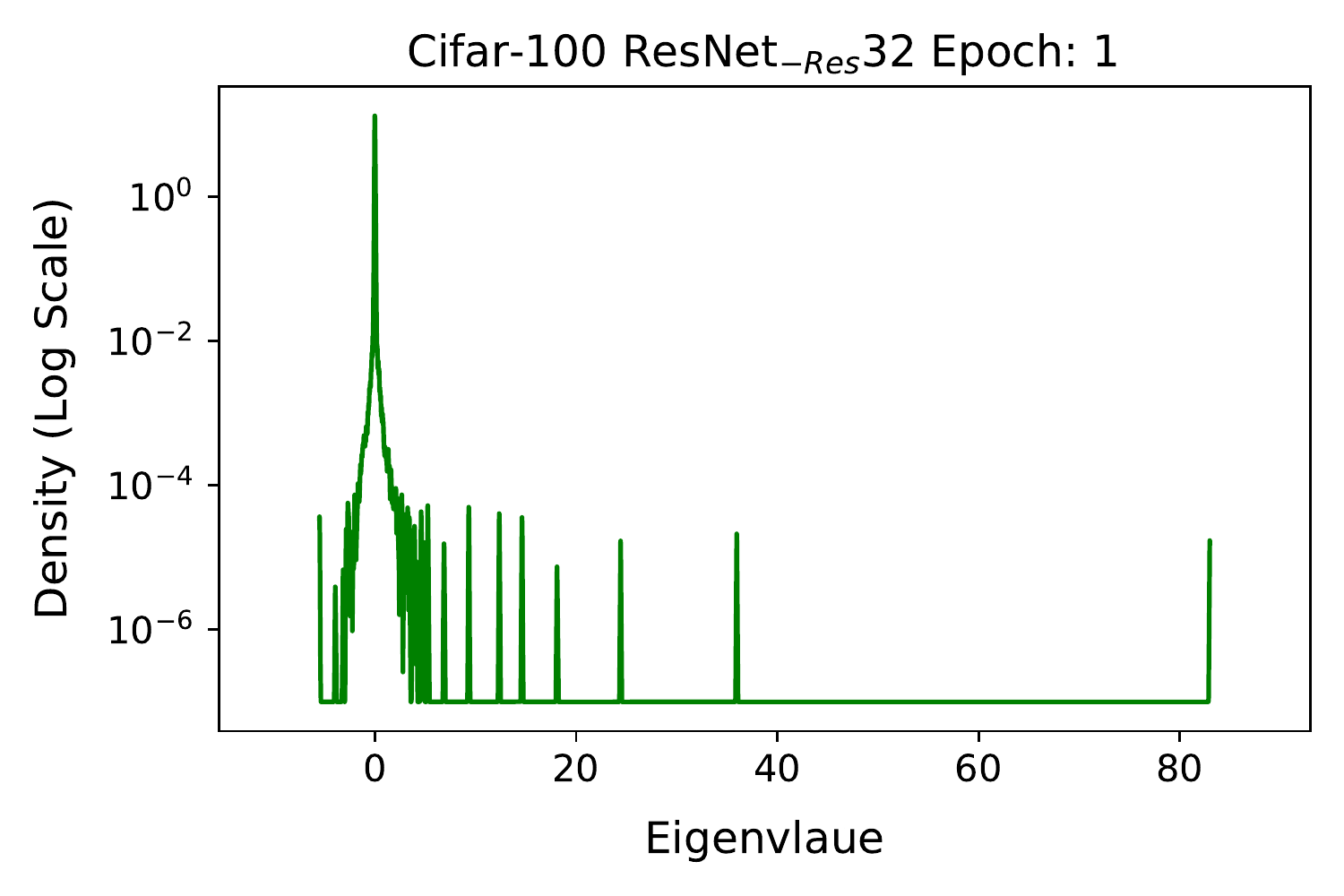}\\
\includegraphics[width=0.295\textwidth]{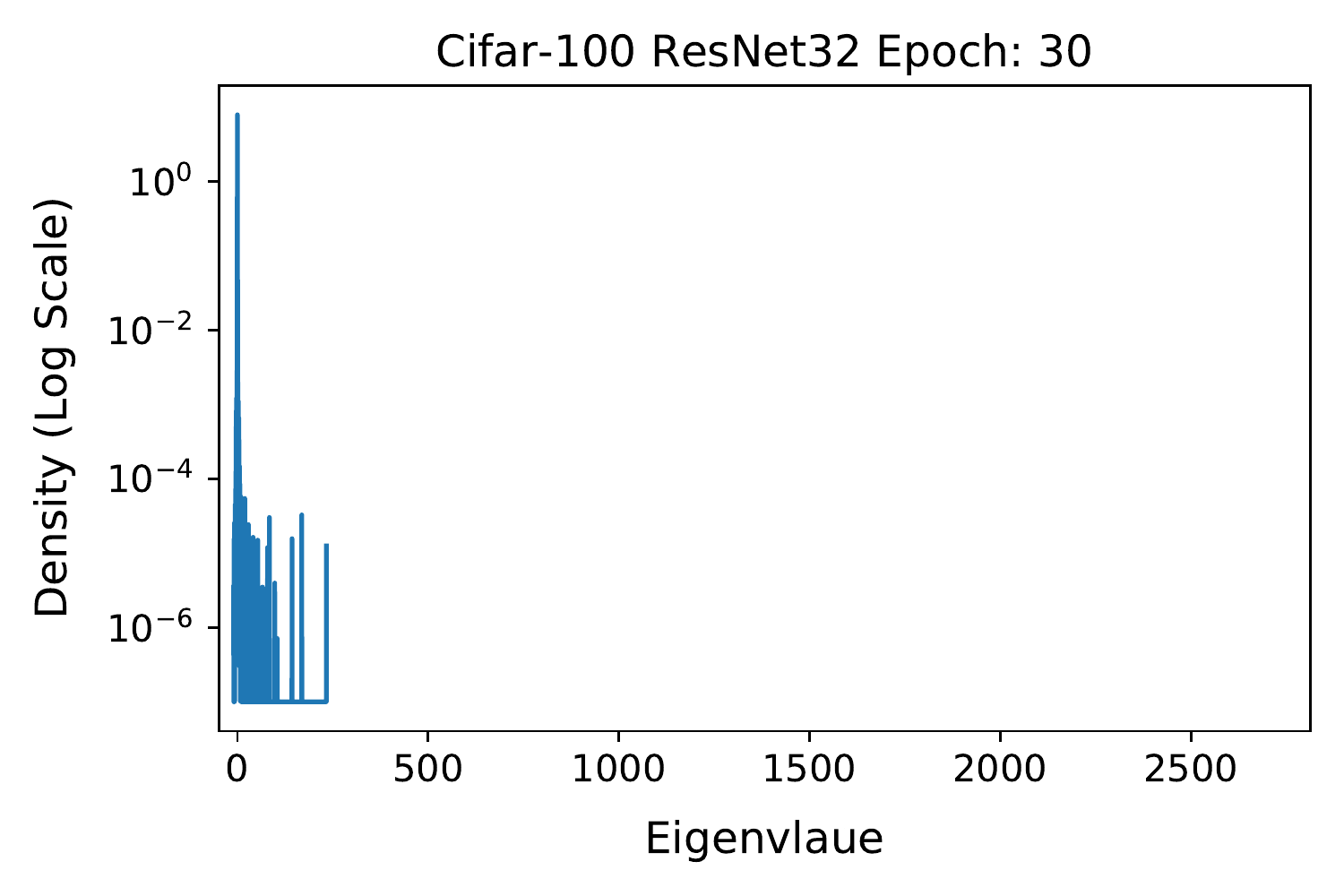}
\includegraphics[width=0.295\textwidth]{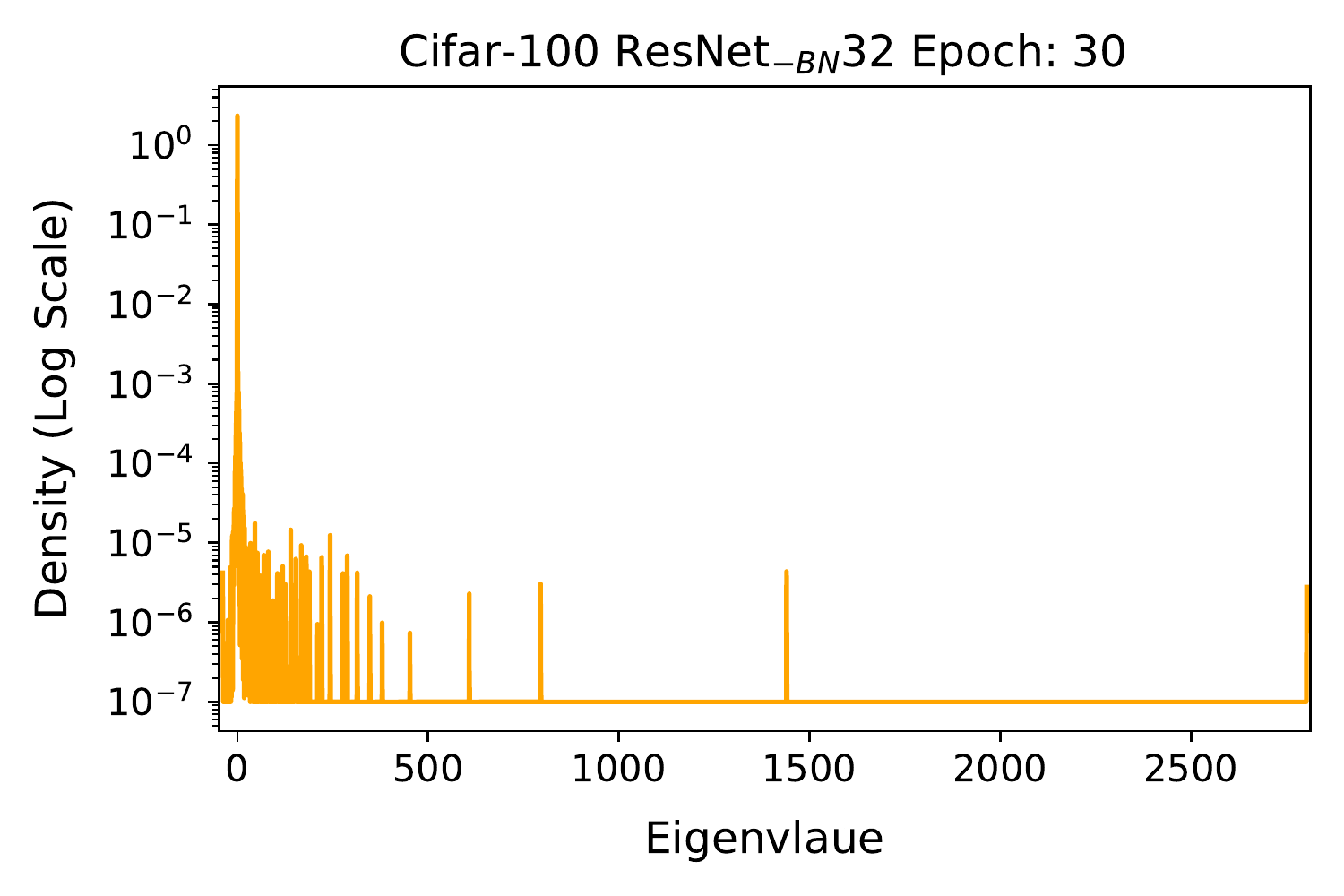}
\includegraphics[width=0.295\textwidth]{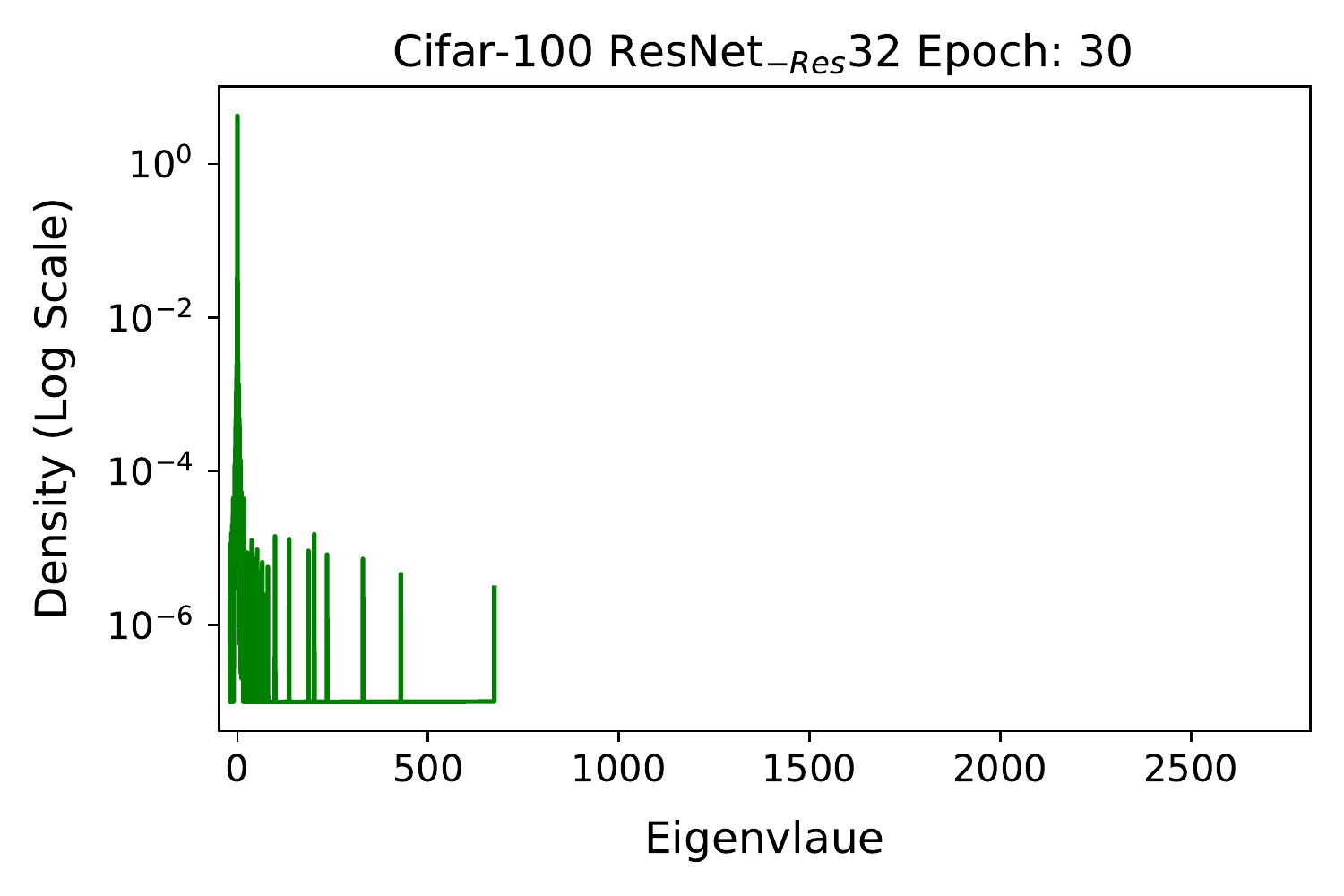}\\
\includegraphics[width=0.295\textwidth]{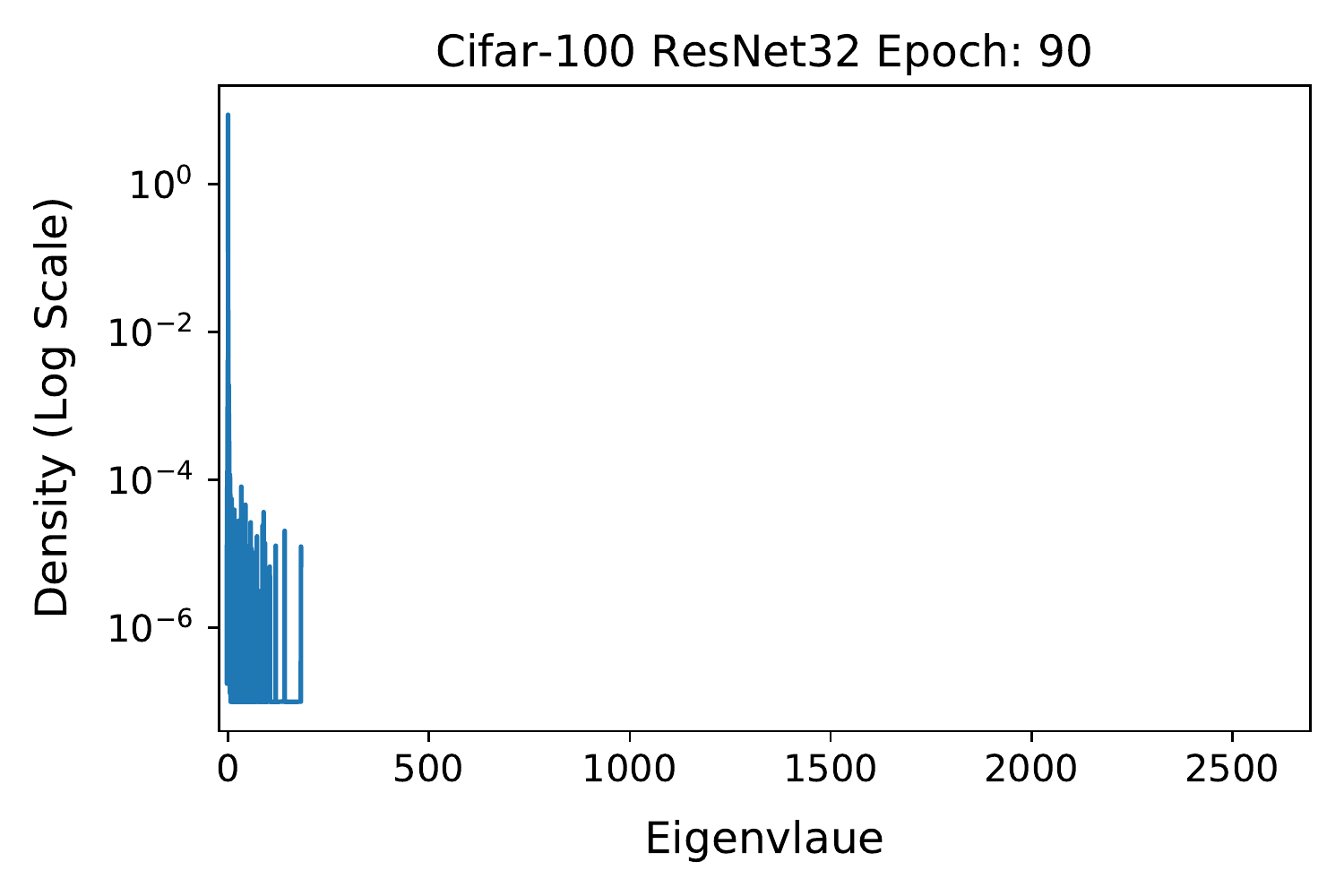}
\includegraphics[width=0.295\textwidth]{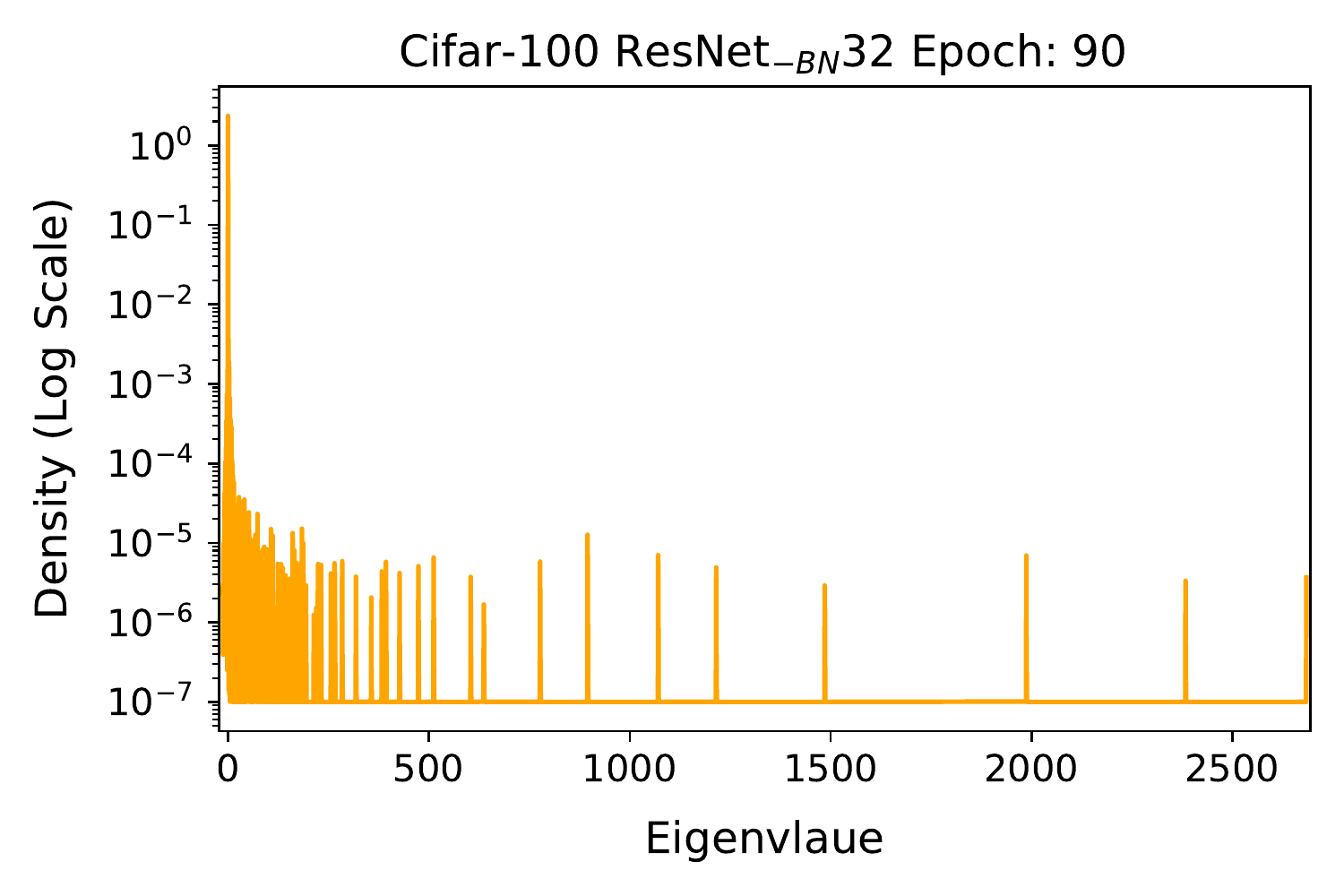}
\includegraphics[width=0.295\textwidth]{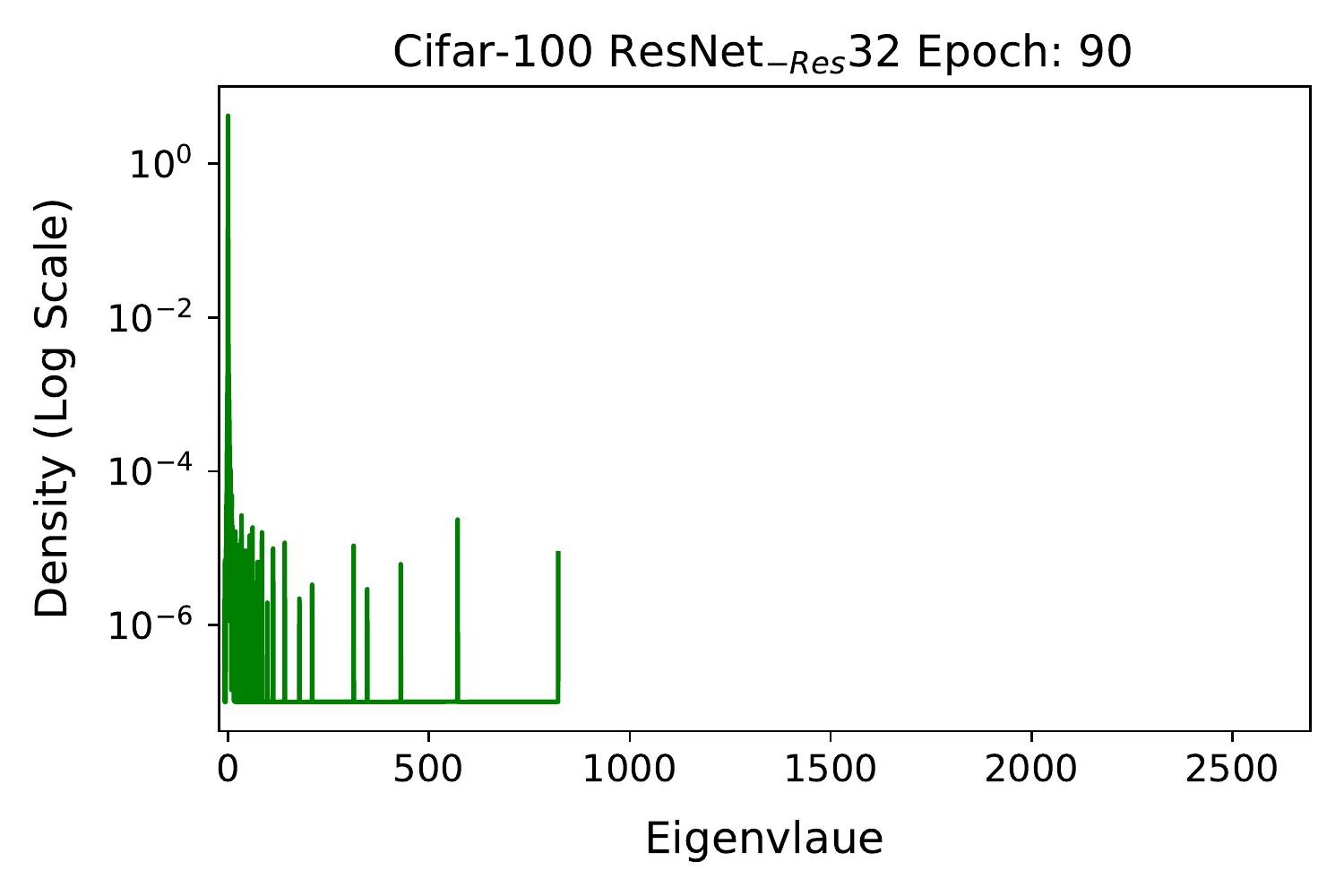}\\
\includegraphics[width=0.295\textwidth]{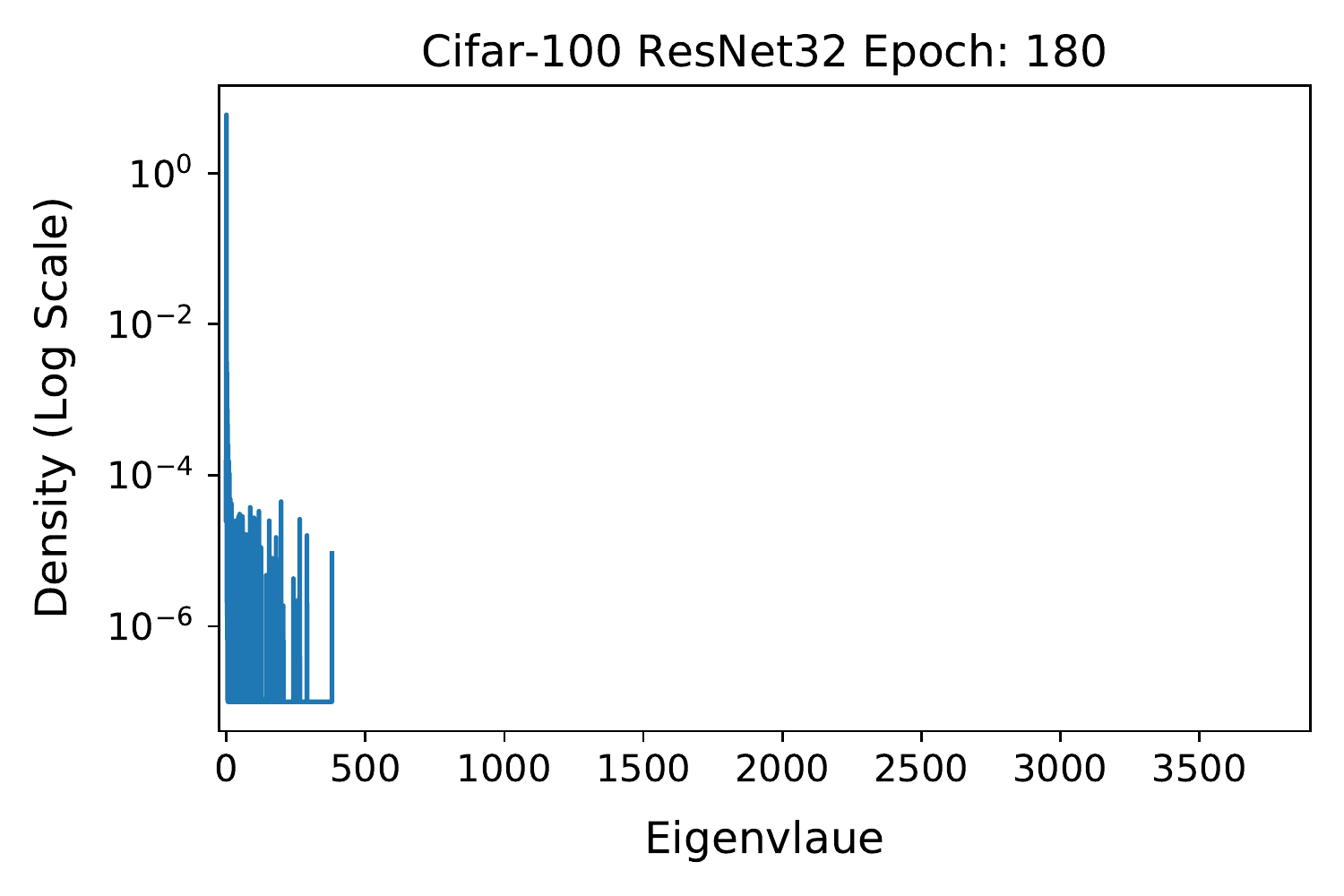}
\includegraphics[width=0.295\textwidth]{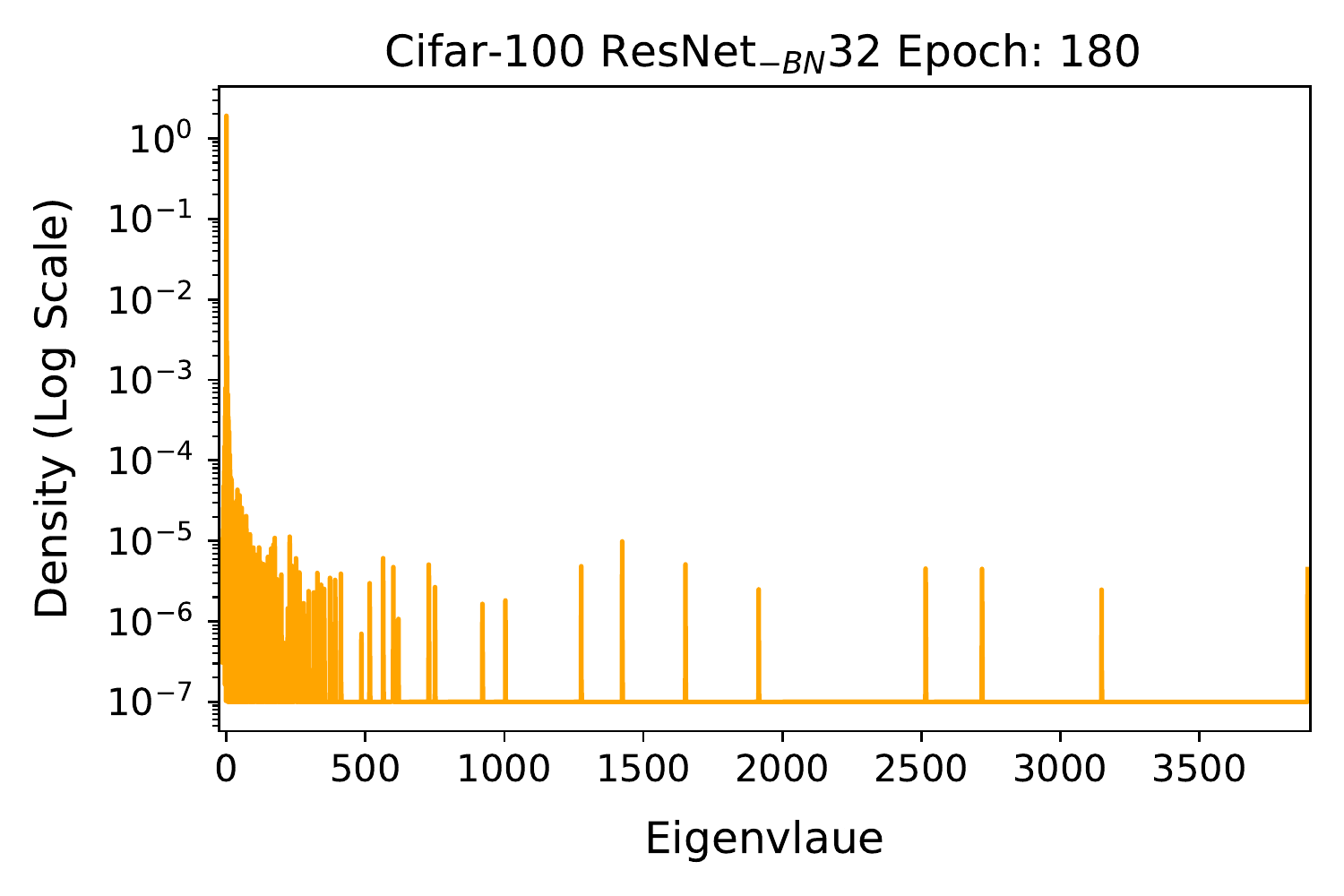}
\includegraphics[width=0.295\textwidth]{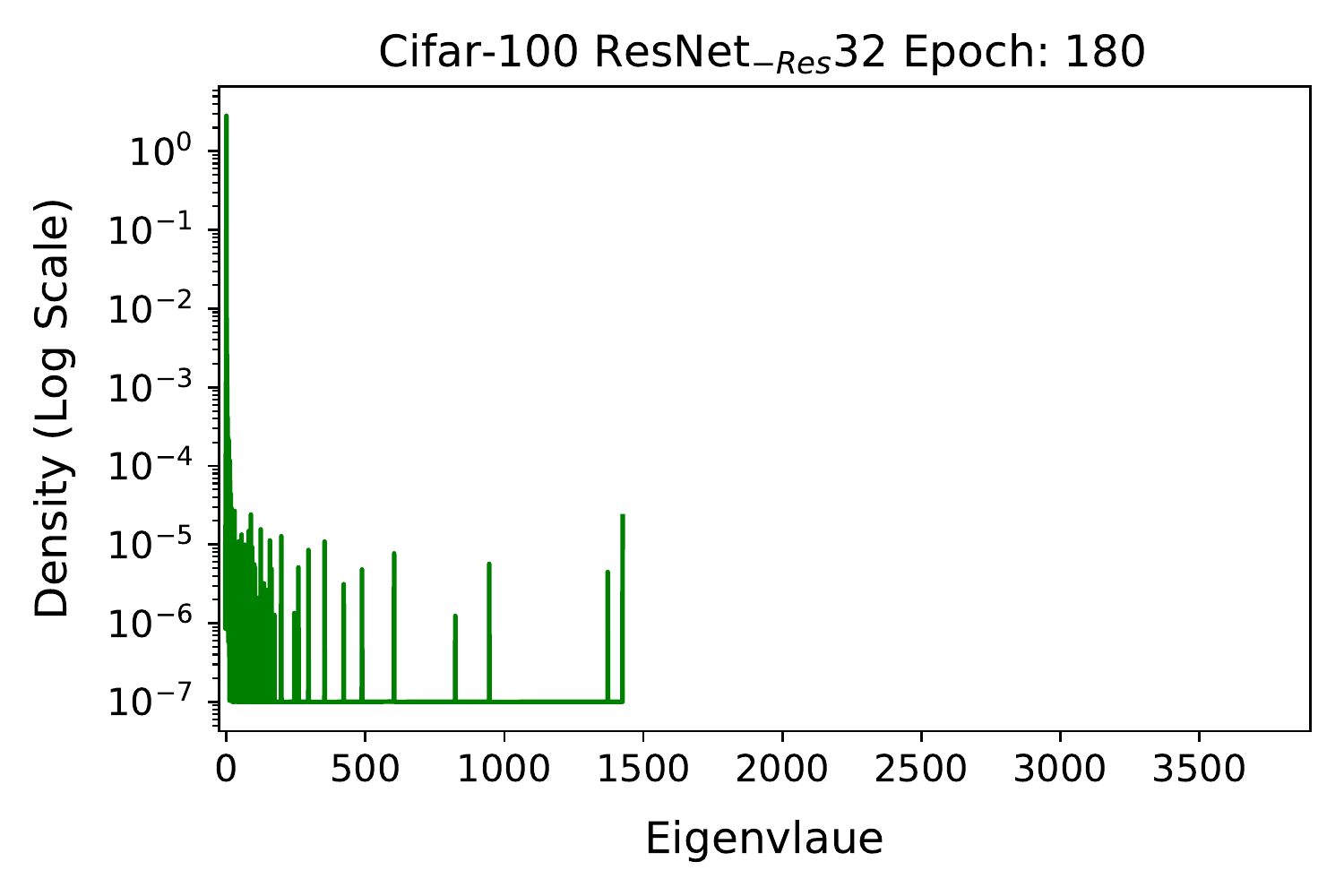}\\
\caption{
Hessian ESD of the entire network for ResNet/\ResNetBN/\ResNetRes with depth 32 on Cifar-100 with Hessian batch size 50000. 
This figure shows the Hessian ESD throughout the training process.
One notable thing here is that the Hessian ESD of \ResNetBN32 centers around zero (at least) until epoch 5. 
This clearly shows that training without BN is indeed harder. 
}
  \label{fig:resnet32-slq-full-net-all-cifar100}
\end{figure*}

\begin{figure*}[!htbp]
\centering
\includegraphics[width=0.295\textwidth]{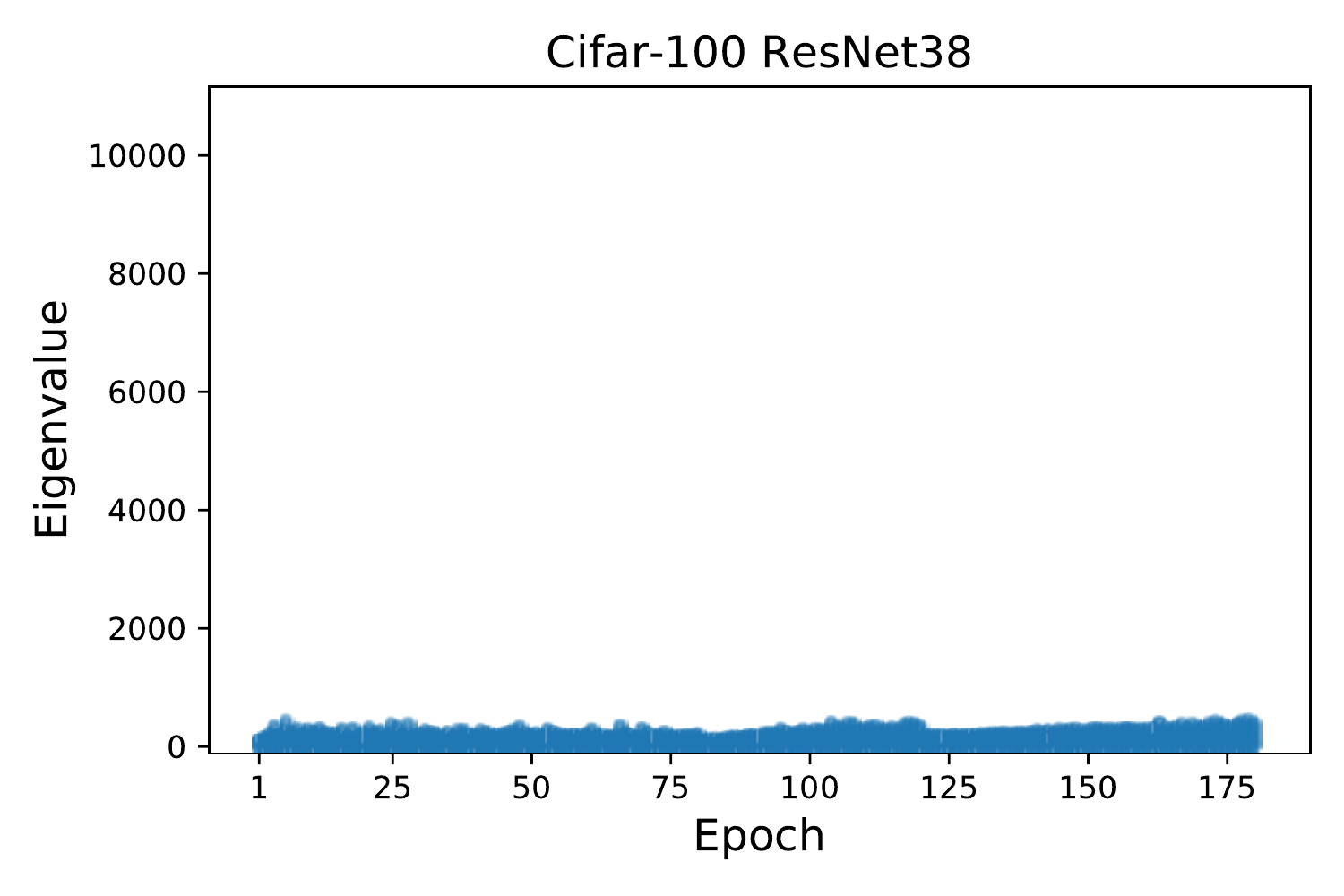}
\includegraphics[width=0.295\textwidth]{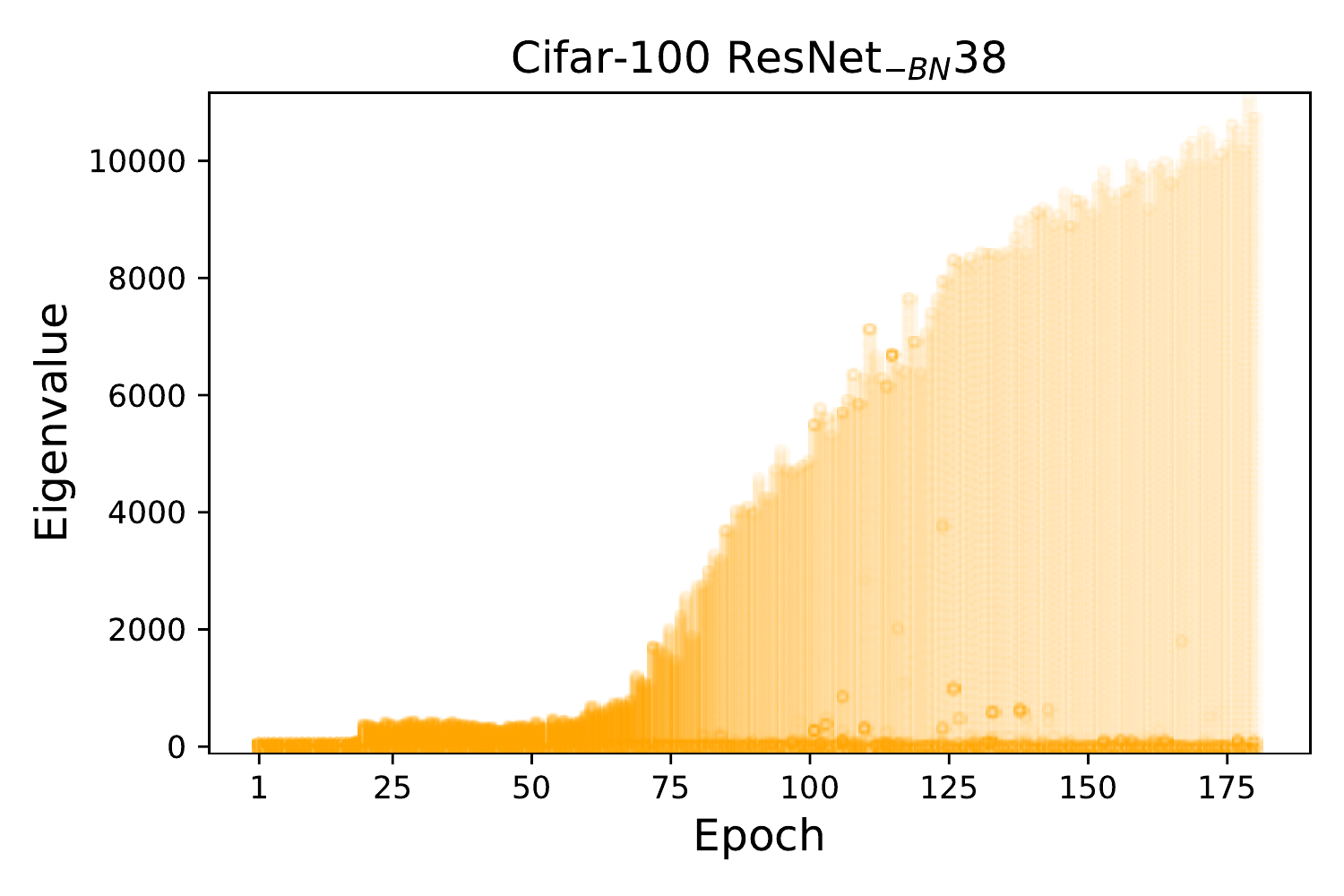}
\includegraphics[width=0.295\textwidth]{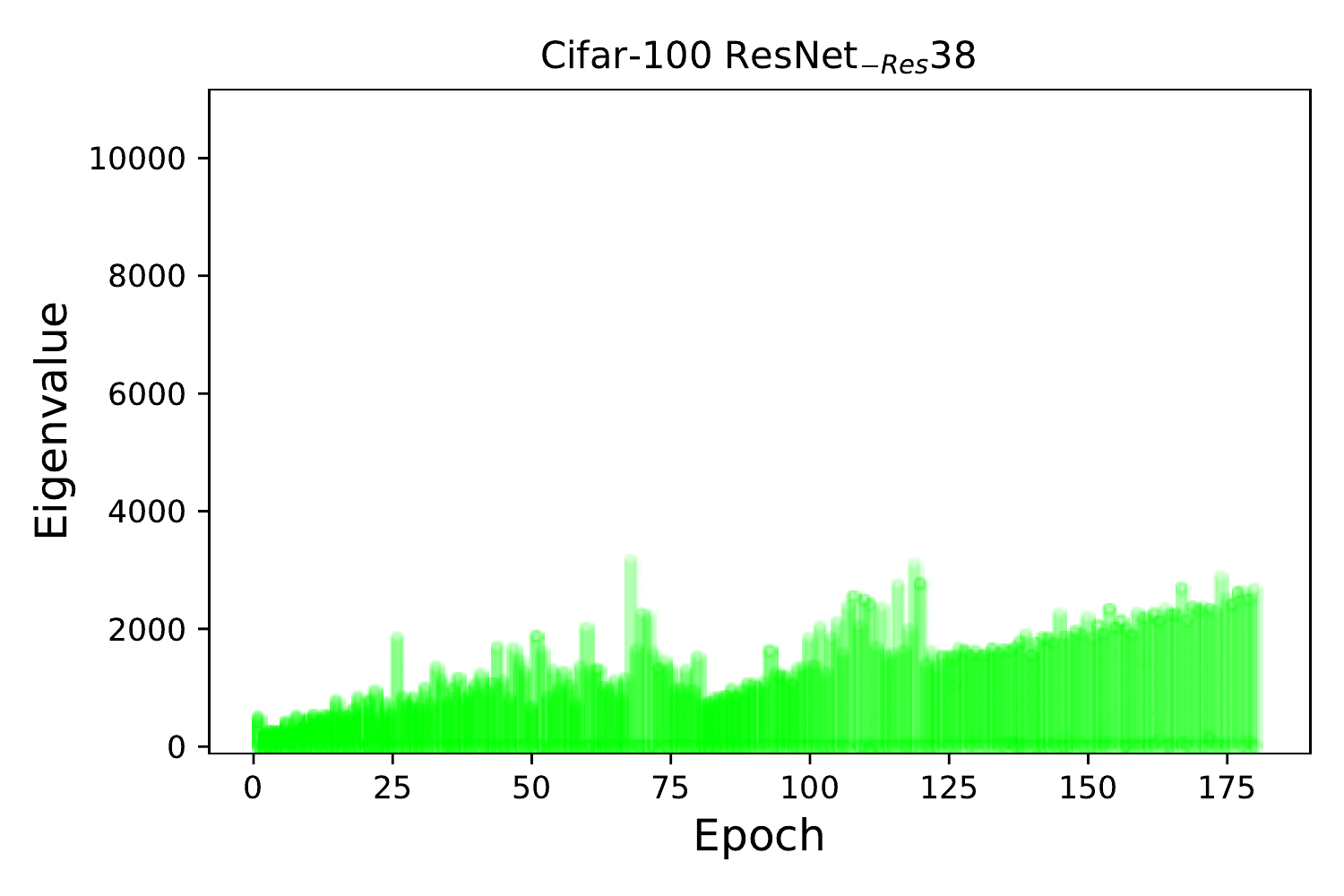}\\
\includegraphics[width=0.295\textwidth]{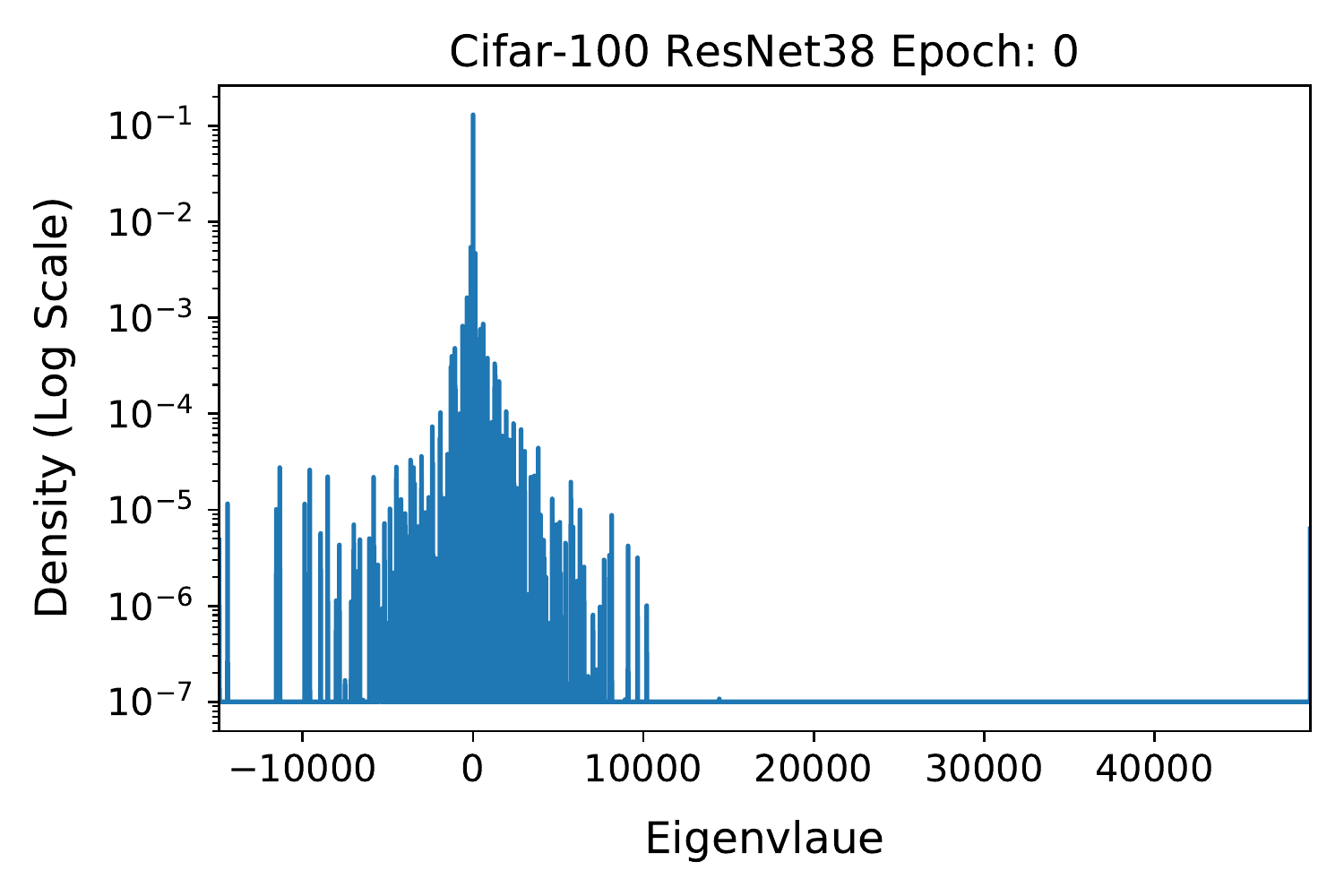}
\includegraphics[width=0.295\textwidth]{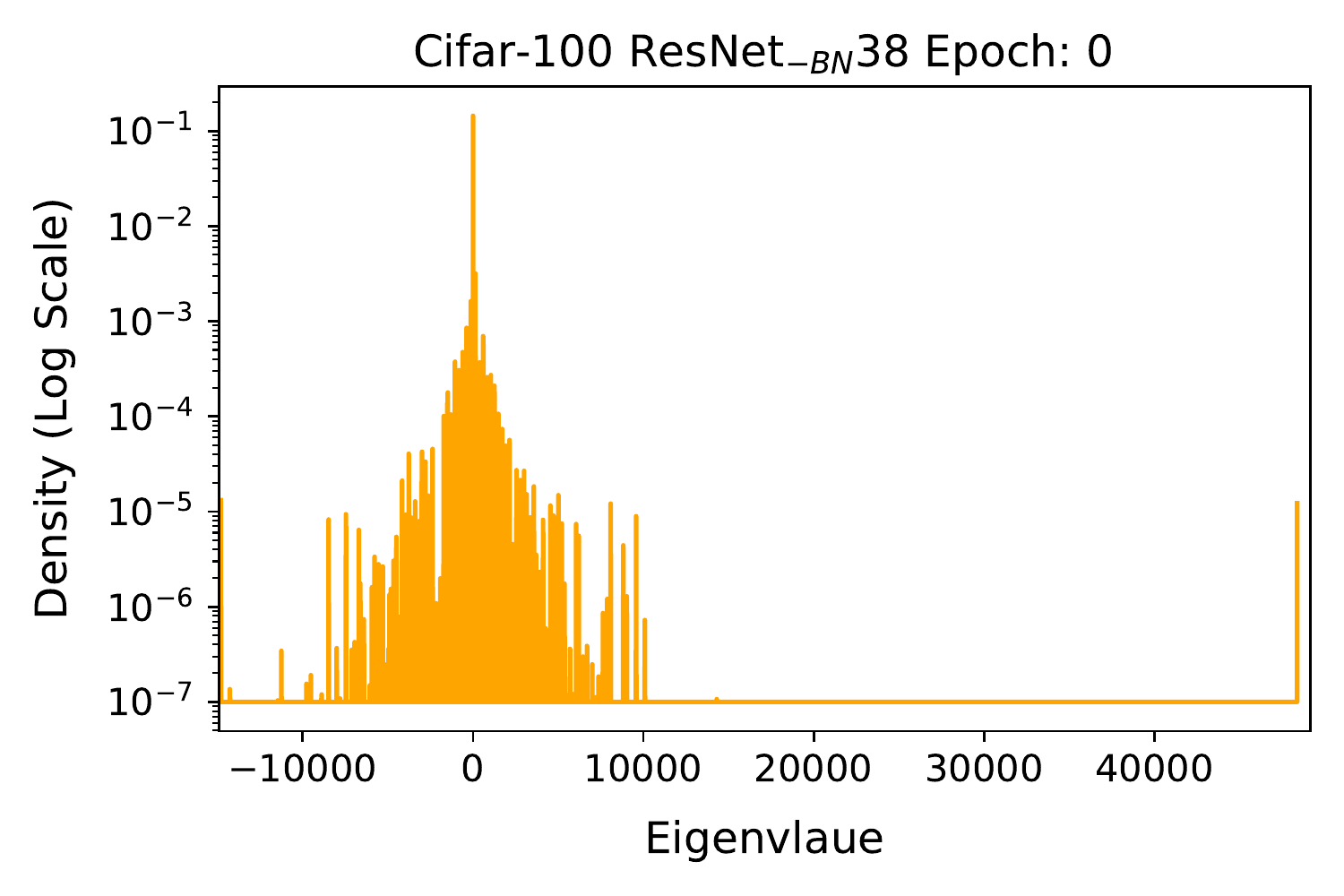}
\includegraphics[width=0.295\textwidth]{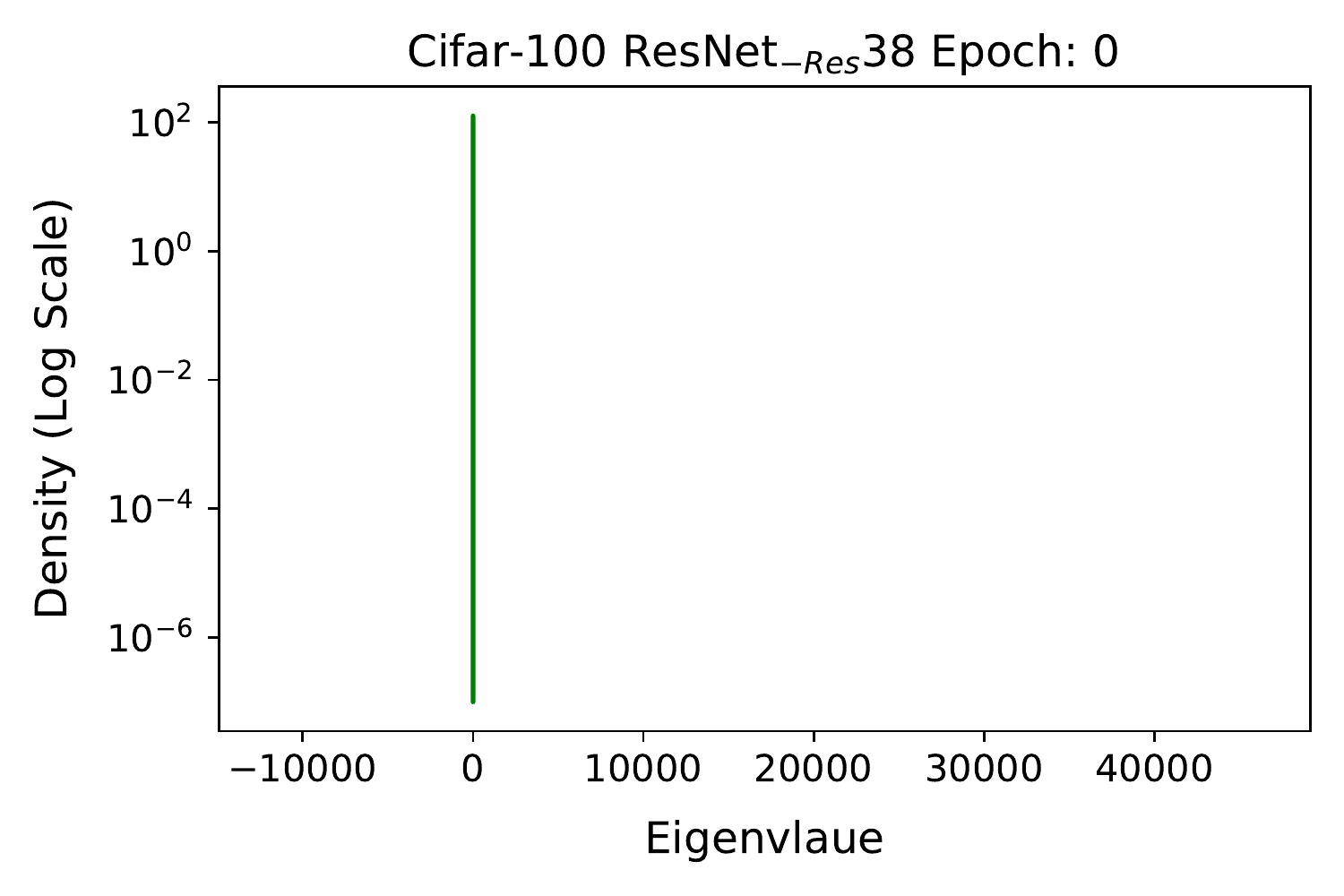}\\
\includegraphics[width=0.295\textwidth]{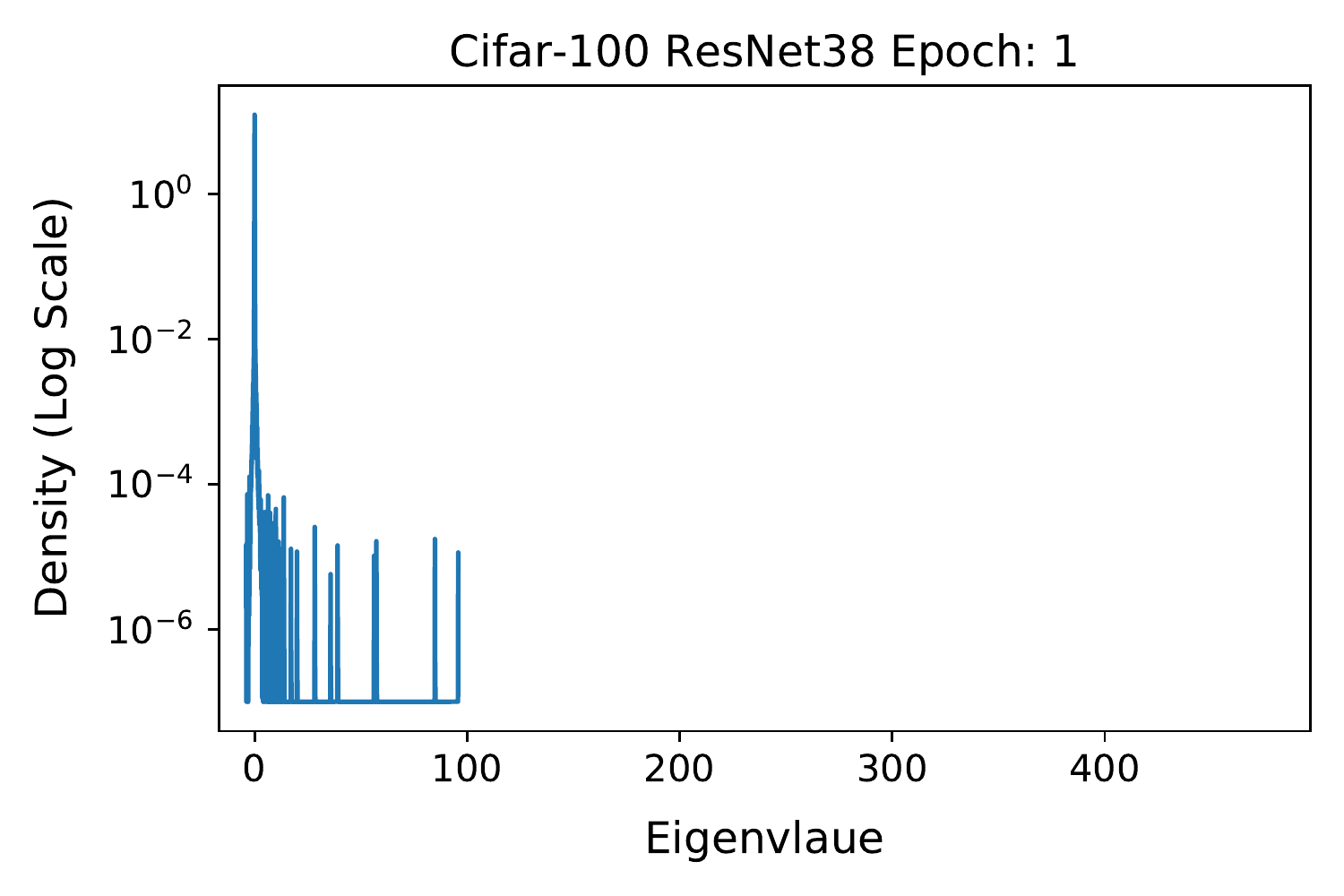}
\includegraphics[width=0.295\textwidth]{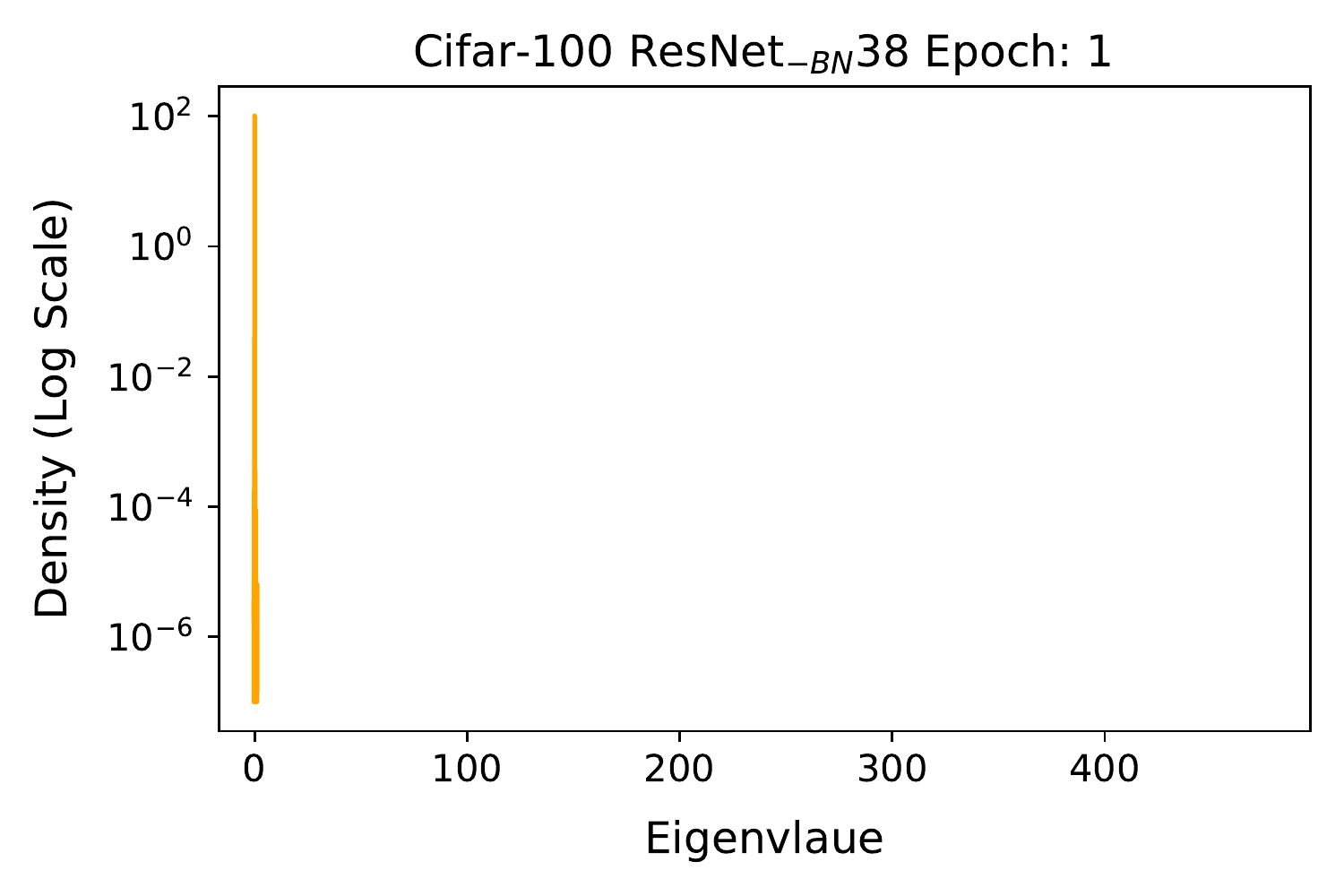}
\includegraphics[width=0.295\textwidth]{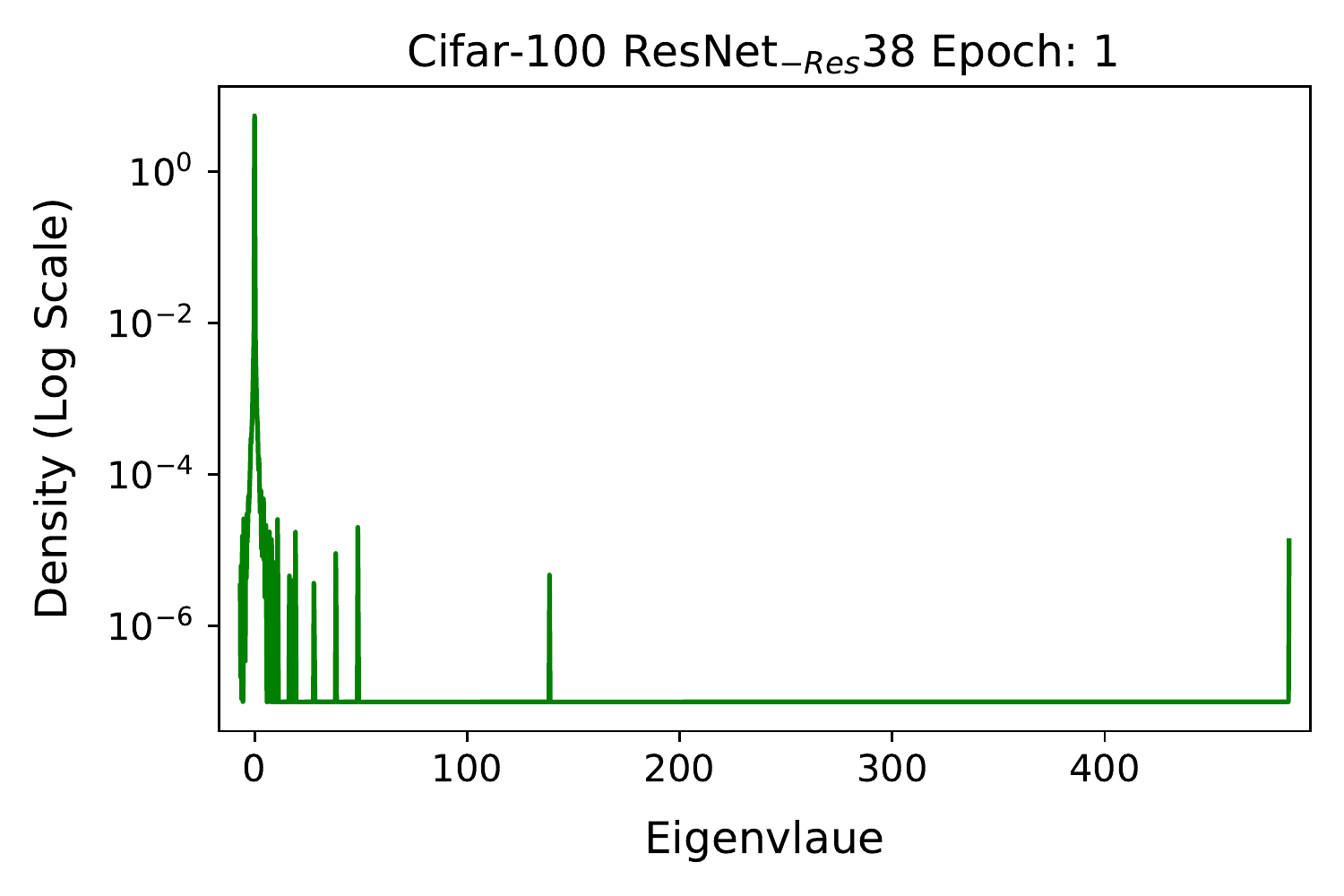}\\
\includegraphics[width=0.295\textwidth]{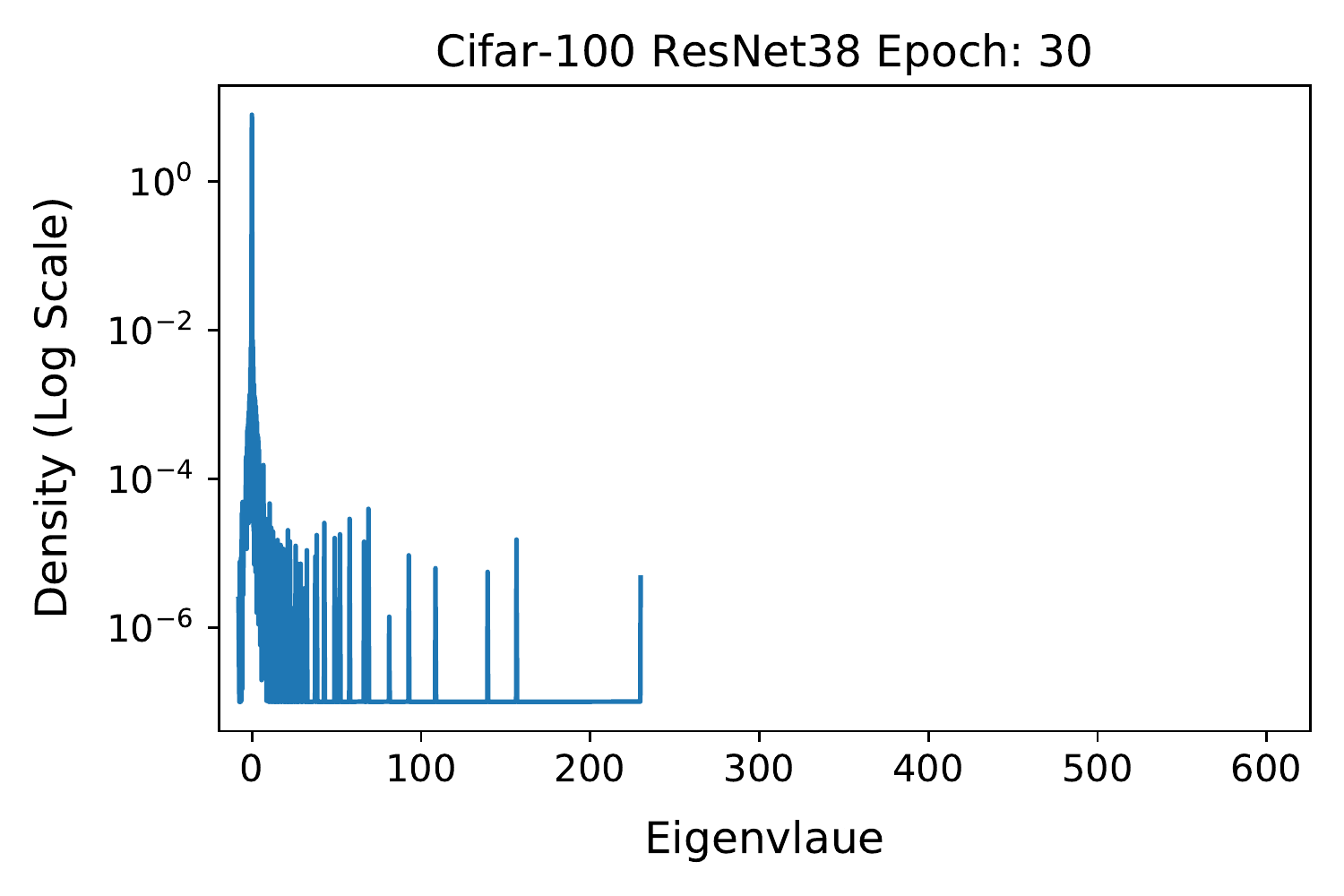}
\includegraphics[width=0.295\textwidth]{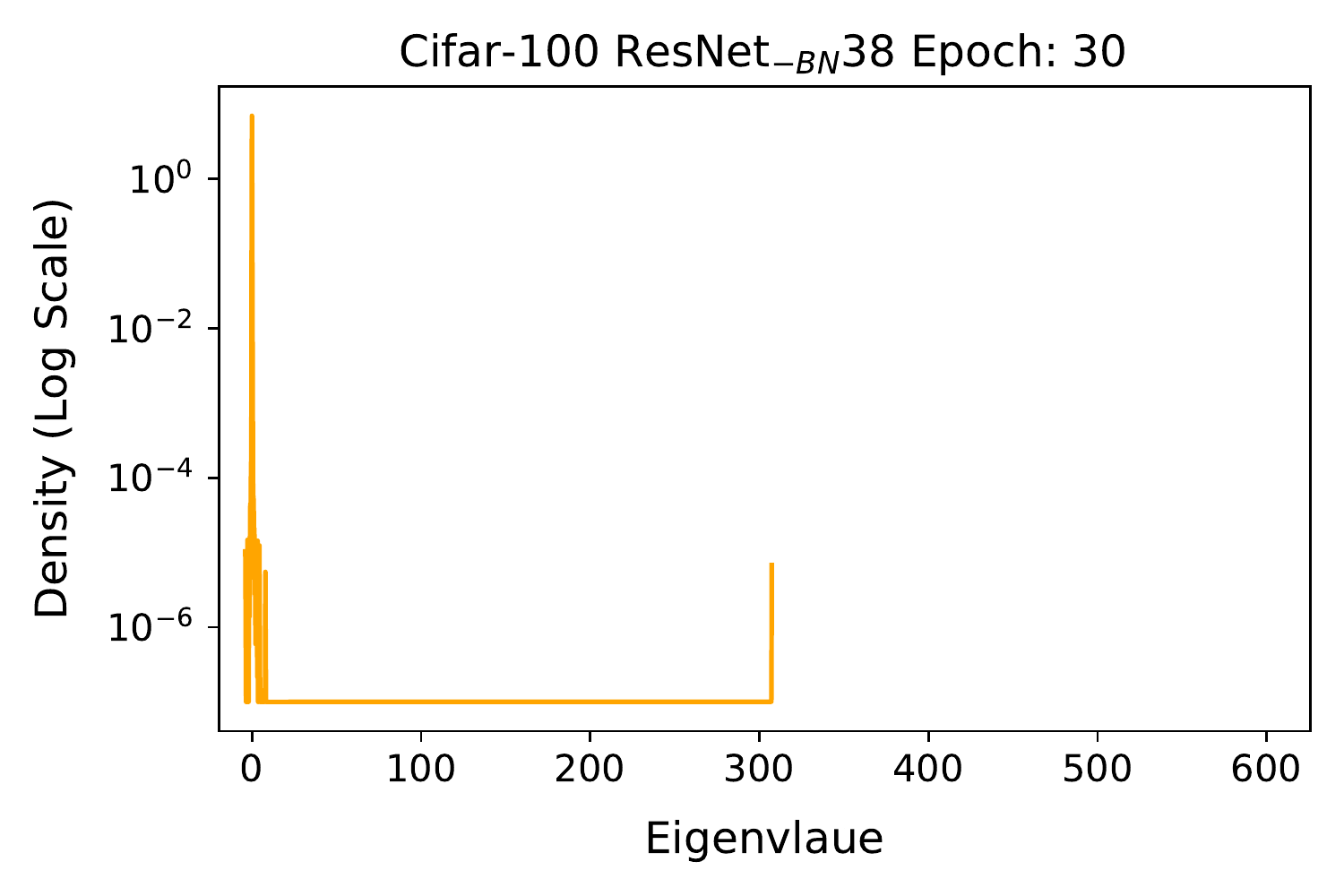}
\includegraphics[width=0.295\textwidth]{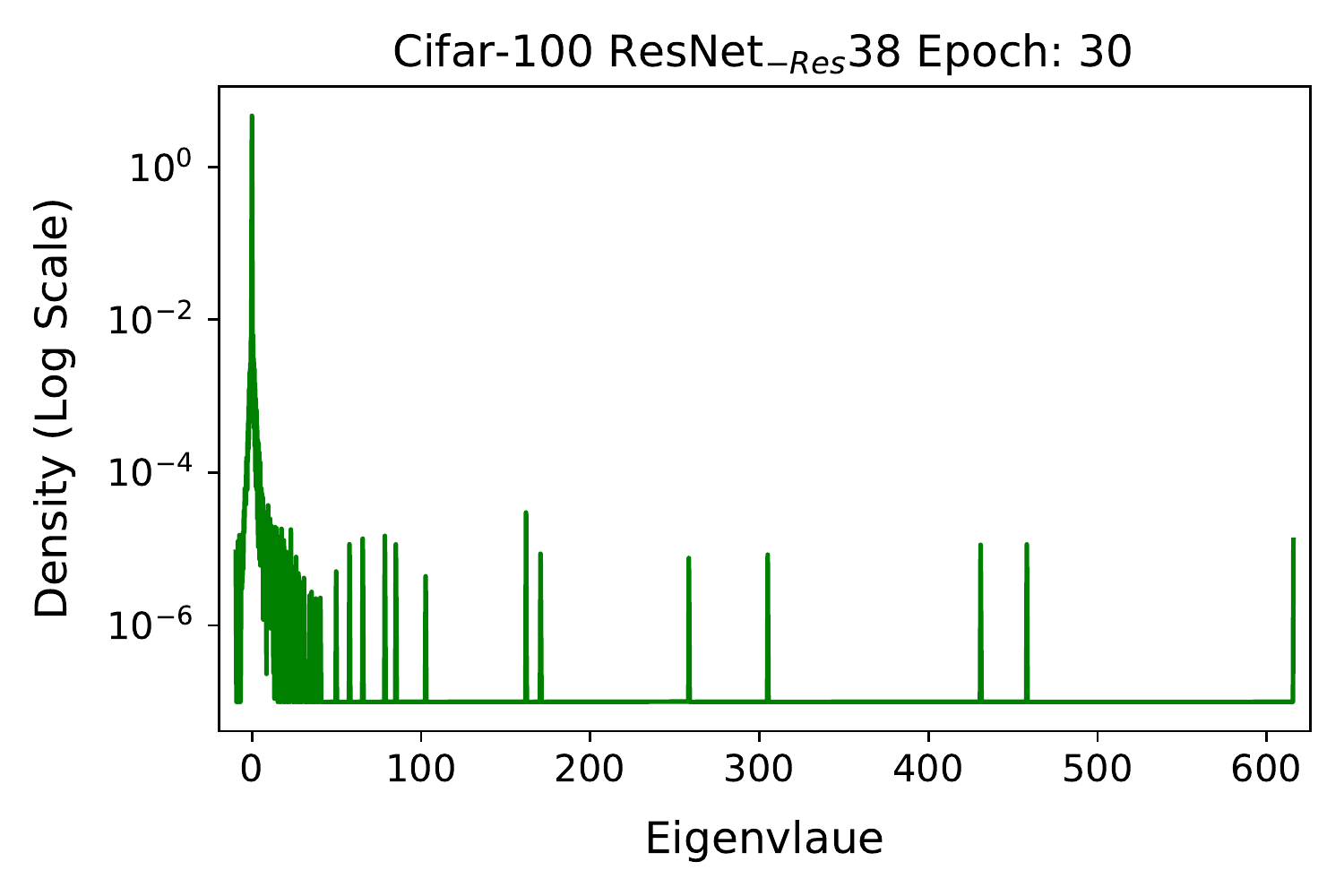}\\
\includegraphics[width=0.295\textwidth]{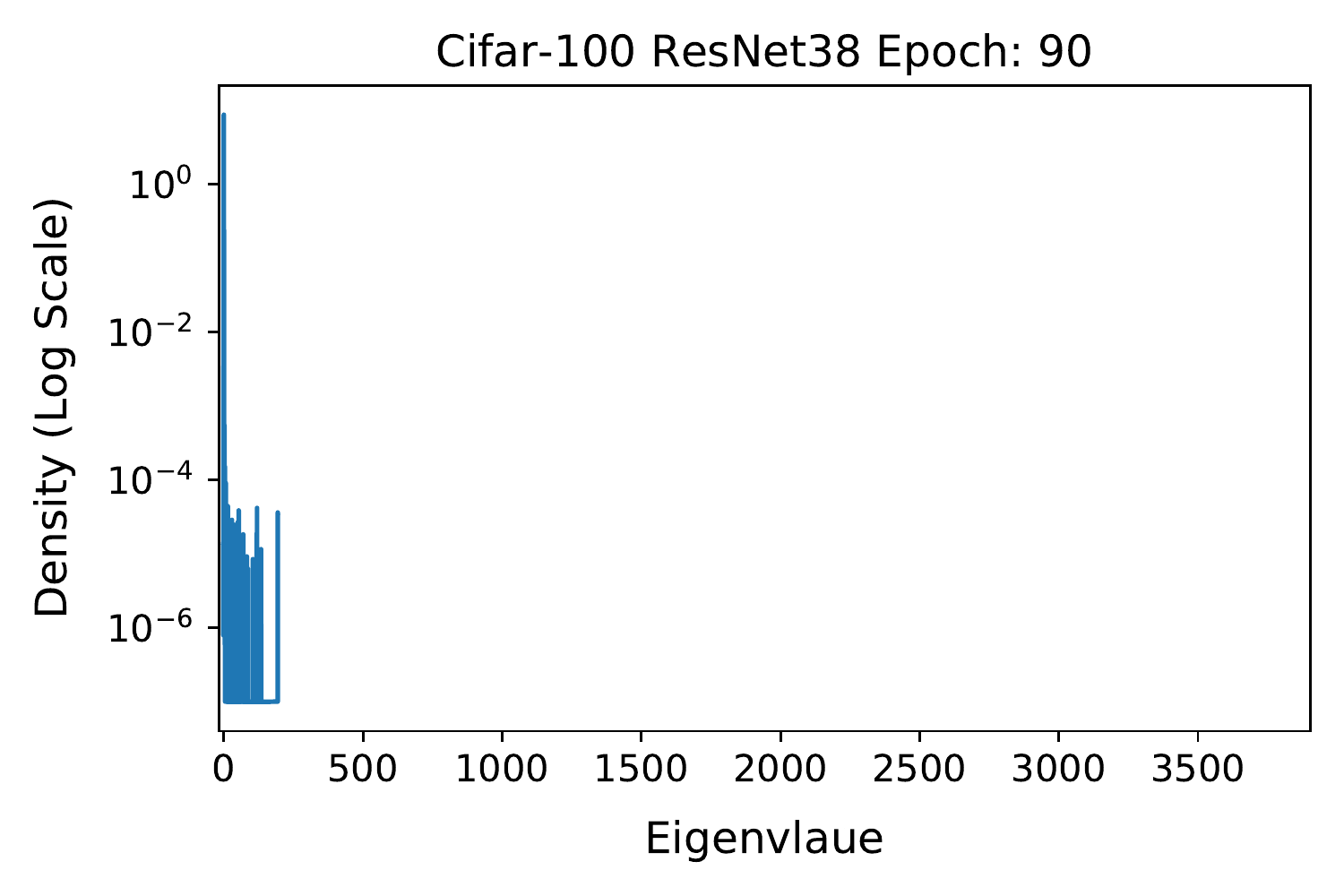}
\includegraphics[width=0.295\textwidth]{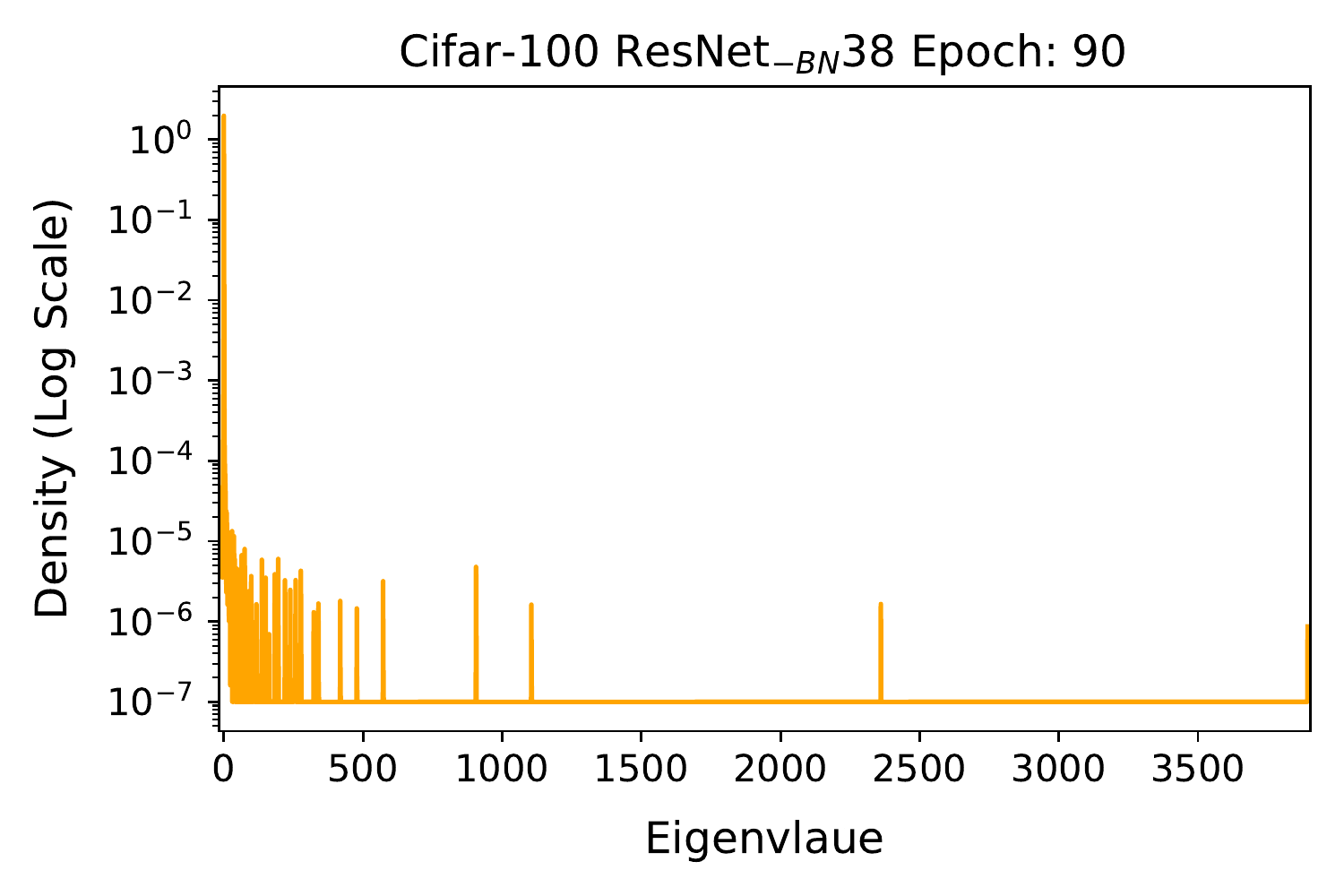}
\includegraphics[width=0.295\textwidth]{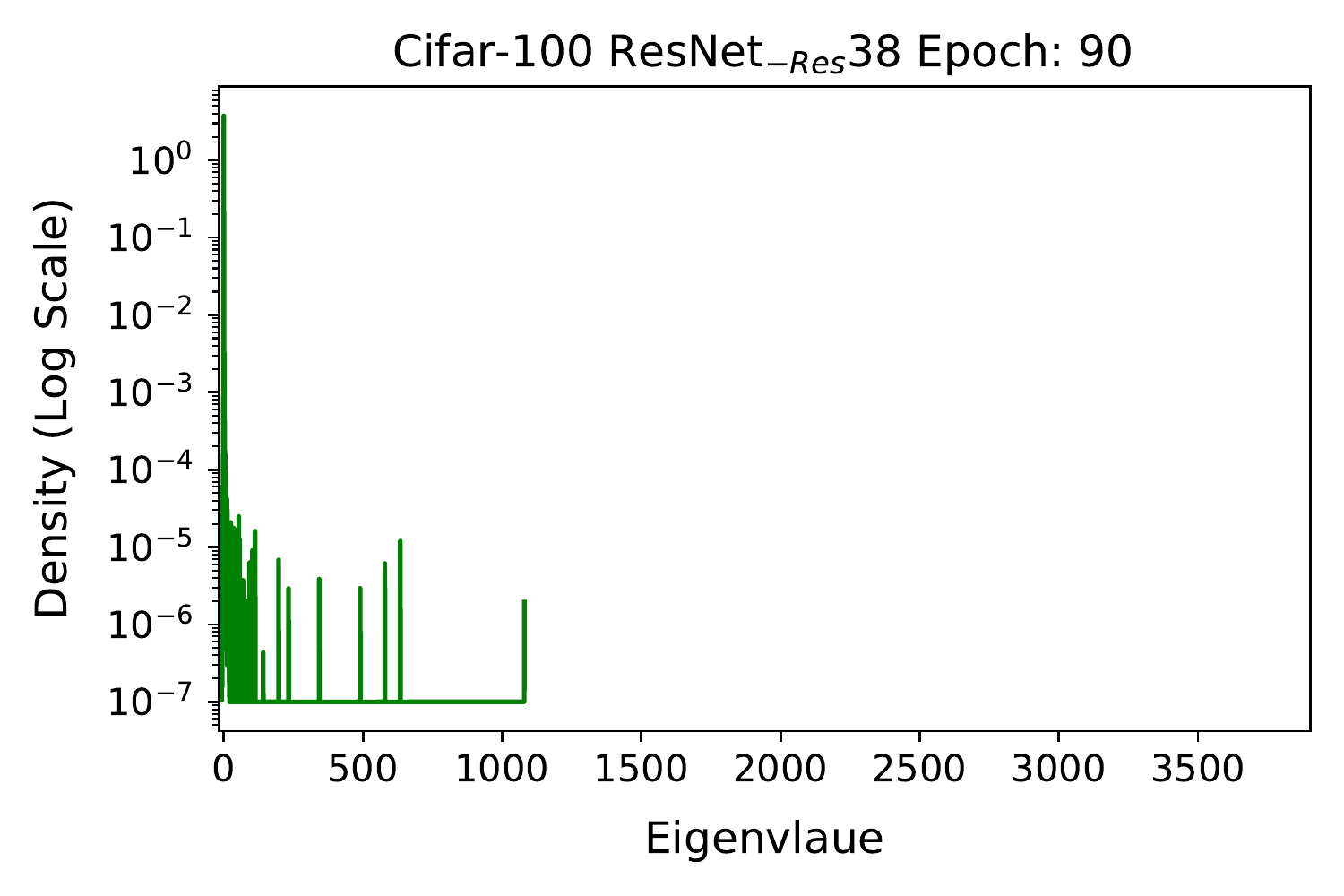}\\
\includegraphics[width=0.295\textwidth]{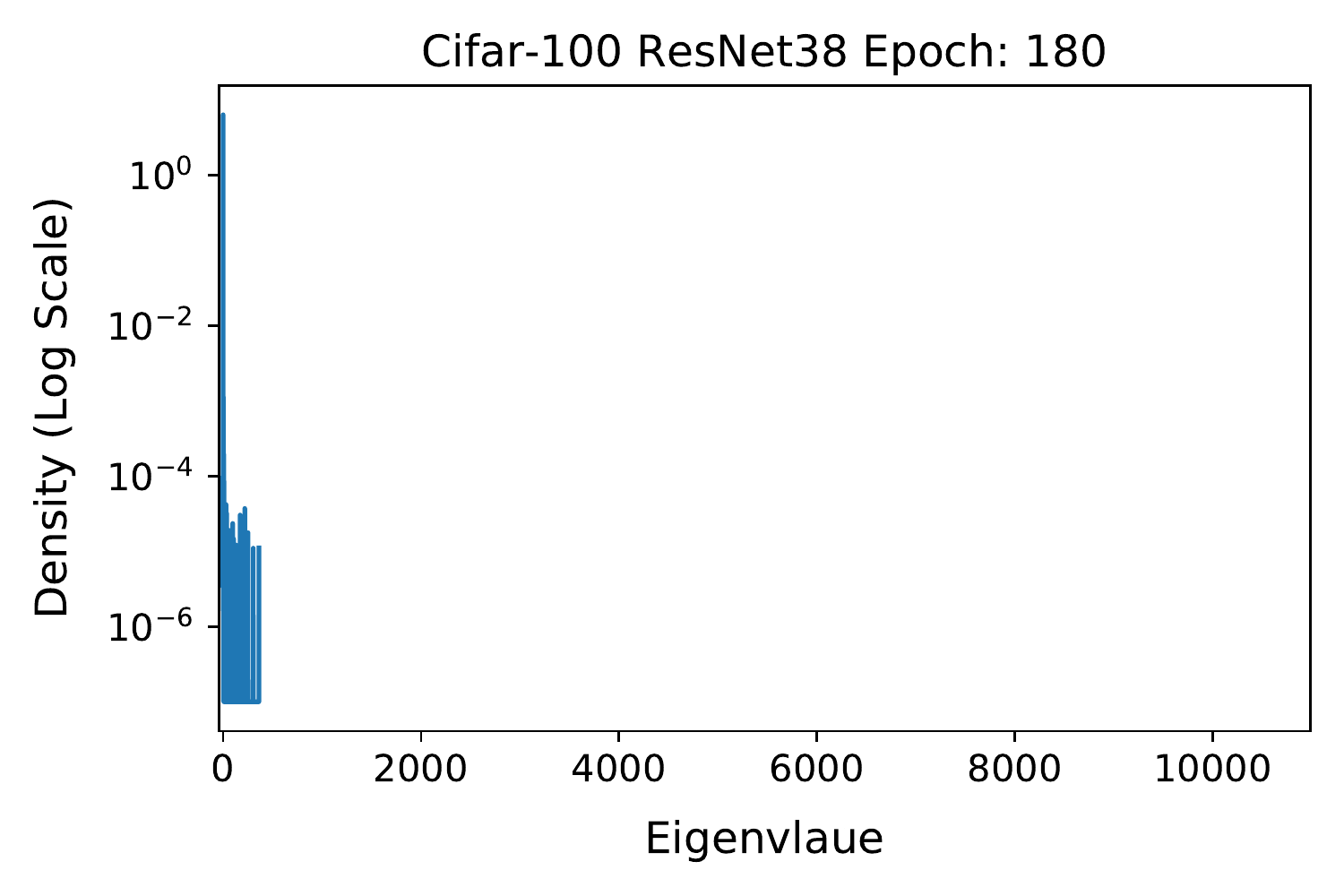}
\includegraphics[width=0.295\textwidth]{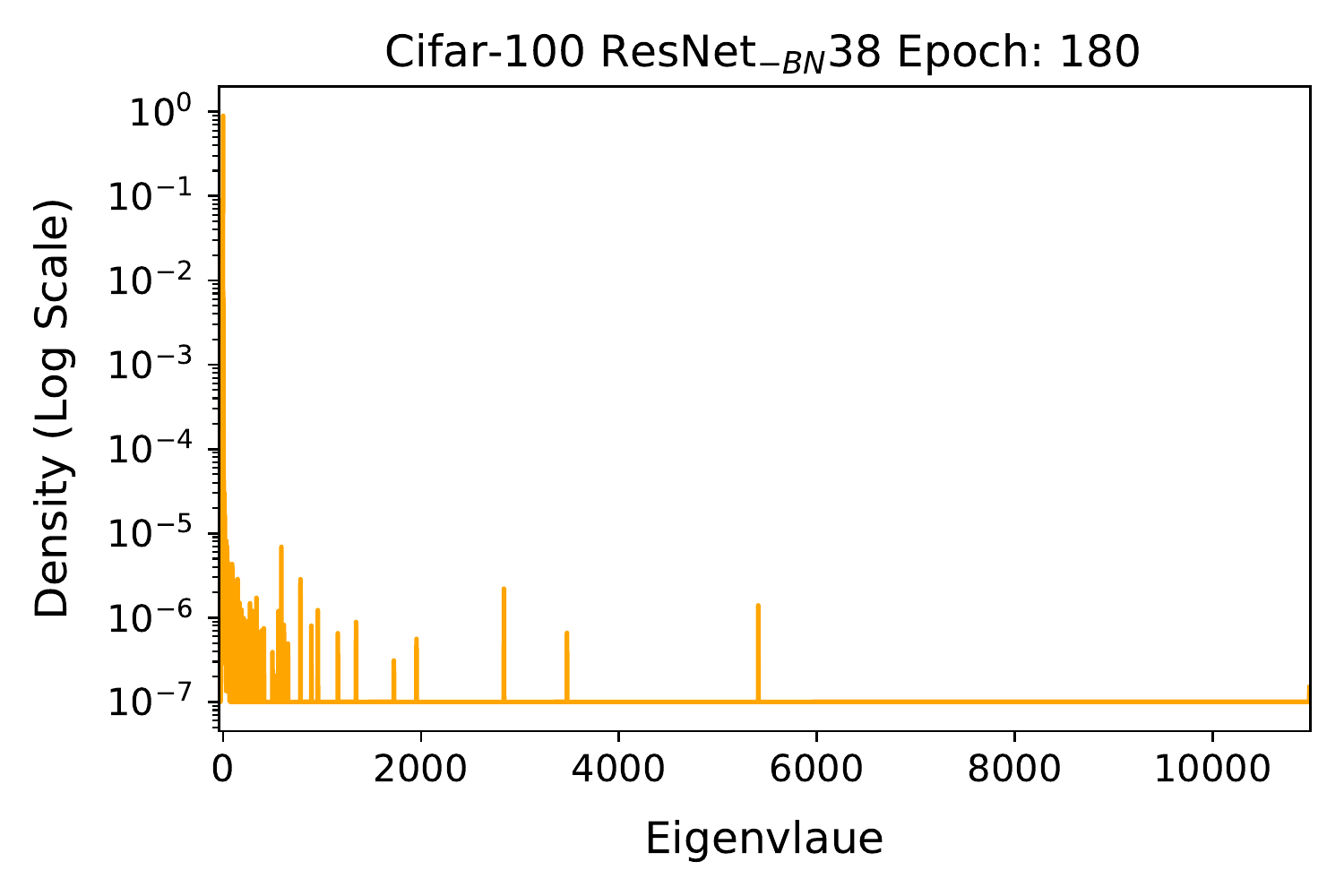}
\includegraphics[width=0.295\textwidth]{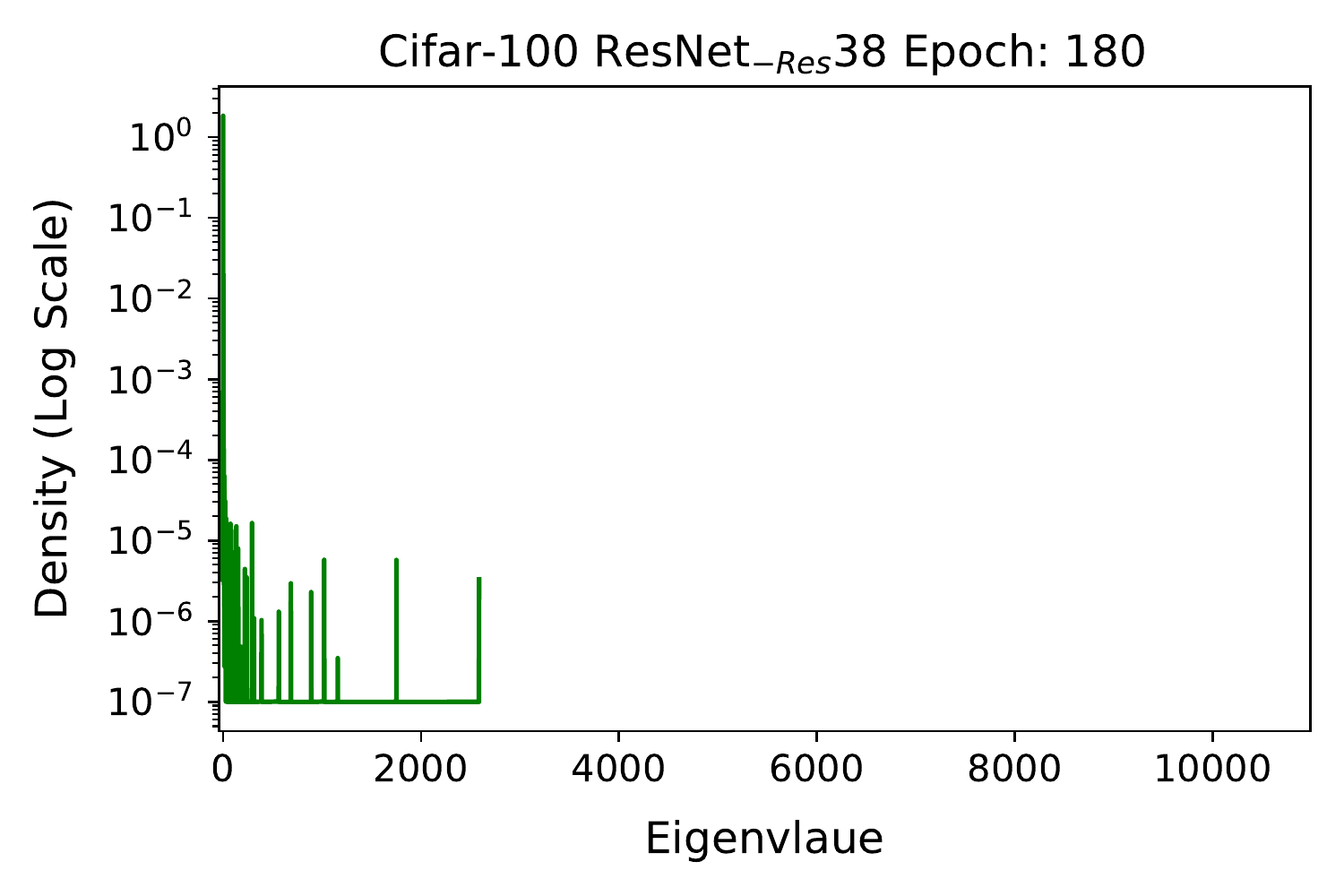}\\
\caption{
Hessian ESD of the entire network for ResNet/\ResNetBN/\ResNetRes with depth 38 on Cifar-100 with Hessian batch size 50000. 
This figure shows the Hessian ESD throughout the training process.
One notable thing here is that the Hessian ESD of \ResNetBN38 centers around zero (at least) until epoch 5. 
This clearly shows that training without BN is indeed harder. 
}
  \label{fig:resnet38-slq-full-net-all-cifar100}
\end{figure*}

\begin{figure*}[!htbp]
\centering
\includegraphics[width=0.295\textwidth]{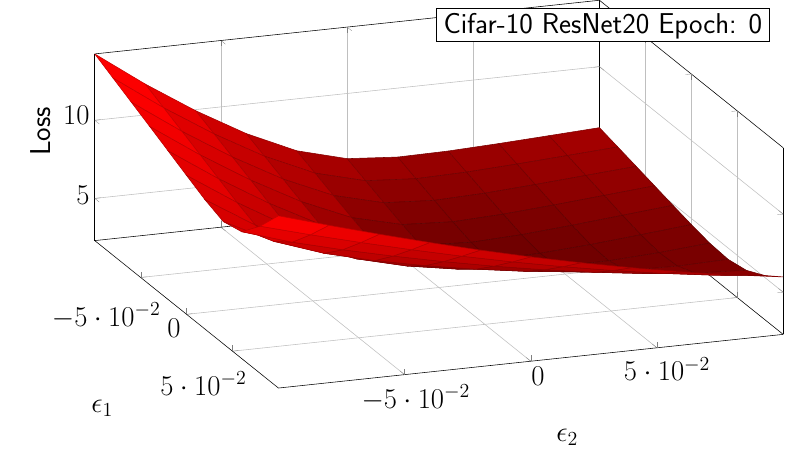}
\includegraphics[width=0.295\textwidth]{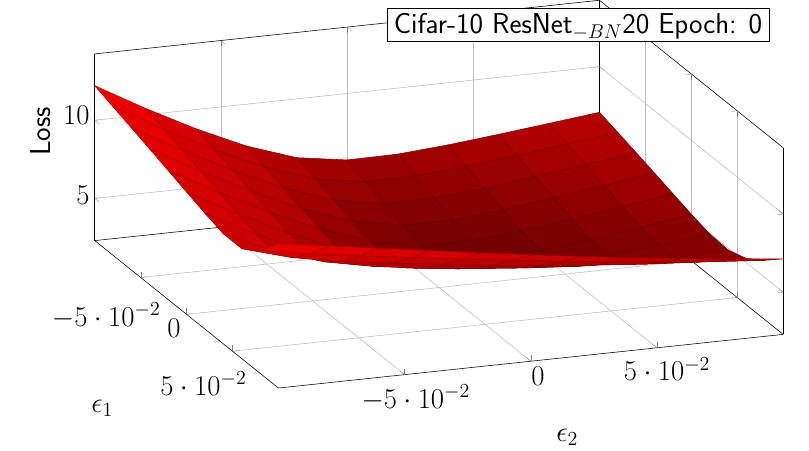}
\includegraphics[width=0.295\textwidth]{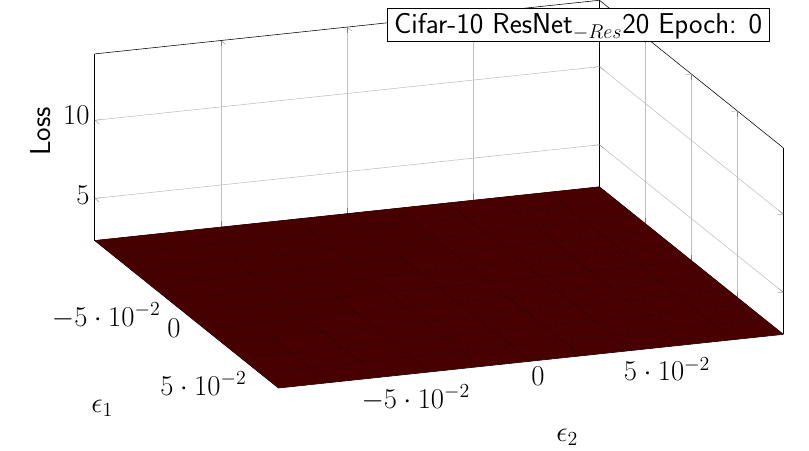}\\
\includegraphics[width=0.295\textwidth]{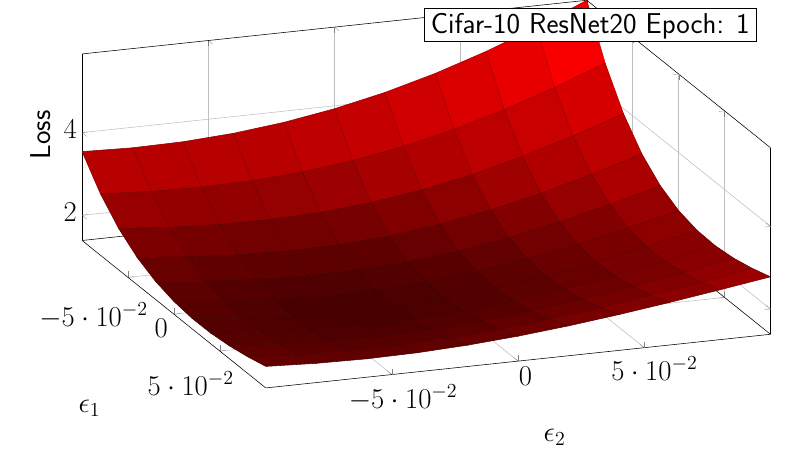}
\includegraphics[width=0.295\textwidth]{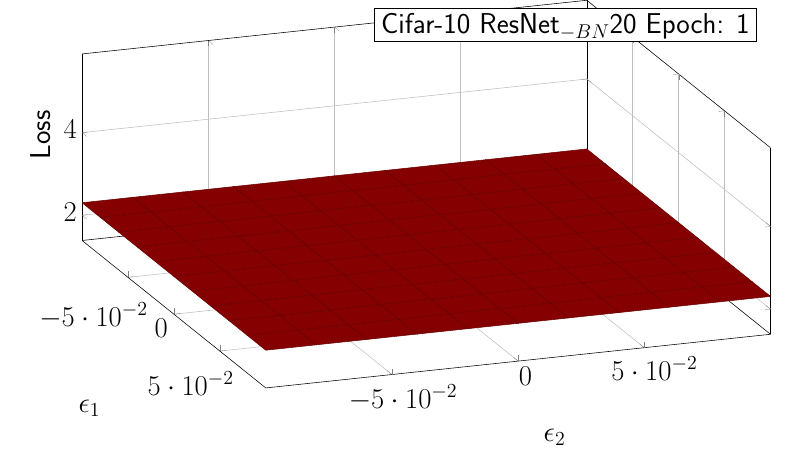}
\includegraphics[width=0.295\textwidth]{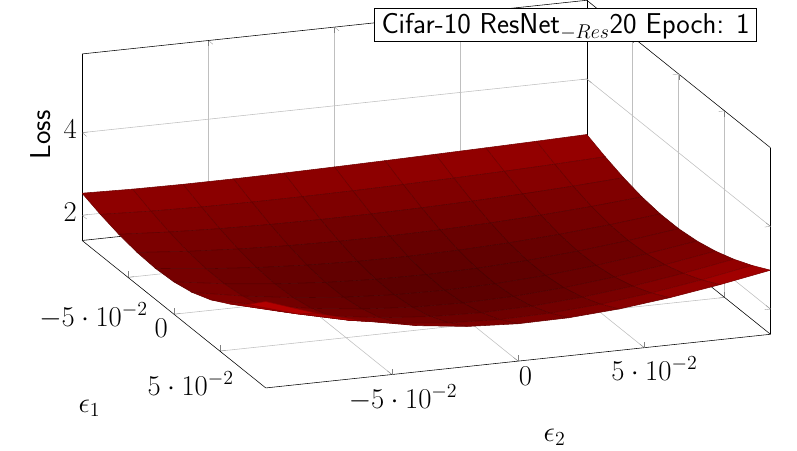}\\
\includegraphics[width=0.295\textwidth]{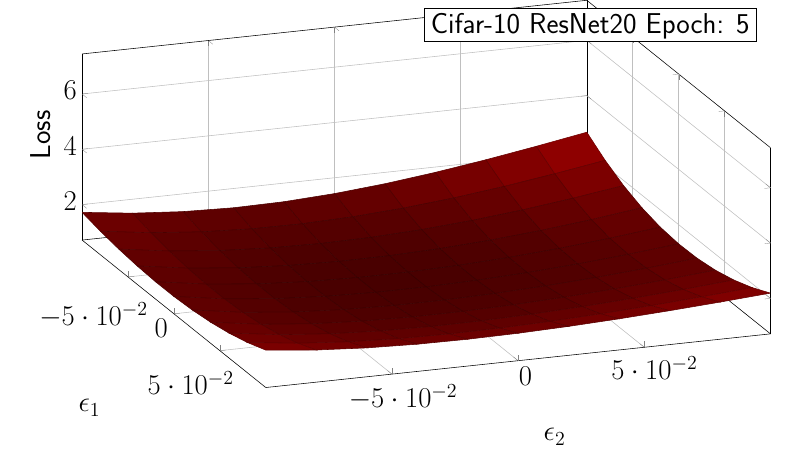}
\includegraphics[width=0.295\textwidth]{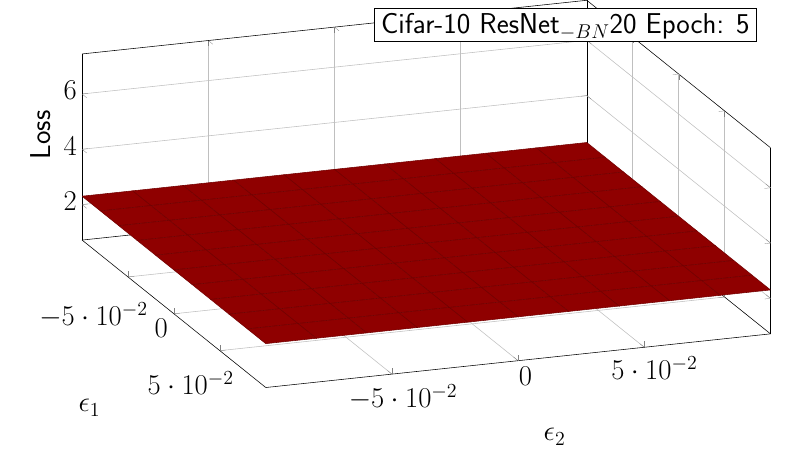}
\includegraphics[width=0.295\textwidth]{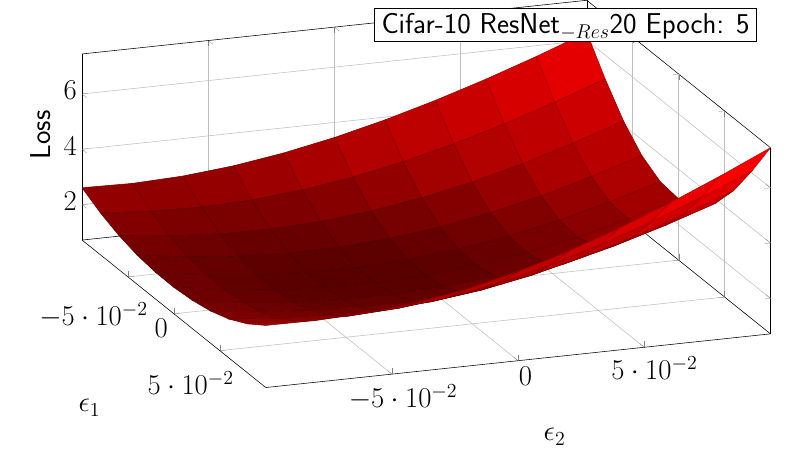}\\
\includegraphics[width=0.295\textwidth]{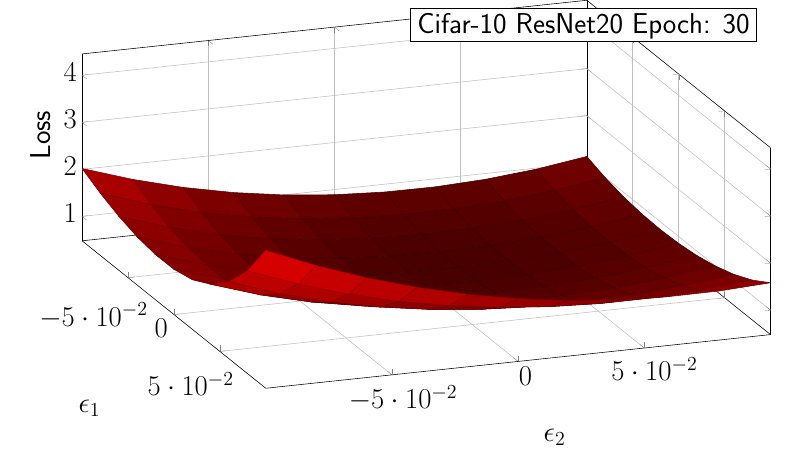}
\includegraphics[width=0.295\textwidth]{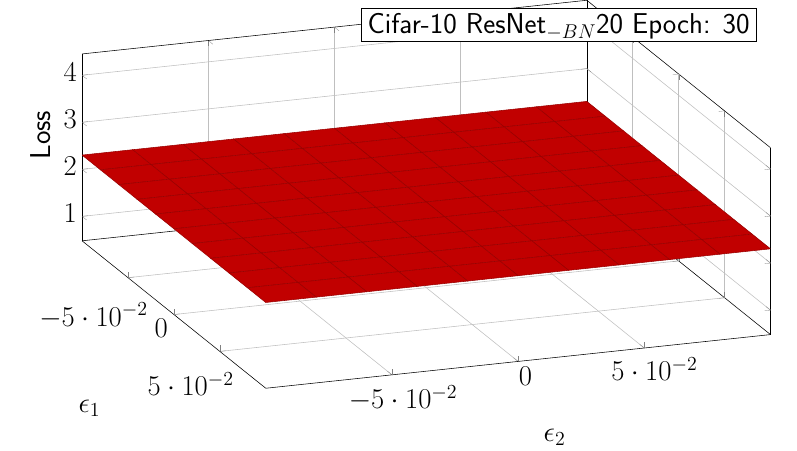}
\includegraphics[width=0.295\textwidth]{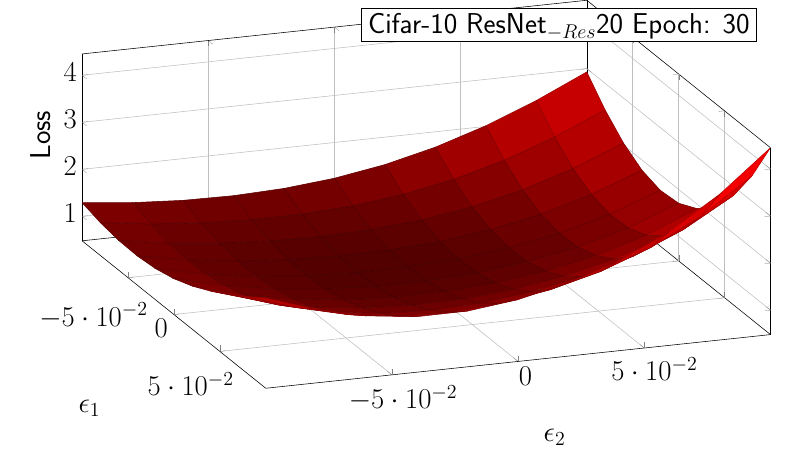}\\
\includegraphics[width=0.295\textwidth]{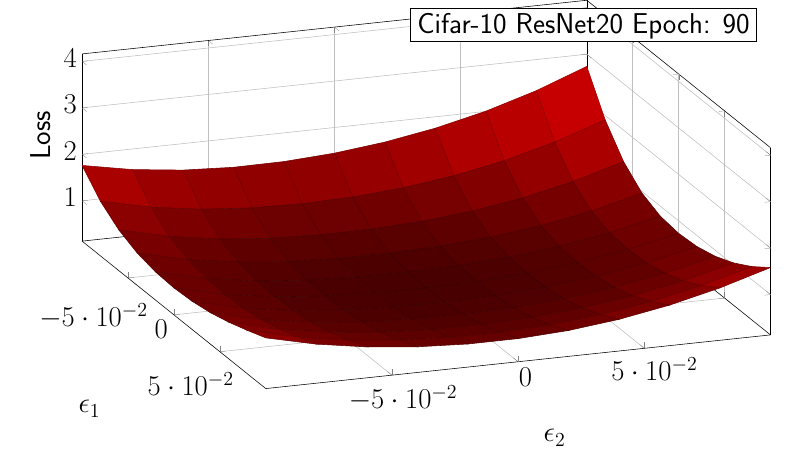}
\includegraphics[width=0.295\textwidth]{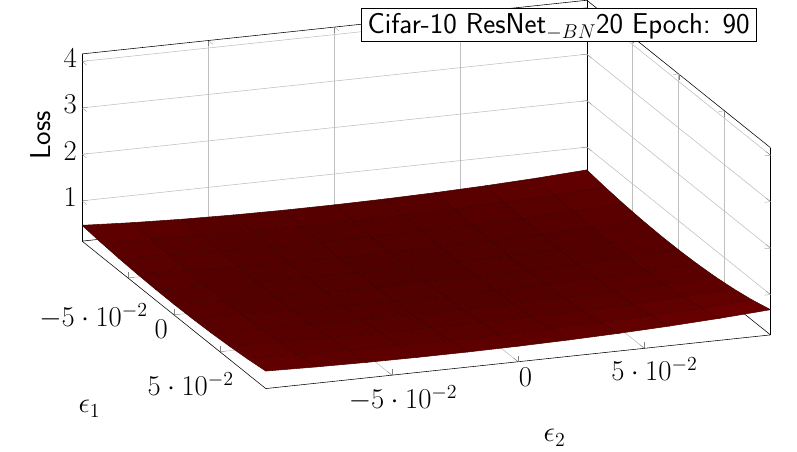}
\includegraphics[width=0.295\textwidth]{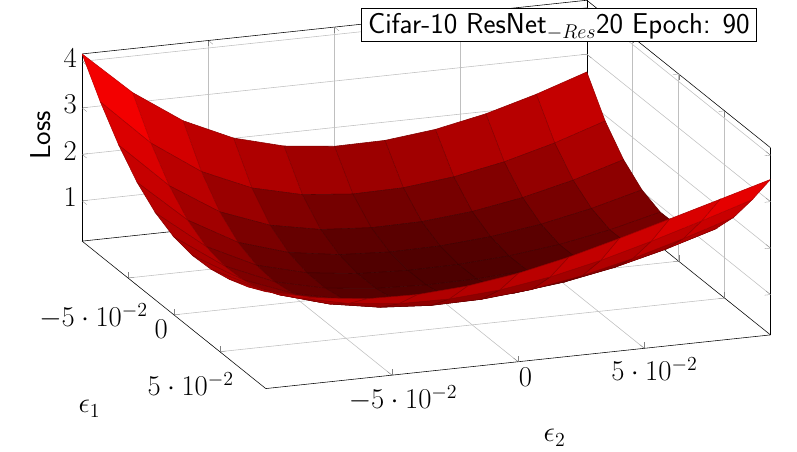}\\
\includegraphics[width=0.295\textwidth]{figures/resnet20/surface/w_w_epoch_180.pdf}
\includegraphics[width=0.295\textwidth]{figures/resnet20/surface/wi_w_epoch_180.pdf}
\includegraphics[width=0.295\textwidth]{figures/resnet20/surface/w_wi_epoch_180.pdf}\\
\caption{
Loss landscape of ResNet/\ResNetBN/\ResNetRes20 on Cifar-10 with batch size 4096 by perturbing the parameters along the first two dominant eigenvectors of the Hessian. 
The loss landscape of \ResNetBN20 (\ResNetRes20) is indeed smoother (sharper) than that of \ResNet20, which is align with the trace plot in~\fref{fig:resnet20/32/38-hut-full-net} and the Hessian ESD plot in~\fref{fig:resnet20-slq-full-net-part}. 
}
  \label{fig:resnet20-loss-landscape-all}
\end{figure*}

\begin{figure*}[!htbp]
\centering
\includegraphics[width=0.295\textwidth]{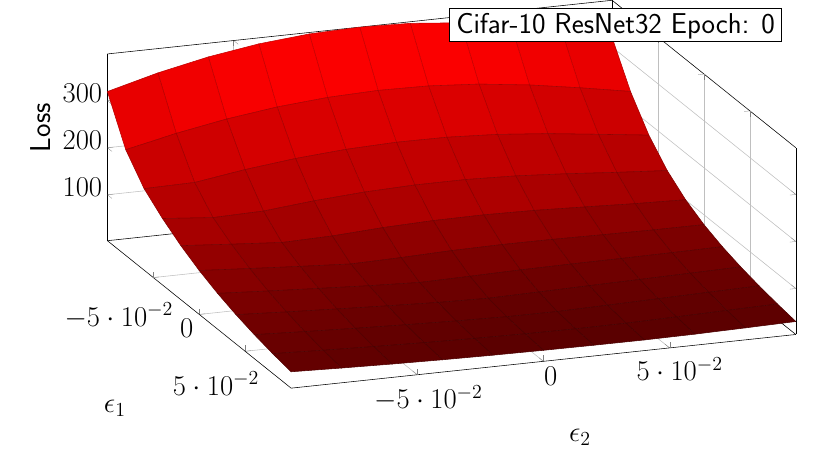}
\includegraphics[width=0.295\textwidth]{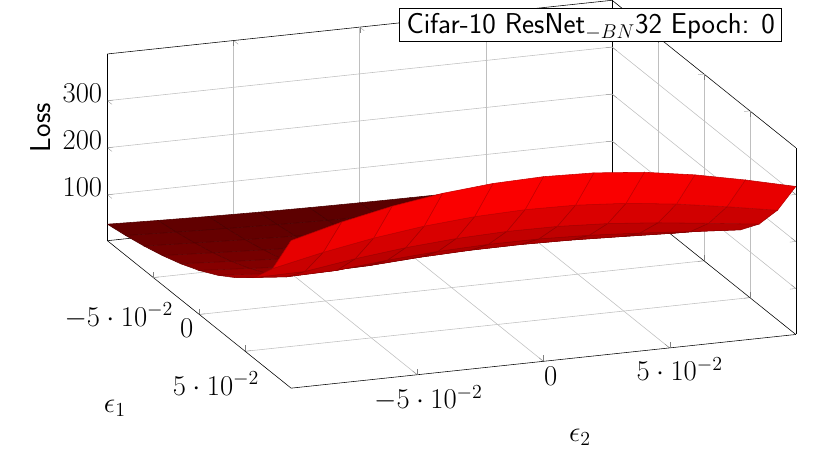}
\includegraphics[width=0.295\textwidth]{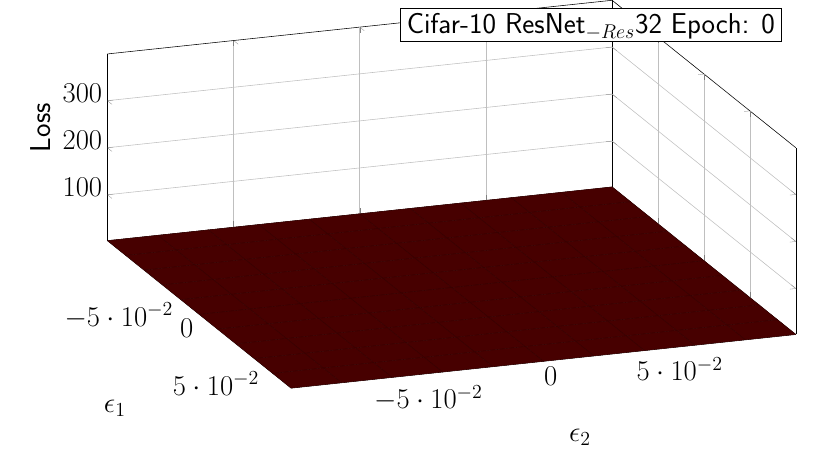}\\
\includegraphics[width=0.295\textwidth]{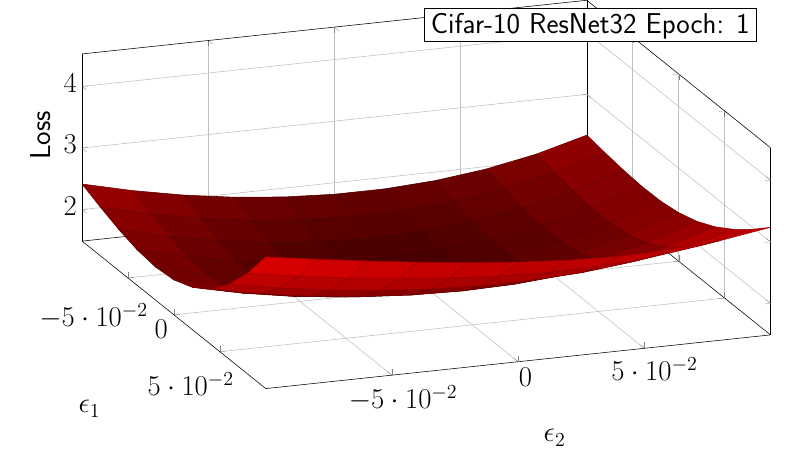}
\includegraphics[width=0.295\textwidth]{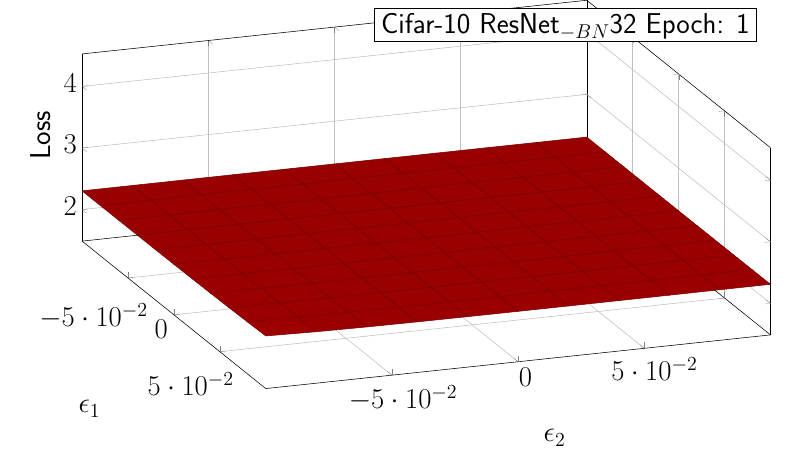}
\includegraphics[width=0.295\textwidth]{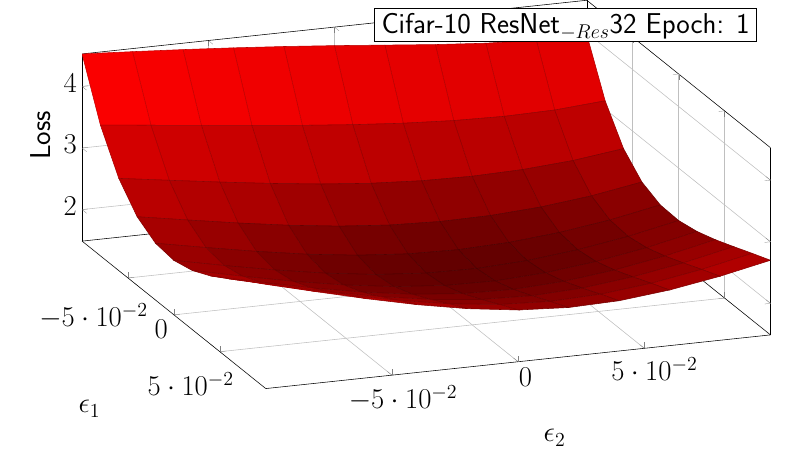}\\
\includegraphics[width=0.295\textwidth]{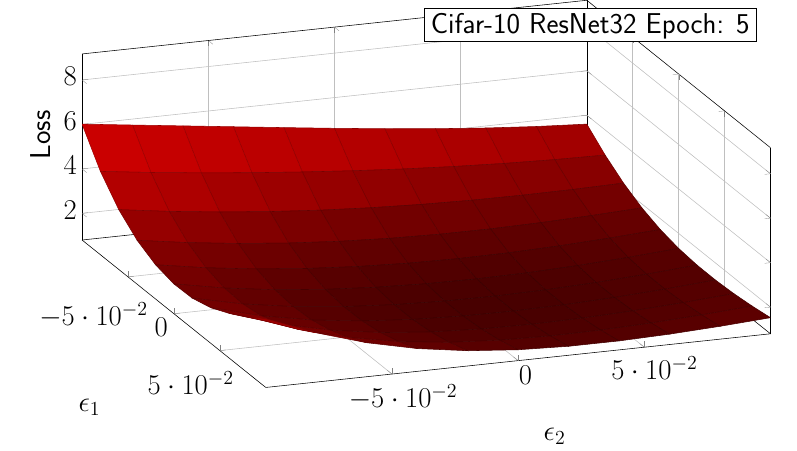}
\includegraphics[width=0.295\textwidth]{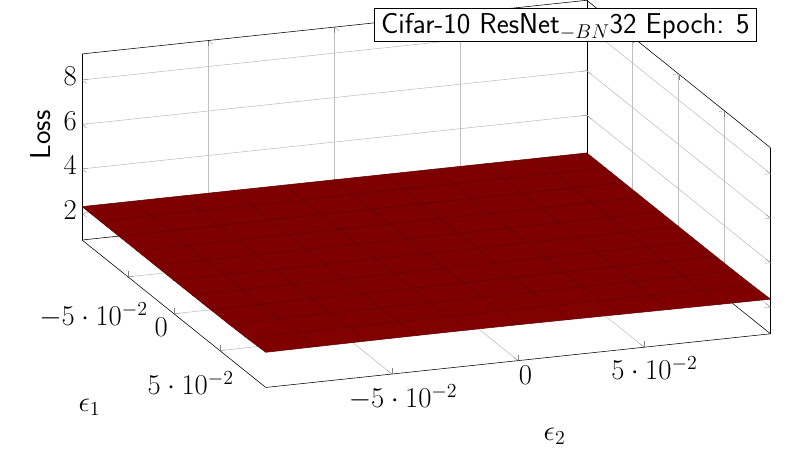}
\includegraphics[width=0.295\textwidth]{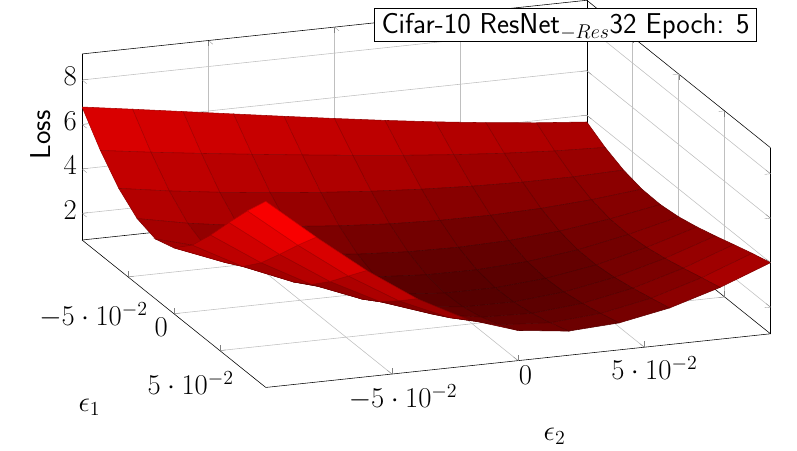}\\
\includegraphics[width=0.295\textwidth]{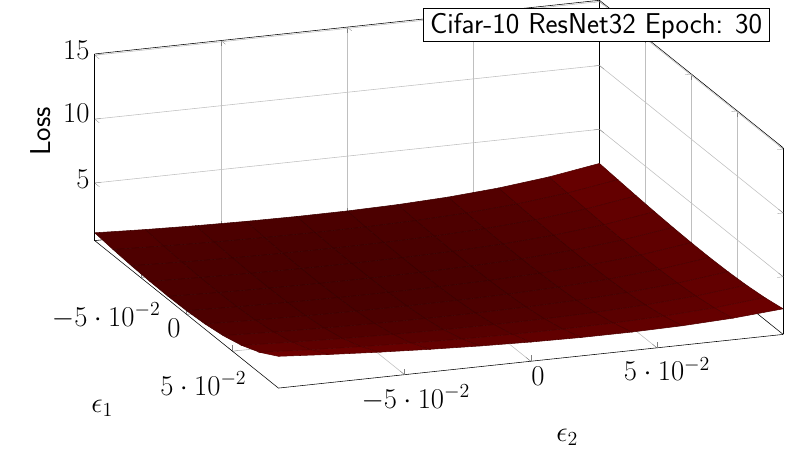}
\includegraphics[width=0.295\textwidth]{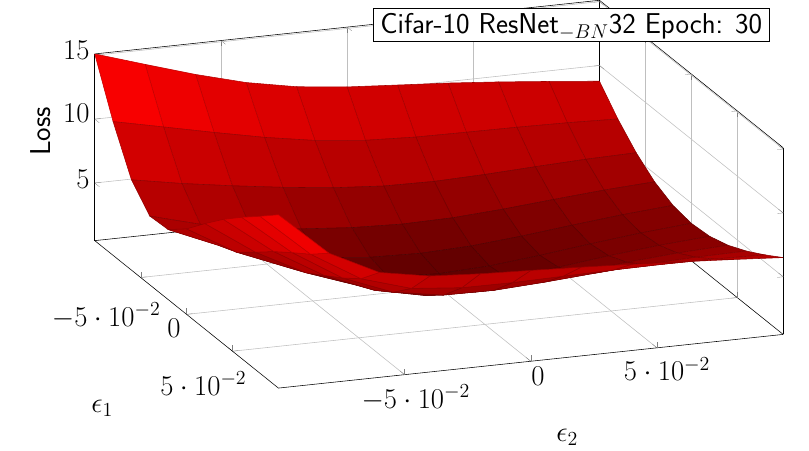}
\includegraphics[width=0.295\textwidth]{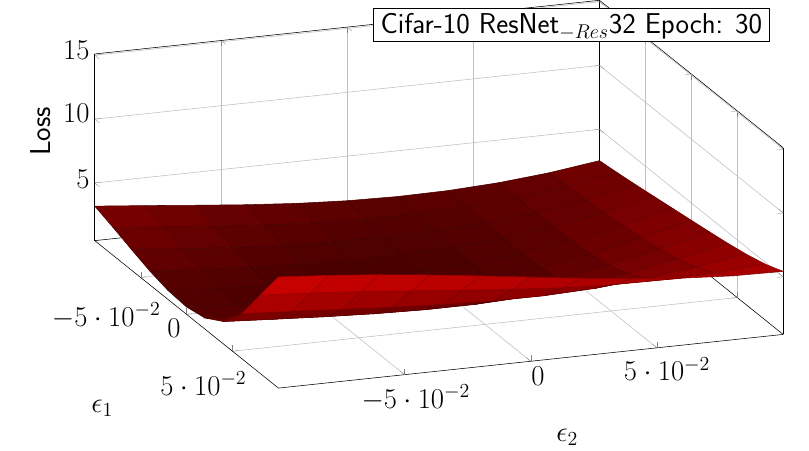}\\
\includegraphics[width=0.295\textwidth]{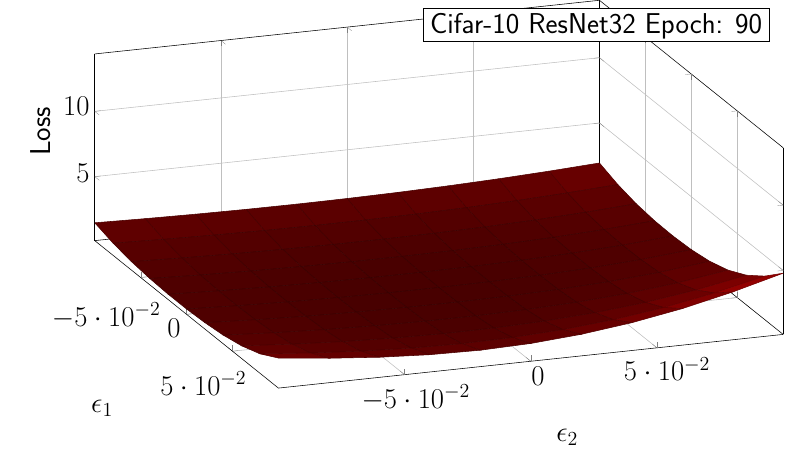}
\includegraphics[width=0.295\textwidth]{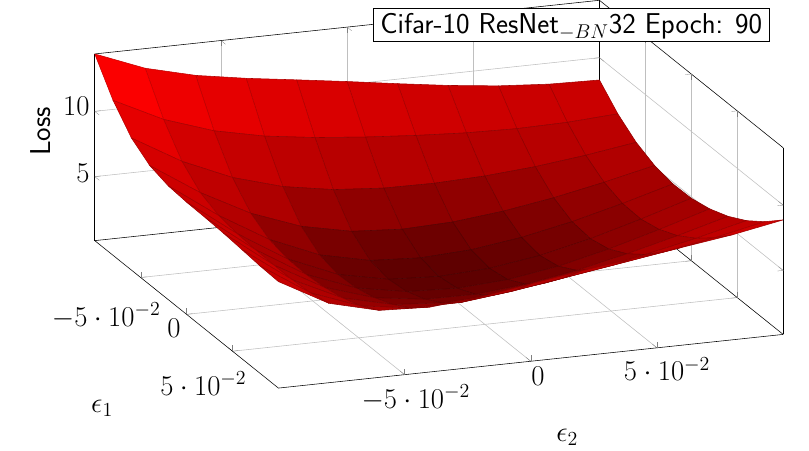}
\includegraphics[width=0.295\textwidth]{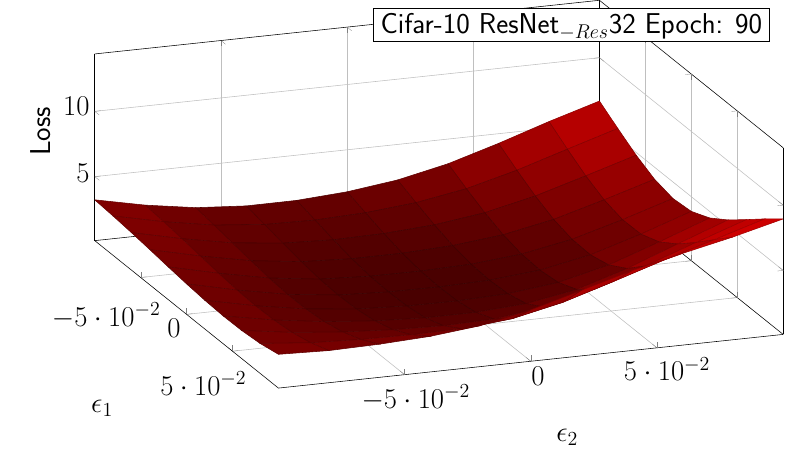}\\
\includegraphics[width=0.295\textwidth]{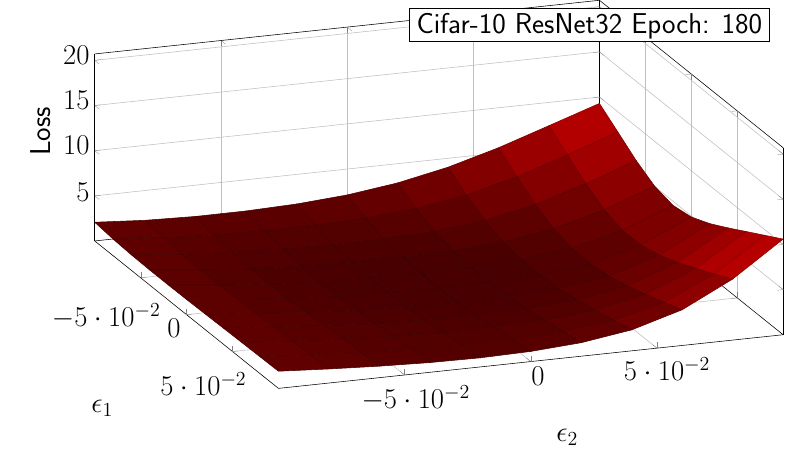}
\includegraphics[width=0.295\textwidth]{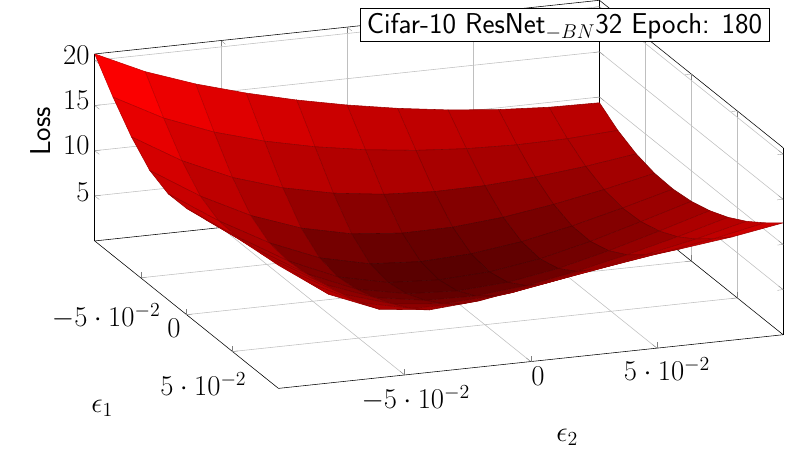}
\includegraphics[width=0.295\textwidth]{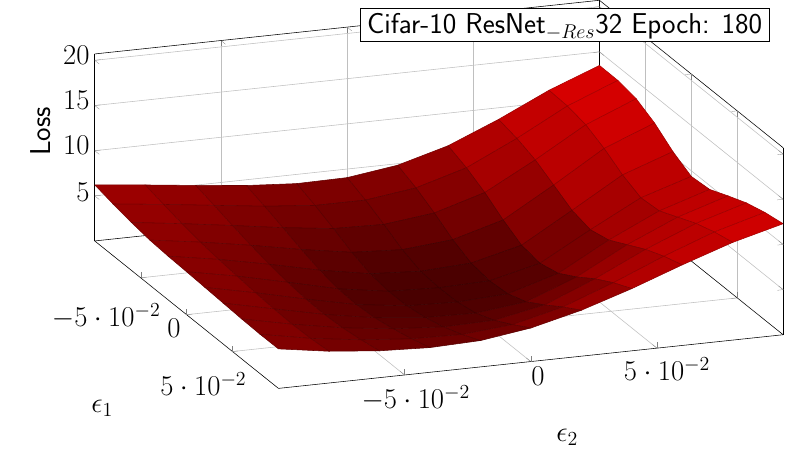}\\
\caption{
Loss landscape of ResNet/\ResNetBN/\ResNetRes32 on Cifar-10 with batch size 4096 by perturbing the parameters along the first two dominant eigenvectors of the Hessian. 
The loss landscape of \ResNetBN32/\ResNetRes32 is indeed sharper than that of \ResNet32, which is align with the trace plot in~\fref{fig:resnet20/32/38-hut-full-net} and the Hessian ESD plot in~\fref{fig:resnet32-slq-full-net-all}.
}
  \label{fig:resnet32-loss-landscape-all}
\end{figure*}

\begin{figure*}[!htbp]
\centering
\includegraphics[width=0.295\textwidth]{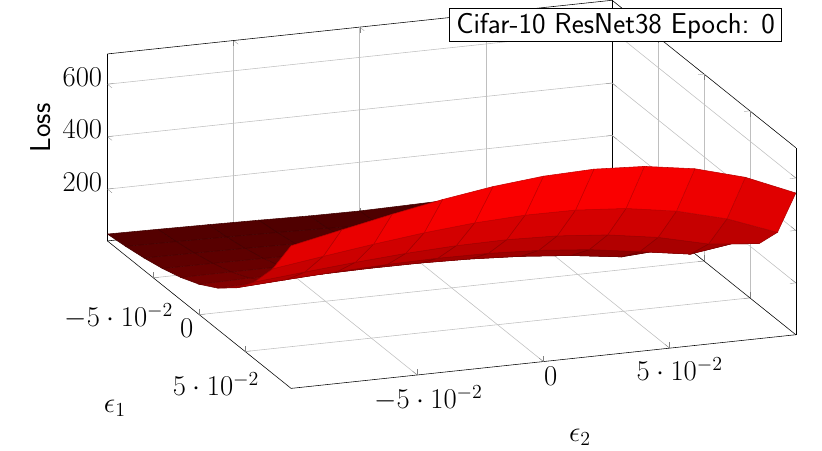}
\includegraphics[width=0.295\textwidth]{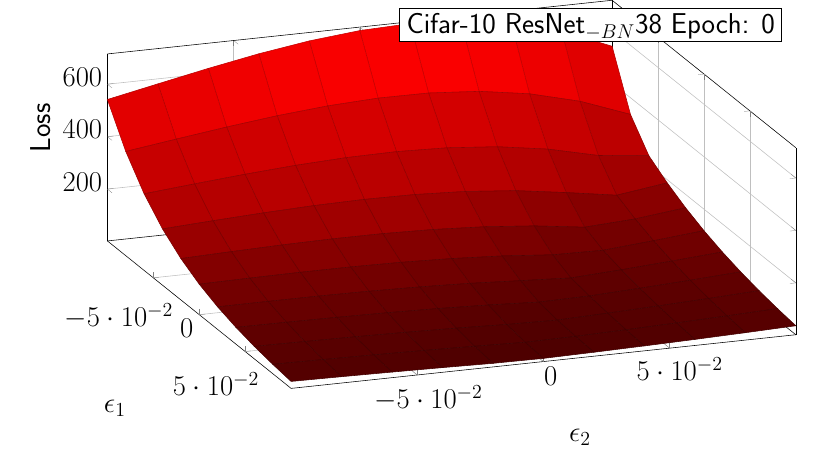}
\includegraphics[width=0.295\textwidth]{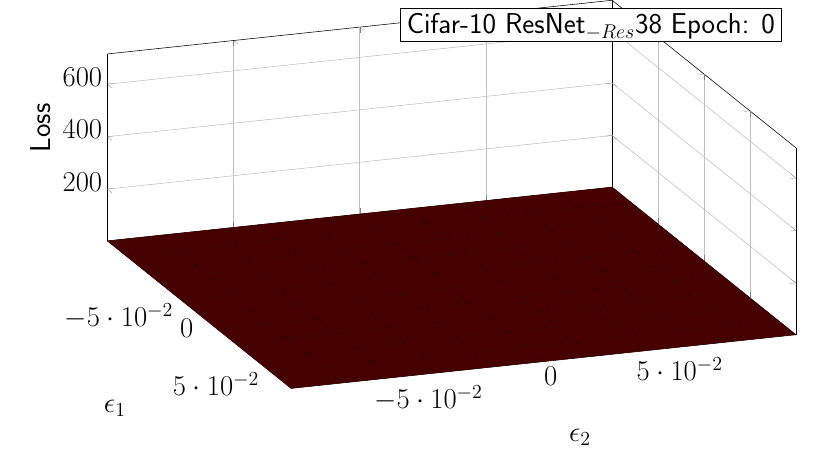}\\
\includegraphics[width=0.295\textwidth]{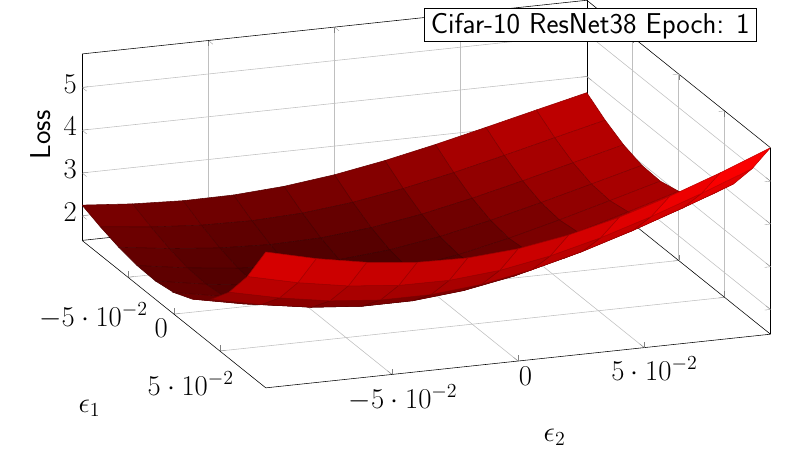}
\includegraphics[width=0.295\textwidth]{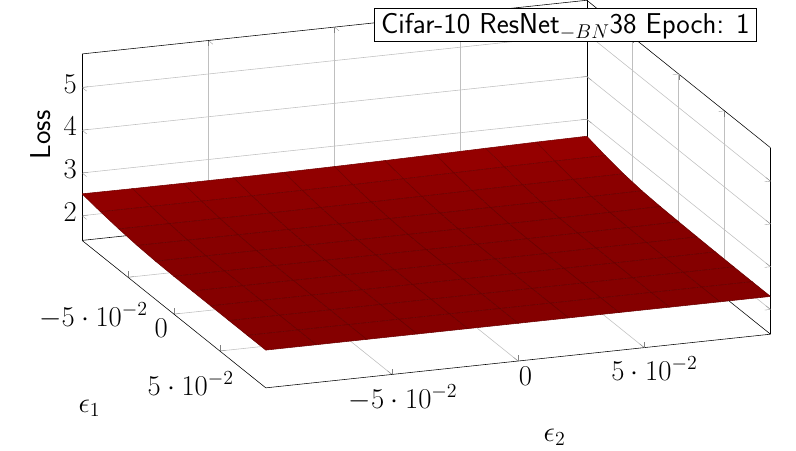}
\includegraphics[width=0.295\textwidth]{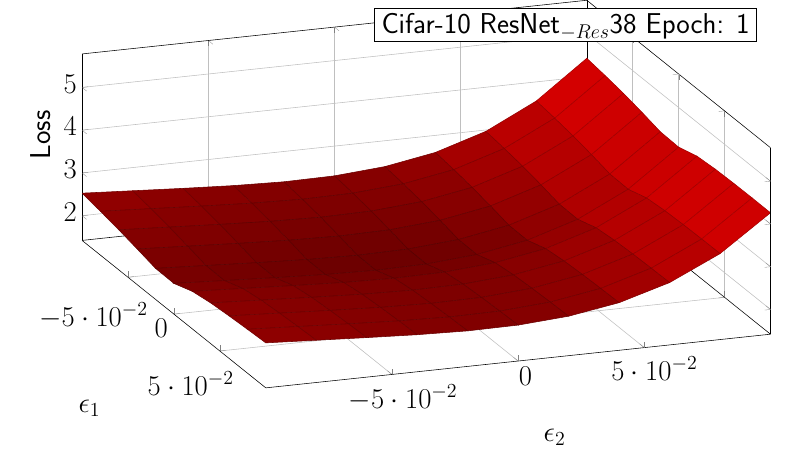}\\
\includegraphics[width=0.295\textwidth]{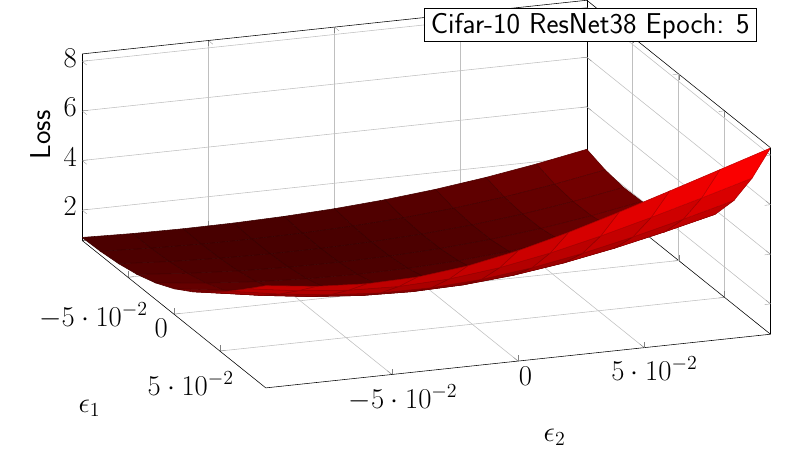}
\includegraphics[width=0.295\textwidth]{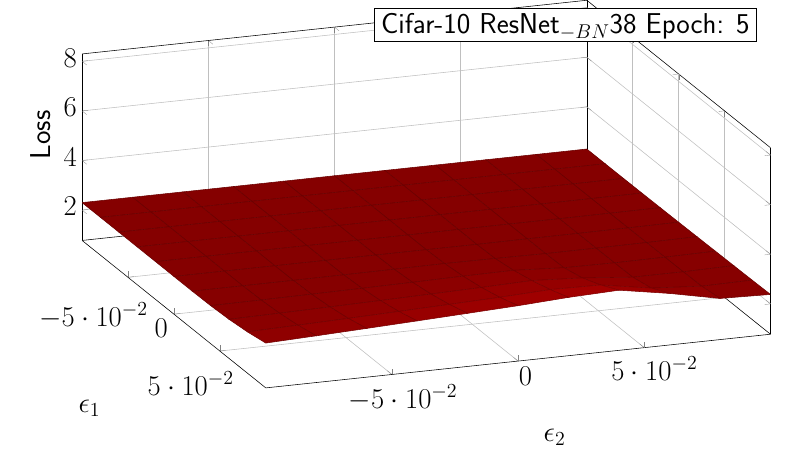}
\includegraphics[width=0.295\textwidth]{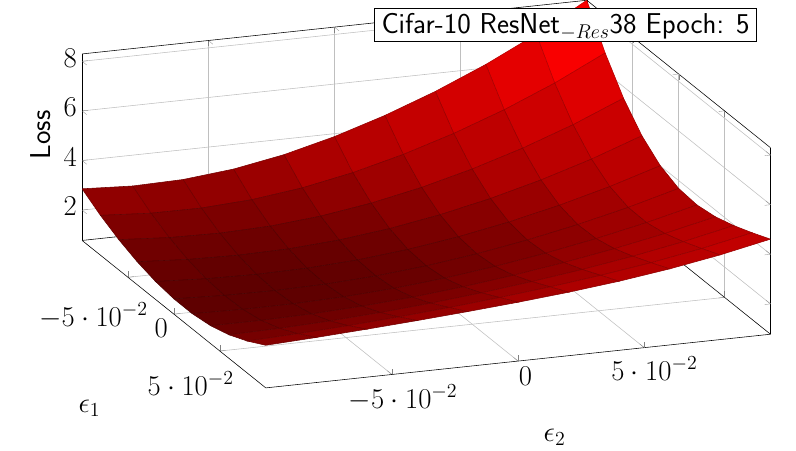}\\
\includegraphics[width=0.295\textwidth]{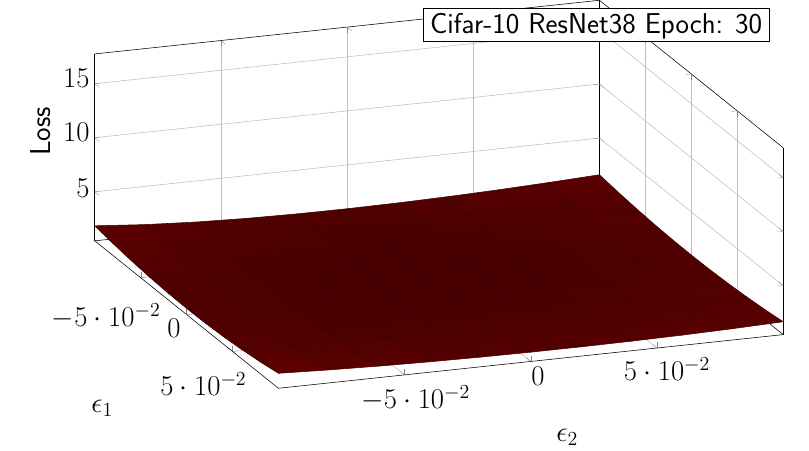}
\includegraphics[width=0.295\textwidth]{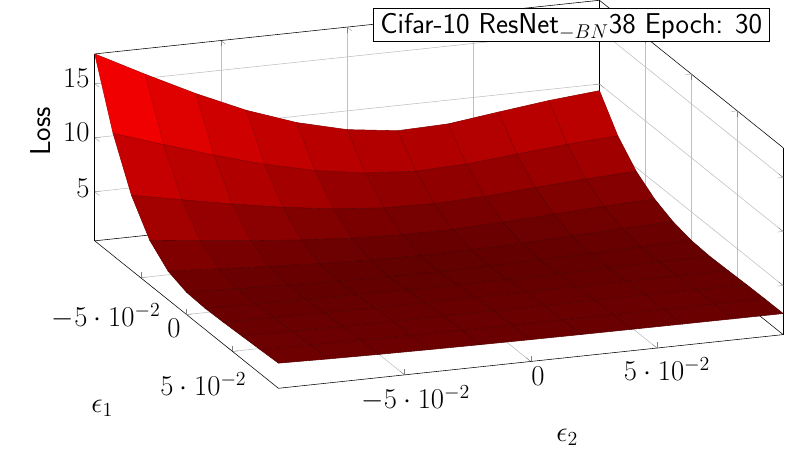}
\includegraphics[width=0.295\textwidth]{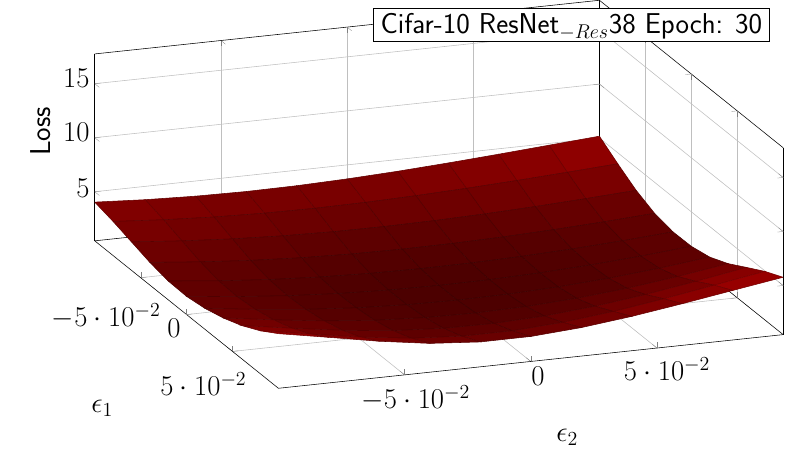}\\
\includegraphics[width=0.295\textwidth]{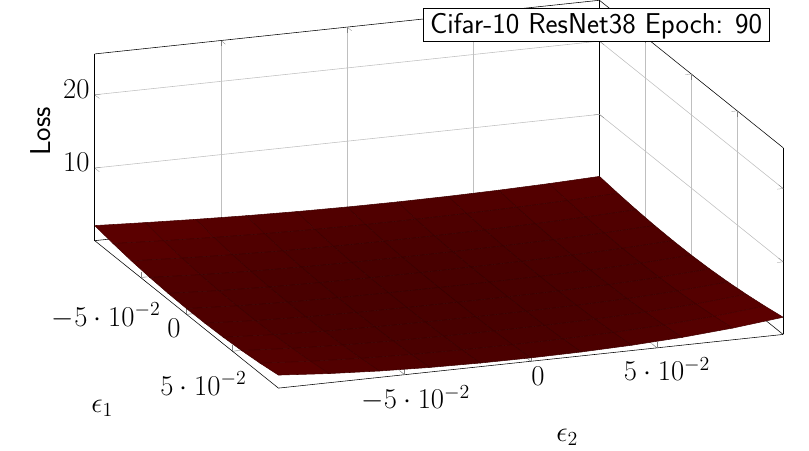}
\includegraphics[width=0.295\textwidth]{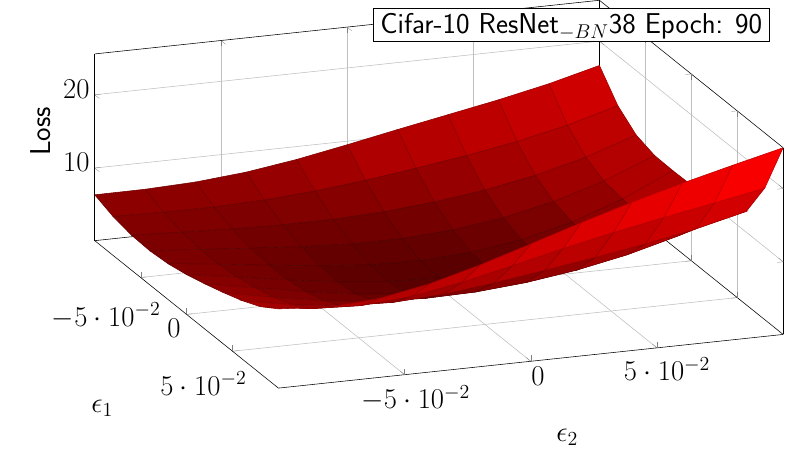}
\includegraphics[width=0.295\textwidth]{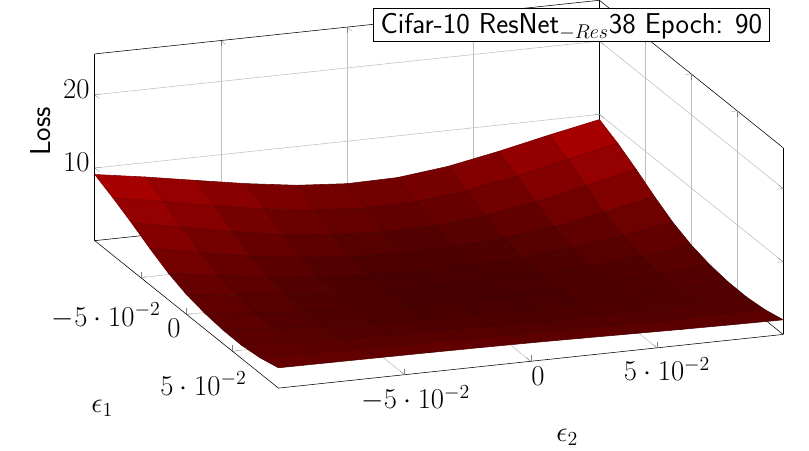}\\
\includegraphics[width=0.295\textwidth]{figures/resnet38/surface/w_w_epoch_180.pdf}
\includegraphics[width=0.295\textwidth]{figures/resnet38/surface/wi_w_epoch_180.pdf}
\includegraphics[width=0.295\textwidth]{figures/resnet38/surface/w_wi_epoch_180.pdf}\\
\caption{
Loss landscape of ResNet/\ResNetBN/\ResNetRes38 on Cifar-10 with batch size 4096 by perturbing the parameters along the first two dominant eigenvectors of the Hessian. 
The loss landscape of \ResNetBN38/\ResNetRes38 is indeed sharper than that of \ResNet38, which is align with the trace plot in~\fref{fig:resnet20/32/38-hut-full-net} and the Hessian ESD plot in~\fref{fig:resnet38-slq-full-net-all}.
}
  \label{fig:resnet38-loss-landscape-all}
\end{figure*}

\begin{figure*}[!htbp]
\centering
\includegraphics[width=0.295\textwidth]{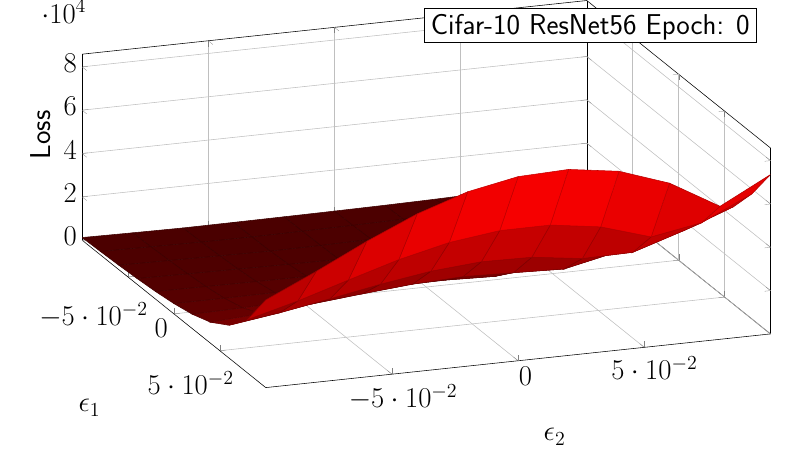}
\includegraphics[width=0.295\textwidth]{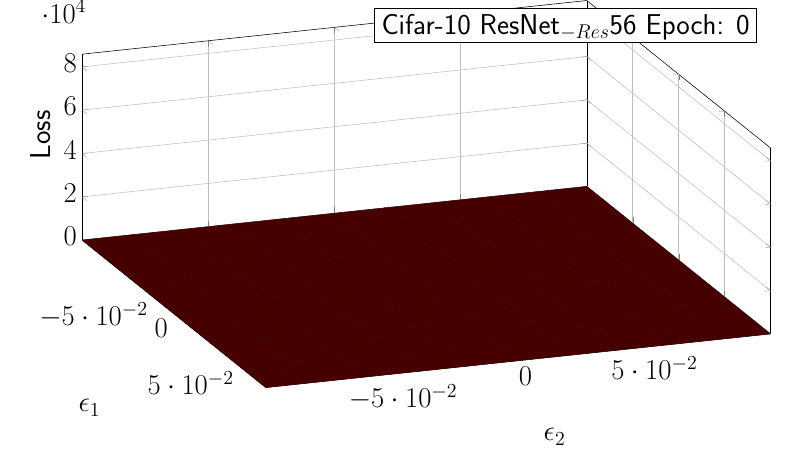}\\
\includegraphics[width=0.295\textwidth]{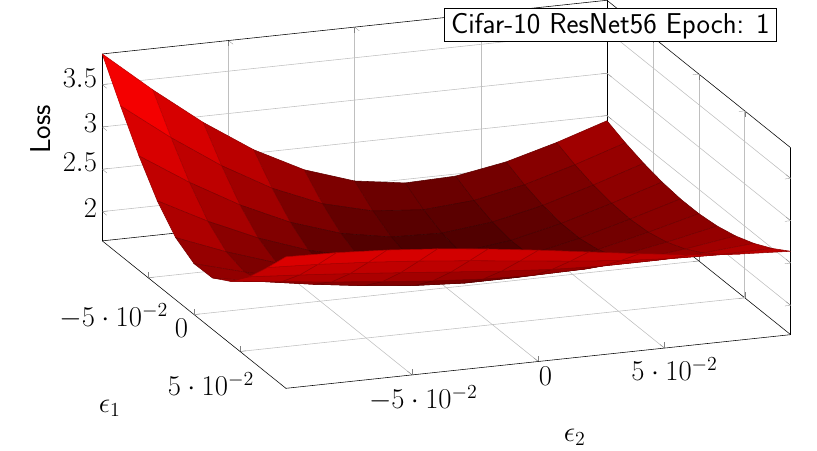}
\includegraphics[width=0.295\textwidth]{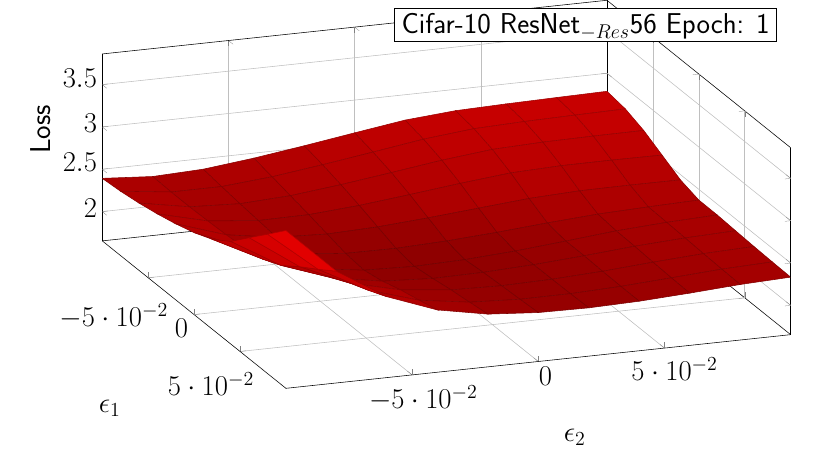}\\
\includegraphics[width=0.295\textwidth]{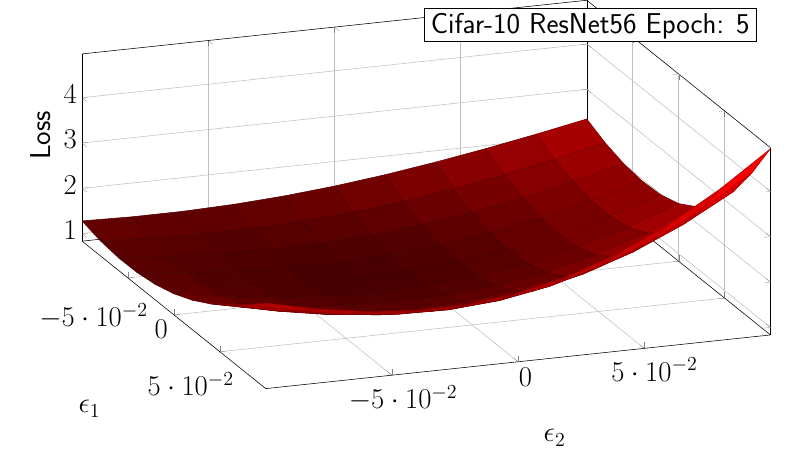}
\includegraphics[width=0.295\textwidth]{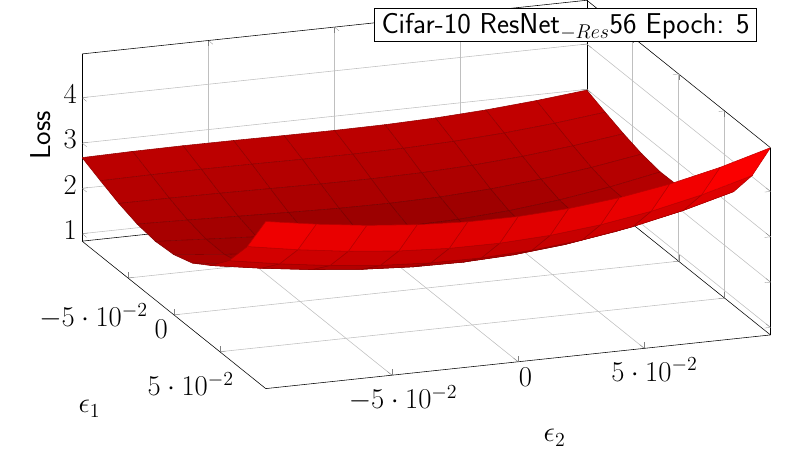}\\
\includegraphics[width=0.295\textwidth]{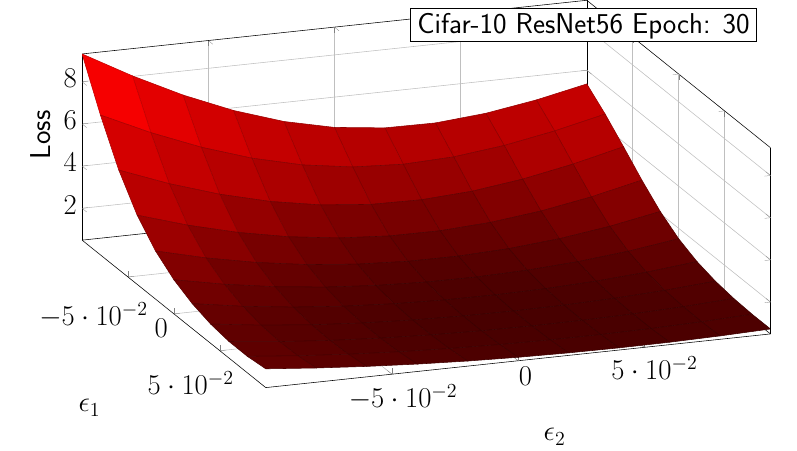}
\includegraphics[width=0.295\textwidth]{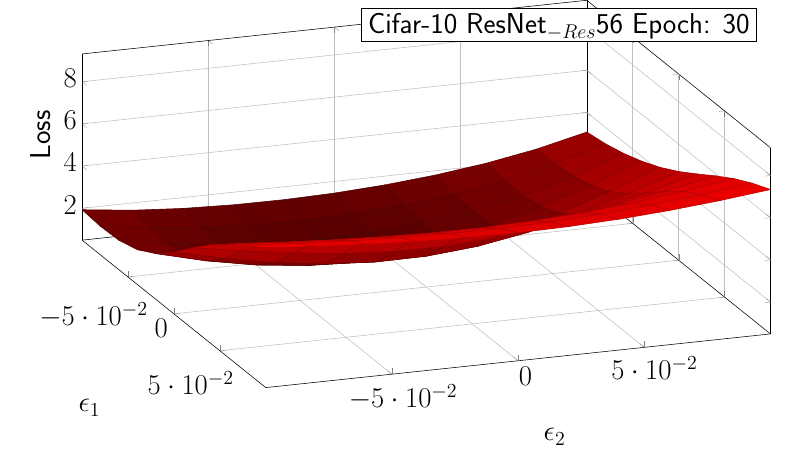}\\
\includegraphics[width=0.295\textwidth]{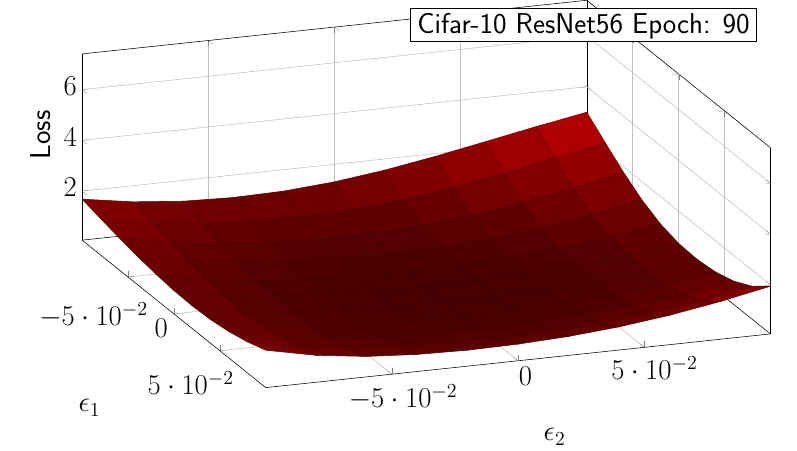}
\includegraphics[width=0.295\textwidth]{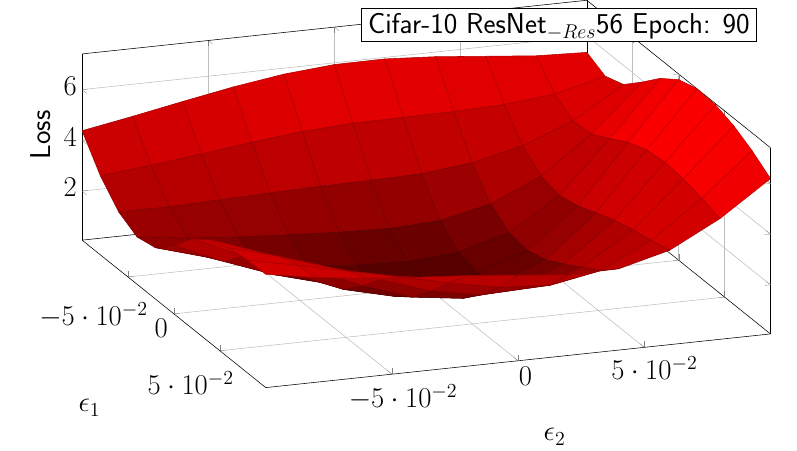}\\
\includegraphics[width=0.295\textwidth]{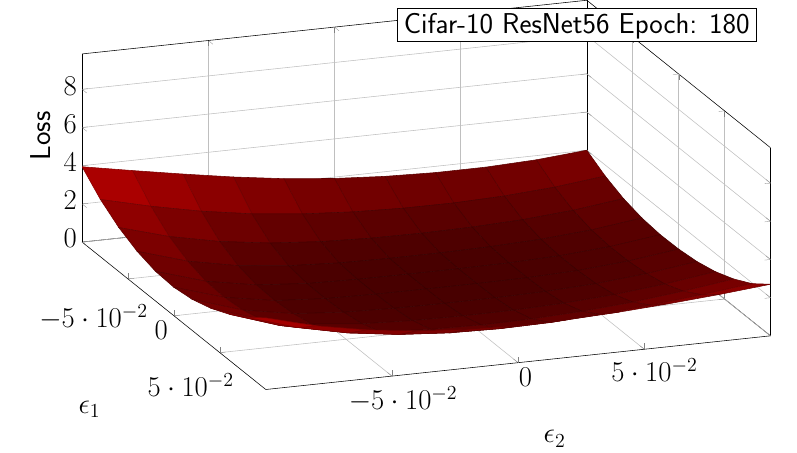}
\includegraphics[width=0.295\textwidth]{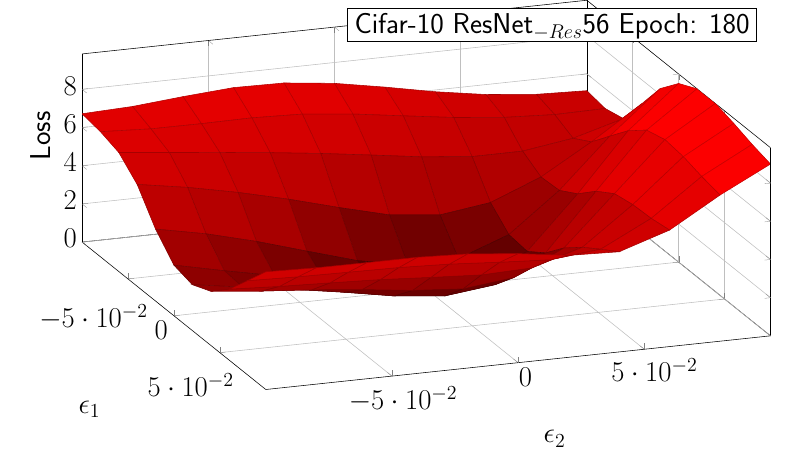}\\
\caption{
Loss landscape of ResNet/\ResNetRes56 on Cifar-10 with batch size 4096 by perturbing the parameters along the first two dominant eigenvectors of the Hessian. Note that the z-axis of ResNet56 at epoch 0 has different range than all the others. 
The loss landscape of \ResNetRes56 is indeed sharper than that of \ResNet56, which is align with the trace plot in~\fref{fig:resnet20/32/38-hut-full-net} and the Hessian ESD plot in~\fref{fig:resnet56-slq-full-net-all}.
}
  \label{fig:resnet56-loss-landscape-all}
\end{figure*}

\begin{figure*}[!htbp]
\centering
\includegraphics[width=0.295\textwidth]{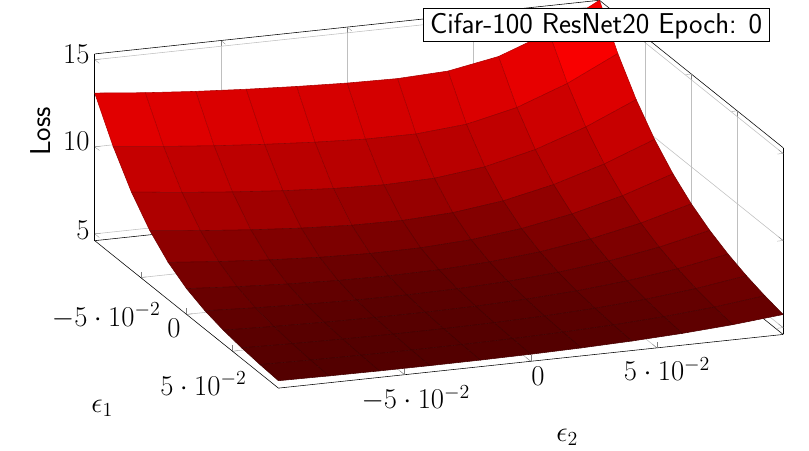}
\includegraphics[width=0.295\textwidth]{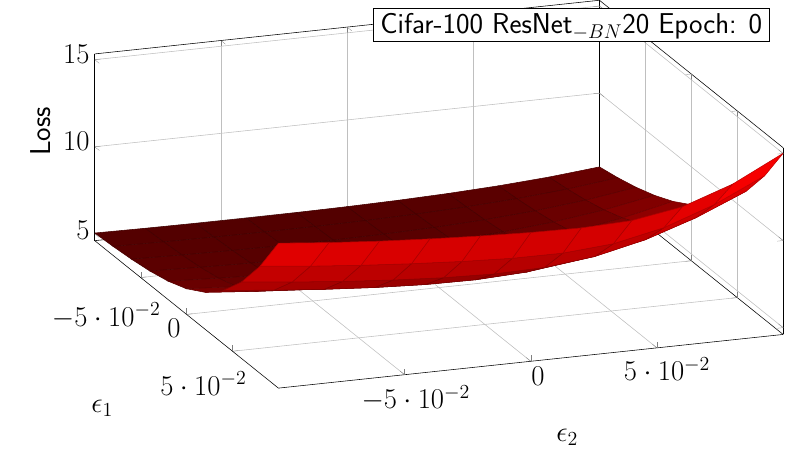}
\includegraphics[width=0.295\textwidth]{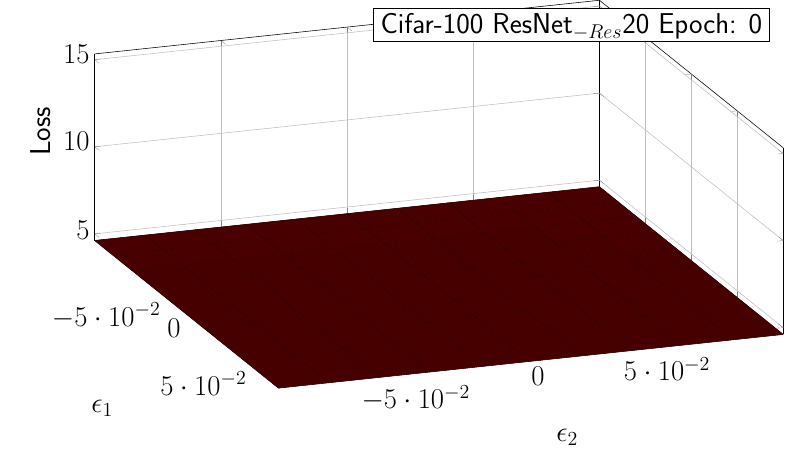}\\
\includegraphics[width=0.295\textwidth]{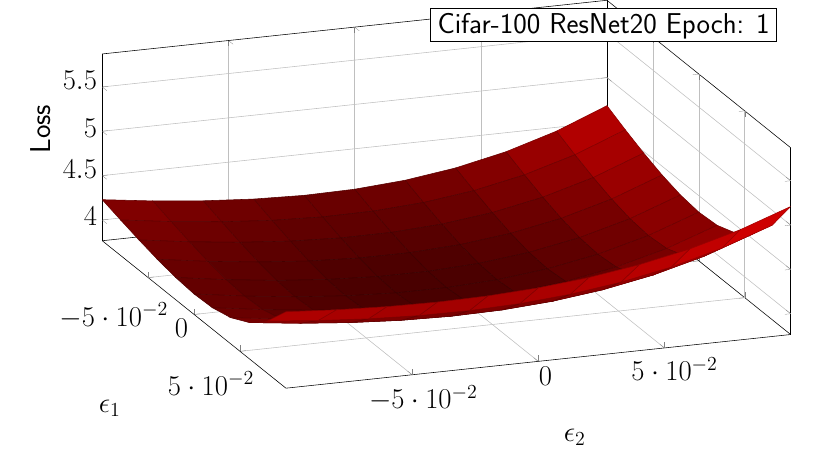}
\includegraphics[width=0.295\textwidth]{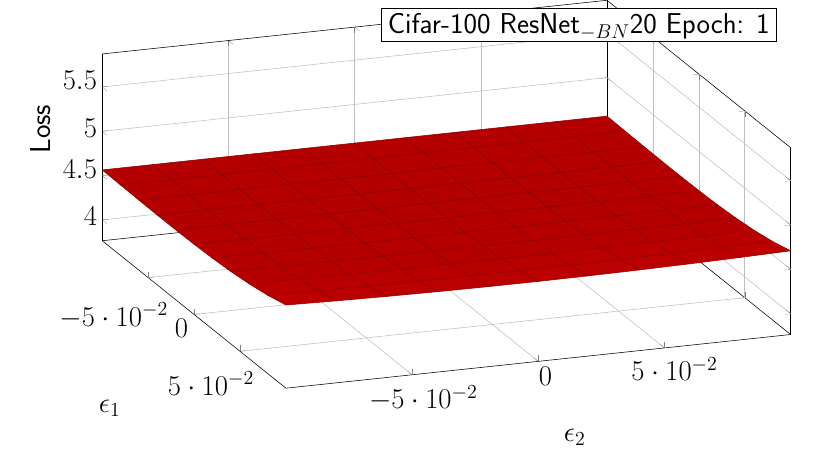}
\includegraphics[width=0.295\textwidth]{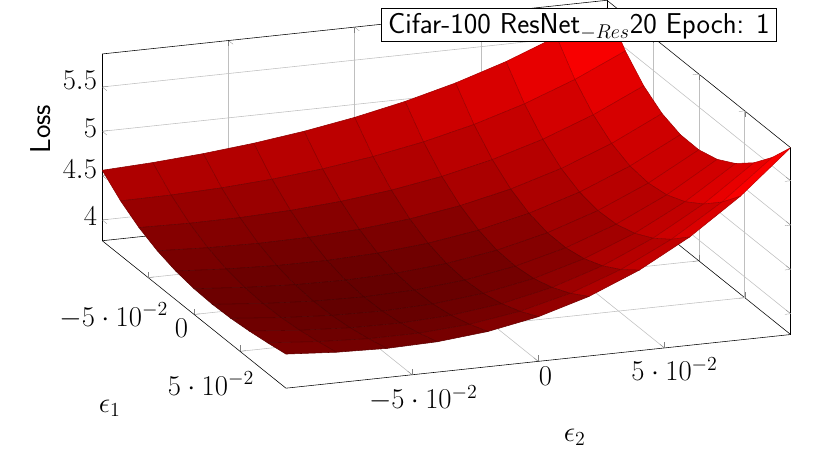}\\
\includegraphics[width=0.295\textwidth]{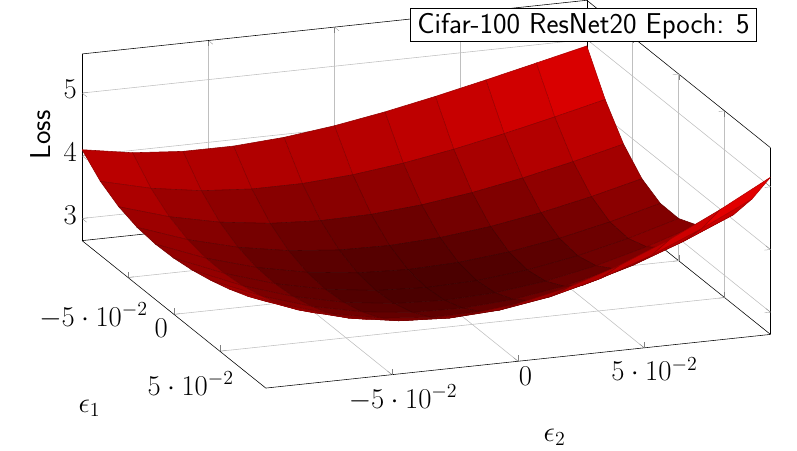}
\includegraphics[width=0.295\textwidth]{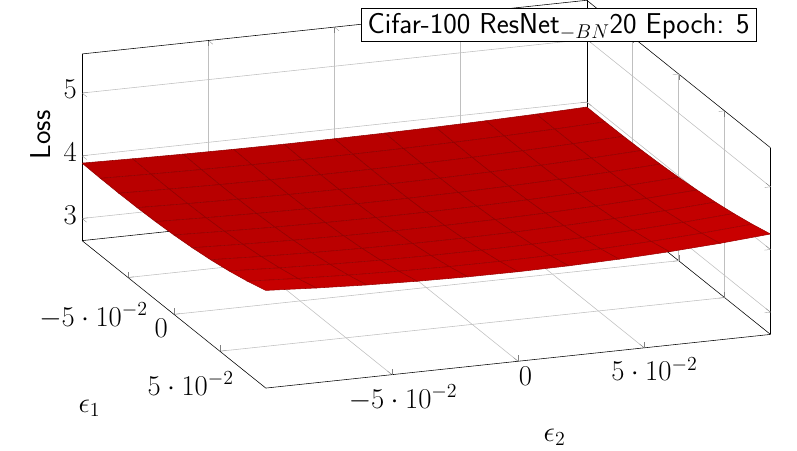}
\includegraphics[width=0.295\textwidth]{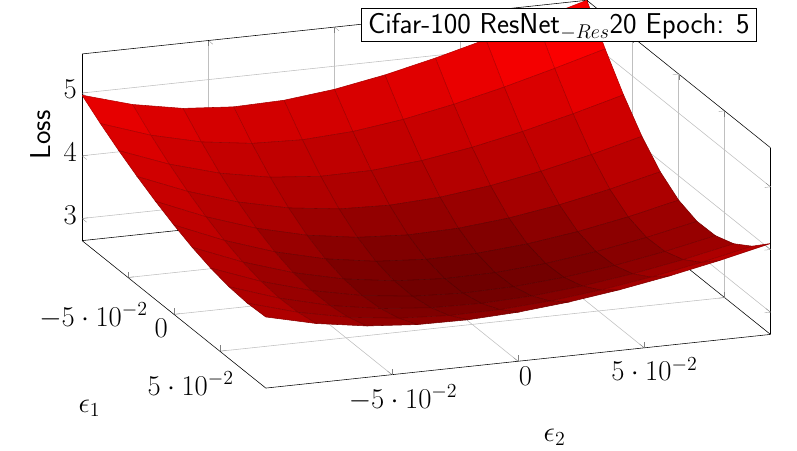}\\
\includegraphics[width=0.295\textwidth]{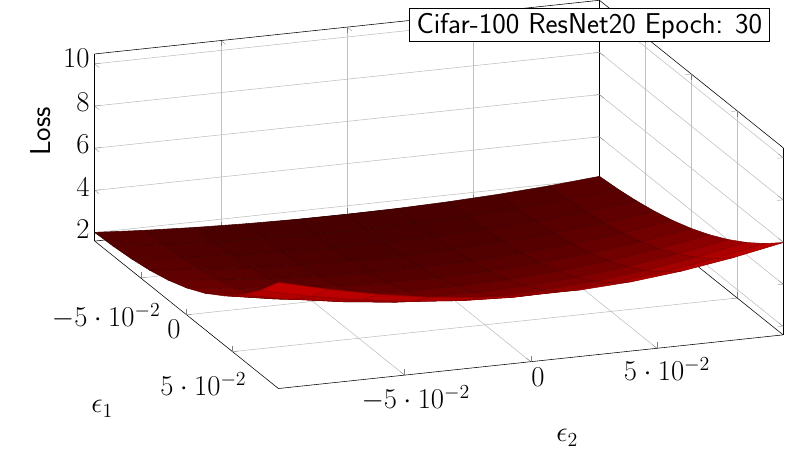}
\includegraphics[width=0.295\textwidth]{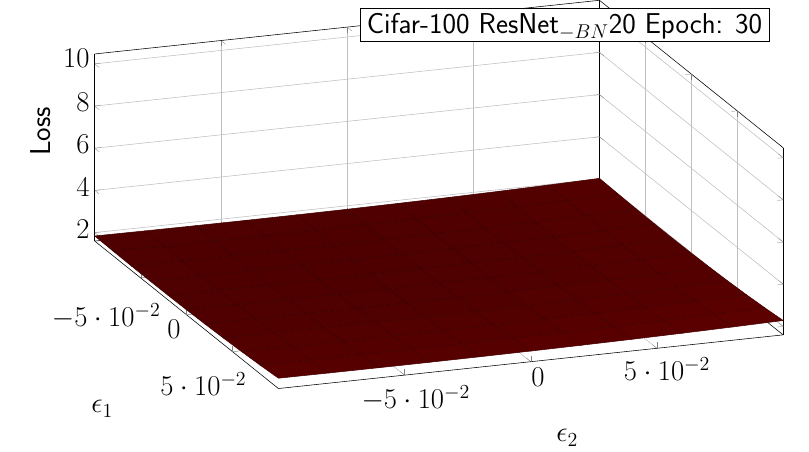}
\includegraphics[width=0.295\textwidth]{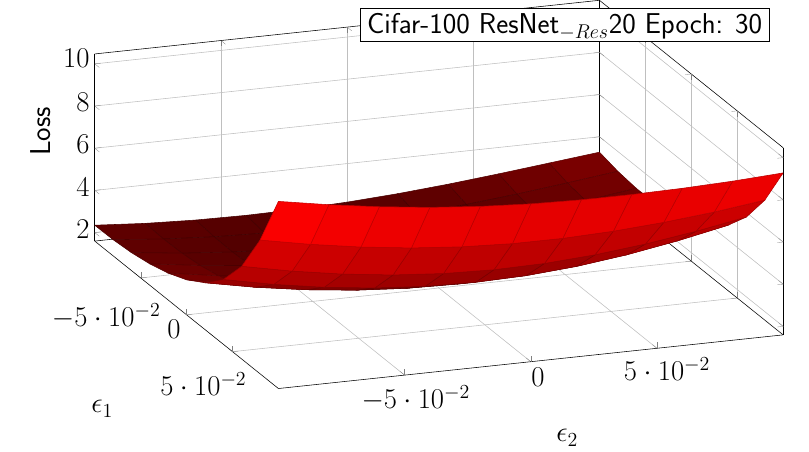}\\
\includegraphics[width=0.295\textwidth]{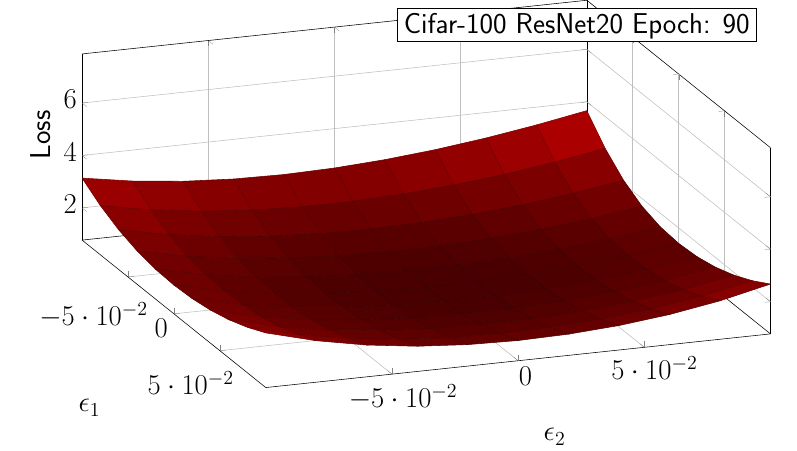}
\includegraphics[width=0.295\textwidth]{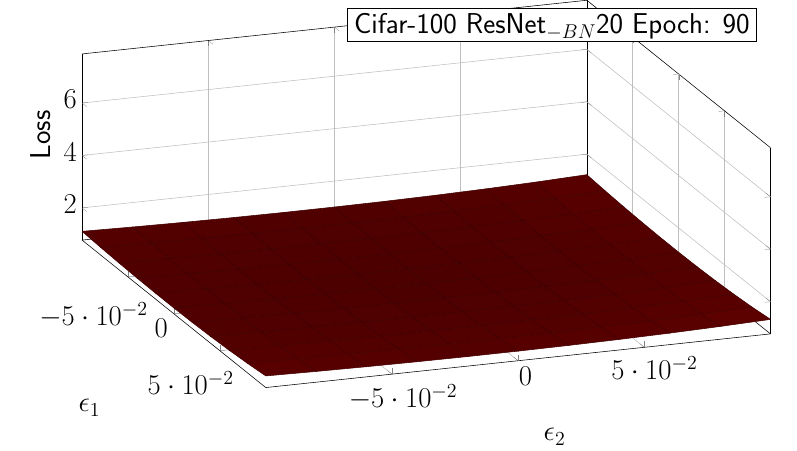}
\includegraphics[width=0.295\textwidth]{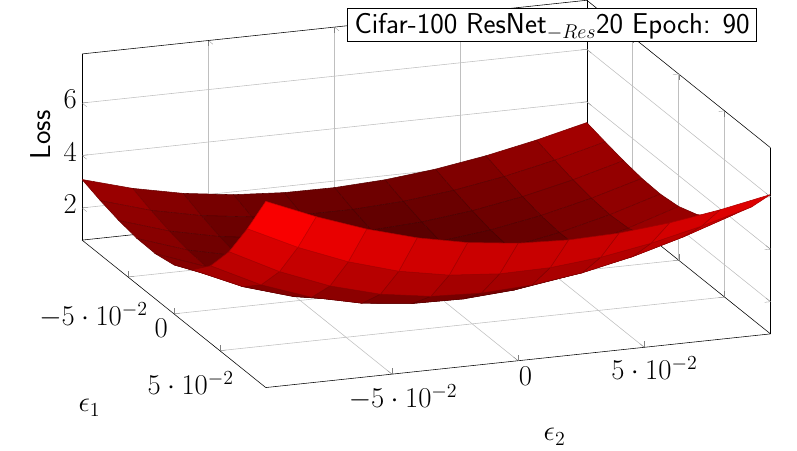}\\
\includegraphics[width=0.295\textwidth]{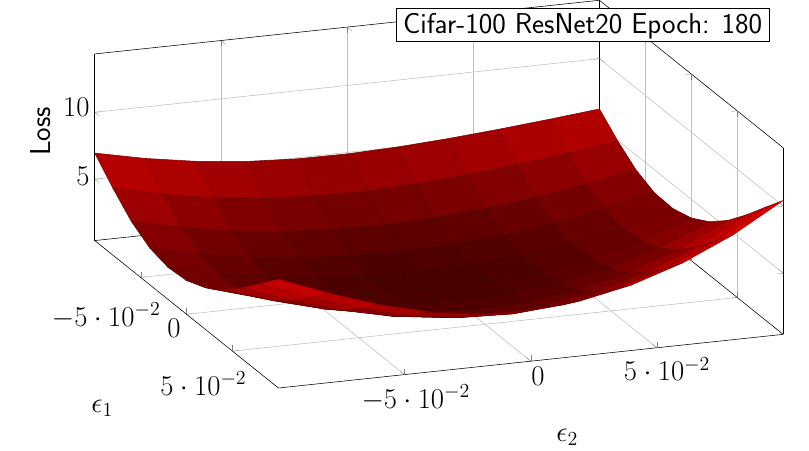}
\includegraphics[width=0.295\textwidth]{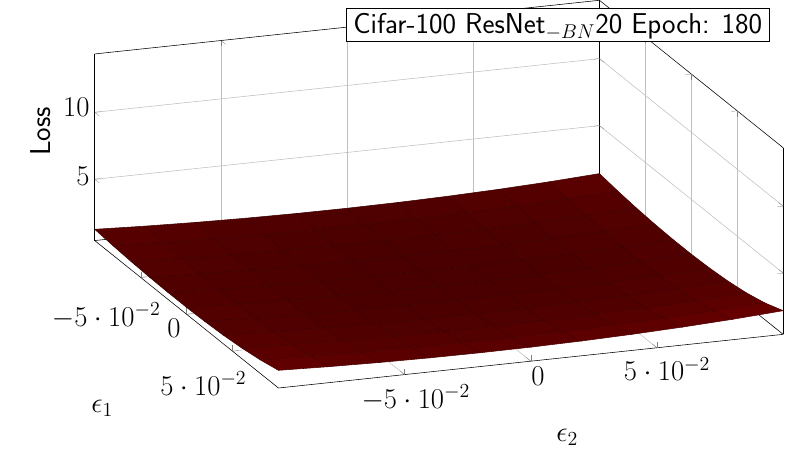}
\includegraphics[width=0.295\textwidth]{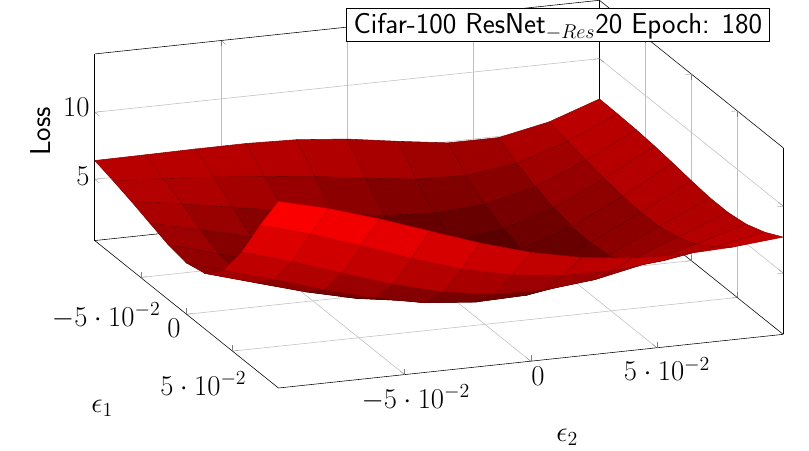}\\
\caption{
Loss landscape of ResNet/\ResNetBN/\ResNetRes20 on Cifar-100 with batch size 4096 by perturbing the parameters along the first two dominant eigenvectors of the Hessian. 
The loss landscape of \ResNetBN20 (\ResNetRes20) is indeed smoother (sharper) than that of \ResNet20, which is align with the trace plot in~\fref{fig:resnet20/32/38-hut-full-net-cifar100} and the Hessian ESD plot in~\fref{fig:resnet20-slq-full-net-all-cifar100}. 
}
  \label{fig:resnet20-loss-landscape-all-cifar100}
\end{figure*}

\begin{figure*}[!htbp]
\centering
\includegraphics[width=0.295\textwidth]{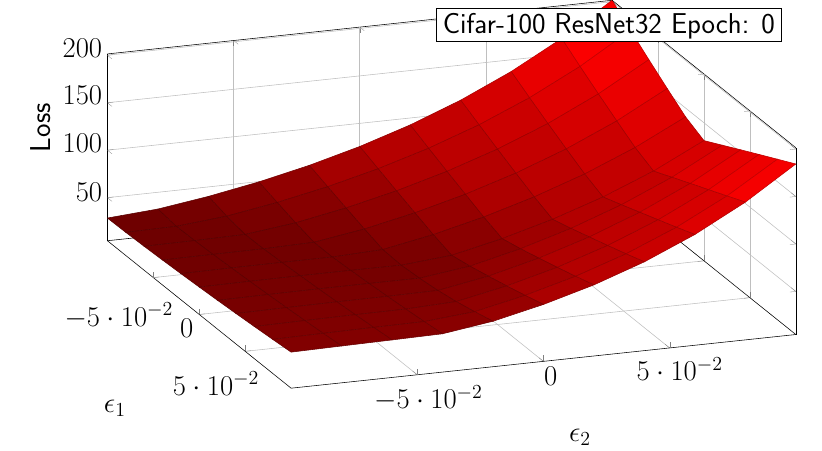}
\includegraphics[width=0.295\textwidth]{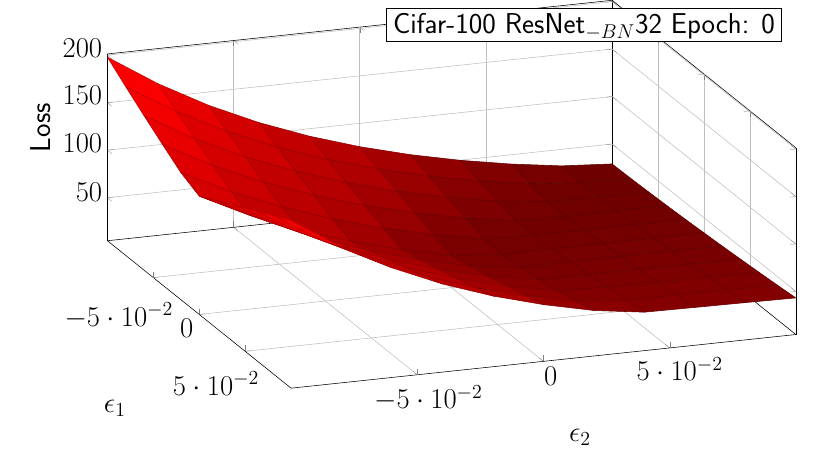}
\includegraphics[width=0.295\textwidth]{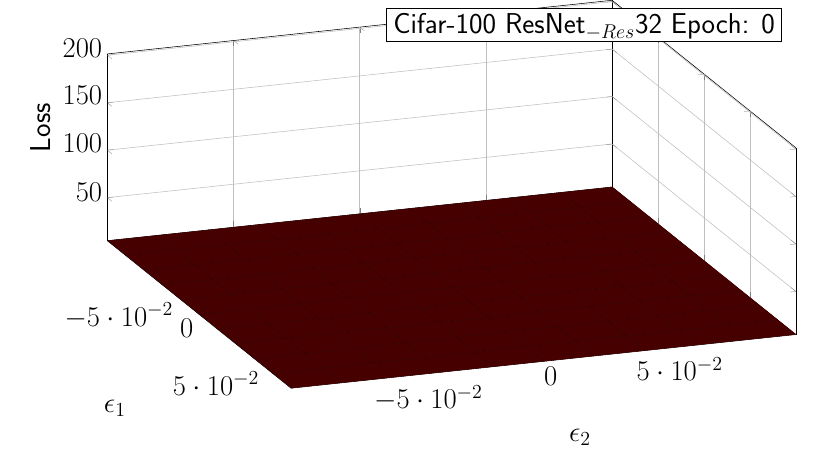}\\
\includegraphics[width=0.295\textwidth]{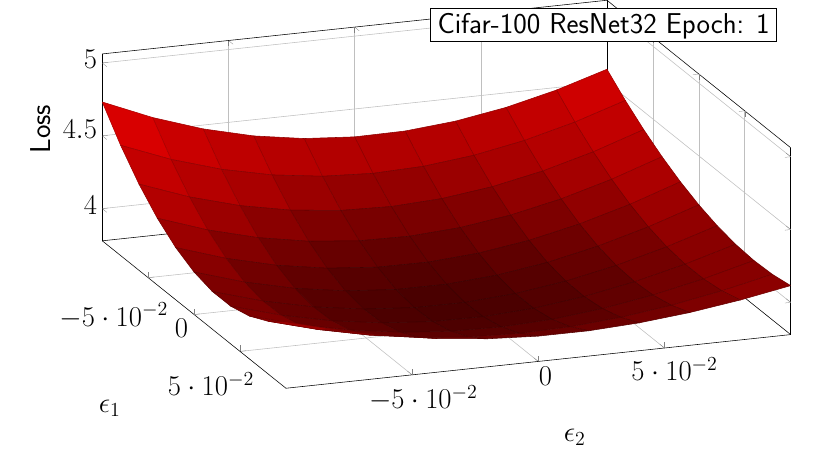}
\includegraphics[width=0.295\textwidth]{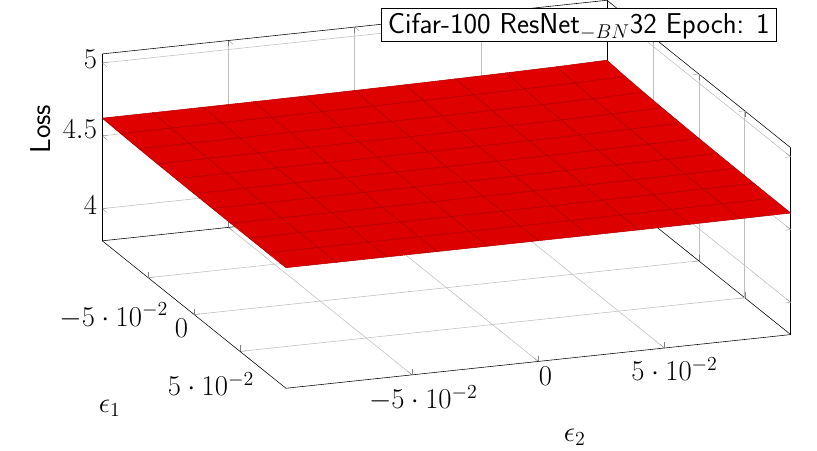}
\includegraphics[width=0.295\textwidth]{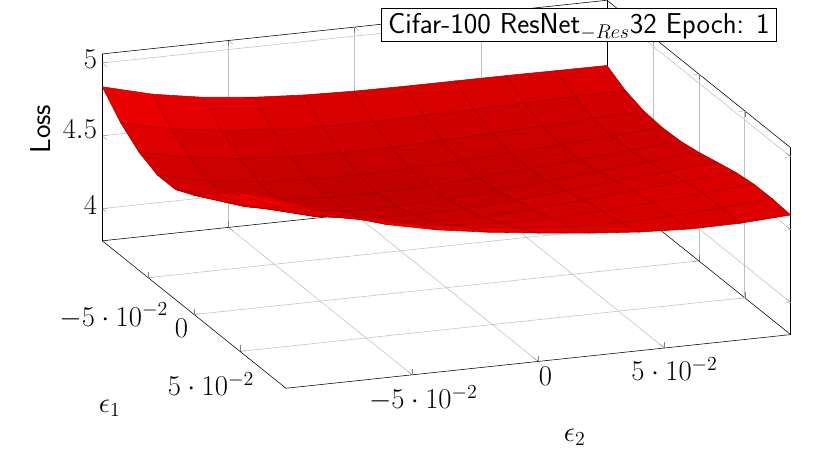}\\
\includegraphics[width=0.295\textwidth]{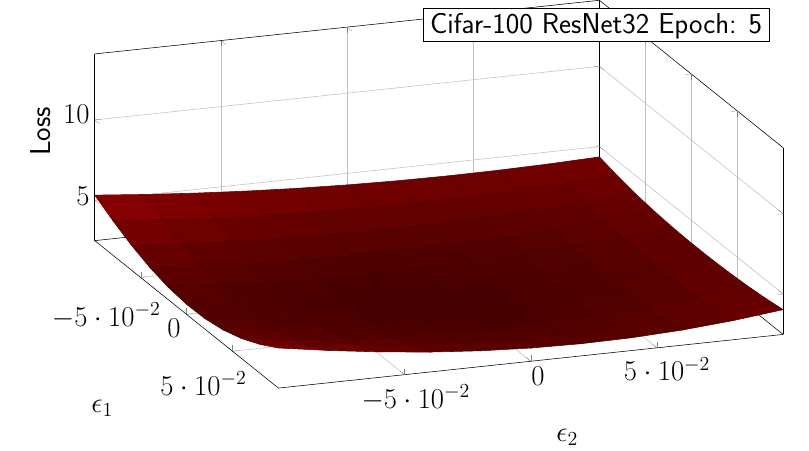}
\includegraphics[width=0.295\textwidth]{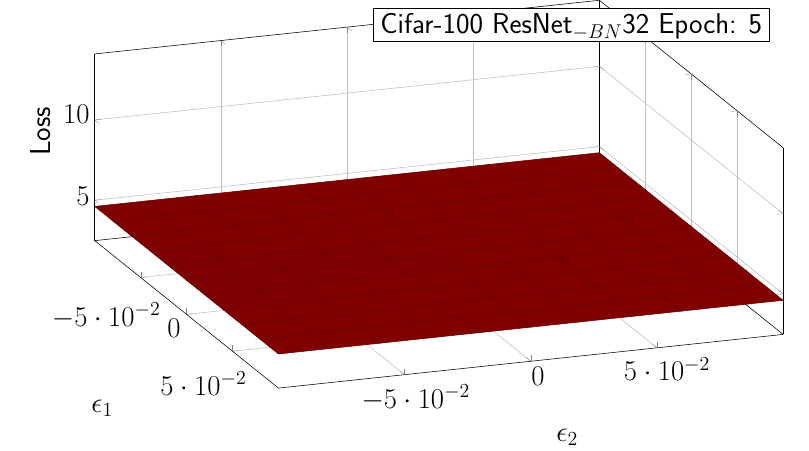}
\includegraphics[width=0.295\textwidth]{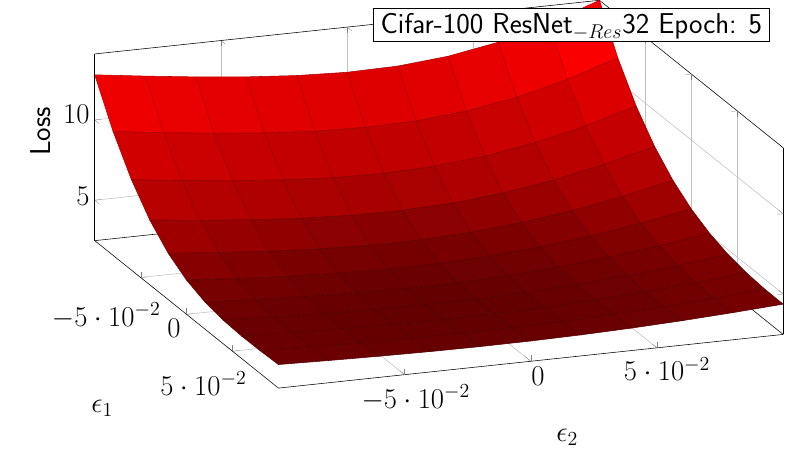}\\
\includegraphics[width=0.295\textwidth]{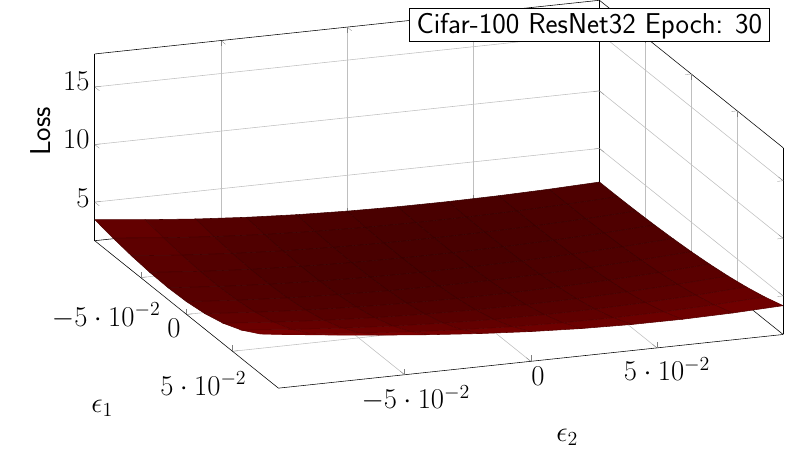}
\includegraphics[width=0.295\textwidth]{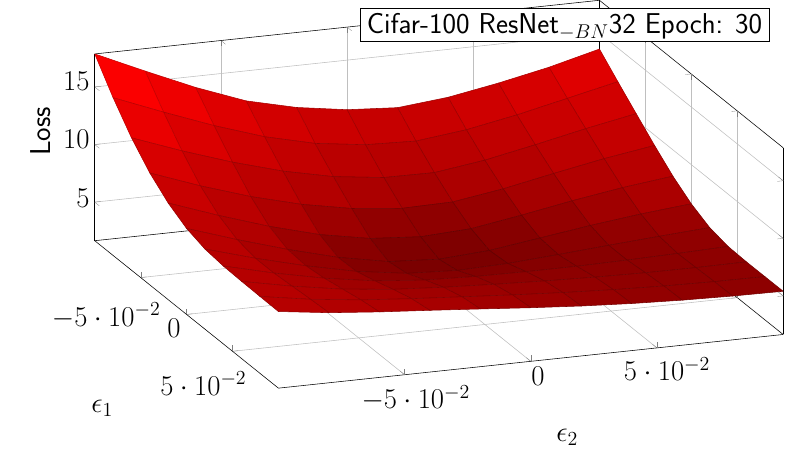}
\includegraphics[width=0.295\textwidth]{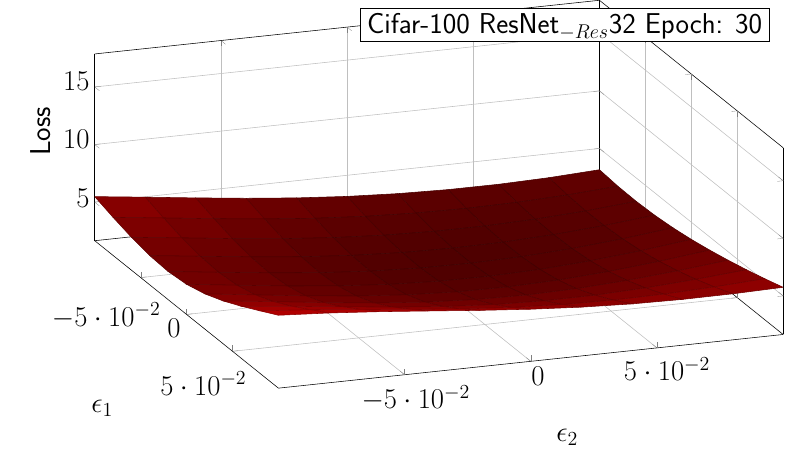}\\
\includegraphics[width=0.295\textwidth]{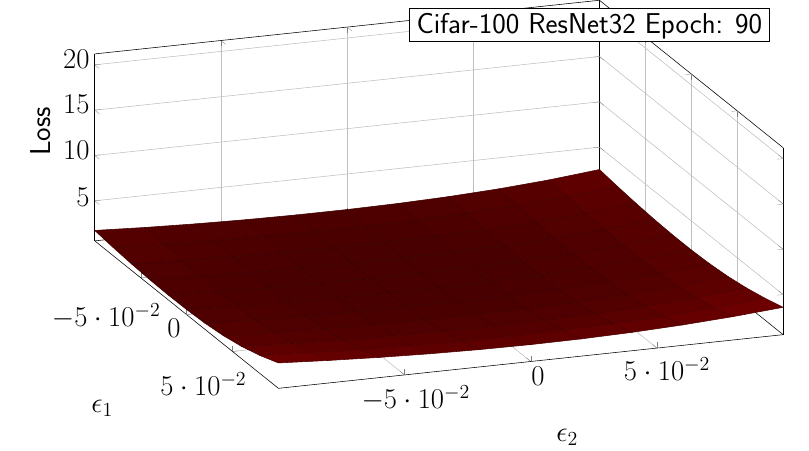}
\includegraphics[width=0.295\textwidth]{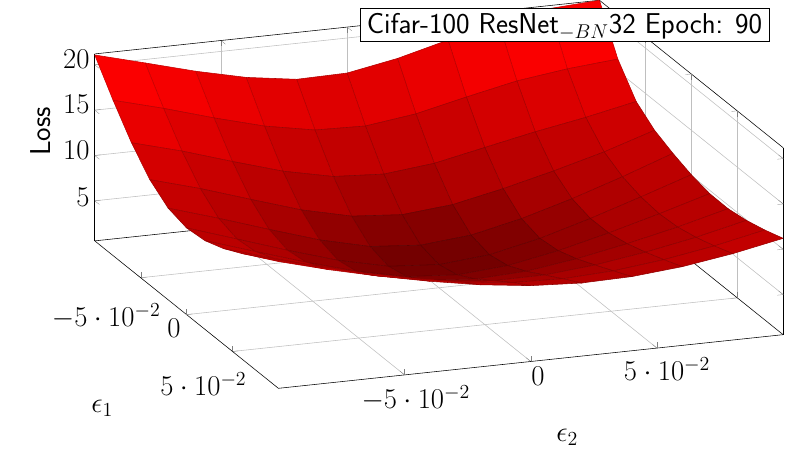}
\includegraphics[width=0.295\textwidth]{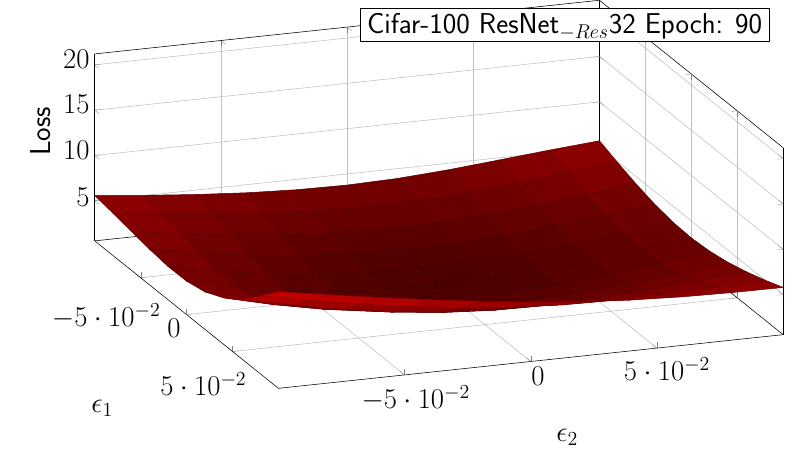}\\
\includegraphics[width=0.295\textwidth]{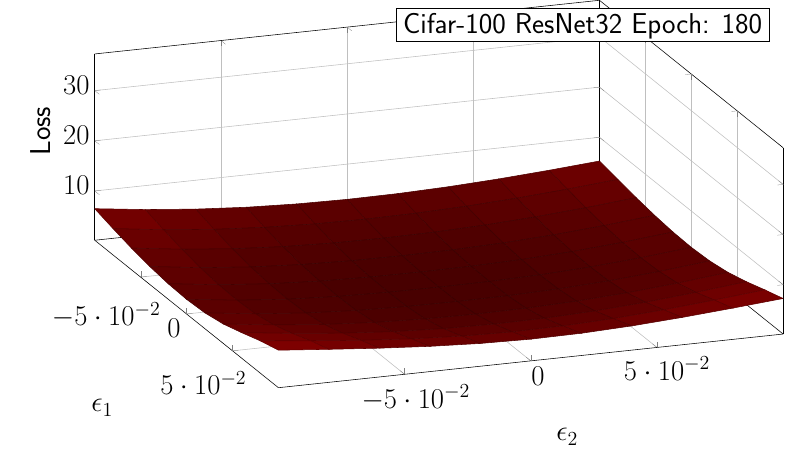}
\includegraphics[width=0.295\textwidth]{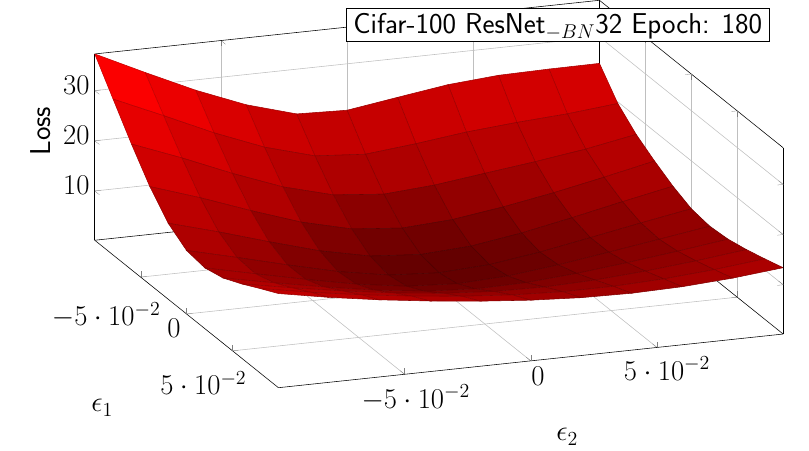}
\includegraphics[width=0.295\textwidth]{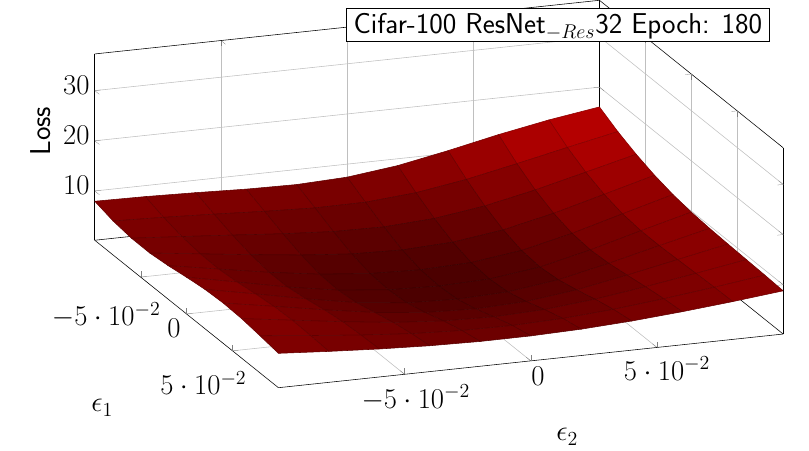}\\
\caption{
Loss landscape of ResNet/\ResNetBN/\ResNetRes32 on Cifar-100 with batch size 4096 by perturbing the parameters along the first two dominant eigenvectors of the Hessian. 
The loss landscape of \ResNetBN32/\ResNetRes32 is indeed sharper than that of \ResNet32, which is align with the trace plot in~\fref{fig:resnet20/32/38-hut-full-net-cifar100} and the Hessian ESD plot in~\fref{fig:resnet32-slq-full-net-all-cifar100}.
}
  \label{fig:resnet32-loss-landscape-all-cifar100}
\end{figure*}

\begin{figure*}[!htbp]
\centering
\includegraphics[width=0.295\textwidth]{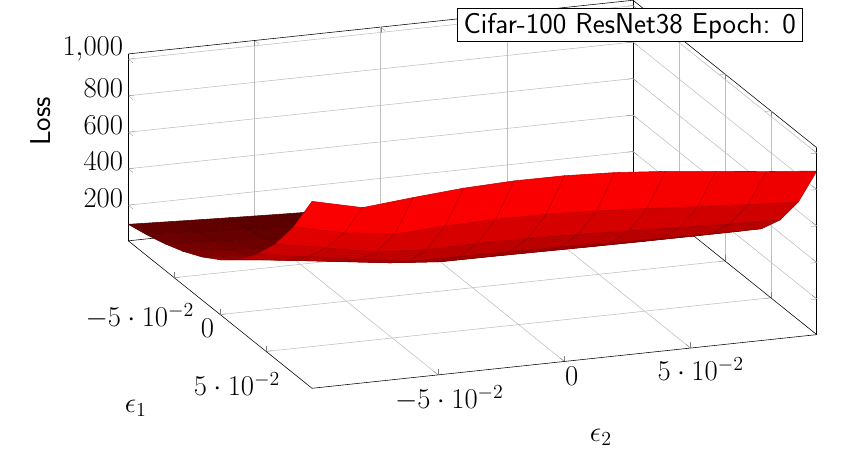}
\includegraphics[width=0.295\textwidth]{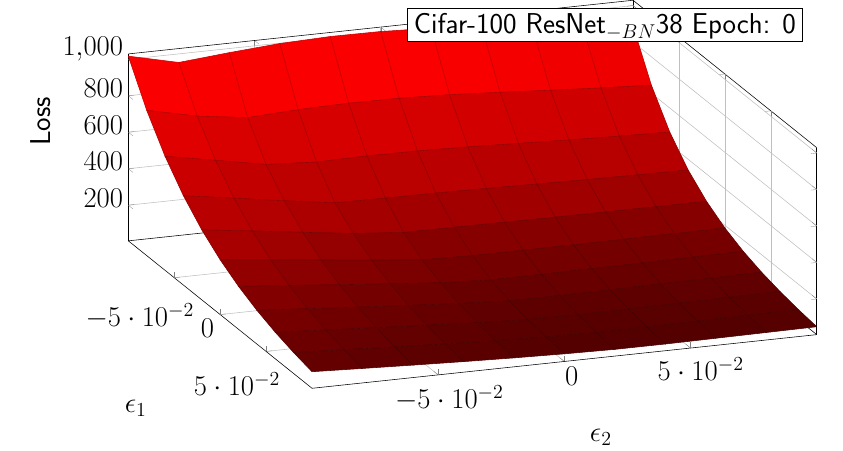}
\includegraphics[width=0.295\textwidth]{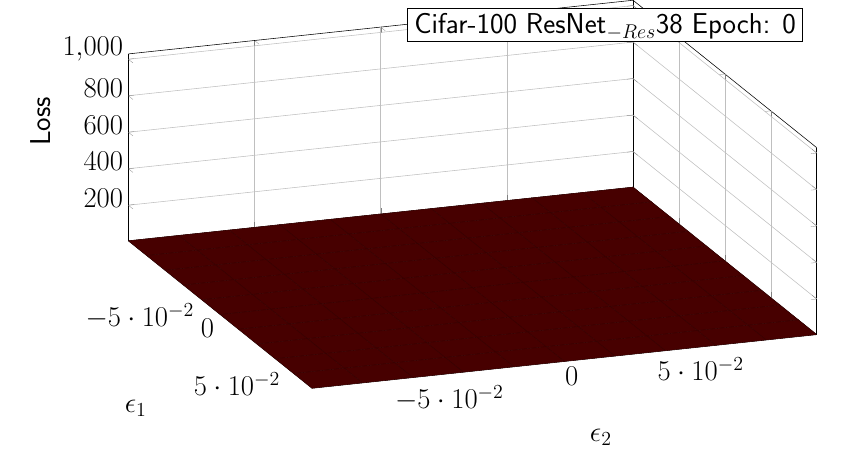}\\
\includegraphics[width=0.295\textwidth]{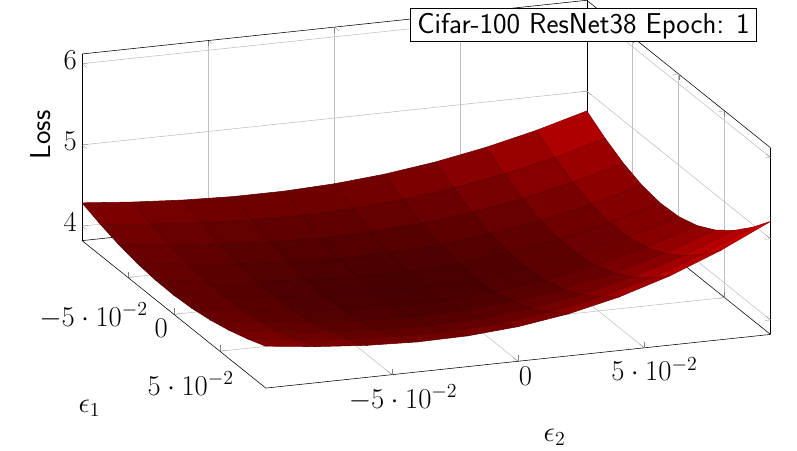}
\includegraphics[width=0.295\textwidth]{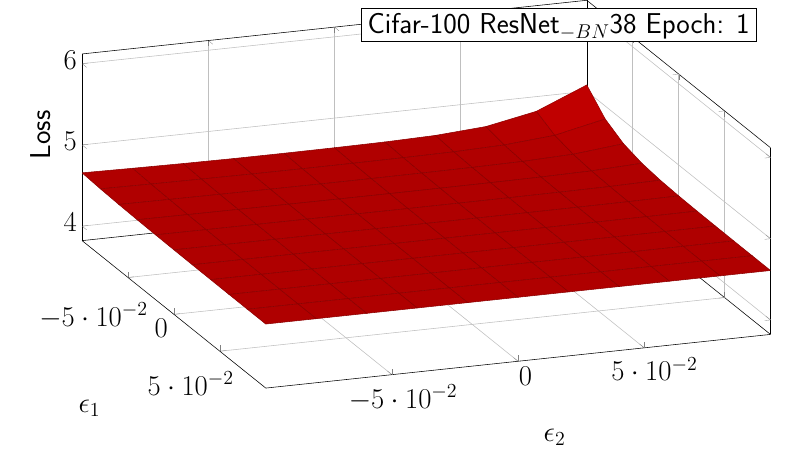}
\includegraphics[width=0.295\textwidth]{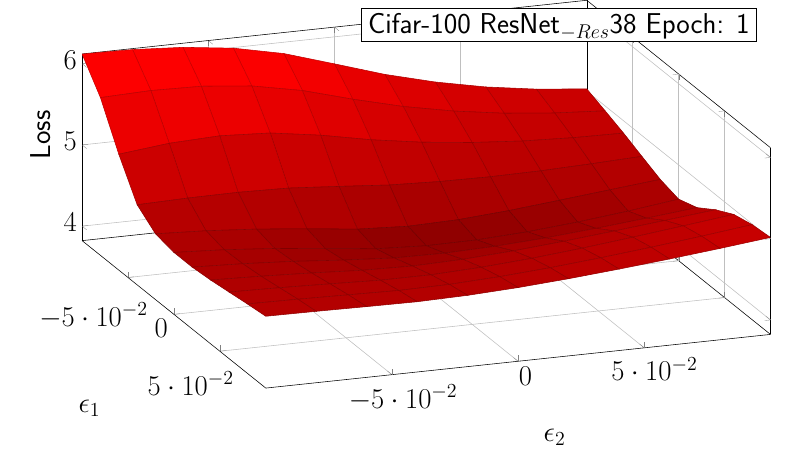}\\
\includegraphics[width=0.295\textwidth]{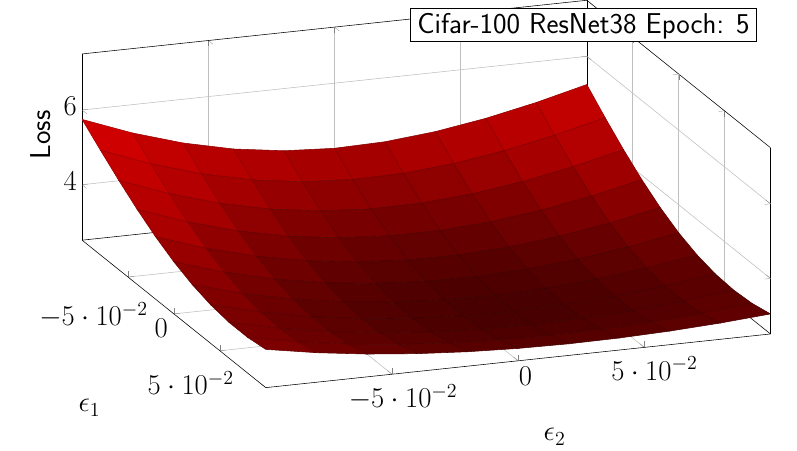}
\includegraphics[width=0.295\textwidth]{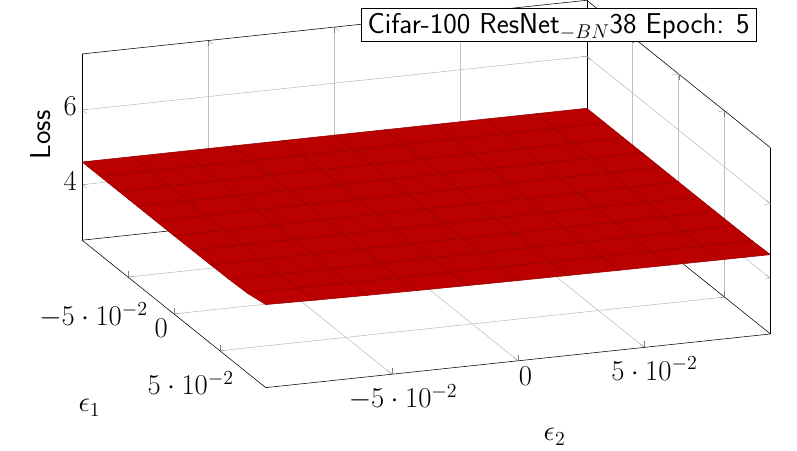}
\includegraphics[width=0.295\textwidth]{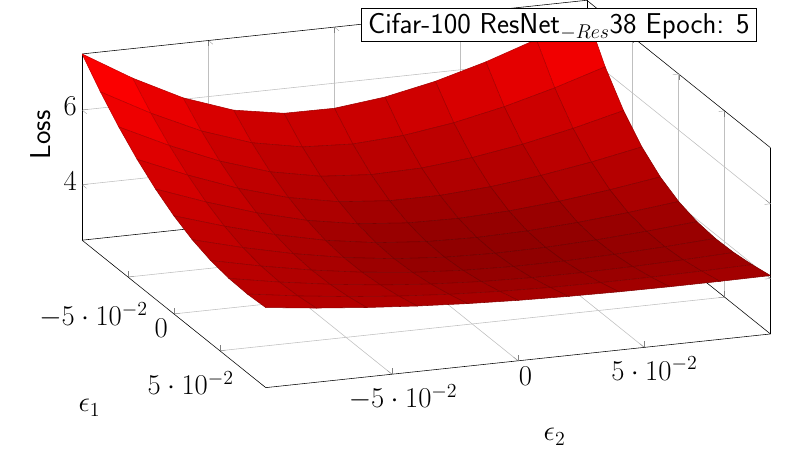}\\
\includegraphics[width=0.295\textwidth]{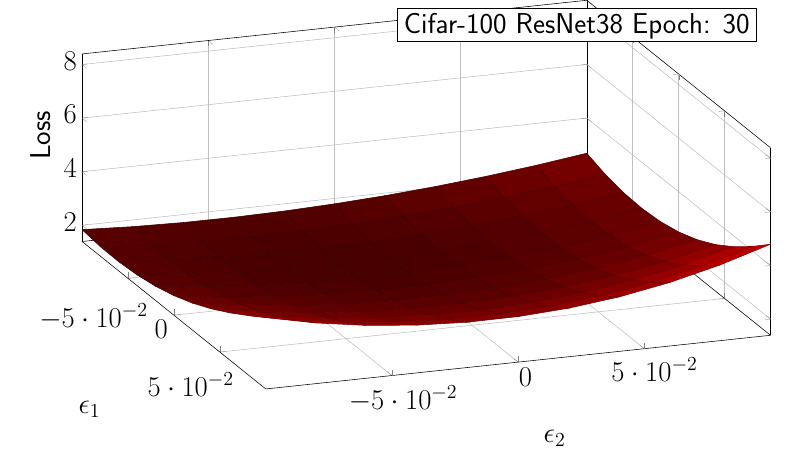}
\includegraphics[width=0.295\textwidth]{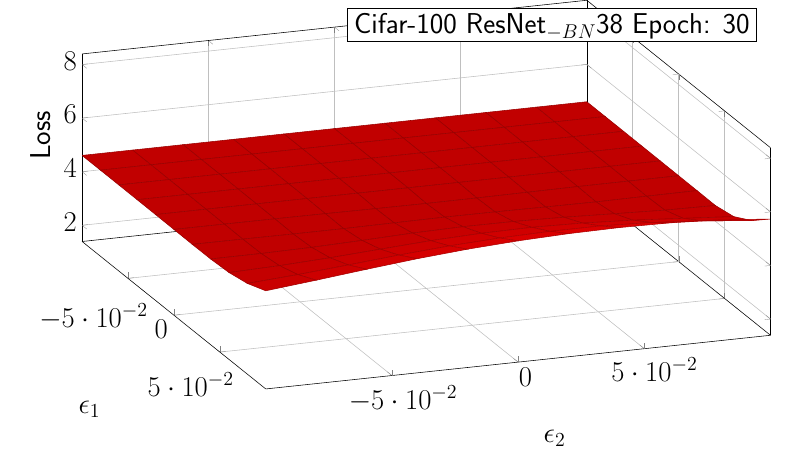}
\includegraphics[width=0.295\textwidth]{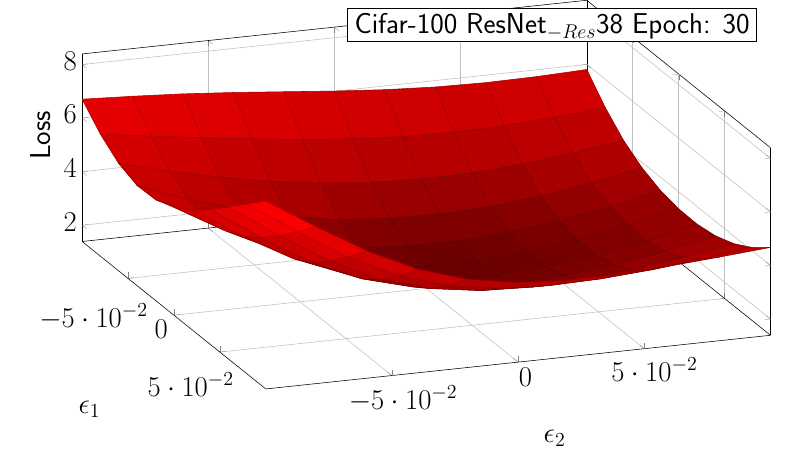}\\
\includegraphics[width=0.295\textwidth]{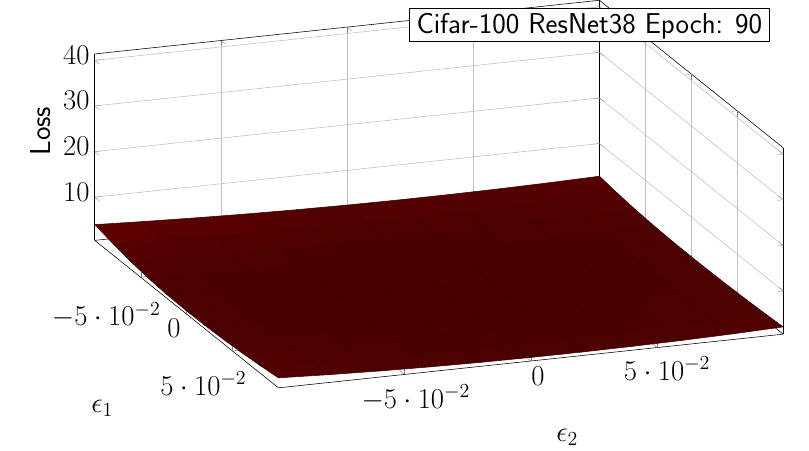}
\includegraphics[width=0.295\textwidth]{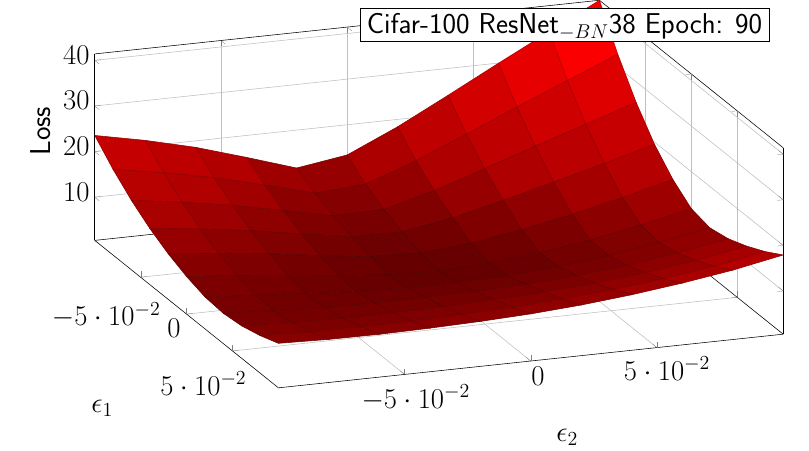}
\includegraphics[width=0.295\textwidth]{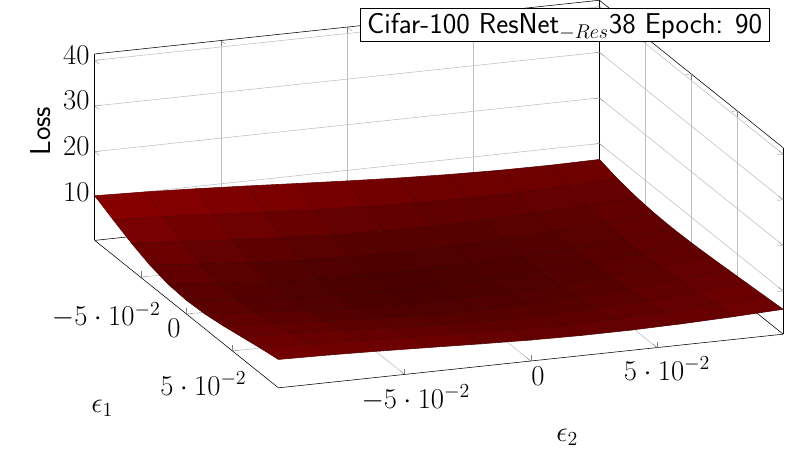}\\
\includegraphics[width=0.295\textwidth]{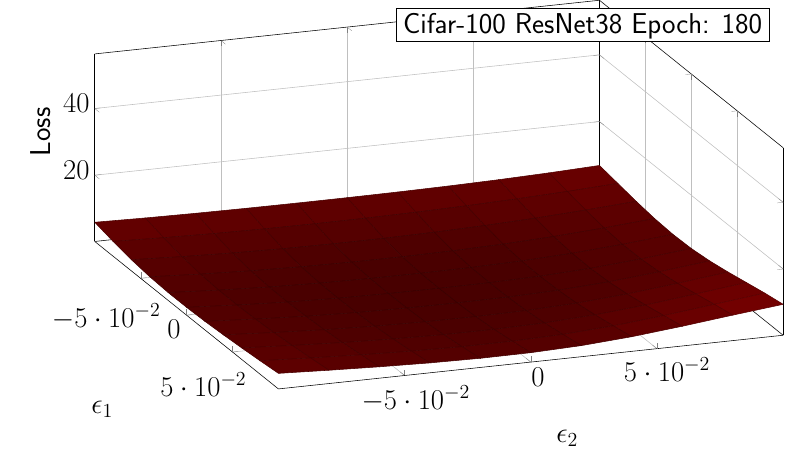}
\includegraphics[width=0.295\textwidth]{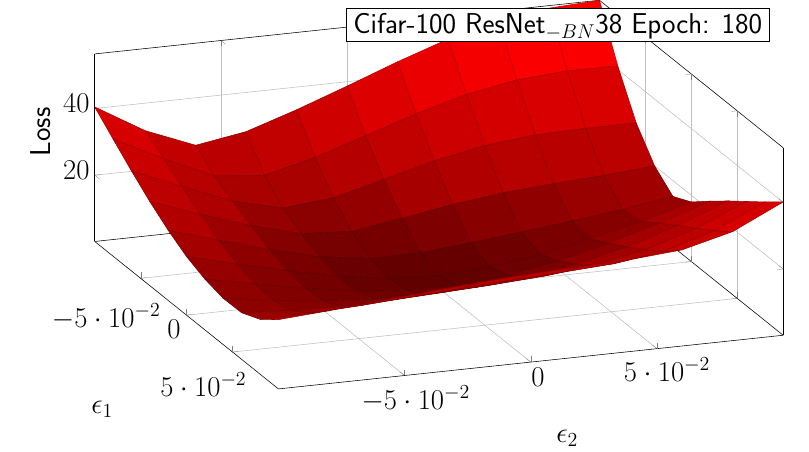}
\includegraphics[width=0.295\textwidth]{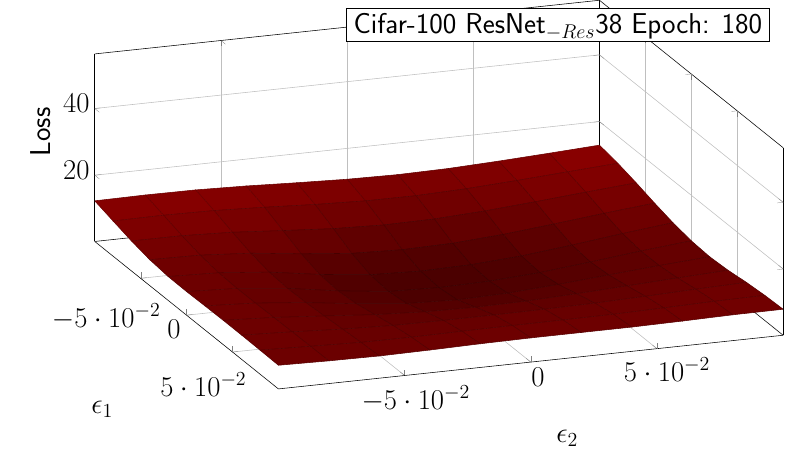}\\
\caption{
Loss landscape of ResNet/\ResNetBN/\ResNetRes38 on Cifar-100 with batch size 4096 by perturbing the parameters along the first two dominant eigenvectors of the Hessian. 
The loss landscape of \ResNetBN38/\ResNetRes38 is indeed sharper than that of \ResNet38, which is align with the trace plot in~\fref{fig:resnet20/32/38-hut-full-net-cifar100} and the Hessian ESD plot in~\fref{fig:resnet38-slq-full-net-all-cifar100}.
}
  \label{fig:resnet38-loss-landscape-all-cifar100}
\end{figure*}

\begin{table}[!ht]
\caption{
Accuracy of ResNet models on Cifar-100 with different depths is shown in the first row.
In the second through the last rows, we report the accuracy of the corresponding architectures, but with BN layer removed from one of the stages, respectively. 
(See~\fref{fig:resnet20_illustration} for stage definition.)
For instance, the last row reports ResNet model with no BN layer in the third~stage.
}
\small
\setlength\tabcolsep{2.35pt}
\label{tab:model_acc_rm_bn-cifar100}
\centering
\begin{tabular}{lcccccccccccccc} \toprule
 Model\textbackslash Depth  & 20       & 32       & 38     \\ 
\midrule 
\hc ResNet                 & 66.47\%  & 68.26\%  & 69.06\%  \\
\ha RM BN stage 1          & 65.69\%  & 65.74\%  & 67.31\% \\
\hc RM BN stage 2          & 65.62\%  & 64.68\%  & 66.46\% \\
\ha RM BN stage 3          & 65.63\%  & 64.57\%  & 61.04\% \\
     \bottomrule 
\end{tabular}
\end{table}

\begin{table}[!htbp]
\caption{
Accuracy of ResNet on Cifar-100 is reported for baseline (first row), along with architectures where the residual connection is removed at different stages.
}
\small
\setlength\tabcolsep{2.35pt}
\label{tab:model_acc_rm_res_cifar100}
\centering
\begin{tabular}{lcccccccccccccc} \toprule
 Model\textbackslash Depth  & 20       & 32       & 38      \\ 
\midrule 
\hc ResNet                 & 66.47\%  & 68.26\% & 69.06\%  \\
\ha RM Res stage 1         & 66.46\%  &66.94\%  & 67.61\%\\
\hc RM Res stage 2         & 65.70\%  &66.05\%  & 66.70\%\\
\ha RM Res stage 3         & 66.21\%  &66.38\%  & 66.03\%\\
     \bottomrule 
\end{tabular}
\end{table}


\end{document}